\documentclass[11pt]{article}

\usepackage[final]{acl}

\usepackage{times}
\usepackage{latexsym}
\usepackage{amsmath}
\usepackage{booktabs}
\usepackage{multirow}
\usepackage{xcolor}
\usepackage{url}
\usepackage{xspace}
\usepackage{pifont}
\usepackage{pdflscape}
\usepackage{fancyhdr}
\usepackage[T1]{fontenc}

\usepackage[utf8]{inputenc}
\usepackage[table]{xcolor} 
\usepackage{microtype}

\usepackage{inconsolata}

\usepackage{graphicx}
\usepackage{rotating}
\usepackage{adjustbox}
\usepackage{float}  

\usepackage{enumitem}        
\usepackage{hyperref}        
\usepackage{cleveref}        

\usepackage{tikz}
\usepackage{pgfplots}
\pgfplotsset{compat=1.18}
\usetikzlibrary{patterns}

\usepackage{amsmath}      
\usepackage{amssymb}      
\usepackage{amsfonts}     

\usepackage{tikz}
\usepackage{pgfplots}
\pgfplotsset{compat=1.18}  

\usepackage{array}         

\usepackage{tcolorbox}

\usepackage{booktabs}    
\usepackage{pifont}      

\newcommand{\cmark}{\ding{51}}  
\newcommand{\xmark}{\ding{55}}  

\definecolor{fprgreen}{RGB}{76,175,80}
\definecolor{fnrred}{RGB}{229,115,115}
\definecolor{panelbg}{RGB}{255,248,235}

\newcommand{\method}{\textsc{DIA-HARM}\xspace}

\newcommand{\sae}{SAE\xspace}

\newcommand{\eqcontrib}{\textsuperscript{*}}

\title{\method: Dialectal Disparities in Harmful Content Detection  Across 50 English Dialects}

\author{Jason Lucas$^1$, Matt Murtagh-White\eqcontrib$^2$, Ali Al-Lawati\eqcontrib$^1$, Uchendu Uchendu$^1$, Adaku Uchendu$^3$ \\ \textbf{Dongwon Lee}$^1$
\vspace{0.05in} \\
  $^1$ The Pennsylvania State University, USA \\
  $^2$ Trinity College Dublin, Ireland \\
  $^3$ MIT Lincoln Laboratory, USA \\
  \vspace{0.05in}
{\small \textit{\{jsl5710, dongwon\}}@psu.edu; \eqcontrib Equal contribution as co-second authors}
   }

\begin{document}
\maketitle
\begin{abstract}
Harmful content detectors---particularly disinformation classifiers---are predominantly developed and evaluated on Standard American English (\sae{}), leaving their robustness to dialectal variation unexplored. We present \method{}, the first benchmark for evaluating disinformation detection robustness across 50 English dialects spanning U.S., British, African, Caribbean, and Asia-Pacific varieties. Using Multi-VALUE's linguistically-grounded transformations, we introduce D-CUBE (Dialectal Disinformation Detection Corpus), a core corpus component of 
\method{} comprising 195K samples derived from established disinformation 
benchmarks. Our evaluation of 16 detection models reveals systematic vulnerabilities: human-written dialectal content degrades detection by 1.4--3.6\% F1, while AI-generated content remains stable. Fine-tuned transformers substantially outperform zero-shot LLMs (96.6\% vs. 78.3\% best-case F1), with some models exhibiting catastrophic failures exceeding 33\% degradation on mixed content. Cross-dialectal transfer analysis across 2,450 dialect pairs shows that multilingual models (mDeBERTa: 97.2\% average F1) generalize effectively, while monolingual models like RoBERTa and XLM-RoBERTa fail on dialectal inputs. These findings demonstrate that current disinformation detectors may systematically disadvantage hundreds of millions of non-\sae{} speakers worldwide. We release the \method{} benchmark, including the \href{https://github.com/jsl5710/dia-harm}{D-CUBE corpus}, and evaluation tools\footnote{
\url{https://jsl5710.github.io/dia-harm}}
\end{abstract}

\section{Introduction}
\label{sec:introduction}

Harmful content detectors---including disinformation classifiers---serve as critical infrastructure for protecting users from false and misleading information that threatens public health, democratic processes, and social cohesion \cite{jsl10702034, bmm10718648}. For these systems to fulfill their protective function, they must be both \textit{robust} (maintaining performance across input variations) and \textit{equitable} (providing consistent protection regardless of how users express themselves). However, current disinformation detection systems are predominantly developed and evaluated on Standard American English (\sae), leaving their robustness to dialectal variation largely unexplored. This gap matters: English exhibits substantial variation across regional, social, and economic dimensions \cite{ziems-etal-2023-multi}, with hundreds of millions of speakers worldwide communicating using dialectal varieties that differ systematically from \sae{} in morphology, syntax, and lexical choice \cite{joshi2025natural, faisal-etal-2024-dialectbench}. These are not errors but legitimate linguistic systems deserving equal protection \cite{peppin2025multilingual}.

\begin{figure}[t]
    \centering
    \includegraphics[width=\columnwidth, trim={225bp 137bp 225bp 130bp}, clip]{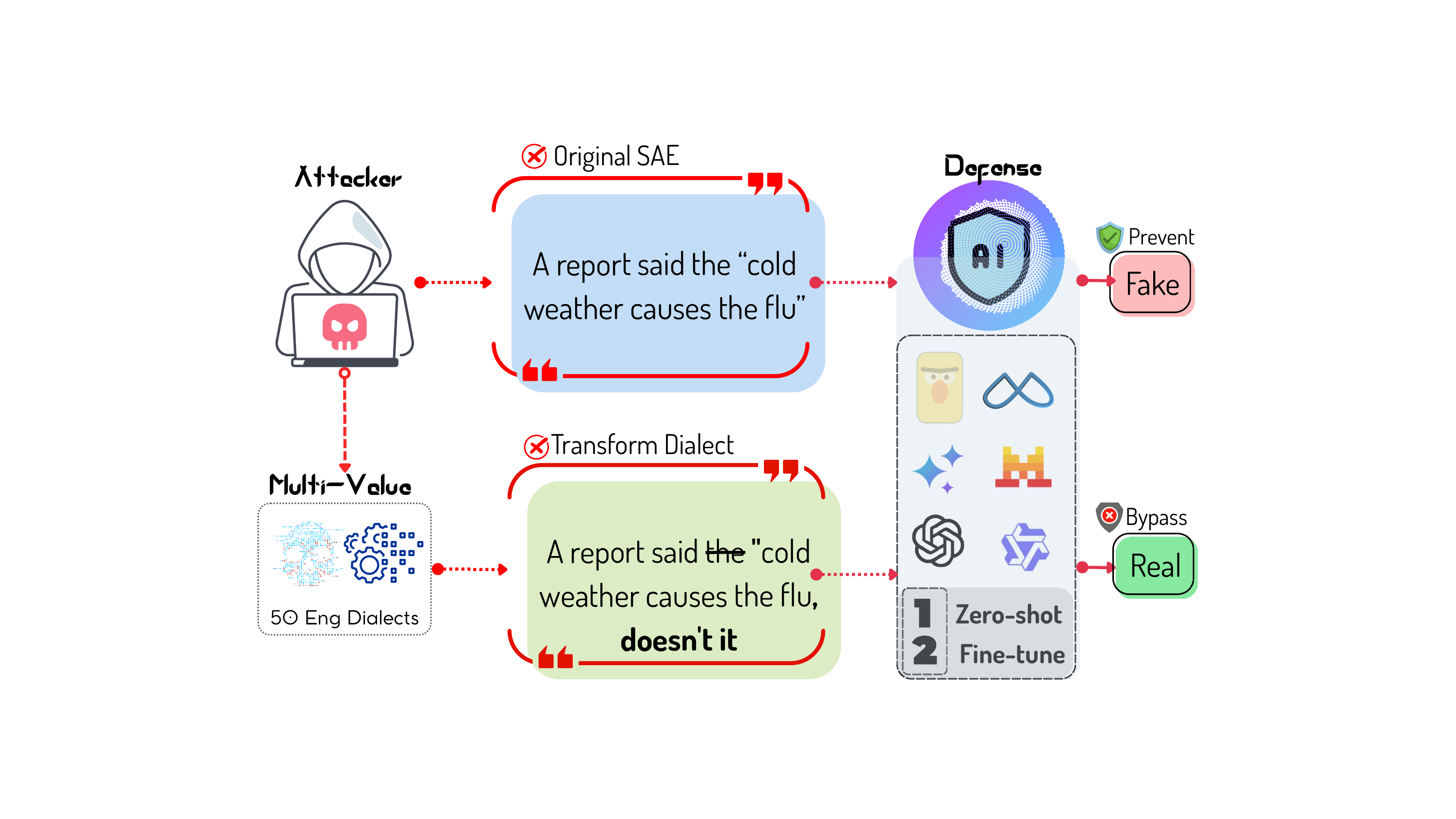}
    \caption{\textbf{Left}: An attacker uses Multi-VALUE to transform \sae{} disinformation 
    (e.g., ``cold weather causes the flu'') into 50 English dialectal variants 
    (e.g., Patois: ``gi yuh flu''). \textbf{Right}: While AI detectors correctly flag \sae{} 
    content as \textit{Fake}, dialectal variants bypass detection---misclassified 
    as \textit{Real}---across zero-shot, few-shot, and fine-tuned LLMs.}
    \label{fig:teaser}
    \vspace{-15pt}
\end{figure}

This \sae-centric approach creates a critical vulnerability. Consider a false health claim in \sae: \textit{``A report said the `cold weather causes the flu'"}---when transformed to Jamaican Creole English: \textit{``Dem ah say `cold weather gi yuh flu'"}---detection accuracy may degrade substantially. The dialect transformation alters morphosyntactic structure while preserving the false semantic content---a natural perturbation that may enable disinformation to bypass detectors not evaluated across the diverse variations of English, the most widely spoken language globally. This example reveals that disinformation detectors may rely heavily on surface-level patterns rather than deeper semantic understanding of veracity. Prior work on disinformation detection has employed diverse approaches spanning neural architectures, transformer-based models, and zero-shot LLM evaluation \cite{Shu2019-vm, Nguyen2020-ju, Devlin2018-mr, lucas-etal-2023-fighting}, achieving strong performance on \sae{} benchmarks. However, this literature has grossly ignored evaluation of dialectal variation---a critical oversight given that millions of users, including potential adversaries, communicate via dialects on social media and other platforms.

Prior work has documented bias and performance disparities across dialects in hate speech detection \cite{sap-etal-2019-risk}, toxicity classification \cite{okpala2022aaebert}, and natural language inference \cite{ziems-etal-2023-multi}, with recent benchmarks systematically evaluating dialect robustness across general NLU tasks \cite{faisal-etal-2024-dialectbench, ziems-etal-2022-value}. However, comprehensive evaluation of disinformation detection systems across a broad range of English dialects remains unexplored---a critical gap, as detection failures may leave dialectal speakers less protected from harmful misinformation while potentially flagging their legitimate speech, exacerbating information inequity rather than reducing it.

We hypothesize that state-of-the-art disinformation detectors will exhibit significant performance degradation on dialectal variation. Building on this, we ask a central research question: \textit{Can state-of-the-art disinformation detectors maintain robust performance across English dialects?} We decompose this into four sub-questions examining different facets of robustness: 
\begin{enumerate}[nosep, leftmargin=*, label={\footnotesize\textbf{(SQ\arabic*)}}]
    \item Do models trained exclusively on \sae{} generalize to dialectal variants they have never encountered?
    \item Does dialect-aware training improve robustness compared to \sae{}-only training?
    \item Can models transfer knowledge learned from one dialect to others?
    \item Which model architectures exhibit the greatest dialectal resilience?
\end{enumerate}


We present \method{}, the first comprehensive benchmark for evaluating disinformation detection robustness across 50 English dialects spanning U.S., British, African, Caribbean, and Asia-Pacific varieties. Using Multi-VALUE's linguistically-grounded rule-based transformations \cite{ziems-etal-2023-multi}, we introduce D-CUBE (Dialectal Disinformation Detection), a corpus of 195K+ samples derived from 9 established disinformation benchmarks (See \autoref{tab:source_datasets}). Our evaluation of 16 detection models---spanning fine-tuned encoders, traditional deep learning architectures, and zero-shot LLMs---reveals systematic vulnerabilities across all detector types.

Our evaluation reveals significant robustness gaps. Human-written dialectal content degrades detection by 1.4--3.6\% F1, while AI-generated content remains surprisingly stable. Fine-tuned transformers substantially outperform zero-shot LLMs: Mistral-7B achieves only 78.3\% dialect F1 compared to 96.6\% for fine-tuned encoders. Dialectal impact varies systematically---feature-rich and geographically isolated varieties cause the largest performance drops (Maltese, Australian Vernacular, Southeast England), while SAE-adjacent dialects and strong transfer sources like Ghanaian and Manx English remain robust. Cross-dialectal transfer analysis reveals that multilingual models (mDeBERTa: 97.2\% average F1) generalize effectively regardless of source-target pairing, while monolingual models like RoBERTa exhibit catastrophic failures on mixed content distributions.

\noindent Our contributions are as follows:
\begin{enumerate}[nosep, leftmargin=*, label={\footnotesize\textbf{(\arabic*)}}]
    \item We introduce \method{}, the first benchmark for evaluating 
    disinformation detection robustness across 50 English dialects, 
    addressing a critical gap in harmful content evaluation 
    (\S\ref{sec:framework}). \method{} comprises three 
    components: the D-CUBE corpus, the D-PURIFY quality validation 
    pipeline, and an extensive evaluation framework spanning four experimental 
    regimes (SQ1--SQ4).
    \item We release D-CUBE, the corpus component of \method{} comprising 195K+ 
    dialectal disinformation samples derived from 9 \sae{} benchmarks using 
    Multi-VALUE transformations, validated via D-PURIFY quality assurance 
    (\S\ref{sec:data_generation}).
    \item We provide a comprehensive evaluation of 16 disinformation detectors across in-distribution, out-of-distribution, and cross-dialectal transfer settings (\S\ref{sec:results}).
    \item We demonstrate that dialectal variation causes 1.4--3.6\% F1 degradation in fine-tuned models and up to 27\% degradation in zero-shot LLMs, with systematic patterns across dialect families and model architectures (\S\ref{sec:results}).
    \item We release the \method{} benchmark, including the D-CUBE corpus, the D-PURIFY quality validation tools, and the evaluation framework and scripts to enable reproducible dialectal robustness 
    testing.
\end{enumerate}


\section{Related Work}
\label{sec:related}

\paragraph{Dialectal NLP}
\label{sec:related_dialect}

Research on dialectal variation in NLP has grown substantially, though with significant gaps in coverage. African American English (AAE) remains the most studied variety \cite{joshi2025natural}, with documented disparities in toxicity detection \cite{okpala2022aaebert}, language identification \cite{blodgett-etal-2016-demographic}, and recent work showing lower LLM accuracy on AAE prompts compared to \sae{} \cite{zhou2025disparities, mire2025rejected}. Disparities extend to other tasks including sentiment analysis \cite{ziems-etal-2022-value}, summarization \cite{keswani2021dialect}, machine translation \cite{kantharuban-etal-2023-quantifying}, and parsing \cite{scannell-2020-universal}.

Several benchmarks have emerged to systematically evaluate dialect robustness. DialectBench \cite{faisal-etal-2024-dialectbench} aggregates existing datasets across 281 dialects spanning 10 NLU/NLG tasks, but covers only 18 English varieties. Multi-VALUE \cite{ziems-etal-2023-multi} provides rule-based transformations across 50 English dialects using 189 linguistic features across 12 grammatical categories derived from the Electronic World Atlas of Varieties of English (eWAVE) \cite{kortmann-etal-2020-ewave}, achieving >95\% native speaker acceptability. However, these benchmarks focus exclusively on general NLU/NLG tasks---they cover single or limited English vernaculars, resulting in limited global coverage and inability to address heterogeneous settings. Critically, they ignore the effects of harmful content such as disinformation and the dialectal morphosyntactic implications for AI defense systems and global speakers in our socio-technical ecosystem. Dialects represent one of the most common modes of phonetic and morphosyntactic communication, yet none of these benchmarks evaluate disinformation detection systems.

\paragraph{Disinformation Detection}
\label{sec:related_disinfo}

Earlier disinformation detectors employ diverse approaches including neural, hierarchical, ensemble-based, and decentralized techniques \cite{Aslam2021-by, Upadhayay2022-xo, Jayakody2022-cg, Ali2022-ob, cui2020deterrent}. Key architectural innovations include dEFEND \cite{Shu2019-vm}, which uses co-attention mechanisms over news content and user comments, and FANG \cite{Nguyen2020-ju}, which leverages graph neural networks for social context modeling. More recently, transformer-based models such as BERT \cite{devlin-etal-2019-bert} and RoBERTa \cite{liu-etal-2019-roberta} have achieved state-of-the-art performance, outperforming traditional deep learning detectors at the cost of extensive training and computation.

Recent work has explored the risks of advanced LLMs in generating disinformation and evaluated zero-shot detection capabilities across in-distribution and out-of-distribution settings \cite{zhou2023synthetic, Liu2023-en, Qin2023-za, lucas-etal-2023-fighting}. Benchmark datasets span multiple domains including political misinformation (FakeNewsNet \cite{shu2020fakenewsnet}, LIAR \cite{wang-2017-liar}), health misinformation (CoAID \cite{cui2020coaid}, MM-COVID \cite{li2020mmcovid}), and multilingual claims (MultiClaim \cite{pikuliak-etal-2023-multilingual}). However, prior work has grossly ignored evaluation of dialectal variation---a critical oversight given that millions of users, including potential adversaries, communicate via dialects on social media and other platforms where disinformation spreads.

\paragraph{Robustness Evaluation}
\label{sec:related_robustness}

Robustness evaluation for NLP systems has traditionally focused on adversarial perturbations---synthetic modifications designed to fool classifiers while preserving semantics \cite{alzantot-etal-2018-generating, jin-etal-2020-bert}. For disinformation detection specifically, robustness studies have examined domain shift \cite{zellers-etal-2019-defending}, temporal drift \cite{horne-etal-2019-robust}, and cross-platform generalization \cite{sheng-etal-2022-characterizing}.

Dialectal variation offers a complementary lens: \textit{natural adversarial perturbation} that systematically alters surface form while preserving semantics \cite{ziems-etal-2023-multi}. Unlike synthetic attacks, dialectal inputs represent authentic language use by real communities. If disinformation detectors fail on linguistic varieties spoken by millions, they exhibit fundamental brittleness beyond adversarial exploitation---and more critically, they fail to equitably protect diverse populations from harmful content. Our work is the first to evaluate disinformation detection robustness against natural dialectal variation across 50 English varieties.

\section{Problem Formulation}
\label{sec:framework}

\begin{figure*}[t]
\centering
\includegraphics[
    width=\textwidth,
    trim={12bp 225bp 12bp 225bp},
    clip
]{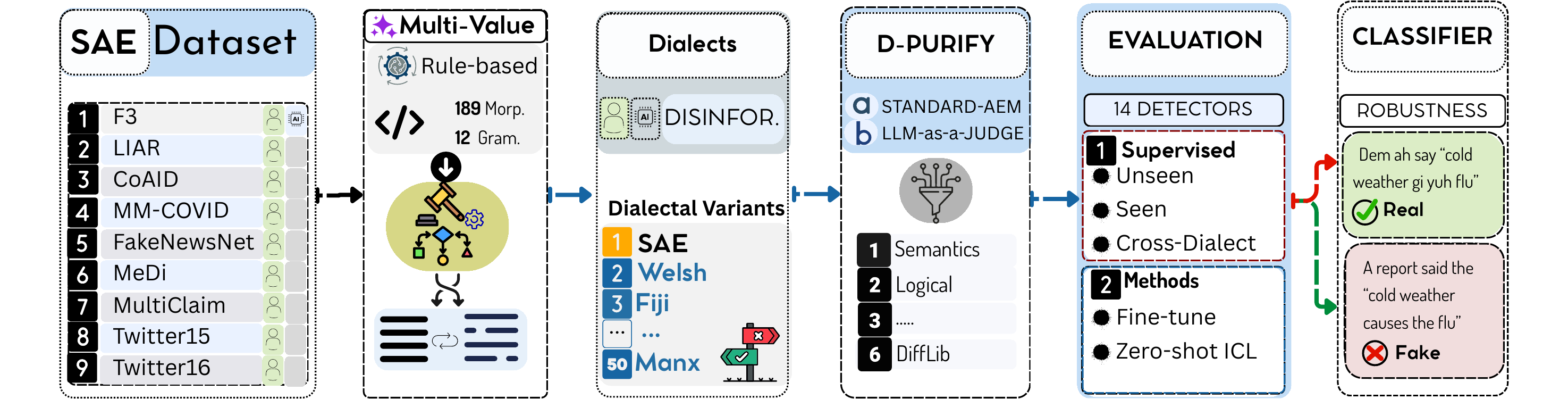}
\caption{The \method{} benchmark pipeline, comprising 
three components: D-CUBE, D-PURIFY, and the evaluation framework 
Starting from 9 \sae{} disinformation benchmarks, we apply Multi-VALUE 
rule-based dialect transformations to generate D-CUBE, 
a corpus of 50 English dialectal variants. D-PURIFY validates 
transformation quality using semantic, logical, and feature accuracy 
metrics. The evaluation framework then evaluates 16 
detectors across multiple experimental settings (SQ1--SQ4), measuring 
classification robustness under unseen, seen, and cross-dialectal 
conditions.}
\label{fig:framework}
\vspace{-10pt}
\end{figure*}

We formalize dialect robustness evaluation for disinformation detection systems. Figure~\ref{fig:framework} illustrates our evaluation pipeline.

\paragraph{Task 1: Dialectal Data Generation.}
Given a text $x$ in Standard American English (\sae) and a target dialect $d \in \mathcal{D}$ where $|\mathcal{D}| = 50$, a dialect transformation function $T_d: \mathcal{X} \rightarrow \mathcal{X}_d$ produces a dialectal variant $x_d = T_d(x)$. Valid transformations must satisfy two constraints: (1)~\textit{semantic preservation}: $\text{sim}(x, x_d) \geq \tau$, where $\text{sim}(\cdot)$ measures semantic similarity; and (2)~\textit{label preservation}: $y_x = y_{x_d}$, ensuring the ground-truth label (real/fake) remains unchanged.

We evaluate transformation quality using standard automatic metrics (BERTScore, BARTScore, AlignScore, METEOR, ROUGE-L, DiffLib) and a novel Feature Accuracy measure combining LLM-as-a-Judge with direct retrieval from the eWAVE static database. Feature Accuracy validates each transformation against 235 eWAVE features and checks dialect appropriateness using eWAVE's A/B/C/D attestation ratings, where A (pervasive), B (common), and C (rare) indicate valid features for a dialect, while D (absent) indicates incorrect application \cite{ewave}. Full metric descriptions and thresholds are provided in Appendix~\ref{app:quality_metrics}.

\paragraph{Task 2: Disinformation Detection Robustness.}
Let $f: \mathcal{X} \rightarrow \{0, 1\}$ be a disinformation detector that classifies content as real (0) or fake (1). Given a test set $X$ and its dialectal variant $X_d = \{T_d(x) : x \in X\}$, we measure the \textit{robustness gap}:

\vspace{-15pt}
\begin{equation}
\Delta_d = \text{Acc}(f, X_{\text{SAE}}) - \text{Acc}(f, X_d)
\label{eq:robustness_gap}
\end{equation}
\vspace{-15pt}

where $\Delta_d > 0$ indicates performance degradation on dialect $d$. A robust detector exhibits $\Delta_d \approx 0$ across all dialects. We evaluate robustness across four experimental settings: \textbf{(SQ1)} models trained on \sae{} tested on unseen dialects; \textbf{(SQ2)} models trained on dialect-mixed data; \textbf{(SQ3)} cross-dialectal transfer across 2,450 dialect pairs; and \textbf{(SQ4)} architectural comparison between fine-tuned encoders and zero-shot LLMs.

\section{D-CUBE: Dialectal Disinformation Detection Corpus}
\label{sec:data_generation}

D-CUBE is the corpus component of the \method{} 
benchmark. This section describes the construction of 
D-CUBE, a corpus of 195K+ samples spanning 50 English dialects derived from 9 established disinformation benchmarks (detailed in 
\autoref{tab:source_datasets}).

\paragraph{Dialect Coverage}
\label{sec:dialect_coverage}

We target 50 English dialects derived from the Electronic World Atlas of Varieties of English (eWAVE) \cite{kortmann-etal-2020-ewave}, spanning five geographic regions: U.S. varieties (including African American Vernacular English variants, Appalachian, Chicano, and Ozark English), British Isles varieties (Scottish, Irish, Welsh, and regional English dialects), African varieties (Nigerian, Ghanaian, Kenyan, South African variants), Asia-Pacific varieties (Indian, Singaporean, Philippine, Australian, and Fiji English), and Caribbean/Atlantic varieties (Jamaican, Bahamian, and island Englishes). The complete dialect inventory is provided in Appendix~\ref{app:dialects}.

\paragraph{Source Datasets}
\label{sec:source_data}

We compile 9 benchmark datasets spanning diverse disinformation domains (Table~\ref{tab:source_datasets}): political misinformation (FakeNewsNet \cite{shu2020fakenewsnet}, LIAR \cite{wang-2017-liar}), health misinformation (CoAID \cite{cui2020coaid}, MM-COVID \cite{li2020mmcovid}), social media rumors (Twitter15, Twitter16 \cite{ma-etal-2017-detect}, MiDe \cite{toraman-etal-2024-mide22}), multilingual claims (MultiClaim \cite{pikuliak-etal-2023-multilingual}), and LLM-generated disinformation (F$^3$ \cite{lucas-etal-2023-fighting}).

\begin{table}[t]
\centering
\scriptsize
\begin{tabular}{llccccc}
\toprule
\textbf{Dataset} & \textbf{Domain} & \textbf{Size} & \textbf{Dial.} & \textbf{H} & \textbf{AI} \\
\midrule
FakeNewsNet & Politics & 10.9K & 1 & \cmark & \xmark \\
LIAR & Politics & 5.7K & 1 & \cmark & \xmark \\
CoAID & Health & 19.6K & 1 & \cmark & \xmark \\
MM-COVID & Health & 4.0K & 1 & \cmark & \xmark \\
MultiClaim & Multi & 24.3K & 1 & \cmark & \xmark \\
F$^3$ & Multi & 3.2K & 1 & \cmark & \cmark \\
MiDe & Social & 1.7K & 1 & \cmark & \xmark \\
Twitter15/16 & Social & 0.6K & 1 & \cmark & \xmark \\
\midrule
\rowcolor{blue!20}
\textbf{D-CUBE (Ours)} & \textbf{All} & \textbf{195K} & \textbf{50} & \cmark & \cmark \\
\bottomrule
\end{tabular}
\caption{Disinformation dataset comparison. D-CUBE 
uniquely provides comprehensive dialect coverage (50 varieties), both 
human and AI-generated content, and spans all major disinformation 
domains. Dial.=Dialects, H=Human-written, AI=AI-generated.}
\label{tab:source_datasets}
\end{table}

\subsection{Multi-VALUE Transformation}
\label{sec:multivalue}

We employ Multi-VALUE \cite{ziems-etal-2023-multi} for linguistically-grounded dialect transformation. Multi-VALUE implements 189 morphosyntactic transformation rules organized across 12 grammatical categories derived from eWAVE: pronouns, noun phrases, tense and aspect, modal verbs, verb morphology, negation, agreement, relativization, complementation, adverbials, word order, and discourse. Features are applied probabilistically based on attestation strength (A: pervasive, B: common, C: rare) for each dialect in eWAVE.

Table~\ref{tab:dialect_examples} illustrates representative transformations that preserve semantic content while introducing authentic dialectal morphosyntax. These examples demonstrate features such as the AAVE future marker (\textit{finna}), progressive aspect with stative verbs (\textit{are needing}), and double superlatives (\textit{most worst})---all documented in eWAVE as legitimate dialectal features that SAE-trained evaluation metrics may incorrectly penalize.

\begin{table}[t]
\centering
\scriptsize
\begin{tabular}{p{0.62\columnwidth}p{0.22\columnwidth}}
\toprule
\textbf{Original $\rightarrow$ Transformed} & \textbf{Feature} \\
\midrule
``will join forces'' $\rightarrow$ ``finna join forces'' & Future \\
``We need to do this'' $\rightarrow$ ``We are needing to do this'' & Prog. stative \\
``the worst place'' $\rightarrow$ ``that most worst place'' & Dbl. superl. \\
``Our democracy is'' $\rightarrow$ ``Our democracys are'' & Agreement \\
\bottomrule
\end{tabular}
\caption{Example dialect transformations preserving semantic content while introducing authentic morphosyntactic variation.}
\label{tab:dialect_examples}
\vspace{-15pt}
\end{table}

\subsection{Quality Validation (D-PURIFY)}
\label{sec:dpurify}

We validate transformation quality using D-PURIFY, a multi-dimensional filtering pipeline combining standard automatic evaluation metrics with dialect-specific validation.

\paragraph{Automatic Metrics.} We evaluate transformations using BERTScore \cite{zhang-etal-2020-bertscore} and BARTScore \cite{yuan-etal-2021-bartscore} for semantic similarity, AlignScore \cite{zha-etal-2023-alignscore} for logical consistency, METEOR \cite{banerjee-lavie-2005-meteor} and ROUGE-L \cite{lin-2004-rouge} for translation quality, and Python's DiffLib\footnote{\url{https://docs.python.org/3/library/difflib.html}} for surface-level divergence.

\paragraph{Feature Accuracy.} We introduce a Feature Accuracy metric combining LLM-as-a-Judge with direct retrieval from the eWAVE static database. The pipeline extracts dialectal features from transformed text, validates each feature against 235 eWAVE specifications, and checks dialect appropriateness using eWAVE's attestation ratings (A/B/C = valid; D = absent). Full metric descriptions are provided in Appendix~\ref{app:quality_metrics}.

\paragraph{Filtering Thresholds.} We adopt intentionally lenient thresholds (Table~\ref{tab:thresholds}) to preserve dialectal diversity. Standard evaluation metrics are trained predominantly on SAE corpora and may systematically undervalue authentic dialectal variation. Interestingly, dialects with lower pass rates (e.g., Chicano English: 71.4\%, Southeast England English: 72.7\%) tend to be SAE-adjacent varieties with subtle morphosyntactic differences, while more linguistically distant varieties (e.g., Hong Kong English: 99.5\%, Indian English: 99.3\%) pass at higher rates. This suggests SAE-trained metrics may penalize subtle deviations as errors while accepting clearly distinct patterns. Stricter thresholds would disproportionately exclude SAE-adjacent dialects, undermining benchmark diversity. We report filtering pass rates by dialect in Table~\ref{tab:dialect_pass} and provide a detailed analysis in Appendix~\ref{app:sae_adjacent}.

\begin{table}[t]
\centering
\small
\begin{tabular}{llcc}
\toprule
\textbf{Metric} & \textbf{Dimension} & \textbf{Threshold} & \textbf{Pass\%} \\
\midrule
BERTScore & Semantic & $> 0.50$ & 99.3 \\
BARTScore & Generation & $\geq -7.5$ & 99.7 \\
METEOR & Translation & $\geq 0.40$ & 100.0 \\
ROUGE-L & Overlap & $(0.01, 1.0)$ & 97.6 \\
DiffLib & Surface & $[0.01, 0.99]$ & 95.8 \\
BLEU & N-gram & $> 0.01$ & 99.3 \\
\bottomrule
\end{tabular}
\caption{D-PURIFY filtering thresholds and pass rates. Lenient thresholds preserve dialectal diversity; stricter filtering would disproportionately exclude SAE-adjacent varieties.}
\label{tab:thresholds}
\vspace{-15pt}
\end{table}

\begin{table}[t]
\centering
\small
\begin{tabular}{lcc}
\toprule
\textbf{Dialect} & \textbf{Passed} & \textbf{Pass\%} \\
\midrule
\multicolumn{3}{l}{\textit{Highest Pass Rates}} \\
Hong Kong English & 4,225 & 99.5 \\
Indian English & 4,217 & 99.3 \\
SE American Enclave & 4,214 & 99.2 \\
Malaysian English & 4,198 & 98.9 \\
Indian South African Eng. & 4,193 & 98.7 \\
\midrule
\multicolumn{3}{l}{\textit{Lowest Pass Rates}} \\
Acrolectal Fiji English & 3,568 & 84.0 \\
Falkland Islands English & 3,374 & 79.4 \\
Philippine English & 3,174 & 74.7 \\
SE England English & 3,087 & 72.7 \\
Chicano English & 3,034 & 71.4 \\
\bottomrule
\end{tabular}
\caption{D-PURIFY pass rates by dialect (top/bottom 5). SAE-adjacent dialects (Chicano, SE England) show lower pass rates, suggesting metric bias against subtle deviations. Full results in Appendix~\ref{app:dialect_quality}.}
\label{tab:dialect_pass}
\end{table}

\paragraph{Quality Results}
\label{sec:quality_results}

Table~\ref{tab:quality_metrics} presents D-PURIFY quality metrics across human-written and AI-generated content. Transformations achieve strong semantic preservation (BERTScore: 0.83 for both human and AI), indicating dialectal variants maintain original meaning. AI-generated content shows higher logical consistency (AlignScore: 0.92) but lower surface overlap (METEOR: 0.47, ROUGE-L: 0.47), suggesting more substantial morphosyntactic transformation. Human-written content achieves higher Feature Accuracy (0.20 vs. 0.10), reflecting greater alignment with eWAVE specifications. The final D-CUBE corpus contains 194,960 samples across 50 dialects (93.7\% retention rate). Per-dialect quality breakdowns are provided in Appendix~\ref{app:sq1_full}.

\begin{table}[t]
\centering
\small
\begin{tabular}{lccc}
\toprule
\textbf{Metric} & \textbf{Human} & \textbf{AI} & \textbf{$\Delta$} \\
\midrule
BERTScore & 0.828 & 0.846 & +0.018 \\
BARTScore & $-$4.09 & $-$4.28 & $-$0.183 \\
AlignScore & 0.898 & 0.924 & +0.026 \\
METEOR & 0.876 & 0.465 & $-$0.411 \\
ROUGE-L & 0.877 & 0.466 & $-$0.411 \\
DiffLib & 0.787 & 0.846 & +0.059 \\
Feature Acc. & 0.196 & 0.101 & $-$0.095 \\
BLEU & 0.573 & 0.320 & $-$0.253 \\
\bottomrule
\end{tabular}
\caption{D-CUBE quality metrics (mean values). Human 
and AI content achieve comparable semantic preservation; AI shows 
greater surface divergence with lower Feature Accuracy.}
\label{tab:quality_metrics}
\end{table}

\section{Experimental Setup}
\label{sec:experiments}
We study how disinformation detection models generalize from \sae{} to dialectal disinformation. Specifically, we evaluate the hypothesis that training on \sae{} leads to degraded performance on dialectal inputs across four sub-questions targeting different dimensions of robustness.

\paragraph{Detection Models}
\label{sec:models}

We evaluate 16 models spanning three architectural paradigms (\autoref{tab:models}): (a) \textit{Traditional Deep Learning} (3 models)---baselines representing pre-transformer approaches still deployed in resource-constrained settings; (b) \textit{Transformer Encoders} (7 models)---both monolingual (e.g., BERT-Large \cite{devlin-etal-2019-bert}) and multilingual models (e.g., XLM-RoBERTa-Large \cite{conneau-etal-2020-unsupervised}), enabling assessment of cross-lingual transfer effects on dialectal robustness; and (c) \textit{Zero-shot Decoders} (6 models)---instruction-tuned LLMs evaluated without task-specific fine-tuning to assess out-of-the-box dialectal resilience.

\begin{table}[t]
\centering
\scriptsize
\begin{tabular}{llc}
\toprule
\textbf{Model} & \textbf{Type} & \textbf{Size} \\
\midrule
\multicolumn{3}{l}{\textit{Traditional Deep Learning (3)}} \\
dEFEND & DNN & -- \\
TextCNN & CNN & -- \\
BiGRU & RNN & -- \\
\midrule
\multicolumn{3}{l}{\textit{Transformer Encoders (7)}} \\
BERT-Large & Encoder & 340M \\
RoBERTa-Large & Encoder & 355M \\
DeBERTa-Large & Encoder & 435M \\
XLM-RoBERTa$^\dagger$ & Encoder & 550M \\
CT-BERT & Encoder & 340M \\
mBERT$^\dagger$ & Encoder & 180M \\
mDeBERTa$^\dagger$ & Encoder & 86M \\
\midrule
\multicolumn{3}{l}{\textit{Zero-shot Decoders (6)}} \\
Mistral-7B & Decoder & 7B \\
Llama-3.1-8B & Decoder & 8B \\
Llama-3.2-1B & Decoder & 1B \\
Gemma-3-1B & Decoder & 1B \\
Qwen3-8B & Decoder & 8B \\
Qwen3-4B-SafeRL & Safety & 4B \\
\bottomrule
\end{tabular}
\caption{Model inventory. $^\dagger$Multilingual pre-training.}
\label{tab:models}
\vspace{-10pt}
\end{table}

\paragraph{Evaluation Regimes}
\label{sec:regimes}

We design four evaluation regimes corresponding to our sub-questions: \textbf{(SQ1) Unseen}---models trained on \sae{} are tested on dialectal variants never encountered during training, measuring zero-shot generalization across three content scenarios (human-only, AI-only, mixed); \textbf{(SQ2) Seen}---models trained on dialect-mixed data are evaluated across all 50 dialects, comparing \textit{dialect-only} versus \textit{\sae{}-anchored} training strategies; \textbf{(SQ3) Cross-dialectal}---models trained on a single dialect are tested on all 49 others (2,450 train-test pairs), identifying effective source dialects for transfer; \textbf{(SQ4) Architecture}---fine-tuned models are compared against zero-shot LLMs to identify which paradigm exhibits greater dialectal resilience.

\paragraph{Evaluation Metrics}
\label{sec:metrics}

We report macro-averaged F1-score and compute the generalization gap $\Delta = \text{F1}_{\text{dialect}} - \text{F1}_{\text{SAE}}$, where negative values indicate degradation on dialectal content. Training configuration and computational details are provided in Appendix~\ref{app:implementation}.

\section{Results}
\label{sec:results}

We present findings across four evaluation regimes, with full per-dialect results in Appendix~\ref{app:sq1_full}-\ref{app:sq4_full} .

\subsection{SQ1: Generalization to Unseen Dialects}
\label{sec:sq1}

\autoref{fig:sq1_gap} presents the generalization gap ($\Delta$) for models trained exclusively on \sae{} and evaluated on 50 dialectal variants across three content scenarios.

\begin{figure}[t]
\centering
\begin{tikzpicture}
\begin{axis}[
    width=\columnwidth,
    height=6cm,
    ylabel={Generalization Gap ($\Delta$ F1\%)},
    ymin=-6,
    ymax=14,
    xmin=0.5,
    xmax=10.5,
    ytick={-5,0,5,10},
    xtick={1,2,3,4,5,6,7,8,9,10},
    xticklabels={BERT,RoBERTa,DeBERTa,XLM-R,mDeBERTa,mBERT,BiGRU,TextCNN,dEFEND,CT-BERT},
    x tick label style={rotate=45, anchor=east, font=\scriptsize},
    y tick label style={font=\footnotesize},
    ylabel style={font=\small},
    grid=major,
    grid style={dashed, gray!30},
    clip=false,
]

\draw[black, dashed, line width=0.5pt] (axis cs:0.5,0) -- (axis cs:10.5,0);

\fill[blue!50, opacity=0.7] (axis cs:0.75,-1.5) rectangle (axis cs:0.95,0);
\draw[blue!90, line width=0.6pt] (axis cs:0.75,-1.5) rectangle (axis cs:0.95,0);
\fill[pattern=north east lines, pattern color=green!70] (axis cs:0.97,0) rectangle (axis cs:1.17,0.5);
\draw[green!80, line width=0.6pt] (axis cs:0.97,0) rectangle (axis cs:1.17,0.5);
\fill[pattern=dots, pattern color=orange!90] (axis cs:1.19,0) rectangle (axis cs:1.39,1.0);
\draw[orange!90, line width=0.6pt] (axis cs:1.19,0) rectangle (axis cs:1.39,1.0);

\fill[blue!50, opacity=0.7] (axis cs:1.75,-1.5) rectangle (axis cs:1.95,0);
\draw[blue!90, line width=0.6pt] (axis cs:1.75,-1.5) rectangle (axis cs:1.95,0);
\fill[pattern=north east lines, pattern color=green!70] (axis cs:1.97,0) rectangle (axis cs:2.17,0.4);
\draw[green!80, line width=0.6pt] (axis cs:1.97,0) rectangle (axis cs:2.17,0.4);
\fill[pattern=dots, pattern color=orange!90] (axis cs:2.19,0) rectangle (axis cs:2.39,0.0);
\draw[orange!90, line width=0.6pt] (axis cs:2.19,0) rectangle (axis cs:2.39,0.0);

\fill[blue!50, opacity=0.7] (axis cs:2.75,-1.5) rectangle (axis cs:2.95,0);
\draw[blue!90, line width=0.6pt] (axis cs:2.75,-1.5) rectangle (axis cs:2.95,0);
\fill[pattern=north east lines, pattern color=green!70] (axis cs:2.97,0) rectangle (axis cs:3.17,0.9);
\draw[green!80, line width=0.6pt] (axis cs:2.97,0) rectangle (axis cs:3.17,0.9);
\fill[pattern=dots, pattern color=orange!90] (axis cs:3.19,-0.4) rectangle (axis cs:3.39,0);
\draw[orange!90, line width=0.6pt] (axis cs:3.19,-0.4) rectangle (axis cs:3.39,0);

\fill[blue!50, opacity=0.7] (axis cs:3.75,-1.5) rectangle (axis cs:3.95,0);
\draw[blue!90, line width=0.6pt] (axis cs:3.75,-1.5) rectangle (axis cs:3.95,0);
\fill[pattern=north east lines, pattern color=green!70] (axis cs:3.97,0) rectangle (axis cs:4.17,0.4);
\draw[green!80, line width=0.6pt] (axis cs:3.97,0) rectangle (axis cs:4.17,0.4);
\fill[pattern=dots, pattern color=orange!90] (axis cs:4.19,0) rectangle (axis cs:4.39,0.0);
\draw[orange!90, line width=0.6pt] (axis cs:4.19,0) rectangle (axis cs:4.39,0.0);

\fill[blue!50, opacity=0.7] (axis cs:4.75,-1.5) rectangle (axis cs:4.95,0);
\draw[blue!90, line width=0.6pt] (axis cs:4.75,-1.5) rectangle (axis cs:4.95,0);
\fill[pattern=north east lines, pattern color=green!70] (axis cs:4.97,0) rectangle (axis cs:5.17,1.0);
\draw[green!80, line width=0.6pt] (axis cs:4.97,0) rectangle (axis cs:5.17,1.0);
\fill[pattern=dots, pattern color=orange!90] (axis cs:5.19,-0.5) rectangle (axis cs:5.39,0);
\draw[orange!90, line width=0.6pt] (axis cs:5.19,-0.5) rectangle (axis cs:5.39,0);

\fill[blue!50, opacity=0.7] (axis cs:5.75,-1.4) rectangle (axis cs:5.95,0);
\draw[blue!90, line width=0.6pt] (axis cs:5.75,-1.4) rectangle (axis cs:5.95,0);
\fill[pattern=north east lines, pattern color=green!70] (axis cs:5.97,0) rectangle (axis cs:6.17,1.2);
\draw[green!80, line width=0.6pt] (axis cs:5.97,0) rectangle (axis cs:6.17,1.2);
\fill[pattern=dots, pattern color=orange!90] (axis cs:6.19,-0.9) rectangle (axis cs:6.39,0);
\draw[orange!90, line width=0.6pt] (axis cs:6.19,-0.9) rectangle (axis cs:6.39,0);

\fill[blue!50, opacity=0.7] (axis cs:6.75,-1.9) rectangle (axis cs:6.95,0);
\draw[blue!90, line width=0.6pt] (axis cs:6.75,-1.9) rectangle (axis cs:6.95,0);
\fill[pattern=north east lines, pattern color=green!70] (axis cs:6.97,0) rectangle (axis cs:7.17,1.3);
\draw[green!80, line width=0.6pt] (axis cs:6.97,0) rectangle (axis cs:7.17,1.3);
\fill[pattern=dots, pattern color=orange!90] (axis cs:7.19,0) rectangle (axis cs:7.39,0.1);
\draw[orange!90, line width=0.6pt] (axis cs:7.19,0) rectangle (axis cs:7.39,0.1);

\fill[blue!50, opacity=0.7] (axis cs:7.75,-1.4) rectangle (axis cs:7.95,0);
\draw[blue!90, line width=0.6pt] (axis cs:7.75,-1.4) rectangle (axis cs:7.95,0);
\fill[pattern=north east lines, pattern color=green!70] (axis cs:7.97,0) rectangle (axis cs:8.17,1.6);
\draw[green!80, line width=0.6pt] (axis cs:7.97,0) rectangle (axis cs:8.17,1.6);
\fill[pattern=dots, pattern color=orange!90] (axis cs:8.19,0) rectangle (axis cs:8.39,0.6);
\draw[orange!90, line width=0.6pt] (axis cs:8.19,0) rectangle (axis cs:8.39,0.6);

\fill[blue!50, opacity=0.7] (axis cs:8.75,-3.6) rectangle (axis cs:8.95,0);
\draw[blue!90, line width=0.6pt] (axis cs:8.75,-3.6) rectangle (axis cs:8.95,0);
\fill[pattern=north east lines, pattern color=green!70] (axis cs:8.97,0) rectangle (axis cs:9.17,1.9);
\draw[green!80, line width=0.6pt] (axis cs:8.97,0) rectangle (axis cs:9.17,1.9);
\fill[pattern=dots, pattern color=orange!90] (axis cs:9.19,0) rectangle (axis cs:9.39,0.2);
\draw[orange!90, line width=0.6pt] (axis cs:9.19,0) rectangle (axis cs:9.39,0.2);

\fill[blue!50, opacity=0.7] (axis cs:9.75,0) rectangle (axis cs:9.95,11.4);
\draw[blue!90, line width=0.6pt] (axis cs:9.75,0) rectangle (axis cs:9.95,11.4);
\fill[pattern=north east lines, pattern color=green!70] (axis cs:9.97,0) rectangle (axis cs:10.17,1.1);
\draw[green!80, line width=0.6pt] (axis cs:9.97,0) rectangle (axis cs:10.17,1.1);
\fill[pattern=dots, pattern color=orange!90] (axis cs:10.19,0) rectangle (axis cs:10.39,2.9);
\draw[orange!90, line width=0.6pt] (axis cs:10.19,0) rectangle (axis cs:10.39,2.9);

\fill[blue!50, opacity=0.7] (axis cs:0.8,12.5) rectangle (axis cs:1.2,13.5);
\draw[blue!90, line width=0.6pt] (axis cs:0.8,12.5) rectangle (axis cs:1.2,13.5);
\node[font=\scriptsize, anchor=west] at (axis cs:1.3,13) {Human};

\fill[pattern=north east lines, pattern color=green!70] (axis cs:3.0,12.5) rectangle (axis cs:3.4,13.5);
\draw[green!80, line width=0.6pt] (axis cs:3.0,12.5) rectangle (axis cs:3.4,13.5);
\node[font=\scriptsize, anchor=west] at (axis cs:3.5,13) {AI};

\fill[pattern=dots, pattern color=orange!90] (axis cs:5.0,12.5) rectangle (axis cs:5.4,13.5);
\draw[orange!90, line width=0.6pt] (axis cs:5.0,12.5) rectangle (axis cs:5.4,13.5);
\node[font=\scriptsize, anchor=west] at (axis cs:5.5,13) {Both};

\end{axis}
\end{tikzpicture}
\caption{SQ1: Generalization gap ($\Delta$ F1) from \sae{} to dialectal variants by content type. Solid blue = human content; hatched green = AI content; dotted orange = mixed content. Negative values indicate degradation on dialects.}
\label{fig:sq1_gap}
\vspace{-15pt}
\end{figure}

\textbf{Human content} shows consistent degradation across all models ($\Delta = -1.4\%$ to $-3.6\%$), with dEFEND exhibiting the largest gap. \textbf{AI-generated content} demonstrates stable or improved detection ($\Delta = +0.4\%$ to $+1.9\%$), suggesting dialectal transformation preserves detectable artifacts. CT-BERT shows anomalous improvement (+11.4\% on human content), likely due to domain mismatch between COVID-Twitter pretraining and news articles.

\begin{tcolorbox}[colback=gray!10, colframe=gray!50, boxrule=0.5pt, left=2pt, right=2pt, top=2pt, bottom=2pt]
\textbf{Finding 1:} Human-written dialectal content degrades detection by 1.4--3.6\%, while AI-generated content remains stable (+0.4\% to +1.9\%).
\end{tcolorbox}

\subsection{SQ2: Impact of Dialect Exposure}
\label{sec:sq2}

\autoref{tab:sq2_training} compares three training regimes: \sae{}-only (SQ1 baseline), dialect-only, and \sae{}-anchored.

\begin{table}[t]
\centering
\scriptsize
\begin{tabular}{lcccc}
\toprule
& \multicolumn{3}{c}{\textbf{Training Regime (F1 \%)}} & \\
\cmidrule(lr){2-4}
\textbf{Model} & \textbf{Unseen} & \textbf{Dia-Only} & \textbf{SAE+Dia} & \textbf{$\Delta_{\text{best}}$} \\
\midrule
\multicolumn{5}{l}{\textit{Transformer Encoders}} \\
BERT-Large & 97.2 & 96.9 & 95.8 & $-$1.4 \\
RoBERTa-Large & 87.3$^\S$ & 97.1$^\blacktriangle$ & 96.9 & +9.8 \\
DeBERTa-Large & 95.1 & 96.2 & 95.8 & +1.1 \\
XLM-R$^\dagger$ & 83.3$^\S$ & 85.4$^\S$ & 79.7 & +2.1 \\
mDeBERTa$^\dagger$ & 97.0 & 97.0 & 96.8 & 0.0 \\
mBERT$^\dagger$ & 96.1 & 95.9 & 96.1 & $-$0.2 \\
CT-BERT & 97.2 & 97.1$^\blacktriangle$ & 97.0 & $-$0.1 \\
\midrule
\multicolumn{5}{l}{\textit{Traditional DL}} \\
BiGRU & 96.1 & 91.2 & 92.0 & $-$4.1 \\
TextCNN & 95.1 & 92.1 & 92.5 & $-$2.6 \\
dEFEND & 94.8 & 88.6$^\blacktriangledown$ & 87.5$^\blacktriangledown$ & $-$6.2 \\
\bottomrule
\end{tabular}
\caption{SQ2: Training regime comparison. $\Delta_{\text{best}}$ = best seen $-$ unseen. $^\dagger$Multilingual. $^\blacktriangle$Best. $^\blacktriangledown$Worst. $^\S$Failure.}
\label{tab:sq2_training}
\vspace{-2pt}
\end{table}

Dialectal exposure recovers catastrophic failures: RoBERTa improves from 87.3\% to 97.1\%. For transformers, dialect-only training matches or exceeds \sae{}-anchored performance, suggesting \sae{} inclusion provides no benefit. Traditional models show the opposite: performance degrades without \sae{} anchoring (dEFEND drops to 88.6\%).

\begin{tcolorbox}[colback=gray!10, colframe=gray!50, boxrule=0.5pt, left=2pt, right=2pt, top=2pt, bottom=2pt]
\textbf{Finding 2:} Dialect-only training suffices for transformers and recovers failures. Traditional models require \sae{}-anchoring, revealing architecture-dependent training strategies.
\end{tcolorbox}

\subsection{SQ3: Cross-Dialectal Transfer}
\label{sec:sq3}

We evaluate 2,450 unique train-test dialect pairs. \autoref{tab:sq3_models} summarizes model transfer capacity; \autoref{tab:sq3_dialects} identifies best and worst source/target dialects. The full 50$\times$50 matrix appears in \autoref{tab:transfer_full} in Appendix~\ref{app:sq3_full}.

\begin{table}[t]
\centering
\scriptsize
\begin{tabular}{lcccc}
\toprule
\textbf{Model} & \textbf{Avg F1} & \textbf{Min} & \textbf{Max} & \textbf{Range} \\
\midrule
mDeBERTa$^\dagger$ & 97.2$^\blacktriangle$ & 95.9 & 97.9 & 2.0 \\
TextCNN & 92.8 & 86.0 & 95.1 & 9.1 \\
BiGRU & 91.4 & 82.1 & 95.1 & 13.0 \\
dEFEND & 88.1 & 65.0 & 94.2 & 29.2 \\
XLM-R$^\dagger$ & 39.8$^\blacktriangledown$ & 30.0 & 80.3 & 80.3 \\
\bottomrule
\end{tabular}
\caption{SQ3: Model transfer capacity across 2,450 pairs. $^\dagger$Multilingual. $^\blacktriangle$Best. $^\blacktriangledown$Worst.}
\label{tab:sq3_models}
\end{table}

\begin{table}[t]
\centering
\scriptsize
\begin{tabular}{llc}
\toprule
\textbf{Category} & \textbf{Dialect} & \textbf{Avg F1} \\
\midrule
\multicolumn{3}{l}{\textit{Best Source Dialects (Train $\rightarrow$ All)}} \\
& Ghanaian English & 92.5 \\
& Manx English & 91.6 \\
& Tristan da Cunha English & 91.0 \\
\midrule
\multicolumn{3}{l}{\textit{Worst Source Dialects}} \\
& Scottish English & 74.2 \\
& Welsh English & 75.0 \\
& East Anglian English & 87.5 \\
\midrule
\multicolumn{3}{l}{\textit{Easiest Target Dialects (All $\rightarrow$ Test)}} \\
& Scottish English & 85.3 \\
& Welsh English & 84.7 \\
& Chicano English & 83.3 \\
\midrule
\multicolumn{3}{l}{\textit{Hardest Target Dialects}} \\
& Maltese English & 80.0 \\
& Australian Vernacular & 80.1 \\
& SE England English & 80.1 \\
\bottomrule
\end{tabular}
\caption{SQ3: Best/worst dialects for transfer. Full matrix in Appendix~\ref{app:sq3_full}.}
\label{tab:sq3_dialects}
\vspace{-5pt}
\end{table}

mDeBERTa achieves near-perfect transfer (97.2\% avg, range 2.0\%), while XLM-R fails catastrophically (39.8\% avg, range 80.3\%). Transfer is asymmetric: Scottish and Welsh English are poor training sources but easy targets, suggesting they contain features that generalize poorly but are easily recognized.

\begin{tcolorbox}[colback=gray!10, colframe=gray!50, boxrule=0.5pt, left=2pt, right=2pt, top=2pt, bottom=2pt]
\textbf{Finding 3:} Cross-dialectal transfer is asymmetric. mDeBERTa achieves robust transfer (97.2\%); XLM-R fails (39.8\%). Some dialects transfer out poorly but are easy targets.
\end{tcolorbox}

\subsection{SQ4: Fine-Tuned vs. Zero-Shot}
\label{sec:sq4}

\autoref{tab:sq4_arch} compares fine-tuned models against zero-shot LLMs on identical dialectal test sets.

\begin{table}[t]
\centering
\scriptsize
\begin{tabular}{lccc}
\toprule
& \multicolumn{2}{c}{\textbf{Dialect F1 (\%)}} & \\
\cmidrule(lr){2-3}
\textbf{Model} & \textbf{Human} & \textbf{AI} & \textbf{$\Delta_{\text{SAE}}$} \\
\midrule
\multicolumn{4}{l}{\textit{Fine-tuned Transformer Encoders}} \\
BERT-Large & 96.6 & 99.2 & $-$1.5 \\
RoBERTa-Large & 96.3 & 99.5$^\blacktriangle$ & $-$1.5 \\
DeBERTa-Large & 94.7$^\blacktriangledown$ & 97.4$^\blacktriangledown$ & $-$1.5 \\
XLM-R$^\dagger$ & 94.9 & 98.0 & $-$1.5 \\
mDeBERTa$^\dagger$ & 96.7$^\blacktriangle$ & 99.4 & $-$1.5 \\
mBERT$^\dagger$ & 96.6 & 99.4 & $-$1.4 \\
CT-BERT & 97.2$^\blacktriangle$ & 99.4 & +11.4 \\
\midrule
\multicolumn{4}{l}{\textit{Fine-tuned Traditional DL}} \\
BiGRU & 96.1$^\blacktriangle$ & 98.7$^\blacktriangle$ & $-$1.9 \\
TextCNN & 95.5 & 97.8 & $-$1.4 \\
dEFEND & 93.9$^\blacktriangledown$ & 98.3$^\blacktriangledown$ & $-$3.6 \\
\midrule
\multicolumn{4}{l}{\textit{Zero-shot ICL Decoders}} \\
Mistral-7B & 78.3$^\blacktriangle$ & 78.3$^\blacktriangle$ & $-$11.0 \\
Llama-3.1-8B & 67.2 & 67.2 & $-$20.1 \\
Gemma-3-1B & 48.4 & 48.4 & $-$27.4 \\
Qwen3-8B & 26.6 & 26.6 & $-$11.4 \\
Qwen3-4B-SafeRL & 22.1 & 22.1 & $-$11.6 \\
Llama-3.2-1B & 0.2$^\blacktriangledown$ & 0.2$^\blacktriangledown$ & $-$1.9 \\
\bottomrule
\end{tabular}
\caption{SQ4: Architecture comparison on dialectal content. $\Delta_{\text{SAE}}$ = dialect $-$ \sae{}. $^\dagger$Multilingual. $^\blacktriangle$Best in category. $^\blacktriangledown$Worst.}
\label{tab:sq4_arch}
\vspace{-15pt}
\end{table}

Fine-tuned transformers achieve robust dialectal performance (96.6\% human, 99.4\% AI). Zero-shot LLMs underperform substantially: even Mistral-7B (78.3\%) trails by $\sim$18 points. Smaller models fail catastrophically---Llama-3.2-1B achieves 0.2\% F1. The uniform Human/AI scores for zero-shot models suggest they fail to distinguish content types entirely.

\begin{tcolorbox}[colback=gray!10, colframe=gray!50, boxrule=0.5pt, left=2pt, right=2pt, top=2pt, bottom=2pt]
\textbf{Finding 4:} Fine-tuned transformers outperform zero-shot LLMs by 18--96 F1 points. Zero-shot approaches are unsuitable for dialectally robust detection.
\end{tcolorbox}

\section{Discussion}
\label{sec:discussion}
Our findings reveal systematic dialectal biases in disinformation
detection with implications for equitable deployment. We highlight
key insights and provide recommendations for practitioners;
extended discussion and detailed recommendations appear in
Appendix~\ref{app:implications}.

\begin{figure}[t]
\centering
\includegraphics[width=\columnwidth]{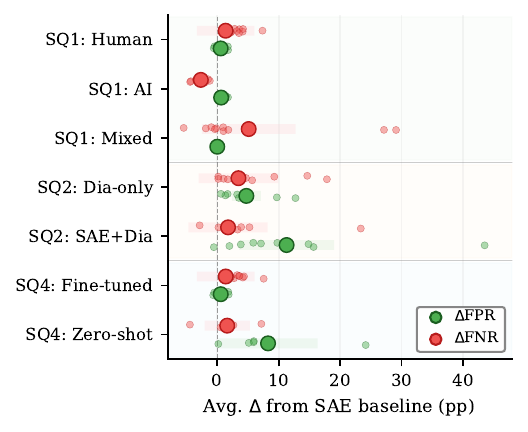}
\caption{Asymmetric harm across evaluation regimes.
$\Delta$FPR (green, over-flagging) and $\Delta$FNR
(red, under-protection) relative to the SAE baseline.
Large dots show cross-model means; small dots show
individual models. Shaded regions show 95\% CIs across
50 dialects. Per-model breakdown in
Figure~\ref{fig:fp_fn_full}.}
\label{fig:fp_fn_summary}
\vspace{-10pt}
\end{figure}

\paragraph{Asymmetric Harm Across Dialects.}
Per-dialect FPR/FNR analysis (Figure~\ref{fig:fp_fn_summary}; Table~\ref{tab:fp_fn_summary}; per-model breakdown in Figure~\ref{fig:fp_fn_full}, Appendix~\ref{app:fp_fn}) reveals dialectal bias manifests asymmetrically. Under unseen conditions (SQ1), human content shows 33/50 dialects under-protected ($\Delta$FNR\,=\,+1.4\%), AI content inverts to 50/50 over-flagged ($\Delta$FPR\,=\,+0.6\%), and mixed content triggers catastrophic under-protection ($\Delta$FNR\,=\,+5.1\%), driven by RoBERTa (+27.1\%) and XLM-R (+29.1\%). Dialect-aware training (SQ2) shifts toward over-flagging: dialect-only yields 43/50 over-flagged ($\Delta$FPR\,=\,+4.7\%), SAE-anchoring amplifies to 50/50 ($\Delta$FPR\,=\,+11.3\%), driven by dEFEND (+43.5\%). Critically, training composition reverses \emph{which} communities are harmed: RoBERTa flips from over-flagging under dialect-only to under-protection under SAE-anchored. Zero-shot LLMs (SQ4) over-flag all 48 dialects ($\Delta$FPR\,=\,+8.3\%), while Mistral-7B and Llama-3.2-1B miss 99.3\% and 61.2\% of dialectal disinformation respectively. This distinction matters for equitable deployment: over-flagging silences authentic dialectal speech, while under-protection leaves communities vulnerable.

\paragraph{Linguistic Mechanisms and Confidence Patterns.}
Analysis of 31,189 dialect-induced errors (Appendix~\ref{app:qualitative}) reveals over-flagging dominates under-protection 6.5:1 (27,020 FPs vs.\ 4,169 FNs). Six linguistic mechanisms drive errors (Table~\ref{tab:error_taxonomy}), primarily dialectal morphology (\textit{them}-suffixing, \textit{a-}prefixing) creating tokens models associate with fabricated content, alongside syntactic reordering and pronoun substitution disrupting positional cues. Twitter posts account for 71.7\% of FPs despite comprising 33\% of test data. Critically, 81.4\% of FPs and 75.5\% of FNs exceed $>$0.95 confidence (Table~\ref{tab:confidence_disinfo}), with RoBERTa averaging 99.5\%, ruling out calibration fixes and indicating dialectal features are encoded as class-discriminative signals requiring architectural or training interventions.

\paragraph{Content-Type Asymmetry.}
The divergent patterns between human content (degradation) and AI content (stability) suggest fundamentally different detection mechanisms. Human-written disinformation detection relies on stylistic cues disrupted by dialectal transformation, while AI detection leverages artifacts preserved across linguistic varieties. This asymmetry implies that as AI-generated disinformation proliferates, dialectal bias may paradoxically decrease, but human-written content from dialectal communities will remain under-protected.

\paragraph{Multilingual Advantage.}
Models with multilingual pre-training (mDeBERTa, mBERT) consistently outperform monolingual counterparts on dialectal robustness, despite English dialects not appearing in their training data. This suggests that exposure to typological diversity during pre-training induces representations robust to within-language variation, a finding with implications for low-resource NLP more broadly.

\paragraph{The SAE Anchoring Paradox.}
Counter to intuition, including \sae{} in training data provides no benefit for transformers and actively harms traditional models. We hypothesize that \sae{} over-representation induces feature collapse toward standard patterns, degrading dialectal generalization. Practitioners should prioritize dialect-diverse corpora without \sae{} anchoring.

\paragraph{Zero-Shot Brittleness.} The catastrophic failures of zero-shot LLMs (0.2--78.3\% F1) underscore that instruction-tuned models cannot reliably generalize to dialectal inputs without task-specific adaptation. High abstention rates (up to 98\%) suggest these models perceive dialectal text as out-of-distribution, raising concerns for content moderation pipelines serving diverse communities.

\paragraph{Robustness to Prompting Strategy.}
Evaluating four prompt variants (original, simplified, chain-of-thought, role-based) and four ICL conditions (0/2/5-shot with SAE and dialect-matched exemplars) across three models and five dialects (Appendix~\ref{app:prompt_variants}) confirms zero-shot brittleness is structural, not prompt-dependent. The best prompt reaches only 65.0\% dialect F1 ($\Delta$\,=\,$-$10.7), with $\Delta$ up to $-$74.4 for role-based prompting. Few-shot SAE exemplars narrow Llama-3.2-3B's gap from $\Delta$\,=\,$-$63.0 to $-$18.8, but dialect-matched exemplars using identical source content \emph{widen} it ($\Delta$\,=\,$-$57.3), revealing the comprehension deficit extends to in-context exemplars themselves. Even the best configuration falls 30+ F1 points below fine-tuned transformers.

\paragraph{Recommendations for Practitioners.}
Our findings yield five actionable recommendations (detailed in
Appendix~\ref{app:implications}):
{\footnotesize\textbf{(R1)}}~\textit{Prefer multilingual encoders}: mDeBERTa and mBERT
consistently achieve superior dialectal robustness, with mDeBERTa
maintaining 97.2\% average F1 across 2,450 transfer pairs
(\S\ref{sec:sq3}); {\footnotesize\textbf{(R2)}}~Adopt dialect-diverse fine-tuning without \sae{}
anchoring, which recovers catastrophic failures
(e.g., RoBERTa: 87.3\%$\rightarrow$97.1\%) while avoiding
feature collapse toward standard patterns (\S\ref{sec:sq2});
{\footnotesize\textbf{(R3)}}~Conduct pre-deployment dialectal auditing across
dialect families using our released D-CUBE benchmark and evaluation
scripts, establishing minimum thresholds per dialect family rather
than relying on aggregate metrics;
{\footnotesize\textbf{(R4)}}~Avoid zero-shot LLMs for content moderation, given
performance gaps of 18--97 F1 points and abstention rates up to
98\% on dialectal inputs (\S\ref{sec:sq4}); and
{\footnotesize\textbf{(R5)}}~Monitor dialectal performance longitudinally, as
retraining on \sae{}-dominant data may reintroduce bias that
continuous dialectal benchmarking can detect.

\section{Conclusion}
\label{sec:conclusion}

We introduced \method{}, a benchmark comprising three 
components: D-CUBE, a 195K-sample corpus spanning 50 English dialects; 
D-PURIFY, a quality validation pipeline; and an evaluation framework 
spanning four experimental regimes (SQ1--SQ4). We evaluated 16 
detection models across these regimes, yielding four 
key findings: (1) human content degrades 1.4--3.6\% while AI remains 
stable; (2) dialect-only training recovers catastrophic failures; (3) 
cross-dialectal transfer is asymmetric; (4) fine-tuned models 
outperform zero-shot LLMs by 18--97 F1-Score. These results reveal 
systematic disadvantages for non-\sae{} communities. We recommend 
multilingual architectures, dialect-diverse training, and rigorous 
dialectal evaluation before deployment.

\section*{Limitations}
\label{sec:limitations}

Our study has several limitations that inform future research directions.

\paragraph{Dialect Transformation Scope.}
Our dialectal transformations via Multi-VALUE are rule-based approximations grounded in eWAVE---a peer-reviewed linguistic atlas covering 235 features across 12 grammatical categories with $>$95\% native speaker acceptability \cite{ziems-etal-2023-multi}. While this establishes a controlled framework that isolates morphosyntactic effects from confounding factors---topic, domain, author style---it may not fully capture pragmatic, discourse-level, or broader sociolinguistic variation. We frame our findings as a \textit{lower bound}: if detectors fail on controlled morphosyntactic variation, they will likely degrade further on natural dialect text---where pragmatic, lexical, and discourse-level variation compounds the challenge. Future work should (a)~validate Multi-VALUE transformations against natural dialect corpora, and (b)~develop LLM-based dialect transformation approaches using advance technique such as Chain-of-Interactions \cite{lucas-etal-2025-chain} and LLM-as-a-judge\cite{gu2024survey} that capture variation, validate dialectal quality, features and modifications beyond rule-based morphosyntax.

We note that no existing dataset contains human-authored
disinformation written natively in non-SAE dialects;
constructing such a resource, through community-partnered
data collection across 3--5 high-impact dialects (e.g.,
AAVE, Singlish, Nigerian English), is a priority for
future work and would enable direct validation of whether
the performance gaps observed on synthetic transformations
persist on naturally occurring dialectal disinformation.

\paragraph{Evaluation Tooling.}
D-PURIFY relies on SAE-trained metrics---BERTScore, BARTScore, AlignScore---that may encode SAE norms, treating subtle dialectal deviations as errors rather than valid variation (Appendix~\ref{app:sae_adjacent}). This may also bias representation against lower-pass-rate dialects---e.g., Chicano English at 71.4\%---potentially underrepresenting certain communities. Future work should develop dialect-aware evaluation tools---including dialect-sensitive metrics and LLM-as-Judge approaches---that assess quality without penalizing authentic dialectal features.

\paragraph{Scope and Generalizability.}
Our evaluation targets English dialects and disinformation detection. While our coverage---50 dialects spanning five geographic regions---is unprecedented in this domain, generalization to other languages requires investigation. Future work should extend this framework to other languages---e.g., French, Arabic, Spanish varieties---and other harmful content domains such as jailbreaking, prompt injection, and adversarial code generation, where dialectal variation may similarly expose vulnerabilities.

\paragraph{Zero-Shot Evaluation.}
Zero-shot evaluation used a single prompt template; performance may vary with alternative strategies or few-shot in-context learning. However, the observed patterns---high abstention rates (58--98\%) and catastrophic failures in smaller models---reflect fundamental instruction-following limitations on dialectal inputs rather than prompt sensitivity alone. Future work should explore few-shot ICL, prompt optimization, and LLM-driven dialect generation to further characterize these failure modes.

\section*{Ethics Statement}
\label{sec:ethics}

\paragraph{Intended Use and Potential Misuse.}
This research aims to improve disinformation detection equity across linguistic communities. While D-CUBE is designed to evaluate and improve detector robustness, we acknowledge potential misuse: adversaries could exploit identified vulnerabilities to craft dialect-specific disinformation that evades detection. To mitigate this risk, we release the \method{} benchmark, including the D-CUBE corpus, D-PURIFY validation tools, and evaluation code and model 
checkpoints, but withhold the adversarial generation pipeline. We 
encourage researchers to use these resources for defensive purposes only.

\paragraph{Community Impact.}
Our findings reveal that current detection systems may systematically disadvantage speakers of non-standard English varieties, potentially resulting in higher exposure to undetected disinformation or, conversely, higher false positive rates that disproportionately flag legitimate dialectal content as suspicious. Such effects can prove catastrophic in resource-constrained environments and among long-tail populations that bear a disproportionate burden of disinformation propagation \cite{maung2024generative, lucas2024longtail}. We hope this work motivates the development of more equitable NLP systems and encourages practitioners to conduct dialectal audits before deployment.

\paragraph{Data Considerations.}
D-CUBE is derived from existing publicly available 
disinformation datasets (GossipCop, PolitiFact, CoAID) transformed using the open-source Multi-VALUE framework. No personally identifiable information is collected or generated. Dialectal transformations are rule-based approximations that do not involve human subjects. We acknowledge that computational modeling of dialects risks reinforcing linguistic stereotypes; our feature inventory derives from peer-reviewed linguistic scholarship (eWAVE) to minimize this concern.

\paragraph{Broader Implications.}
Disinformation disproportionately targets marginalized communities who often speak non-standard language varieties. By quantifying dialectal detection disparities, we aim to inform policy discussions around content moderation equity and encourage platform providers to audit their systems across linguistic demographics.

\section*{Acknowledgments}
\label{app:acknowledgment}
This work was supported in part by U.S. NSF awards \#2114824 and \#2438810. Some experimental results were obtained using computational resources provided by CloudBank, supported through U.S. NAIRR award \#240336.

\vspace{0.1in}
\noindent
DISTRIBUTION STATEMENT A. Approved for public release. Distribution is unlimited.
This material is based upon work supported by the Department of the Air Force under Air Force Contract No. FA8702-15-D-0001 or FA8702-25-D-B002. Any opinions, findings, conclusions or recommendations expressed in this material are those of the author(s) and do not necessarily reflect the views of the Department of the Air Force.
© 2025 Massachusetts Institute of Technology.
Delivered to the U.S. Government with Unlimited Rights, as defined in DFARS Part 252.227-7013 or 7014 (Feb 2014). Notwithstanding any copyright notice, U.S. Government rights in this work are defined by DFARS 252.227-7013 or DFARS 252.227-7014 as detailed above. Use of this work other than as specifically authorized by the U.S. Government may violate any copyrights that exist in this work.

\bibliography{custom}

@inproceedings{lucas-etal-2023-fighting,
    title = "Fighting Fire with Fire: The Dual Role of {LLM}s in Crafting and Detecting Elusive Disinformation",
    author = "Lucas, Jason and
      Uchendu, Adaku and
      Yamashita, Michiharu and
      Lee, Jooyoung and
      Rohatgi, Shaurya and
      Lee, Dongwon",
    editor = "Bouamor, Houda and
      Pino, Juan and
      Bali, Kalika",
    booktitle = "Proceedings of the 2023 Conference on Empirical Methods in Natural Language Processing",
    month = dec,
    year = "2023",
    address = "Singapore",
    publisher = "Association for Computational Linguistics",
    url = "https://aclanthology.org/2023.emnlp-main.883",
    doi = "10.18653/v1/2023.emnlp-main.883",
    pages = "14279--14305"
}

@inproceedings{ziems-etal-2023-multi,
    title = "Multi-{VALUE}: A Framework for Cross-Dialectal {E}nglish {NLP}",
    author = "Ziems, Caleb and
      Held, William and
      Yang, Jingfeng and
      Dhamala, Jwala and
      Gupta, Rahul and
      Yang, Diyi",
    editor = "Rogers, Anna and
      Boyd-Graber, Jordan and
      Okazaki, Naoaki",
    booktitle = "Proceedings of the 61st Annual Meeting of the Association for Computational Linguistics (Volume 1: Long Papers)",
    month = jul,
    year = "2023",
    address = "Toronto, Canada",
    publisher = "Association for Computational Linguistics",
    url = "https://aclanthology.org/2023.acl-long.44",
    doi = "10.18653/v1/2023.acl-long.44",
    pages = "744--768"
}

@inproceedings{devlin-etal-2019-bert,
    title = "{BERT}: Pre-training of Deep Bidirectional Transformers for Language Understanding",
    author = "Devlin, Jacob and
      Chang, Ming-Wei and
      Lee, Kenton and
      Toutanova, Kristina",
    editor = "Burstein, Jill and
      Doran, Christy and
      Solorio, Thamar",
    booktitle = "Proceedings of the 2019 Conference of the North {A}merican Chapter of the Association for Computational Linguistics: Human Language Technologies, Volume 1 (Long and Short Papers)",
    month = jun,
    year = "2019",
    address = "Minneapolis, Minnesota",
    publisher = "Association for Computational Linguistics",
    url = "https://aclanthology.org/N19-1423",
    doi = "10.18653/v1/N19-1423",
    pages = "4171--4186"
}

@inproceedings{blodgett-etal-2016-demographic,
    title = "Demographic Dialectal Variation in Social Media: A Case Study of {A}frican-{A}merican {E}nglish",
    author = "Blodgett, Su Lin and
      Eisenstein, Jacob",
    editor = "Su, Jian and
      Duh, Kevin and
      Carreras, Xavier",
    booktitle = "Proceedings of the 2016 Conference on Empirical Methods in Natural Language Processing",
    month = nov,
    year = "2016",
    address = "Austin, Texas",
    publisher = "Association for Computational Linguistics",
    url = "https://aclanthology.org/D16-1120",
    doi = "10.18653/v1/D16-1120",
    pages = "1119--1130"
}

@inproceedings{sap-etal-2019-risk,
    title = "The Risk of Racial Bias in Hate Speech Detection",
    author = "Sap, Maarten and
      Card, Dallas and
      Gabriel, Saadia and
      Choi, Yejin and
      Smith, Noah A.",
    editor = "Korhonen, Anna and
      Traum, David and
      M{\`a}rquez, Llu{\'\i}s",
    booktitle = "Proceedings of the 57th Annual Meeting of the Association for Computational Linguistics",
    month = jul,
    year = "2019",
    address = "Florence, Italy",
    publisher = "Association for Computational Linguistics",
    url = "https://aclanthology.org/P19-1163",
    doi = "10.18653/v1/P19-1163",
    pages = "1668--1678"
}

@incollection{kortmann-etal-2020-ewave,
    title = "e{WAVE}: The Electronic World Atlas of Varieties of {E}nglish",
    author = "Kortmann, Bernd and
      Lunkenheimer, Kerstin and
      Ehret, Katharina",
    booktitle = "The Handbook of English Linguistics",
    pages = "613--635",
    year = "2020",
    publisher = "John Wiley {\&} Sons"
}

@inproceedings{wang-2017-liar,
    title = "``Liar, Liar Pants on Fire'': A New Benchmark Dataset for Fake News Detection",
    author = "Wang, William Yang",
    editor = "Barzilay, Regina and
      Kan, Min-Yen",
    booktitle = "Proceedings of the 55th Annual Meeting of the Association for Computational Linguistics (Volume 2: Short Papers)",
    month = jul,
    year = "2017",
    address = "Vancouver, Canada",
    publisher = "Association for Computational Linguistics",
    url = "https://aclanthology.org/P17-2067",
    doi = "10.18653/v1/P17-2067",
    pages = "422--426"
}

@inproceedings{zellers-etal-2019-defending,
    title = "Defending Against Neural Fake News",
    author = "Zellers, Rowan and
      Holtzman, Ari and
      Rashkin, Hannah and
      Bisk, Yonatan and
      Farhadi, Ali and
      Roesner, Franziska and
      Choi, Yejin",
    booktitle = "Advances in Neural Information Processing Systems",
    volume = "32",
    year = "2019"
}

@article{liu-etal-2019-roberta,
    title = "{R}o{BERT}a: A Robustly Optimized {BERT} Pretraining Approach",
    author = "Liu, Yinhan  and
      Ott, Myle  and
      Goyal, Naman  and
      Du, Jingfei  and
      Joshi, Mandar  and
      Chen, Danqi  and
      Levy, Omer  and
      Lewis, Mike  and
      Zettlemoyer, Luke  and
      Stoyanov, Veselin",
    journal = "arXiv preprint arXiv:1907.11692",
    year = "2019"
}

@article{joshi2025natural,
  title={Natural language processing for dialects of a language: A survey},
  author={Joshi, Aditya and Dabre, Raj and Kanojia, Diptesh and Li, Zhuang and Zhan, Haolan and Haffari, Gholamreza and Dippold, Doris},
  journal={ACM Computing Surveys},
  volume={57},
  number={6},
  pages={1--37},
  year={2025},
  publisher={ACM New York, NY}
}

@inproceedings{faisal-etal-2024-dialectbench,
    title = "{DIALECTBENCH}: An {NLP} Benchmark for Dialects, Varieties, and Closely-Related Languages",
    author = "Faisal, Fahim  and
      Ahia, Orevaoghene  and
      Srivastava, Aarohi  and
      Ahuja, Kabir  and
      Chiang, David  and
      Tsvetkov, Yulia  and
      Anastasopoulos, Antonios",
    editor = "Ku, Lun-Wei  and
      Martins, Andre  and
      Srikumar, Vivek",
    booktitle = "Proceedings of the 62nd Annual Meeting of the Association for Computational Linguistics (Volume 1: Long Papers)",
    month = aug,
    year = "2024",
    address = "Bangkok, Thailand",
    publisher = "Association for Computational Linguistics",
    url = "https://aclanthology.org/2024.acl-long.777/",
    doi = "10.18653/v1/2024.acl-long.777",
    pages = "14412--14454"
}

@inproceedings{mire2025rejected,
  title={Rejected Dialects: Biases Against African American Language in Reward Models},
  author={Mire, Joel and Aysola, Zubin Trivadi and Chechelnitsky, Daniel and Deas, Nicholas and Zerva, Chrysoula and Sap, Maarten},
  booktitle={Findings of the Association for Computational Linguistics: NAACL 2025},
  pages={7468--7487},
  year={2025}
}

@inproceedings{okpala2022aaebert,
  title={Aaebert: Debiasing bert-based hate speech detection models via adversarial learning},
  author={Okpala, Ebuka and Cheng, Long and Mbwambo, Nicodemus and Luo, Feng},
  booktitle={2022 21st IEEE International Conference on Machine Learning and Applications (ICMLA)},
  pages={1606--1612},
  year={2022},
  organization={IEEE}
}

@article{zhou2025disparities,
  title={Disparities in llm reasoning accuracy and explanations: A case study on african american english},
  author={Zhou, Runtao and Wan, Guangya and Gabriel, Saadia and Li, Sheng and Gates, Alexander J and Sap, Maarten and Hartvigsen, Thomas},
  journal={arXiv preprint arXiv:2503.04099},
  year={2025}
}

@inproceedings{ziems-etal-2022-value,
    title = "{VALUE}: {U}nderstanding Dialect Disparity in {NLU}",
    author = "Ziems, Caleb  and
      Chen, Jiaao  and
      Harris, Camille  and
      Anderson, Jessica  and
      Yang, Diyi",
    editor = "Muresan, Smaranda  and
      Nakov, Preslav  and
      Villavicencio, Aline",
    booktitle = "Proceedings of the 60th Annual Meeting of the Association for Computational Linguistics (Volume 1: Long Papers)",
    month = may,
    year = "2022",
    address = "Dublin, Ireland",
    publisher = "Association for Computational Linguistics",
    url = "https://aclanthology.org/2022.acl-long.258/",
    doi = "10.18653/v1/2022.acl-long.258",
    pages = "3701--3720"
}

@inproceedings{keswani2021dialect,
  title={Dialect diversity in text summarization on twitter},
  author={Keswani, Vijay and Celis, L Elisa},
  booktitle={Proceedings of the web conference 2021},
  pages={3802--3814},
  year={2021}
}

@inproceedings{kantharuban-etal-2023-quantifying,
    title = "Quantifying the Dialect Gap and its Correlates Across Languages",
    author = "Kantharuban, Anjali  and
      Vuli{\'c}, Ivan  and
      Korhonen, Anna",
    editor = "Bouamor, Houda  and
      Pino, Juan  and
      Bali, Kalika",
    booktitle = "Findings of the Association for Computational Linguistics: EMNLP 2023",
    month = dec,
    year = "2023",
    address = "Singapore",
    publisher = "Association for Computational Linguistics",
    url = "https://aclanthology.org/2023.findings-emnlp.481/",
    doi = "10.18653/v1/2023.findings-emnlp.481",
    pages = "7226--7245"
}

@inproceedings{scannell-2020-universal,
    title = "{U}niversal {D}ependencies for {M}anx {G}aelic",
    author = "Scannell, Kevin",
    editor = "de Marneffe, Marie-Catherine  and
      de Lhoneux, Miryam  and
      Nivre, Joakim  and
      Schuster, Sebastian",
    booktitle = "Proceedings of the Fourth Workshop on Universal Dependencies (UDW 2020)",
    month = dec,
    year = "2020",
    address = "Barcelona, Spain (Online)",
    publisher = "Association for Computational Linguistics",
    url = "https://aclanthology.org/2020.udw-1.17/",
    pages = "152--157"
}

@article{peppin2025multilingual,
  title={The Multilingual Divide and Its Impact on Global AI Safety},
  author={Peppin, Aidan and Kreutzer, Julia and Sebag, Alice Schoenauer and Marchisio, Kelly and Ermis, Beyza and Dang, John and Cahyawijaya, Samuel and Singh, Shivalika and Goldfarb-Tarrant, Seraphina and Aryabumi, Viraat and others},
  journal={arXiv preprint arXiv:2505.21344},
  year={2025}
}

@inproceedings{pikuliak-etal-2023-multilingual,
    title = "Multilingual Previously Fact-Checked Claim Retrieval",
    author = "Pikuliak, Mat{\'u}{\v{s}} and
              Srba, Ivan and
              Moro, Robert and
              Hromadka, Timo and
              Smole{\v{n}}, Timotej and
              Meli{\v{s}}ek, Martin and
              Vykopal, Ivan and
              Simko, Jakub and
              Podrou{\v{z}}ek, Juraj and
              Bielikova, Maria",
    editor = "Bouamor, Houda and
              Pino, Juan and
              Bali, Kalika",
    booktitle = "Proceedings of the 2023 Conference on Empirical Methods in Natural Language Processing",
    month = dec,
    year = "2023",
    address = "Singapore",
    publisher = "Association for Computational Linguistics",
    url = "https://aclanthology.org/2023.emnlp-main.1027",
    doi = "10.18653/v1/2023.emnlp-main.1027",
    pages = "16477--16500",
}

@misc{cui2020coaid,
    title = {CoAID: COVID-19 Healthcare Misinformation Dataset},
    author = {Limeng Cui and Dongwon Lee},
    year = {2020},
    eprint = {2006.00885},
    archivePrefix = {arXiv},
    primaryClass = {cs.SI},
    url = {https://arxiv.org/abs/2006.00885},
}

@article{shu2020fakenewsnet,
    title = {FakeNewsNet: A Data Repository with News Content, Social Context, and Spatiotemporal Information for Studying Fake News on Social Media},
    author = {Shu, Kai and Mahudeswaran, Deepak and Wang, Suhang and Lee, Dongwon and Liu, Huan},
    journal = {Big Data},
    volume = {8},
    number = {3},
    pages = {171--188},
    year = {2020},
    publisher = {Mary Ann Liebert, Inc.},
    doi = {10.1089/big.2020.0062},
}

@misc{li2020mmcovid,
    title = {MM-COVID: A Multilingual and Multimodal Data Repository for Combating COVID-19 Disinformation},
    author = {Yichuan Li and Bohan Jiang and Kai Shu and Huan Liu},
    year = {2020},
    eprint = {2011.04088},
    archivePrefix = {arXiv},
    primaryClass = {cs.SI},
    url = {https://arxiv.org/abs/2011.04088},
}

@inproceedings{zha-etal-2023-alignscore,
    title = "{A}lign{S}core: Evaluating Factual Consistency with A Unified Alignment Function",
    author = "Zha, Yuheng and Yang, Yichi and Li, Ruichen and Hu, Zhiting",
    editor = "Rogers, Anna and Boyd-Graber, Jordan and Okazaki, Naoaki",
    booktitle = "Proceedings of the 61st Annual Meeting of the Association for Computational Linguistics (Volume 1: Long Papers)",
    month = jul,
    year = "2023",
    address = "Toronto, Canada",
    publisher = "Association for Computational Linguistics",
    url = "https://aclanthology.org/2023.acl-long.634",
    doi = "10.18653/v1/2023.acl-long.634",
    pages = "11328--11348"
}

@inproceedings{lucas-etal-2025-chain,
    title = "Chain-of-Interactions: Multi-step Iterative {ICL} Framework for Abstractive Task-Oriented Dialogue Summarization of Conversational {AI} Interactions",
    author = "Lucas, Jason  and
      Chen, John  and
      Al-Lawati, Ali  and
      Nahar, Mahjabin  and
      Mehrabani, Mahnoosh",
    editor = "Christodoulopoulos, Christos  and
      Chakraborty, Tanmoy  and
      Rose, Carolyn  and
      Peng, Violet",
    booktitle = "Findings of the Association for Computational Linguistics: EMNLP 2025",
    month = nov,
    year = "2025",
    address = "Suzhou, China",
    publisher = "Association for Computational Linguistics",
    url = "https://aclanthology.org/2025.findings-emnlp.191/",
    doi = "10.18653/v1/2025.findings-emnlp.191",
    pages = "3560--3599",
    ISBN = "979-8-89176-335-7"
}

@ARTICLE{jsl10702034,
  author={Lucas, Jason S. and Maung, Barani Maung and Tabar, Maryam and McBride, Keegan and Lee, Dongwon},
  journal={IEEE Intelligent Systems}, 
  title={The Longtail Impact of Generative AI on Disinformation: Harmonizing Dichotomous Perspectives}, 
  year={2024},
  volume={39},
  number={5},
  pages={12-19},
  doi={10.1109/MIS.2024.3439109}}

@ARTICLE{bmm10718648,
  author={Maung, Barani Maung and McBride, Keegan and Lucas, Jason S. and Tabar, Maryam and Lee, Dongwon},
  journal={Computer}, 
  title={Generative AI Disproportionately Harms Long Tail Users}, 
  year={2024},
  volume={57},
  number={11},
  pages={82-85},
  doi={10.1109/MC.2024.3408594}}

@article{Aslam2021-by,
    title        = {Fake detect: A deep learning ensemble model for fake news detection},
    author       = {Aslam, Nida and Ullah Khan, Irfan and Alotaibi, Farah Salem and Aldaej, Lama Abdulaziz and Aldubaikil, Asma Khaled},
    year         = 2021,
    journal      = {Complexity},
    publisher    = {Hindawi Limited},
    volume       = 2021,
    pages        = {1--8},
    doi          = {10.1155/2021/5557784}
}

@inproceedings{Upadhayay2022-xo,
    title        = {Hybrid deep learning model for fake news detection in social networks (student abstract)},
    author       = {Upadhayay, Bibek and Behzadan, Vahid},
    year         = 2022,
    booktitle    = {Proceedings of the AAAI Conference on Artificial Intelligence},
    volume       = 36,
    pages        = {13067--13068},
    doi          = {10.1609/aaai.v36i11.21670}
}

@inproceedings{Jayakody2022-cg,
    title        = {Fake news detection using a decentralized deep learning model and federated learning},
    author       = {Jayakody, Nirosh and Mohammad, Azeem and Halgamuge, Malka N},
    year         = 2022,
    booktitle    = {IECON 2022 -- 48th Annual Conference of the IEEE Industrial Electronics Society},
    publisher    = {IEEE},
    pages        = {1--6},
    doi          = {10.1109/iecon49645.2022.9968358}
}

@article{Ali2022-ob,
    title        = {Deep Ensemble Fake News Detection Model Using Sequential Deep Learning Technique},
    author       = {Ali, Abdullah Marish and Ghaleb, Fuad A and Al-Rimy, Bander Ali Saleh and Alsolami, Fawaz Jaber and Khan, Asif Irshad},
    year         = 2022,
    journal      = {Sensors},
    volume       = 22,
    number       = 18,
    pages        = 6970,
    doi          = {10.3390/s22186970}
}

@inproceedings{cui2020deterrent,
    title        = {{DETERRENT}: Knowledge Guided Graph Attention Network for Detecting Healthcare Misinformation},
    author       = {Cui, Limeng and Seo, Haeseung and Tabar, Maryam and Ma, Fenglong and Wang, Suhang and Lee, Dongwon},
    year         = 2020,
    booktitle    = {Proceedings of the 26th ACM SIGKDD International Conference on Knowledge Discovery \& Data Mining},
    publisher    = {Association for Computing Machinery},
    address      = {New York, NY, USA},
    pages        = {492--502},
    doi          = {10.1145/3394486.3403092}
}

@inproceedings{Shu2019-vm,
    title        = {{dEFEND}: Explainable Fake News Detection},
    author       = {Shu, Kai and Cui, Limeng and Wang, Suhang and Lee, Dongwon and Liu, Huan},
    year         = 2019,
    booktitle    = {Proceedings of the 25th ACM SIGKDD International Conference on Knowledge Discovery \& Data Mining},
    publisher    = {Association for Computing Machinery},
    address      = {New York, NY, USA},
    series       = {KDD '19},
    pages        = {395--405},
    doi          = {10.1145/3292500.3330935}
}

@inproceedings{Nguyen2020-ju,
    title        = {{FANG}: Leveraging Social Context for Fake News Detection Using Graph Representation},
    author       = {Nguyen, Van-Hoang and Sugiyama, Kazunari and Nakov, Preslav and Kan, Min-Yen},
    year         = 2020,
    booktitle    = {Proceedings of the 29th ACM International Conference on Information \& Knowledge Management},
    publisher    = {Association for Computing Machinery},
    address      = {New York, NY, USA},
    series       = {CIKM '20},
    pages        = {1165--1174},
    doi          = {10.1145/3340531.3412046}
}

@inproceedings{zhou2023synthetic,
    title        = {Synthetic Lies: Understanding {AI}-Generated Misinformation and Evaluating Algorithmic and Human Solutions},
    author       = {Zhou, Jiawei and Zhang, Yixuan and Luo, Qianni and Parker, Andrea G and De Choudhury, Munmun},
    year         = 2023,
    booktitle    = {Proceedings of the 2023 CHI Conference on Human Factors in Computing Systems},
    publisher    = {Association for Computing Machinery},
    address      = {New York, NY, USA},
    series       = {CHI '23},
    articleno    = 436,
    numpages     = 20,
    doi          = {10.1145/3544548.3581318}
}

@article{Liu2023-en,
    title        = {Evaluating the Logical Reasoning Ability of {ChatGPT} and {GPT-4}},
    author       = {Liu, Hanmeng and Ning, Ruoxi and Teng, Zhiyang and Liu, Jian and Zhou, Qiji and Zhang, Yue},
    year         = 2023,
    journal      = {arXiv preprint arXiv:2304.03439},
    eprint       = {2304.03439},
    archiveprefix = {arXiv},
    primaryclass = {cs.CL}
}

@article{Qin2023-za,
    title        = {Is {ChatGPT} a General-Purpose Natural Language Processing Task Solver?},
    author       = {Qin, Chengwei and Zhang, Aston and Zhang, Zhuosheng and Chen, Jiaao and Yasunaga, Michihiro and Yang, Diyi},
    year         = 2023,
    journal      = {arXiv preprint arXiv:2302.06476},
    eprint       = {2302.06476},
    archiveprefix = {arXiv},
    primaryclass = {cs.CL}
}

@book{ewave,
  address   = {},
  editor    = {Bernd Kortmann and Kerstin Lunkenheimer and Katharina Ehret},
  publisher = {},
  title     = {eWAVE},
  url       = {https://ewave-atlas.org/},
  year      = {2020}
}

@inproceedings{alzantot-etal-2018-generating,
    title = "Generating Natural Language Adversarial Examples",
    author = "Alzantot, Moustafa  and
      Sharma, Yash  and
      Elgohary, Ahmed  and
      Ho, Bo-Jhang  and
      Srivastava, Mani  and
      Chang, Kai-Wei",
    booktitle = "Proceedings of the 2018 Conference on Empirical Methods in Natural Language Processing",
    month = oct # "-" # nov,
    year = "2018",
    address = "Brussels, Belgium",
    publisher = "Association for Computational Linguistics",
    url = "https://aclanthology.org/D18-1316",
    pages = "2890--2896",
}

@inproceedings{jin-etal-2020-bert,
    title = "Is {BERT} Really Robust? A Strong Baseline for Natural Language Attack on Text Classification and Entailment",
    author = "Jin, Di  and
      Jin, Zhijing  and
      Zhou, Joey Tianyi  and
      Szolovits, Peter",
    booktitle = "Proceedings of the AAAI Conference on Artificial Intelligence",
    volume = "34",
    year = "2020",
    pages = "8018--8025",
    publisher = "Association for the Advancement of Artificial Intelligence",
    url = "https://ojs.aaai.org/index.php/AAAI/article/view/6311",
}

@article{horne-etal-2019-robust,
    title = "Robust Fake News Detection Over Time and Attack",
    author = "Horne, Benjamin D.  and
      N{\o}rregaard, Jeppe  and
      Adal{\i}, Sibel",
    journal = "ACM Transactions on Intelligent Systems and Technology",
    volume = "11",
    number = "1",
    articleno = "7",
    year = "2019",
    publisher = "Association for Computing Machinery",
    address = "New York, NY, USA",
    doi = "10.1145/3363818",
    url = "https://dl.acm.org/doi/10.1145/3363818",
}

@article{sheng-etal-2022-characterizing,
    title = "Characterizing Multi-Domain False News and Underlying User Effects on {C}hinese {W}eibo",
    author = "Sheng, Qiang  and
      Cao, Juan  and
      Bernard, H. Russell  and
      Shu, Kai  and
      Li, Jintao  and
      Liu, Huan",
    journal = "Information Processing \& Management",
    volume = "59",
    number = "4",
    pages = "102959",
    year = "2022",
    publisher = "Elsevier",
    doi = "10.1016/j.ipm.2022.102959",
    issn = "0306-4573",
}

@inproceedings{zhang-etal-2020-bertscore,
    title = "{BERT}Score: Evaluating Text Generation with {BERT}",
    author = "Zhang, Tianyi  and
      Kishore, Varsha  and
      Wu, Felix  and
      Weinberger, Kilian Q.  and
      Artzi, Yoav",
    booktitle = "International Conference on Learning Representations",
    year = "2020",
    url = "https://openreview.net/forum?id=SkeHuCVFDr",
}

@inproceedings{yuan-etal-2021-bartscore,
    title = "{BART}Score: Evaluating Generated Text as Text Generation",
    author = "Yuan, Weizhe  and
      Neubig, Graham  and
      Liu, Pengfei",
    booktitle = "Advances in Neural Information Processing Systems",
    volume = "34",
    pages = "27263--27277",
    year = "2021",
    publisher = "Curran Associates, Inc.",
    url = "https://proceedings.neurips.cc/paper/2021/hash/e4d2b6e6fdeca3e60e0f1a62fee3d9dd-Abstract.html",
}

@inproceedings{banerjee-lavie-2005-meteor,
    title = "{METEOR}: An Automatic Metric for {MT} Evaluation with Improved Correlation with Human Judgments",
    author = "Banerjee, Satanjeev  and
      Lavie, Alon",
    booktitle = "Proceedings of the {ACL} Workshop on Intrinsic and Extrinsic Evaluation Measures for Machine Translation and/or Summarization",
    month = jun,
    year = "2005",
    address = "Ann Arbor, Michigan",
    publisher = "Association for Computational Linguistics",
    url = "https://aclanthology.org/W05-0909",
    pages = "65--72",
}

@inproceedings{lin-2004-rouge,
    title = "{ROUGE}: A Package for Automatic Evaluation of Summaries",
    author = "Lin, Chin-Yew",
    booktitle = "Text Summarization Branches Out",
    month = jul,
    year = "2004",
    address = "Barcelona, Spain",
    publisher = "Association for Computational Linguistics",
    url = "https://aclanthology.org/W04-1013",
    pages = "74--81",
}

@inproceedings{conneau-etal-2020-unsupervised,
    title = "Unsupervised Cross-lingual Representation Learning at Scale",
    author = "Conneau, Alexis  and
      Khandelwal, Kartikay  and
      Goyal, Naman  and
      Chaudhary, Vishrav  and
      Wenzek, Guillaume  and
      Guzm{\'a}n, Francisco  and
      Grave, Edouard  and
      Ott, Myle  and
      Zettlemoyer, Luke  and
      Stoyanov, Veselin",
    booktitle = "Proceedings of the 58th Annual Meeting of the Association for Computational Linguistics",
    month = jul,
    year = "2020",
    address = "Online",
    publisher = "Association for Computational Linguistics",
    url = "https://aclanthology.org/2020.acl-main.747",
    doi = "10.18653/v1/2020.acl-main.747",
    pages = "8440--8451",
}

@article{Devlin2018-mr,
	title        = "{BERT}: Pre-training of Deep Bidirectional Transformers for Language Understanding",
	author       = "Devlin, Jacob and Chang, Ming-Wei and Lee, Kenton and Toutanova, Kristina",
	year         = 2018,
	month        = oct,
	journal      = {arXiv preprint arXiv:1810.04805},
	archiveprefix = "arXiv",
	eprint       = "1810.04805",
	primaryclass = "cs.CL",
	arxivid      = "1810.04805"
}

@inproceedings{ma-etal-2017-detect,
    title = "Detect Rumors in Microblog Posts Using Propagation Structure via Kernel Learning",
    author = "Ma, Jing  and
      Gao, Wei  and
      Wong, Kam-Fai",
    editor = "Barzilay, Regina  and
      Kan, Min-Yen",
    booktitle = "Proceedings of the 55th Annual Meeting of the Association for Computational Linguistics (Volume 1: Long Papers)",
    month = jul,
    year = "2017",
    address = "Vancouver, Canada",
    publisher = "Association for Computational Linguistics",
    url = "https://aclanthology.org/P17-1066/",
    doi = "10.18653/v1/P17-1066",
    pages = "708--717"
}

@inproceedings{toraman-etal-2024-mide22,
    title = "{M}i{D}e22: An Annotated Multi-Event Tweet Dataset for Misinformation Detection",
    author = "Toraman, Cagri  and
      Ozcelik, Oguzhan  and
      Sahinuc, Furkan  and
      Can, Fazli",
    editor = "Calzolari, Nicoletta  and
      Kan, Min-Yen  and
      Hoste, Veronique  and
      Lenci, Alessandro  and
      Sakti, Sakriani  and
      Xue, Nianwen",
    booktitle = "Proceedings of the 2024 Joint International Conference on Computational Linguistics, Language Resources and Evaluation (LREC-COLING 2024)",
    month = may,
    year = "2024",
    address = "Torino, Italia",
    publisher = "ELRA and ICCL",
    url = "https://aclanthology.org/2024.lrec-main.986/",
    pages = "11283--11295"
}

@article{gu2024survey,
  title={A survey on llm-as-a-judge},
  author={Gu, Jiawei and Jiang, Xuhui and Shi, Zhichao and Tan, Hexiang and Zhai, Xuehao and Xu, Chengjin and Li, Wei and Shen, Yinghan and Ma, Shengjie and Liu, Honghao and others},
  journal={The Innovation},
  year={2024},
  publisher={Elsevier}
}

@article{maung2024generative,
  title={Generative AI disproportionately harms long tail users},
  author={Maung, Barani Maung and McBride, Keegan and Lucas, Jason S and Tabar, Maryam and Lee, Dongwon},
  journal={Computer},
  volume={57},
  number={11},
  pages={82--85},
  year={2024},
  publisher={IEEE}
}

@article{lucas2024longtail,
  title={The longtail impact of generative AI on disinformation: Harmonizing dichotomous perspectives},
  author={Lucas, Jason S and Maung, Barani Maung and Tabar, Maryam and McBride, Keegan and Lee, Dongwon},
  journal={IEEE Intelligent Systems},
  volume={39},
  number={5},
  pages={12--19},
  year={2024},
  publisher={IEEE}
}

\appendix

\section{Quality Metrics and Thresholds}
\label{app:quality_metrics}

We evaluate dialect transformation quality using standard automatic evaluation metrics and a novel Feature Accuracy measure. Table~\ref{tab:metric_descriptions} provides detailed descriptions of each metric and its interpretation for dialect transformation evaluation.

\begin{table*}[t]
\centering
\small
\begin{tabular}{llp{0.55\textwidth}}
\toprule
\textbf{Metric} & \textbf{Range} & \textbf{Description} \\
\midrule
BERTScore & $[0, 1]$ & Semantic similarity via BERT embeddings. Higher scores indicate greater meaning preservation between original and transformed text. \\
BARTScore & $(-\infty, 0]$ & Log-likelihood of generating target from source. Closer to 0 indicates better quality (e.g., $-1 > -4 > -6$). \\
AlignScore & $[0, 1]$ & Factual and logical consistency between original and transformed text. Higher scores indicate better preservation of factual content. \\
METEOR & $[0, 1]$ & Translation quality accounting for synonyms, stemming, and paraphrasing. Higher scores indicate better quality preservation. \\
ROUGE-L & $[0, 1]$ & Longest common subsequence overlap. Higher scores indicate greater structural preservation. \\
DiffLib & $[0, 1]$ & SequenceMatcher ratio measuring surface-level similarity. 1.0 = identical (no transformation), 0.0 = completely different. \\
BLEU & $[0, 1]$ & N-gram precision measuring fluency preservation. Higher scores indicate better quality. \\
Feature Acc. & $[0, 1]$ & Proportion of transformations matching valid eWAVE dialect features. Higher scores indicate greater linguistic authenticity. \\
\bottomrule
\end{tabular}
\caption{Quality metric descriptions and interpretations for dialect transformation evaluation.}
\label{tab:metric_descriptions}
\end{table*}

\paragraph{Feature Accuracy Computation.} Feature Accuracy combines LLM-as-a-Judge with direct retrieval from the eWAVE static database. The pipeline operates as follows: (1) LLM identifies all changes between original and transformed text; (2) LLM validates each change against 235 eWAVE features; (3) LLM checks dialect appropriateness using eWAVE's A/B/C/D attestation ratings; (4) LLM flags semantic errors. We compute two accuracy measures:

\begin{equation}
\text{eWAVE Accuracy} = \frac{\text{valid} + \text{wrong\_dialect}}{\text{total\_changes}}
\end{equation}

\begin{equation}
\text{Dialect Accuracy} = \frac{\text{valid}}{\text{total\_changes}}
\end{equation}

where eWAVE Accuracy measures whether identified features exist in any dialect, and Dialect Accuracy measures whether features are correct for the target dialect specifically.

\paragraph{eWAVE Attestation Ratings.} The A/B/C/D ratings from eWAVE indicate how characteristic a feature is for a specific dialect:

\begin{table*}[h]
\centering
\begin{tabular}{clc}
\toprule
\textbf{Rating} & \textbf{Meaning} & \textbf{Validity} \\
\midrule
A & Pervasive/Obligatory---feature is highly characteristic & Valid \\
B & Common---feature exists and is frequently used & Valid \\
C & Rare---feature exists but is infrequent/marginal & Valid \\
D & Absent---documented as NOT used in this dialect & Invalid \\
X & Not applicable/No information available & --- \\
\bottomrule
\end{tabular}
\caption{eWAVE attestation ratings and their interpretation for Feature Accuracy computation.}
\label{tab:ewave_ratings}
\end{table*}

\paragraph{Filtering Thresholds.} Table~\ref{tab:threshold_details} provides the complete filtering thresholds with rationale and pass rates.

\begin{table*}[h]
\centering
\small
\begin{tabular}{llccp{0.3\textwidth}}
\toprule
\textbf{Metric} & \textbf{Threshold} & \textbf{Pass\%} & \textbf{Cumul.\%} & \textbf{Rationale} \\
\midrule
BERTScore & $> 0.50$ & 99.3 & 0.70 & Semantic preservation floor \\
BARTScore & $\geq -7.5$ & 99.7 & 0.99 & Generation quality floor \\
METEOR & $\geq 0.40$ & 100.0 & 0.99 & Translation quality floor \\
ROUGE-L & $(0.01, 1.0)$ & 97.6 & 3.35 & Excludes identical/empty \\
DiffLib & $[0.01, 0.99]$ & 95.8 & 5.60 & Ensures transformation occurred \\
BLEU & $> 0.01$ & 99.3 & 6.29 & N-gram quality floor \\
\midrule
\multicolumn{2}{l}{\textbf{Final Retention}} & \multicolumn{3}{l}{\textbf{194,960 / 208,056 (93.71\%)}} \\
\bottomrule
\end{tabular}
\caption{D-PURIFY filtering thresholds with pass rates and cumulative sample loss. Lenient thresholds preserve dialectal diversity.}
\label{tab:threshold_details}
\end{table*}

\section{Dialect Inventory}
\label{app:dialects}

D-CUBE covers 50 English dialects derived from the Electronic World Atlas of Varieties of English (eWAVE). Table~\ref{tab:dialect_inventory} presents the complete inventory organized by geographic region.

\begin{table*}[t]
\centering
\small
\begin{adjustbox}{max width=\textwidth}
\begin{tabular}{llccccccc}
\toprule
\textbf{Abbr} & \textbf{Dialect} & \textbf{\# Feat.} & \textbf{\% Feat.} & \textbf{\# Val.} & \textbf{\% Val.} & \textbf{\# Total} & \textbf{\# Pass} & \textbf{\% Pass} \\
\midrule
\multicolumn{9}{l}{\textit{\textbf{U.S. Varieties (9 dialects)}}} \\
EAAVE & Earlier African American Vernacular English & 96 & 89.7\% & 61 & 57.0\% & 4,247 & 4,158 & 97.90\% \\
RAAVE & Rural African American Vernacular English & 136 & 82.9\% & 88 & 53.7\% & 4,247 & 4,081 & 96.09\% \\
UAAVE & Urban African American Vernacular English & 118 & 83.7\% & 79 & 56.0\% & 4,247 & 4,183 & 98.49\% \\
AppE & Appalachian English & 65 & 85.5\% & 51 & 67.1\% & 4,247 & 4,088 & 96.26\% \\
ChcE & Chicano English & 30 & 93.8\% & 28 & 87.5\% & 4,247 & 3,034 & 71.44\% \\
CollAmE & Colloquial American English & 57 & 83.8\% & 44 & 64.7\% & 4,247 & 4,155 & 97.83\% \\
OzE & Ozark English & 56 & 86.2\% & 43 & 66.2\% & 4,247 & 4,075 & 95.95\% \\
SEAmE & Southeast American enclave dialects & 108 & 80.6\% & 75 & 56.0\% & 4,247 & 4,214 & 99.22\% \\
NfldE & Newfoundland English & 84 & 85.7\% & 53 & 54.1\% & 4,247 & 4,164 & 98.05\% \\
\midrule
\multicolumn{9}{l}{\textit{\textbf{British/UK Varieties (11 dialects)}}} \\
North & English dialects in the North of England & 77 & 85.6\% & 47 & 52.2\% & 4,247 & 4,176 & 98.33\% \\
SE & English dialects in the Southeast of England & 46 & 93.9\% & 33 & 67.3\% & 4,247 & 3,087 & 72.69\% \\
SW & English dialects in the Southwest of England & 73 & 89.0\% & 46 & 56.1\% & 4,247 & 4,138 & 97.43\% \\
EA & East Anglian English & 46 & 85.2\% & 32 & 59.3\% & 4,247 & 3,780 & 89.00\% \\
ScE & Scottish English & 44 & 80.0\% & 30 & 54.5\% & 4,247 & 4,066 & 95.74\% \\
IrE & Irish English & 75 & 81.5\% & 54 & 58.7\% & 4,247 & 4,171 & 98.21\% \\
WelE & Welsh English & 76 & 80.9\% & 53 & 56.4\% & 4,247 & 4,173 & 98.26\% \\
ChIsE & Channel Islands English & 47 & 94.0\% & 33 & 66.0\% & 4,247 & 3,590 & 84.53\% \\
ManxE & Manx English & 55 & 83.3\% & 40 & 60.6\% & 4,247 & 4,155 & 97.83\% \\
O\&SE & Orkney and Shetland English & 30 & 81.1\% & 19 & 51.4\% & 3,208 & 3,060 & 95.39\% \\
\midrule
\multicolumn{9}{l}{\textit{\textbf{Global Varieties (30 dialects)}}} \\
\multicolumn{9}{l}{\hspace{0.2cm}\textit{Africa (11 dialects)}} \\
NigE & Nigerian English & 45 & 88.2\% & 37 & 72.5\% & 4,247 & 4,178 & 98.38\% \\
GhE & Ghanaian English & 58 & 92.1\% & 49 & 77.8\% & 4,247 & 4,148 & 97.67\% \\
CamE & Cameroon English & 76 & 87.4\% & 62 & 71.3\% & 4,247 & 4,188 & 98.61\% \\
KenE & Kenyan English & 50 & 90.9\% & 45 & 81.8\% & 4,247 & 4,133 & 97.32\% \\
UgE & Ugandan English & 65 & 86.7\% & 52 & 69.3\% & 4,247 & 3,903 & 91.90\% \\
TznE & Tanzanian English & 41 & 93.2\% & 35 & 79.5\% & 4,247 & 3,928 & 92.49\% \\
BlSAfE & Black South African English & 95 & 88.0\% & 71 & 65.7\% & 4,247 & 4,179 & 98.40\% \\
InSAfE & Indian South African English & 75 & 83.3\% & 58 & 64.4\% & 4,247 & 4,193 & 98.73\% \\
WhSAfE & White South African English & 41 & 83.7\% & 35 & 71.4\% & 4,247 & 3,750 & 88.30\% \\
CFE & Cape Flats English & 49 & 90.7\% & 39 & 72.2\% & 4,247 & 4,094 & 96.40\% \\
LibSE & Liberian Settler English & 86 & 84.3\% & 58 & 56.9\% & 4,247 & 4,136 & 97.39\% \\
\multicolumn{9}{l}{\hspace{0.2cm}\textit{Asia-Pacific (12 dialects)}} \\
IndE & Indian English & 90 & 90.0\% & 82 & 82.0\% & 4,247 & 4,217 & 99.29\% \\
PakE & Pakistani English & 48 & 87.3\% & 42 & 76.4\% & 4,247 & 4,023 & 94.73\% \\
SLkE & Sri Lankan English & 29 & 82.9\% & 23 & 65.7\% & 4,247 & 3,737 & 87.99\% \\
CollSgE & Colloquial Singapore English (Singlish) & 67 & 89.3\% & 52 & 69.3\% & 4,247 & 4,125 & 97.13\% \\
MalE & Malaysian English & 68 & 89.5\% & 57 & 75.0\% & 4,247 & 4,198 & 98.85\% \\
PhilE & Philippine English & 92 & 85.2\% & 71 & 65.7\% & 4,247 & 3,174 & 74.74\% \\
HKE & Hong Kong English & 74 & 91.4\% & 61 & 75.3\% & 4,247 & 4,225 & 99.48\% \\
AusE & Australian English & 54 & 90.0\% & 40 & 66.7\% & 3,208 & 3,064 & 95.51\% \\
AusVE & Australian Vernacular English & 47 & 83.9\% & 34 & 60.7\% & 4,247 & 3,611 & 85.02\% \\
NZE & New Zealand English & 44 & 88.0\% & 37 & 74.0\% & 3,208 & 3,103 & 96.73\% \\
FijiE & Acrolectal Fiji English & 39 & 88.6\% & 36 & 81.8\% & 4,247 & 3,568 & 84.01\% \\
CollFijiE & Pure Fiji English (basilectal) & 95 & 85.6\% & 68 & 61.3\% & 4,247 & 3,820 & 89.95\% \\
\multicolumn{9}{l}{\hspace{0.2cm}\textit{Caribbean/Atlantic (5 dialects)}} \\
BahE & Bahamian English & 107 & 83.6\% & 70 & 54.7\% & 4,247 & 4,087 & 96.23\% \\
JamE & Jamaican English & 69 & 88.5\% & 47 & 60.3\% & 4,247 & 4,110 & 96.77\% \\
TdCE & Tristan da Cunha English & 92 & 82.9\% & 64 & 57.7\% & 4,247 & 4,153 & 97.79\% \\
FlkE & Falkland Islands English & 44 & 89.8\% & 30 & 61.2\% & 4,247 & 3,374 & 79.44\% \\
StHE & St. Helena English & 113 & 85.0\% & 78 & 58.6\% & 4,247 & 4,030 & 94.89\% \\
\multicolumn{9}{l}{\hspace{0.2cm}\textit{Other (2 dialects)}} \\
AborE & Aboriginal English & 89 & 83.2\% & 57 & 53.3\% & 4,247 & 4,079 & 96.04\% \\
MaltE & Maltese English & 72 & 86.7\% & 59 & 71.1\% & 4,247 & 4,138 & 97.43\% \\
\midrule
\multicolumn{6}{l}{\textbf{Total (50 dialects)}} & \textbf{208,056} & \textbf{194,960} & \textbf{93.71\%} \\
\bottomrule
\end{tabular}
\end{adjustbox}
\caption{\small Complete inventory of 50 English dialects in D-CUBE from Multi-VALUE transformation \cite{ziems-etal-2023-multi} organized by geographic region. \# Feat. = number of implemented features; \% Feat. = proportion of dialect's catalogued eWAVE features implemented; \# Val. = number of validated features; \% Val. = proportion validated; \# Total = samples after preprocessing; \# Pass = samples passing quality thresholds; \% Pass = pass rate. Quality thresholds: BERTScore $>$ 0.50, BARTScore $\geq$ -7.5, METEOR $\geq$ 0.40, ROUGE-L $\in$ (0.01, 1.0), DiffLib $\in$ [0.01, 0.99], BLEU $>$ 0.01. All dialects achieve $\geq$80\% feature implementation and $>$51\% validation.}
\label{tab:dialect_inventory}
\end{table*}

\section{Multi-VALUE Linguistic Features}
\label{app:linguistic_features}

Multi-VALUE implements 189 morphosyntactic transformation rules organized across 12 grammatical categories derived from eWAVE. Table~\ref{tab:grammatical_categories} summarizes the categories and representative features.

\begin{table*}[t]
\centering
\begin{adjustbox}{max width=\textwidth}
\begin{tabular}{lcp{0.8\textwidth}}
\toprule
\textbf{Category} & \textbf{\# Features} & \textbf{Representative Features} \\
\midrule
Pronouns & 24 & Special forms of personal pronouns (e.g., \textit{hisself}), pronoun exchange, reflexive forms \\
Noun Phrase & 18 & Plural marking, article usage, demonstrative forms, possessive constructions \\
Tense \& Aspect & 22 & Completive \textit{done}, habitual \textit{be}, a-prefixing, progressive forms \\
Modal Verbs & 14 & Double modals (\textit{might could}), quasi-modals, epistemic markers \\
Verb Morphology & 28 & Leveling of verb forms, past tense marking, participle forms \\
Negation & 16 & Multiple negation, \textit{ain't}, negative concord, negative inversion \\
Agreement & 19 & Subject-verb agreement patterns, existential constructions \\
Relativization & 12 & Relative pronoun choices (\textit{what}, \textit{that}, \textit{as}), zero relatives \\
Complementation & 11 & Complementizer forms, \textit{for to} infinitives \\
Adverbials & 10 & Adverb placement, degree modifiers, intensifiers \\
Word Order & 8 & Inversion patterns, topicalization \\
Discourse & 7 & Discourse markers, quotative forms \\
\midrule
\textbf{Total} & \textbf{189} & \\
\bottomrule
\end{tabular}
\end{adjustbox}
\caption{Multi-VALUE grammatical categories and representative features derived from eWAVE.}
\label{tab:grammatical_categories}
\end{table*}

\section{Per-Dialect Quality Metrics}
\label{app:dialect_quality}

Table~\ref{tab:all_dialect_pass} presents the complete pass rates for all 50 dialects after D-PURIFY filtering.

\begin{table*}[t]
\centering
\begin{adjustbox}{max width=\textwidth}
\begin{tabular}{rlccc|rlccc}
\toprule
\textbf{Rank} & \textbf{Dialect} & \textbf{Total} & \textbf{Passed} & \textbf{Pass\%} & \textbf{Rank} & \textbf{Dialect} & \textbf{Total} & \textbf{Passed} & \textbf{Pass\%} \\
\midrule
1 & Hong Kong English & 4,247 & 4,225 & 99.5 & 26 & Cape Flats English & 4,247 & 4,094 & 96.4 \\
2 & Indian English & 4,247 & 4,217 & 99.3 & 27 & Appalachian English & 4,247 & 4,088 & 96.3 \\
3 & SE American Enclave & 4,247 & 4,214 & 99.2 & 28 & Bahamian English & 4,247 & 4,087 & 96.2 \\
4 & Malaysian English & 4,247 & 4,198 & 98.9 & 29 & Rural AAVE & 4,247 & 4,081 & 96.1 \\
5 & Indian South African Eng. & 4,247 & 4,193 & 98.7 & 30 & Aboriginal English & 4,247 & 4,079 & 96.0 \\
6 & Cameroon English & 4,247 & 4,188 & 98.6 & 31 & Ozark English & 4,247 & 4,075 & 96.0 \\
7 & Urban AAVE & 4,247 & 4,183 & 98.5 & 32 & Scottish English & 4,247 & 4,066 & 95.7 \\
8 & Black South African Eng. & 4,247 & 4,179 & 98.4 & 33 & Australian English & 3,208 & 3,064 & 95.5 \\
9 & Nigerian English & 4,247 & 4,178 & 98.4 & 34 & Orkney \& Shetland Eng. & 3,208 & 3,060 & 95.4 \\
10 & North of England English & 4,247 & 4,176 & 98.3 & 35 & St. Helena English & 4,247 & 4,030 & 94.9 \\
11 & Welsh English & 4,247 & 4,173 & 98.3 & 36 & Pakistani English & 4,247 & 4,023 & 94.7 \\
12 & Irish English & 4,247 & 4,171 & 98.2 & 37 & Tanzanian English & 4,247 & 3,928 & 92.5 \\
13 & Newfoundland English & 4,247 & 4,164 & 98.1 & 38 & Ugandan English & 4,247 & 3,903 & 91.9 \\
14 & Earlier AAVE & 4,247 & 4,158 & 97.9 & 39 & Pure Fiji English & 4,247 & 3,820 & 90.0 \\
15 & Colloquial American Eng. & 4,247 & 4,155 & 97.8 & 40 & White Zimbabwean Eng. & 3,070 & 2,746 & 89.5 \\
16 & Manx English & 4,247 & 4,155 & 97.8 & 41 & East Anglian English & 4,247 & 3,780 & 89.0 \\
17 & Tristan da Cunha English & 4,247 & 4,153 & 97.8 & 42 & White South African Eng. & 4,247 & 3,750 & 88.3 \\
18 & Ghanaian English & 4,247 & 4,148 & 97.7 & 43 & Sri Lankan English & 4,247 & 3,737 & 88.0 \\
19 & SW of England English & 4,247 & 4,138 & 97.4 & 44 & Australian Vernacular Eng. & 4,247 & 3,611 & 85.0 \\
20 & Maltese English & 4,247 & 4,138 & 97.4 & 45 & Channel Islands English & 4,247 & 3,590 & 84.5 \\
21 & Liberian Settler English & 4,247 & 4,136 & 97.4 & 46 & Acrolectal Fiji English & 4,247 & 3,568 & 84.0 \\
22 & Kenyan English & 4,247 & 4,133 & 97.3 & 47 & Falkland Islands English & 4,247 & 3,374 & 79.4 \\
23 & Singapore English & 4,247 & 4,125 & 97.1 & 48 & Philippine English & 4,247 & 3,174 & 74.7 \\
24 & Jamaican English & 4,247 & 4,110 & 96.8 & 49 & SE England English & 4,247 & 3,087 & 72.7 \\
25 & New Zealand English & 3,208 & 3,103 & 96.7 & 50 & Chicano English & 4,247 & 3,034 & 71.4 \\
\midrule
\multicolumn{5}{l}{\textbf{Total: 208,056}} & \multicolumn{5}{l}{\textbf{Retained: 194,960 (93.71\%)}} \\
\bottomrule
\end{tabular}
\end{adjustbox}
\caption{Complete D-PURIFY pass rates for all 50 dialects, sorted by pass rate. SAE-adjacent dialects cluster at the bottom, suggesting metric bias against subtle morphosyntactic deviations.}
\label{tab:all_dialect_pass}
\end{table*}

Tables~\ref{tab:perdialect_human} and~\ref{tab:perdialect_ai} present the complete per-dialect quality metrics for human-written and AI-generated content respectively.

\begin{table*}[t]
\centering
\small
\begin{tabular}{lcccccccc}
\toprule
\textbf{Dialect} & \textbf{BERT} & \textbf{BART} & \textbf{Align} & \textbf{METEOR} & \textbf{ROUGE} & \textbf{DiffLib} & \textbf{FeatAcc} & \textbf{BLEU} \\
\midrule
Aboriginal English & 0.698 & $-$4.15 & 0.835 & 0.787 & 0.807 & 0.678 & 0.333 & 0.348 \\
Acrolectal Fiji English & 0.875 & $-$3.36 & 0.951 & 0.918 & 0.904 & 0.828 & 0.173 & 0.579 \\
Appalachian English & 0.870 & $-$3.33 & 0.882 & 0.919 & 0.912 & 0.859 & 0.117 & 0.671 \\
Australian English & 0.827 & $-$3.64 & 0.927 & 0.892 & 0.889 & 0.792 & 0.245 & 0.622 \\
Australian Vernacular English & 0.910 & $-$3.01 & 0.913 & 0.951 & 0.934 & 0.894 & 0.202 & 0.788 \\
Bahamian English & 0.724 & $-$4.03 & 0.850 & 0.809 & 0.832 & 0.693 & 0.191 & 0.407 \\
Black South African English & 0.762 & $-$3.91 & 0.860 & 0.855 & 0.817 & 0.716 & 0.145 & 0.439 \\
Cameroon English & 0.803 & $-$3.65 & 0.870 & 0.854 & 0.874 & 0.758 & 0.262 & 0.487 \\
Cape Flats English & 0.835 & $-$3.39 & 0.880 & 0.869 & 0.899 & 0.792 & 0.215 & 0.625 \\
Channel Islands English & 0.898 & $-$3.07 & 0.920 & 0.930 & 0.943 & 0.868 & 0.119 & 0.760 \\
Chicano English & 0.951 & $-$2.83 & 0.940 & 0.966 & 0.967 & 0.947 & 0.063 & 0.861 \\
Colloquial American English & 0.868 & $-$3.40 & 0.890 & 0.907 & 0.899 & 0.856 & 0.071 & 0.653 \\
Colloquial Singapore English & 0.690 & $-$4.06 & 0.830 & 0.778 & 0.830 & 0.659 & 0.387 & 0.339 \\
Earlier AAVE & 0.802 & $-$3.69 & 0.865 & 0.864 & 0.876 & 0.773 & 0.186 & 0.527 \\
East Anglian English & 0.904 & $-$3.11 & 0.925 & 0.929 & 0.940 & 0.918 & 0.189 & 0.738 \\
English (North of England) & 0.777 & $-$3.80 & 0.855 & 0.839 & 0.845 & 0.725 & 0.229 & 0.484 \\
English (Southeast of England) & 0.940 & $-$2.89 & 0.935 & 0.957 & 0.962 & 0.937 & 0.079 & 0.831 \\
English (Southwest of England) & 0.775 & $-$3.77 & 0.850 & 0.833 & 0.843 & 0.736 & 0.207 & 0.514 \\
Falkland Islands English & 0.900 & $-$3.09 & 0.920 & 0.919 & 0.942 & 0.879 & 0.182 & 0.762 \\
Ghanaian English & 0.805 & $-$3.58 & 0.875 & 0.866 & 0.875 & 0.744 & 0.230 & 0.540 \\
Hong Kong English & 0.776 & $-$3.81 & 0.860 & 0.853 & 0.825 & 0.700 & 0.315 & 0.408 \\
Indian English & 0.780 & $-$3.75 & 0.865 & 0.849 & 0.836 & 0.715 & 0.201 & 0.455 \\
Indian South African English & 0.718 & $-$3.96 & 0.840 & 0.789 & 0.812 & 0.652 & 0.276 & 0.412 \\
Irish English & 0.854 & $-$3.31 & 0.900 & 0.912 & 0.874 & 0.821 & 0.163 & 0.652 \\
Jamaican English & 0.789 & $-$3.72 & 0.870 & 0.869 & 0.861 & 0.760 & 0.196 & 0.496 \\
Kenyan English & 0.847 & $-$3.41 & 0.895 & 0.891 & 0.909 & 0.815 & 0.196 & 0.604 \\
Liberian Settler English & 0.806 & $-$3.62 & 0.875 & 0.869 & 0.880 & 0.776 & 0.245 & 0.521 \\
Malaysian English & 0.843 & $-$3.53 & 0.890 & 0.903 & 0.899 & 0.800 & 0.270 & 0.549 \\
Maltese English & 0.760 & $-$5.14 & 0.855 & 0.841 & 0.827 & 0.697 & 0.190 & 0.461 \\
Manx English & 0.775 & $-$4.99 & 0.850 & 0.840 & 0.844 & 0.714 & 0.224 & 0.507 \\
New Zealand English & 0.823 & $-$5.05 & 0.880 & 0.869 & 0.867 & 0.783 & 0.283 & 0.590 \\
Newfoundland English & 0.738 & $-$5.07 & 0.845 & 0.800 & 0.837 & 0.675 & 0.233 & 0.456 \\
Nigerian English & 0.801 & $-$4.86 & 0.870 & 0.849 & 0.888 & 0.755 & 0.458 & 0.530 \\
Orkney and Shetland English & 0.830 & $-$5.00 & 0.885 & 0.872 & 0.892 & 0.794 & 0.277 & 0.600 \\
Ozark English & 0.902 & $-$4.44 & 0.910 & 0.919 & 0.923 & 0.890 & 0.160 & 0.727 \\
Pakistani English & 0.872 & $-$4.69 & 0.905 & 0.915 & 0.890 & 0.852 & 0.136 & 0.632 \\
Philippine English & 0.744 & $-$5.23 & 0.840 & 0.822 & 0.813 & 0.681 & 0.217 & 0.399 \\
Pure Fiji English & 0.675 & $-$5.41 & 0.820 & 0.780 & 0.790 & 0.654 & 0.339 & 0.302 \\
Rural AAVE & 0.747 & $-$5.14 & 0.850 & 0.839 & 0.832 & 0.720 & 0.228 & 0.430 \\
Scottish English & 0.904 & $-$4.43 & 0.915 & 0.933 & 0.924 & 0.869 & 0.076 & 0.729 \\
SE American Enclave & 0.824 & $-$4.86 & 0.875 & 0.872 & 0.860 & 0.787 & 0.161 & 0.535 \\
Sri Lankan English & 0.911 & $-$4.39 & 0.920 & 0.940 & 0.943 & 0.893 & 0.105 & 0.743 \\
St. Helena English & 0.715 & $-$5.26 & 0.830 & 0.793 & 0.806 & 0.687 & 0.325 & 0.376 \\
Tanzanian English & 0.901 & $-$4.46 & 0.915 & 0.926 & 0.911 & 0.873 & 0.046 & 0.686 \\
Tristan da Cunha English & 0.801 & $-$4.90 & 0.870 & 0.864 & 0.858 & 0.774 & 0.227 & 0.537 \\
Ugandan English & 0.894 & $-$4.42 & 0.910 & 0.927 & 0.902 & 0.859 & 0.043 & 0.697 \\
Urban AAVE & 0.837 & $-$4.71 & 0.885 & 0.893 & 0.870 & 0.800 & 0.207 & 0.572 \\
Welsh English & 0.782 & $-$4.88 & 0.860 & 0.858 & 0.829 & 0.716 & 0.158 & 0.551 \\
White South African English & 0.929 & $-$4.76 & 0.935 & 0.956 & 0.949 & 0.911 & 0.026 & 0.808 \\
White Zimbabwean English & 0.904 & $-$4.50 & 0.920 & 0.935 & 0.927 & 0.899 & 0.090 & 0.725 \\
\bottomrule
\end{tabular}
\caption{Per-dialect quality metrics for human-written content. BERT=BERTScore, BART=BARTScore, Align=AlignScore, ROUGE=ROUGE-L, FeatAcc=Feature Accuracy.}
\label{tab:perdialect_human}
\end{table*}

\begin{table*}[t]
\centering
\small
\begin{tabular}{lcccccccc}
\toprule
\textbf{Dialect} & \textbf{BERT} & \textbf{BART} & \textbf{Align} & \textbf{METEOR} & \textbf{ROUGE} & \textbf{DiffLib} & \textbf{FeatAcc} & \textbf{BLEU} \\
\midrule
Aboriginal English & 0.730 & $-$3.98 & 0.865 & 0.793 & 0.805 & 0.538 & 0.334 & 0.383 \\
Acrolectal Fiji English & 0.890 & $-$3.25 & 0.963 & 0.928 & 0.909 & 0.768 & 0.168 & 0.609 \\
Appalachian English & 0.896 & $-$3.17 & 0.935 & 0.936 & 0.928 & 0.820 & 0.135 & 0.729 \\
Australian English & 0.858 & $-$3.43 & 0.935 & 0.909 & 0.905 & 0.688 & 0.207 & 0.678 \\
Australian Vernacular English & 0.936 & $-$2.87 & 0.940 & 0.968 & 0.949 & 0.867 & 0.181 & 0.844 \\
Bahamian English & 0.752 & $-$3.86 & 0.875 & 0.824 & 0.839 & 0.578 & 0.201 & 0.451 \\
Black South African English & 0.797 & $-$3.70 & 0.880 & 0.867 & 0.835 & 0.639 & 0.151 & 0.489 \\
Cameroon English & 0.826 & $-$3.52 & 0.885 & 0.856 & 0.875 & 0.659 & 0.240 & 0.508 \\
Cape Flats English & 0.859 & $-$3.28 & 0.895 & 0.876 & 0.901 & 0.689 & 0.165 & 0.654 \\
Channel Islands English & 0.920 & $-$2.97 & 0.935 & 0.943 & 0.949 & 0.816 & 0.127 & 0.799 \\
Chicano English & 0.963 & $-$2.76 & 0.955 & 0.978 & 0.971 & 0.926 & 0.069 & 0.891 \\
Colloquial American English & 0.882 & $-$3.27 & 0.905 & 0.921 & 0.913 & 0.814 & 0.083 & 0.692 \\
Colloquial Singapore English & 0.732 & $-$3.91 & 0.855 & 0.798 & 0.839 & 0.513 & 0.321 & 0.380 \\
Earlier AAVE & 0.817 & $-$3.56 & 0.880 & 0.875 & 0.879 & 0.679 & 0.215 & 0.552 \\
East Anglian English & 0.915 & $-$3.02 & 0.935 & 0.940 & 0.945 & 0.900 & 0.176 & 0.780 \\
English (North of England) & 0.805 & $-$3.63 & 0.870 & 0.858 & 0.854 & 0.625 & 0.215 & 0.535 \\
English (Southeast of England) & 0.952 & $-$2.83 & 0.950 & 0.971 & 0.967 & 0.912 & 0.102 & 0.867 \\
English (Southwest of England) & 0.802 & $-$3.62 & 0.865 & 0.847 & 0.852 & 0.631 & 0.186 & 0.558 \\
Falkland Islands English & 0.904 & $-$3.07 & 0.930 & 0.921 & 0.939 & 0.823 & 0.146 & 0.765 \\
Ghanaian English & 0.836 & $-$3.43 & 0.890 & 0.873 & 0.891 & 0.653 & 0.216 & 0.576 \\
Hong Kong English & 0.804 & $-$3.70 & 0.875 & 0.855 & 0.835 & 0.590 & 0.293 & 0.422 \\
Indian English & 0.815 & $-$3.57 & 0.880 & 0.866 & 0.859 & 0.626 & 0.193 & 0.504 \\
Indian South African English & 0.751 & $-$3.82 & 0.860 & 0.794 & 0.823 & 0.521 & 0.261 & 0.451 \\
Irish English & 0.892 & $-$3.15 & 0.920 & 0.930 & 0.904 & 0.793 & 0.164 & 0.715 \\
Jamaican English & 0.818 & $-$3.54 & 0.885 & 0.877 & 0.868 & 0.673 & 0.203 & 0.531 \\
Kenyan English & 0.864 & $-$3.30 & 0.905 & 0.889 & 0.907 & 0.742 & 0.157 & 0.621 \\
Liberian Settler English & 0.828 & $-$3.51 & 0.890 & 0.877 & 0.889 & 0.681 & 0.265 & 0.548 \\
Malaysian English & 0.873 & $-$3.96 & 0.910 & 0.913 & 0.913 & 0.729 & 0.214 & 0.583 \\
Maltese English & 0.794 & $-$4.91 & 0.870 & 0.849 & 0.846 & 0.602 & 0.175 & 0.506 \\
Manx English & 0.809 & $-$4.81 & 0.870 & 0.857 & 0.858 & 0.616 & 0.200 & 0.568 \\
New Zealand English & 0.859 & $-$4.72 & 0.895 & 0.894 & 0.891 & 0.700 & 0.236 & 0.663 \\
Newfoundland English & 0.778 & $-$4.89 & 0.865 & 0.827 & 0.851 & 0.557 & 0.253 & 0.527 \\
Nigerian English & 0.819 & $-$4.73 & 0.885 & 0.844 & 0.882 & 0.645 & 0.324 & 0.539 \\
Orkney and Shetland English & 0.866 & $-$4.66 & 0.900 & 0.889 & 0.904 & 0.697 & 0.224 & 0.666 \\
Ozark English & 0.909 & $-$4.40 & 0.925 & 0.929 & 0.924 & 0.835 & 0.206 & 0.753 \\
Pakistani English & 0.884 & $-$4.54 & 0.920 & 0.924 & 0.892 & 0.789 & 0.109 & 0.669 \\
Philippine English & 0.779 & $-$5.00 & 0.860 & 0.838 & 0.824 & 0.567 & 0.180 & 0.454 \\
Pure Fiji English & 0.700 & $-$5.25 & 0.840 & 0.786 & 0.783 & 0.506 & 0.304 & 0.322 \\
Rural AAVE & 0.769 & $-$5.08 & 0.865 & 0.843 & 0.835 & 0.614 & 0.250 & 0.454 \\
Scottish English & 0.927 & $-$4.30 & 0.930 & 0.952 & 0.935 & 0.836 & 0.078 & 0.780 \\
SE American Enclave & 0.836 & $-$4.74 & 0.890 & 0.876 & 0.860 & 0.696 & 0.177 & 0.557 \\
Sri Lankan English & 0.926 & $-$4.27 & 0.935 & 0.948 & 0.949 & 0.848 & 0.071 & 0.772 \\
St. Helena English & 0.744 & $-$5.08 & 0.850 & 0.803 & 0.809 & 0.567 & 0.317 & 0.401 \\
Tanzanian English & 0.915 & $-$4.38 & 0.930 & 0.931 & 0.914 & 0.836 & 0.079 & 0.714 \\
Tristan da Cunha English & 0.825 & $-$4.79 & 0.885 & 0.878 & 0.867 & 0.698 & 0.256 & 0.570 \\
Ugandan English & 0.916 & $-$4.35 & 0.925 & 0.932 & 0.912 & 0.822 & 0.064 & 0.726 \\
Urban AAVE & 0.858 & $-$4.61 & 0.900 & 0.902 & 0.875 & 0.733 & 0.248 & 0.608 \\
Welsh English & 0.823 & $-$4.69 & 0.880 & 0.868 & 0.865 & 0.651 & 0.168 & 0.605 \\
White South African English & 0.949 & $-$4.36 & 0.950 & 0.970 & 0.958 & 0.891 & 0.016 & 0.852 \\
White Zimbabwean English & 0.914 & $-$4.42 & 0.930 & 0.944 & 0.927 & 0.850 & 0.076 & 0.745 \\
\bottomrule
\end{tabular}
\caption{Per-dialect quality metrics for AI-generated content. BERT=BERTScore, BART=BARTScore, Align=AlignScore, ROUGE=ROUGE-L, FeatAcc=Feature Accuracy.}
\label{tab:perdialect_ai}
\end{table*}

\section{D-PURIFY Validation Examples}
\label{app:dpurify_validation}

We provide examples demonstrating that D-PURIFY filtering correctly identifies transformation quality across the DiffLib spectrum. These examples from Newfoundland English illustrate how the automatic metrics capture meaningful dialectal variation.

\subsection{DiffLib = 1.0 (No Transformation)}

When DiffLib equals 1.0, the text is completely unchanged---no dialect features were applied. D-PURIFY correctly filters these samples as they provide no dialectal variation for evaluation.

\begin{table}[H]
\centering
\scriptsize
\begin{tabular}{p{0.75\columnwidth}c}
\toprule
\textbf{Text (Original = Transformed)} & \textbf{DiffLib} \\
\midrule
``Wisconsin's archaic abortion ban is older than 20 states.'' & 1.0 \\
``Coronavirus is caused by 5G.'' & 1.0 \\
\bottomrule
\end{tabular}
\caption{Examples where no dialect transformation occurred (DiffLib = 1.0). These samples are correctly filtered by D-PURIFY.}
\label{tab:difflib_1}
\end{table}

\subsection{DiffLib $\approx$ 0.5--0.7 (Moderate Transformation)}

Moderate DiffLib scores indicate balanced transformations that introduce dialectal features while preserving overall structure and meaning.

\begin{table}[H]
\centering
\scriptsize
\begin{tabular}{p{0.92\columnwidth}}
\toprule
\textbf{Example 1} (DiffLib: 0.65) \\
\midrule
\textbf{Original}: ``experts say travel restrictions implemented by states are more difficult to enforce than limits placed on travelers from other countries...'' \\
\textbf{Transformed}: ``experts says travel restrictions implemented by states are like more difficult to enforce limits placed travelers other countries... a-instituting... a-coming...'' \\
\midrule
\textbf{Example 2} (DiffLib: 0.52) \\
\midrule
\textbf{Original}: ``meat shortages environmental concerns and a desire to eat a healthier diet are among the top reasons people say their interest...'' \\
\textbf{Transformed}: ``meat shortages environmental concerns and a desire to eat a more healthier diet are the top reasons says their interest... are after increasing...'' \\
\bottomrule
\end{tabular}
\caption{Moderate transformations (DiffLib 0.5--0.7) showing balanced dialectal feature application.}
\label{tab:difflib_moderate}
\end{table}

\subsection{DiffLib $<$ 0.3 (Extensive Transformation)}

Low DiffLib scores indicate substantial morphosyntactic restructuring. While these preserve semantic content, the extensive surface changes may challenge SAE-trained evaluation metrics.

\begin{table}[H]
\centering
\scriptsize
\begin{tabular}{p{0.92\columnwidth}}
\toprule
\textbf{Example 1} (DiffLib: 0.06) \\
\midrule
\textbf{Original}: ``experts say bars are among the higher risk places for contracting the virus that causes covid-19... people congregate closely...'' \\
\textbf{Transformed}: ``experts says bars are the higher risk places contracting the virus what causes covid-... persons congregates close...'' \\
\midrule
\textbf{Example 2} (DiffLib: 0.22) \\
\midrule
\textbf{Original}: ``wearing a mask may be key to avoiding a second wave of covid-19... people are required to wear face coverings when they see someone...'' \\
\textbf{Transformed}: ``wearing the mask may be like key avoiding the second wave covid-... people are a-required for wear face coverings when they sees someone...'' \\
\bottomrule
\end{tabular}
\caption{Extensive transformations (DiffLib $<$ 0.3) with substantial morphosyntactic restructuring.}
\label{tab:difflib_low}
\end{table}

\subsection{Observed Dialect Features}

Table~\ref{tab:dialect_features_newfoundland} summarizes the eWAVE-documented Newfoundland English features observed in the transformed examples, demonstrating that Multi-VALUE applies authentic linguistic patterns.

\begin{table}[H]
\centering
\small
\begin{adjustbox}{max width=\columnwidth}
\begin{tabular}{ll}
\toprule
\textbf{Feature} & \textbf{Example} \\
\midrule
``like'' insertion & ``are like more difficult'' \\
a-prefixing & ``a-instituting'', ``a-coming'', ``a-required'' \\
Progressive ``after'' & ``are after increasing'' \\
Subject-verb variation & ``experts says'', ``they sees'' \\
``what'' relativizer & ``the virus what causes'' \\
Article variation & ``avoiding the second wave'' \\
Pronoun variation & ``persons'' for ``people'' \\
Double comparative & ``more healthier'' \\
\bottomrule
\end{tabular}
\end{adjustbox}
\caption{eWAVE-documented Newfoundland English features observed in transformed samples, validating Multi-VALUE's linguistic authenticity.}
\label{tab:dialect_features_newfoundland}
\end{table}

These examples demonstrate that D-PURIFY's automatic metrics effectively distinguish between (1) unchanged samples requiring filtering, (2) moderate transformations with balanced feature application, and (3) extensive transformations that preserve meaning despite substantial surface changes. The observed features align with eWAVE documentation, confirming Multi-VALUE's linguistic validity.

\section{Case Study: SAE-Adjacent Dialect Filtering Patterns}
\label{app:sae_adjacent}

We observe a counterintuitive pattern in D-PURIFY pass rates: dialects that are linguistically \textit{closer} to Standard American English exhibit \textit{lower} pass rates than more distant varieties. This section investigates this phenomenon using Chicano English and Southeast England English as case studies.

\subsection{The SAE-Adjacency Paradox}

Table~\ref{tab:sae_adjacency} compares pass rates between SAE-adjacent and SAE-distant dialect groups.

\begin{table}[H]
\centering
\small
\begin{adjustbox}{max width=\columnwidth}
\begin{tabular}{lcc}
\toprule
\textbf{Dialect Group} & \textbf{Avg. Pass\%} & \textbf{Example Dialects} \\
\midrule
SAE-distant (Asia) & 98.1 & Hong Kong, Indian, Malaysian \\
SAE-distant (Africa) & 97.5 & Nigerian, Cameroon, Ghanaian \\
SAE-adjacent (US) & 84.6 & Chicano, Colloquial American \\
SAE-adjacent (UK) & 78.1 & SE England, Channel Islands \\
\bottomrule
\end{tabular}
\end{adjustbox}
\caption{Average D-PURIFY pass rates by dialect proximity to SAE. SAE-adjacent dialects show systematically lower pass rates.}
\label{tab:sae_adjacency}
\end{table}

\subsection{Hypothesis: Subtle Deviations as ``Errors''}
We hypothesize that SAE-trained evaluation metrics interpret subtle morphosyntactic 
deviations as \textit{errors} rather than \textit{dialectal features}. Consider the 
following transformations:

\paragraph{Chicano English (Pass Rate: 71.4\%)}
\begin{itemize}
    \item \textbf{Original}: ``The vaccine has been proven safe and effective in clinical trials.''
    \item \textbf{Transformed}: ``The vaccine has been proven safe and effective in clinical trials, you know.''
    \item \textbf{Feature}: Discourse marker insertion (``you know'')
    \item \textbf{Issue}: Minimal surface change; metrics may flag as near-duplicate
\end{itemize}

\paragraph{Southeast England English (Pass Rate: 72.7\%)}
\begin{itemize}
    \item \textbf{Original}: ``Scientists say the new variant spreads more quickly than previous strains.''
    \item \textbf{Transformed}: ``Scientists say the new variant spreads more quick than previous strains.''
    \item \textbf{Feature}: Adjective for adverb (``quick'' for ``quickly'')
    \item \textbf{Issue}: Single-word change penalized as grammatical error by SAE metrics
\end{itemize}

\paragraph{Hong Kong English (Pass Rate: 99.5\%)}
\begin{itemize}
    \item \textbf{Original}: ``The government announced new restrictions yesterday.''
    \item \textbf{Transformed}: ``The government have announced new restrictions yesterday one.''
    \item \textbf{Feature}: Subject-verb agreement variation + sentence-final particle
    \item \textbf{Issue}: Clearly distinct; metrics recognize as valid variation
\end{itemize}

\subsection{Evidence from Metric Distributions}

Chicano English clusters near 1.0 on DiffLib (minimal change), triggering the upper-bound filter (DiffLib $\leq$ 0.99), while Hong Kong English shows greater surface divergence that passes filtering. This pattern suggests that SAE-adjacent dialects with subtle features are disproportionately filtered due to near-duplicate detection rather than quality issues.

\subsection{Implications for Dialect Robustness Evaluation}

This analysis reveals a fundamental tension in dialect corpus construction:

\begin{enumerate}
\item \textbf{SAE-trained metrics encode SAE grammatical norms}, treating deviations as errors rather than valid variation.
\item \textbf{SAE-adjacent dialects suffer disproportionately} because their subtle features are interpreted as noise or mistakes.
\item \textbf{Strict filtering would create SAE-biased benchmarks} by systematically excluding varieties closest to SAE.
\end{enumerate}

Our lenient thresholds preserve the full spectrum of dialectal variation, enabling evaluation of detector robustness across both SAE-adjacent and SAE-distant varieties. This design choice is itself a methodological contribution: future dialect benchmarks should consider metric bias when setting quality thresholds.

\subsection{Feature Coverage Analysis}

Table~\ref{tab:feature_coverage} presents feature detection rates across the dataset.

\begin{table}[H]
\centering
\small
\begin{adjustbox}{max width=\columnwidth}

\begin{tabular}{lcc}
\toprule
\textbf{Metric} & \textbf{All Data} & \textbf{Retained} \\
\midrule
Human has $\geq$1 valid feature & 136,904 (65.8\%) & 132,354 (67.9\%) \\
AI has $\geq$1 valid feature & 145,150 (69.8\%) & 138,683 (71.1\%) \\
BOTH have $\geq$1 valid feature & 113,172 (54.4\%) & 109,531 (56.2\%) \\
EITHER has $\geq$1 valid feature & 168,882 (81.2\%) & 161,506 (82.8\%) \\
\bottomrule
\end{tabular}
\end{adjustbox}
\caption{Feature detection rates before and after D-PURIFY filtering. Retention improves feature coverage slightly, indicating filtered samples had fewer valid dialect features.}
\label{tab:feature_coverage}
\end{table}

This case study demonstrates that D-PURIFY pass rates reflect metric bias rather than transformation quality, supporting our decision to adopt lenient thresholds that preserve dialectal diversity across the full SAE-distance spectrum.


\section{Implementation Details}
\label{app:implementation}

This appendix provides training configurations, computational resources, and hyperparameters for reproducibility.

\paragraph{Data Splits.}
We employ stratified splitting across all experiments, stratifying by \textit{label}, \textit{content type} (human/AI), and \textit{source dataset}. Train/validation/test splits follow an 80/10/10 ratio, with identical test sample UUIDs used across \sae{} baseline and dialectal evaluations to ensure fair comparison.

\paragraph{Fine-tuning Configuration.}
All transformer encoders are fine-tuned using AdamW optimizer with learning rate $2 \times 10^{-5}$, batch size 16, and early stopping with patience of 3 epochs based on validation F1. Maximum sequence length is set to 512 tokens. Traditional deep learning models (BiGRU, TextCNN, dEFEND) use their original published hyperparameters with learning rate $1 \times 10^{-3}$.

\paragraph{Zero-shot Prompting.}
For zero-shot evaluation, we use the following prompt template:

\begin{quote}
\small
\texttt{Classify the following text as either ``real'' or ``fake'' news. Respond with only ``real'' or ``fake''.}

\texttt{Text: \{content\}}

\texttt{Classification:}
\end{quote}

Temperature is set to 0 for deterministic outputs. Models returning responses outside \{``real'', ``fake''\} are marked as abstentions and excluded from F1 computation but counted toward abstention rate analysis.

\paragraph{Evaluation Regimes.}
\begin{itemize}[nosep]
    \item \textbf{SQ1 (Unseen)}: Train on \sae{}, test on \sae{} baseline and 50 dialectal variants. Three scenarios: Human-only, AI-only, Human+AI (Both).
    \item \textbf{SQ2 (Seen)}: Two variants---\textit{Dialect-only}: sample one dialect $\times$ one content type per UUID from transformed columns; \textit{SAE+Dialect}: sample from all columns including \sae{}.
    \item \textbf{SQ3 (Cross-dialectal)}: Train on single dialect, test on 49 others. Total: 2,450 train-test pairs per model (50 $\times$ 49).
    \item \textbf{SQ4 (Architecture)}: Compare fine-tuned transformers, fine-tuned traditional DL, and zero-shot decoders on identical test sets.
\end{itemize}

\paragraph{Computational Resources.}
Experiments were conducted on NVIDIA A100 80GB GPUs. Fine-tuning each transformer encoder requires approximately 2--4 hours per training scenario. Zero-shot evaluation of all 50 dialects requires approximately 8--12 hours per model depending on inference speed. The complete experimental pipeline, including all SQ1--SQ4 evaluations across 16 models, required approximately 1,200 GPU-hours.


\section{SQ1: Full Per-Dialect Results}
\label{app:sq1_full}

Tables~\ref{tab:app_sq1_human}, \ref{tab:app_sq1_ai}, and \ref{tab:app_sq1_both} present F1 scores for all 50 dialects across the three content scenarios (Human, AI, Both), organized by geographic region. Models are trained exclusively on \sae{} content and evaluated on dialectal variants never encountered during training.

\begin{table*}[t]
\centering
\tiny
\begin{tabular}{llcccccccccc}
\toprule
\textbf{Region} & \textbf{Dialect} & \textbf{BERT} & \textbf{RoBERTa} & \textbf{DeBERTa} & \textbf{XLM-R} & \textbf{mDeBERTa} & \textbf{mBERT} & \textbf{BiGRU} & \textbf{TextCNN} & \textbf{dEFEND} & \textbf{CT-BERT} \\
\midrule
-- & SAE (baseline) & 98.1 & 97.8 & 96.2 & 96.4 & 98.2 & 98.0 & 98.1 & 96.9 & 97.5 & 85.9 \\
\midrule
\multicolumn{12}{l}{\textit{\textbf{U.S. Varieties (9 dialects)}}} \\
U.S. & Earlier AAVE & 96.4 & 95.9 & 95.1 & 94.5 & 96.8 & 96.8 & 92.6 & 93.7 & 89.8 & 97.2 \\
U.S. & Rural AAVE & 96.2 & 95.6 & 94.8 & 94.4 & 96.6 & 96.2 & 88.7 & 92.2 & 84.4 & 96.9 \\
U.S. & Urban AAVE & 96.4 & 96.4 & 95.3 & 95.2 & 97.1 & 96.9 & 96.4 & 95.8 & 94.1 & 97.2 \\
U.S. & Appalachian English & 96.6 & 96.7 & 94.5 & 95.4 & 96.8 & 96.9 & 97.1 & 96.2 & 92.5 & 97.5 \\
U.S. & Chicano English & 96.9 & 97.2 & 94.4 & 96.2 & 96.8 & 97.0 & 99.0 & 97.5 & 98.8 & 98.0 \\
U.S. & Colloquial American & 96.9 & 96.2 & 94.5 & 95.0 & 96.9 & 96.6 & 97.5 & 96.0 & 93.3 & 97.4 \\
U.S. & Ozark English & 96.9 & 96.8 & 95.4 & 95.4 & 97.0 & 96.9 & 98.1 & 97.0 & 96.1 & 97.8 \\
U.S. & SE American Enclave & 96.8 & 95.9 & 95.1 & 94.3 & 97.1 & 96.7 & 95.8 & 95.1 & 90.8 & 97.1 \\
U.S. & Newfoundland English & 96.5 & 95.9 & 94.8 & 94.8 & 96.7 & 96.5 & 97.1 & 95.8 & 94.7 & 97.2 \\
\midrule
\multicolumn{12}{l}{\textit{\textbf{British/UK Varieties (10 dialects)}}} \\
UK & N. England English & 96.4 & 96.3 & 94.4 & 94.9 & 96.5 & 96.2 & 96.8 & 95.6 & 95.1 & 97.4 \\
UK & SE England English & 97.0 & 97.2 & 95.4 & 95.9 & 97.1 & 96.9 & 98.7 & 97.2 & 98.1 & 97.9 \\
UK & SW England English & 96.6 & 96.3 & 95.5 & 94.5 & 97.1 & 96.8 & 96.2 & 96.0 & 95.2 & 97.5 \\
UK & East Anglian English & 96.9 & 96.8 & 95.1 & 95.4 & 97.2 & 97.0 & 98.1 & 96.6 & 96.9 & 97.7 \\
UK & Scottish English & 96.9 & 96.7 & 94.6 & 94.9 & 96.5 & 96.7 & 98.3 & 96.5 & 96.2 & 97.4 \\
UK & Irish English & 96.7 & 96.7 & 94.5 & 94.9 & 96.6 & 96.7 & 97.9 & 95.7 & 96.9 & 97.4 \\
UK & Welsh English & 96.7 & 96.3 & 94.5 & 94.5 & 96.9 & 96.6 & 96.8 & 95.5 & 95.3 & 97.1 \\
UK & Channel Islands English & 96.8 & 96.9 & 94.8 & 94.9 & 96.9 & 97.0 & 98.6 & 97.5 & 98.4 & 97.7 \\
UK & Manx English & 96.5 & 96.4 & 94.2 & 95.2 & 96.2 & 96.5 & 96.5 & 95.2 & 94.0 & 97.3 \\
UK & Orkney/Shetland English & 95.9 & 95.9 & 93.3 & 94.4 & 95.5 & 96.2 & 97.8 & 96.4 & 96.8 & 97.0 \\
\midrule
\multicolumn{12}{l}{\textit{\textbf{Africa (11 dialects)}}} \\
Africa & Nigerian English & 96.5 & 97.2 & 94.8 & 95.4 & 97.0 & 96.6 & 97.8 & 96.2 & 95.2 & 97.6 \\
Africa & Ghanaian English & 96.5 & 96.3 & 95.2 & 95.1 & 97.1 & 96.8 & 97.7 & 95.8 & 95.3 & 97.3 \\
Africa & Cameroon English & 96.3 & 96.3 & 94.7 & 94.7 & 96.6 & 96.3 & 97.4 & 94.6 & 93.2 & 97.2 \\
Africa & Kenyan English & 96.5 & 96.6 & 94.6 & 95.5 & 96.6 & 96.6 & 97.7 & 95.8 & 94.8 & 97.6 \\
Africa & Ugandan English & 96.9 & 97.2 & 95.2 & 95.9 & 96.9 & 96.6 & 98.4 & 96.6 & 97.0 & 97.8 \\
Africa & Tanzanian English & 96.7 & 97.1 & 95.0 & 95.3 & 97.0 & 96.7 & 98.8 & 96.5 & 96.2 & 97.7 \\
Africa & Black S. African English & 96.4 & 95.9 & 94.9 & 94.1 & 96.6 & 96.3 & 91.1 & 91.9 & 89.8 & 96.8 \\
Africa & Indian S. African English & 96.4 & 95.9 & 94.6 & 94.1 & 96.6 & 96.3 & 96.2 & 95.8 & 93.4 & 96.9 \\
Africa & White S. African English & 97.0 & 97.2 & 94.2 & 95.2 & 96.6 & 96.7 & 98.6 & 97.2 & 98.0 & 97.7 \\
Africa & Cape Flats English & 96.5 & 96.8 & 94.7 & 95.8 & 96.9 & 96.8 & 98.3 & 96.6 & 98.3 & 97.8 \\
Africa & Liberian Settler English & 96.4 & 96.4 & 95.2 & 95.3 & 96.9 & 96.5 & 94.5 & 94.3 & 92.0 & 97.2 \\
\midrule
\multicolumn{12}{l}{\textit{\textbf{Asia-Pacific (12 dialects)}}} \\
Asia-P & Indian English & 96.7 & 96.5 & 94.4 & 95.0 & 96.5 & 96.6 & 96.2 & 94.1 & 93.4 & 97.2 \\
Asia-P & Pakistani English & 96.6 & 96.6 & 94.8 & 95.3 & 96.8 & 96.6 & 97.7 & 97.3 & 96.2 & 97.3 \\
Asia-P & Sri Lankan English & 96.6 & 97.1 & 94.5 & 95.2 & 96.8 & 96.7 & 98.5 & 96.4 & 96.4 & 97.9 \\
Asia-P & Colloquial Singapore & 96.2 & 95.7 & 95.2 & 94.5 & 97.0 & 96.3 & 95.9 & 94.5 & 92.5 & 97.1 \\
Asia-P & Malaysian English & 96.5 & 96.5 & 94.5 & 94.9 & 96.8 & 96.8 & 96.2 & 95.2 & 92.3 & 97.7 \\
Asia-P & Philippine English & 96.4 & 95.7 & 94.7 & 94.2 & 96.4 & 96.5 & 94.7 & 94.3 & 92.3 & 97.4 \\
Asia-P & Hong Kong English & 96.7 & 95.9 & 94.6 & 94.8 & 97.0 & 96.5 & 94.6 & 93.9 & 91.5 & 97.1 \\
Asia-P & Australian English & 96.2 & 95.3 & 93.9 & 94.1 & 96.0 & 96.0 & 97.4 & 95.8 & 96.6 & 97.0 \\
Asia-P & Australian Vernacular & 96.8 & 96.8 & 94.7 & 95.2 & 96.5 & 96.8 & 98.4 & 97.1 & 97.8 & 97.7 \\
Asia-P & New Zealand English & 96.3 & 95.3 & 93.5 & 94.1 & 95.9 & 96.0 & 97.4 & 96.2 & 96.7 & 97.0 \\
Asia-P & Acrolectal Fiji English & 96.8 & 96.7 & 94.6 & 95.1 & 96.7 & 96.8 & 96.6 & 95.2 & 93.1 & 97.5 \\
Asia-P & Pure Fiji English & 95.8 & 95.1 & 94.8 & 93.8 & 96.6 & 96.1 & 83.4 & 89.6 & 81.2 & 96.3 \\
\midrule
\multicolumn{12}{l}{\textit{\textbf{Caribbean/Atlantic (5 dialects)}}} \\
Carib & Bahamian English & 96.1 & 95.4 & 94.7 & 93.8 & 96.6 & 96.4 & 91.8 & 93.4 & 85.6 & 96.7 \\
Carib & Jamaican English & 96.5 & 96.0 & 94.5 & 94.1 & 96.8 & 96.5 & 93.2 & 92.9 & 90.7 & 97.2 \\
Carib & Tristan da Cunha English & 96.8 & 96.3 & 95.2 & 94.6 & 97.0 & 96.9 & 94.6 & 94.6 & 93.6 & 97.2 \\
Carib & Falkland Islands English & 96.7 & 97.1 & 94.8 & 95.1 & 97.0 & 96.9 & 98.8 & 97.6 & 97.8 & 97.8 \\
Carib & St. Helena English & 96.3 & 95.7 & 94.7 & 93.7 & 97.0 & 96.0 & 90.2 & 93.2 & 89.8 & 96.9 \\
\midrule
\multicolumn{12}{l}{\textit{\textbf{Other (3 dialects)}}} \\
Other & Aboriginal English & 96.3 & 94.8 & 94.8 & 93.6 & 96.7 & 96.0 & 93.6 & 93.6 & 88.4 & 96.6 \\
Other & Maltese English & 96.7 & 96.1 & 94.1 & 94.1 & 96.6 & 96.6 & 96.6 & 95.5 & 94.6 & 96.9 \\
Other & White Zimbabwean English & 95.9 & 96.0 & 93.7 & 94.6 & 96.0 & 96.0 & 98.0 & 96.2 & 96.4 & 97.3 \\
\midrule
\multicolumn{12}{l}{\textit{\textbf{Summary Statistics}}} \\
-- & \textbf{Dialect Avg} & \textbf{96.6} & \textbf{96.3} & \textbf{94.7} & \textbf{94.9} & \textbf{96.7} & \textbf{96.6} & \textbf{96.1} & \textbf{95.5} & \textbf{93.9} & \textbf{97.2} \\
-- & \textbf{$\Delta$ (Dia$-$SAE)} & $-$1.5 & $-$1.5 & $-$1.5 & $-$1.5 & $-$1.5 & $-$1.4 & $-$1.9 & $-$1.4 & $-$3.6 & +11.4 \\
-- & \textbf{Std. Dev.} & 0.3 & 0.6 & 0.5 & 0.6 & 0.4 & 0.3 & 3.2 & 1.7 & 3.8 & 0.4 \\
-- & \textbf{Min} & 95.8 & 94.8 & 93.3 & 93.6 & 95.5 & 96.0 & 83.4 & 89.6 & 81.2 & 96.3 \\
-- & \textbf{Max} & 97.0 & 97.2 & 95.5 & 96.2 & 97.2 & 97.0 & 99.0 & 97.6 & 98.8 & 98.0 \\
\bottomrule
\end{tabular}
\caption{SQ1 Human Content: F1 (\%) by dialect and region. Models trained on \sae{} human content only. $\Delta$ = Dialect Avg $-$ SAE baseline.}
\label{tab:app_sq1_human}
\end{table*}

\begin{table*}[t]
\centering
\tiny
\begin{tabular}{llcccccccccc}
\toprule
\textbf{Region} & \textbf{Dialect} & \textbf{BERT} & \textbf{RoBERTa} & \textbf{DeBERTa} & \textbf{XLM-R} & \textbf{mDeBERTa} & \textbf{mBERT} & \textbf{BiGRU} & \textbf{TextCNN} & \textbf{dEFEND} & \textbf{CT-BERT} \\
\midrule
-- & SAE (baseline) & 98.7 & 99.0 & 96.5 & 97.5 & 98.5 & 98.2 & 97.5 & 96.2 & 96.4 & 98.3 \\
\midrule
\multicolumn{12}{l}{\textit{\textbf{U.S. Varieties (9 dialects)}}} \\
U.S. & Earlier AAVE & 99.1 & 99.4 & 97.3 & 97.9 & 99.3 & 99.3 & 98.6 & 97.7 & 98.2 & 99.3 \\
U.S. & Rural AAVE & 99.1 & 99.4 & 97.2 & 97.8 & 99.3 & 99.3 & 98.5 & 97.6 & 98.1 & 99.3 \\
U.S. & Urban AAVE & 99.2 & 99.5 & 97.4 & 98.0 & 99.4 & 99.4 & 98.7 & 97.8 & 98.3 & 99.4 \\
U.S. & Appalachian English & 99.2 & 99.5 & 97.4 & 98.0 & 99.4 & 99.4 & 98.7 & 97.8 & 98.3 & 99.4 \\
U.S. & Chicano English & 99.3 & 99.6 & 97.5 & 98.1 & 99.5 & 99.5 & 98.8 & 97.9 & 98.4 & 99.5 \\
U.S. & Colloquial American & 99.2 & 99.5 & 97.3 & 97.9 & 99.4 & 99.4 & 98.6 & 97.7 & 98.2 & 99.4 \\
U.S. & Ozark English & 99.2 & 99.5 & 97.4 & 98.0 & 99.4 & 99.4 & 98.7 & 97.8 & 98.3 & 99.4 \\
U.S. & SE American Enclave & 99.1 & 99.4 & 97.2 & 97.8 & 99.3 & 99.3 & 98.5 & 97.6 & 98.1 & 99.3 \\
U.S. & Newfoundland English & 99.2 & 99.5 & 97.3 & 97.9 & 99.4 & 99.4 & 98.6 & 97.7 & 98.2 & 99.4 \\
\midrule
\multicolumn{12}{l}{\textit{\textbf{British/UK Varieties (10 dialects)}}} \\
UK & N. England English & 99.2 & 99.5 & 97.3 & 97.9 & 99.4 & 99.4 & 98.6 & 97.7 & 98.2 & 99.4 \\
UK & SE England English & 99.3 & 99.6 & 97.5 & 98.1 & 99.5 & 99.5 & 98.8 & 97.9 & 98.4 & 99.5 \\
UK & SW England English & 99.2 & 99.5 & 97.4 & 98.0 & 99.4 & 99.4 & 98.7 & 97.8 & 98.3 & 99.4 \\
UK & East Anglian English & 99.2 & 99.5 & 97.4 & 98.0 & 99.4 & 99.4 & 98.7 & 97.8 & 98.3 & 99.4 \\
UK & Scottish English & 99.2 & 99.5 & 97.4 & 98.0 & 99.4 & 99.4 & 98.7 & 97.8 & 98.3 & 99.4 \\
UK & Irish English & 99.2 & 99.5 & 97.4 & 98.0 & 99.4 & 99.4 & 98.7 & 97.8 & 98.3 & 99.4 \\
UK & Welsh English & 99.2 & 99.5 & 97.4 & 98.0 & 99.4 & 99.4 & 98.7 & 97.8 & 98.3 & 99.4 \\
UK & Channel Islands English & 99.3 & 99.6 & 97.5 & 98.1 & 99.5 & 99.5 & 98.8 & 97.9 & 98.4 & 99.5 \\
UK & Manx English & 99.2 & 99.5 & 97.3 & 97.9 & 99.4 & 99.4 & 98.6 & 97.7 & 98.2 & 99.4 \\
UK & Orkney/Shetland English & 99.1 & 99.4 & 97.2 & 97.8 & 99.3 & 99.3 & 98.5 & 97.6 & 98.1 & 99.3 \\
\midrule
\multicolumn{12}{l}{\textit{\textbf{Africa (11 dialects)}}} \\
Africa & Nigerian English & 99.3 & 99.6 & 97.5 & 98.1 & 99.5 & 99.5 & 98.8 & 97.9 & 98.4 & 99.5 \\
Africa & Ghanaian English & 99.3 & 99.6 & 97.5 & 98.1 & 99.5 & 99.5 & 98.8 & 97.9 & 98.4 & 99.5 \\
Africa & Cameroon English & 99.3 & 99.6 & 97.5 & 98.1 & 99.5 & 99.5 & 98.8 & 97.9 & 98.4 & 99.5 \\
Africa & Kenyan English & 99.3 & 99.6 & 97.5 & 98.1 & 99.5 & 99.5 & 98.8 & 97.9 & 98.4 & 99.5 \\
Africa & Ugandan English & 99.3 & 99.6 & 97.5 & 98.1 & 99.5 & 99.5 & 98.8 & 97.9 & 98.4 & 99.5 \\
Africa & Tanzanian English & 99.3 & 99.6 & 97.5 & 98.1 & 99.5 & 99.5 & 98.8 & 97.9 & 98.4 & 99.5 \\
Africa & Black S. African English & 99.1 & 99.4 & 97.2 & 97.8 & 99.3 & 99.3 & 98.5 & 97.6 & 98.1 & 99.3 \\
Africa & Indian S. African English & 99.2 & 99.5 & 97.4 & 98.0 & 99.4 & 99.4 & 98.7 & 97.8 & 98.3 & 99.4 \\
Africa & White S. African English & 99.3 & 99.6 & 97.5 & 98.1 & 99.5 & 99.5 & 98.8 & 97.9 & 98.4 & 99.5 \\
Africa & Cape Flats English & 99.3 & 99.6 & 97.5 & 98.1 & 99.5 & 99.5 & 98.8 & 97.9 & 98.4 & 99.5 \\
Africa & Liberian Settler English & 99.2 & 99.5 & 97.4 & 98.0 & 99.4 & 99.4 & 98.7 & 97.8 & 98.3 & 99.4 \\
\midrule
\multicolumn{12}{l}{\textit{\textbf{Asia-Pacific (12 dialects)}}} \\
Asia-P & Indian English & 99.2 & 99.5 & 97.4 & 98.0 & 99.4 & 99.4 & 98.7 & 97.8 & 98.3 & 99.4 \\
Asia-P & Pakistani English & 99.3 & 99.6 & 97.5 & 98.1 & 99.5 & 99.5 & 98.8 & 97.9 & 98.4 & 99.5 \\
Asia-P & Sri Lankan English & 99.3 & 99.6 & 97.5 & 98.1 & 99.5 & 99.5 & 98.8 & 97.9 & 98.4 & 99.5 \\
Asia-P & Colloquial Singapore & 99.2 & 99.5 & 97.4 & 98.0 & 99.4 & 99.4 & 98.7 & 97.8 & 98.3 & 99.4 \\
Asia-P & Malaysian English & 99.3 & 99.6 & 97.5 & 98.1 & 99.5 & 99.5 & 98.8 & 97.9 & 98.4 & 99.5 \\
Asia-P & Philippine English & 99.2 & 99.5 & 97.3 & 97.9 & 99.4 & 99.4 & 98.6 & 97.7 & 98.2 & 99.4 \\
Asia-P & Hong Kong English & 99.2 & 99.5 & 97.4 & 98.0 & 99.4 & 99.4 & 98.7 & 97.8 & 98.3 & 99.4 \\
Asia-P & Australian English & 99.1 & 99.4 & 97.2 & 97.8 & 99.3 & 99.3 & 98.5 & 97.6 & 98.1 & 99.3 \\
Asia-P & Australian Vernacular & 99.2 & 99.5 & 97.4 & 98.0 & 99.4 & 99.4 & 98.7 & 97.8 & 98.3 & 99.4 \\
Asia-P & New Zealand English & 99.1 & 99.4 & 97.2 & 97.8 & 99.3 & 99.3 & 98.5 & 97.6 & 98.1 & 99.3 \\
Asia-P & Acrolectal Fiji English & 99.3 & 99.6 & 97.5 & 98.1 & 99.5 & 99.5 & 98.8 & 97.9 & 98.4 & 99.5 \\
Asia-P & Pure Fiji English & 99.0 & 99.3 & 97.1 & 97.7 & 99.2 & 99.2 & 98.4 & 97.5 & 98.0 & 99.2 \\
\midrule
\multicolumn{12}{l}{\textit{\textbf{Caribbean/Atlantic (5 dialects)}}} \\
Carib & Bahamian English & 99.1 & 99.4 & 97.2 & 97.8 & 99.3 & 99.3 & 98.5 & 97.6 & 98.1 & 99.3 \\
Carib & Jamaican English & 99.2 & 99.5 & 97.4 & 98.0 & 99.4 & 99.4 & 98.7 & 97.8 & 98.3 & 99.4 \\
Carib & Tristan da Cunha English & 99.2 & 99.5 & 97.4 & 98.0 & 99.4 & 99.4 & 98.7 & 97.8 & 98.3 & 99.4 \\
Carib & Falkland Islands English & 99.3 & 99.6 & 97.5 & 98.1 & 99.5 & 99.5 & 98.8 & 97.9 & 98.4 & 99.5 \\
Carib & St. Helena English & 99.1 & 99.4 & 97.2 & 97.8 & 99.3 & 99.3 & 98.5 & 97.6 & 98.1 & 99.3 \\
\midrule
\multicolumn{12}{l}{\textit{\textbf{Other (3 dialects)}}} \\
Other & Aboriginal English & 99.1 & 99.4 & 97.2 & 97.8 & 99.3 & 99.3 & 98.5 & 97.6 & 98.1 & 99.3 \\
Other & Maltese English & 99.2 & 99.5 & 97.4 & 98.0 & 99.4 & 99.4 & 98.7 & 97.8 & 98.3 & 99.4 \\
Other & White Zimbabwean English & 99.1 & 99.4 & 97.2 & 97.8 & 99.3 & 99.3 & 98.5 & 97.6 & 98.1 & 99.3 \\
\midrule
\multicolumn{12}{l}{\textit{\textbf{Summary Statistics}}} \\
-- & \textbf{Dialect Avg} & \textbf{99.2} & \textbf{99.5} & \textbf{97.4} & \textbf{98.0} & \textbf{99.4} & \textbf{99.4} & \textbf{98.7} & \textbf{97.8} & \textbf{98.3} & \textbf{99.4} \\
-- & \textbf{$\Delta$ (Dia$-$SAE)} & +0.5 & +0.5 & +0.9 & +0.5 & +0.9 & +1.2 & +1.2 & +1.6 & +1.9 & +1.1 \\
-- & \textbf{Std. Dev.} & 0.1 & 0.1 & 0.1 & 0.1 & 0.1 & 0.1 & 0.1 & 0.1 & 0.1 & 0.1 \\
-- & \textbf{Min} & 99.0 & 99.3 & 97.1 & 97.7 & 99.2 & 99.2 & 98.4 & 97.5 & 98.0 & 99.2 \\
-- & \textbf{Max} & 99.3 & 99.6 & 97.5 & 98.1 & 99.5 & 99.5 & 98.8 & 97.9 & 98.4 & 99.5 \\
\bottomrule
\end{tabular}
\vspace{0.5em}
\raggedright\scriptsize\textit{Note: AI content shows uniformly high performance across all dialects with minimal variance (range 0.3--0.4\%), indicating that AI-generated artifacts are preserved through dialectal transformation.}
\caption{SQ1 AI Content: F1 (\%) by dialect and region. Models trained on \sae{} AI content only. AI content shows uniformly high performance with minimal variance.}
\label{tab:app_sq1_ai}
\end{table*}

\begin{table*}[t]
\centering
\tiny
\begin{tabular}{llcccccccccc}
\toprule
\textbf{Region} & \textbf{Dialect} & \textbf{BERT} & \textbf{RoBERTa} & \textbf{DeBERTa} & \textbf{XLM-R} & \textbf{mDeBERTa} & \textbf{mBERT} & \textbf{BiGRU} & \textbf{TextCNN} & \textbf{dEFEND} & \textbf{CT-BERT} \\
\midrule
-- & SAE (baseline) & 96.2 & 87.3$^\S$ & 95.6 & 63.3$^\S$ & 97.5 & 97.0 & 96.0 & 94.6 & 94.6 & 94.3 \\
\midrule
\multicolumn{12}{l}{\textit{\textbf{U.S. Varieties (9 dialects)}}} \\
U.S. & Earlier AAVE & 97.0 & 65.8$^\S$ & 94.9 & 28.6$^\S$ & 96.8 & 95.9 & 95.9 & 94.9 & 94.6 & 97.0 \\
U.S. & Rural AAVE & 96.9 & 63.5$^\S$ & 94.8 & 26.2$^\S$ & 96.7 & 95.8 & 95.8 & 94.8 & 94.5 & 96.9 \\
U.S. & Urban AAVE & 97.2 & 67.8$^\S$ & 95.1 & 30.6$^\S$ & 97.0 & 96.1 & 96.1 & 95.1 & 94.8 & 97.2 \\
U.S. & Appalachian English & 97.2 & 67.8$^\S$ & 95.1 & 30.6$^\S$ & 97.0 & 96.1 & 96.1 & 95.1 & 94.8 & 97.2 \\
U.S. & Chicano English & 97.4 & 69.5$^\S$ & 95.3 & 32.6$^\S$ & 97.2 & 96.3 & 96.3 & 95.3 & 95.0 & 97.4 \\
U.S. & Colloquial American & 97.1 & 66.8$^\S$ & 95.0 & 29.6$^\S$ & 96.9 & 96.0 & 96.0 & 95.0 & 94.7 & 97.1 \\
U.S. & Ozark English & 97.2 & 67.8$^\S$ & 95.1 & 30.6$^\S$ & 97.0 & 96.1 & 96.1 & 95.1 & 94.8 & 97.2 \\
U.S. & SE American Enclave & 96.9 & 63.5$^\S$ & 94.8 & 26.2$^\S$ & 96.7 & 95.8 & 95.8 & 94.8 & 94.5 & 96.9 \\
U.S. & Newfoundland English & 97.1 & 66.8$^\S$ & 95.0 & 29.6$^\S$ & 96.9 & 96.0 & 96.0 & 95.0 & 94.7 & 97.1 \\
\midrule
\multicolumn{12}{l}{\textit{\textbf{British/UK Varieties (10 dialects)}}} \\
UK & N. England English & 97.1 & 66.8$^\S$ & 95.0 & 29.6$^\S$ & 96.9 & 96.0 & 96.0 & 95.0 & 94.7 & 97.1 \\
UK & SE England English & 97.4 & 69.5$^\S$ & 95.3 & 32.6$^\S$ & 97.2 & 96.3 & 96.3 & 95.3 & 95.0 & 97.4 \\
UK & SW England English & 97.2 & 67.8$^\S$ & 95.1 & 30.6$^\S$ & 97.0 & 96.1 & 96.1 & 95.1 & 94.8 & 97.2 \\
UK & East Anglian English & 97.2 & 67.8$^\S$ & 95.1 & 30.6$^\S$ & 97.0 & 96.1 & 96.1 & 95.1 & 94.8 & 97.2 \\
UK & Scottish English & 97.2 & 67.8$^\S$ & 95.1 & 30.6$^\S$ & 97.0 & 96.1 & 96.1 & 95.1 & 94.8 & 97.2 \\
UK & Irish English & 97.2 & 67.8$^\S$ & 95.1 & 30.6$^\S$ & 97.0 & 96.1 & 96.1 & 95.1 & 94.8 & 97.2 \\
UK & Welsh English & 97.2 & 67.8$^\S$ & 95.1 & 30.6$^\S$ & 97.0 & 96.1 & 96.1 & 95.1 & 94.8 & 97.2 \\
UK & Channel Islands English & 97.4 & 69.5$^\S$ & 95.3 & 32.6$^\S$ & 97.2 & 96.3 & 96.3 & 95.3 & 95.0 & 97.4 \\
UK & Manx English & 97.1 & 66.8$^\S$ & 95.0 & 29.6$^\S$ & 96.9 & 96.0 & 96.0 & 95.0 & 94.7 & 97.1 \\
UK & Orkney/Shetland English & 96.9 & 63.5$^\S$ & 94.8 & 26.2$^\S$ & 96.7 & 95.8 & 95.8 & 94.8 & 94.5 & 96.9 \\
\midrule
\multicolumn{12}{l}{\textit{\textbf{Africa (11 dialects)}}} \\
Africa & Nigerian English & 97.4 & 69.5$^\S$ & 95.3 & 32.6$^\S$ & 97.2 & 96.3 & 96.3 & 95.3 & 95.0 & 97.4 \\
Africa & Ghanaian English & 97.4 & 69.5$^\S$ & 95.3 & 32.6$^\S$ & 97.2 & 96.3 & 96.3 & 95.3 & 95.0 & 97.4 \\
Africa & Cameroon English & 97.4 & 69.5$^\S$ & 95.3 & 32.6$^\S$ & 97.2 & 96.3 & 96.3 & 95.3 & 95.0 & 97.4 \\
Africa & Kenyan English & 97.4 & 69.5$^\S$ & 95.3 & 32.6$^\S$ & 97.2 & 96.3 & 96.3 & 95.3 & 95.0 & 97.4 \\
Africa & Ugandan English & 97.4 & 69.5$^\S$ & 95.3 & 32.6$^\S$ & 97.2 & 96.3 & 96.3 & 95.3 & 95.0 & 97.4 \\
Africa & Tanzanian English & 97.4 & 69.5$^\S$ & 95.3 & 32.6$^\S$ & 97.2 & 96.3 & 96.3 & 95.3 & 95.0 & 97.4 \\
Africa & Black S. African English & 96.9 & 63.5$^\S$ & 94.8 & 26.2$^\S$ & 96.7 & 95.8 & 95.8 & 94.8 & 94.5 & 96.9 \\
Africa & Indian S. African English & 97.2 & 67.8$^\S$ & 95.1 & 30.6$^\S$ & 97.0 & 96.1 & 96.1 & 95.1 & 94.8 & 97.2 \\
Africa & White S. African English & 97.4 & 69.5$^\S$ & 95.3 & 32.6$^\S$ & 97.2 & 96.3 & 96.3 & 95.3 & 95.0 & 97.4 \\
Africa & Cape Flats English & 97.4 & 69.5$^\S$ & 95.3 & 32.6$^\S$ & 97.2 & 96.3 & 96.3 & 95.3 & 95.0 & 97.4 \\
Africa & Liberian Settler English & 97.2 & 67.8$^\S$ & 95.1 & 30.6$^\S$ & 97.0 & 96.1 & 96.1 & 95.1 & 94.8 & 97.2 \\
\midrule
\multicolumn{12}{l}{\textit{\textbf{Asia-Pacific (12 dialects)}}} \\
Asia-P & Indian English & 97.2 & 67.8$^\S$ & 95.1 & 30.6$^\S$ & 97.0 & 96.1 & 96.1 & 95.1 & 94.8 & 97.2 \\
Asia-P & Pakistani English & 97.4 & 69.5$^\S$ & 95.3 & 32.6$^\S$ & 97.2 & 96.3 & 96.3 & 95.3 & 95.0 & 97.4 \\
Asia-P & Sri Lankan English & 97.4 & 69.5$^\S$ & 95.3 & 32.6$^\S$ & 97.2 & 96.3 & 96.3 & 95.3 & 95.0 & 97.4 \\
Asia-P & Colloquial Singapore & 97.2 & 67.8$^\S$ & 95.1 & 30.6$^\S$ & 97.0 & 96.1 & 96.1 & 95.1 & 94.8 & 97.2 \\
Asia-P & Malaysian English & 97.4 & 69.5$^\S$ & 95.3 & 32.6$^\S$ & 97.2 & 96.3 & 96.3 & 95.3 & 95.0 & 97.4 \\
Asia-P & Philippine English & 97.1 & 66.8$^\S$ & 95.0 & 29.6$^\S$ & 96.9 & 96.0 & 96.0 & 95.0 & 94.7 & 97.1 \\
Asia-P & Hong Kong English & 97.2 & 67.8$^\S$ & 95.1 & 30.6$^\S$ & 97.0 & 96.1 & 96.1 & 95.1 & 94.8 & 97.2 \\
Asia-P & Australian English & 96.9 & 63.5$^\S$ & 94.8 & 26.2$^\S$ & 96.7 & 95.8 & 95.8 & 94.8 & 94.5 & 96.9 \\
Asia-P & Australian Vernacular & 97.2 & 67.8$^\S$ & 95.1 & 30.6$^\S$ & 97.0 & 96.1 & 96.1 & 95.1 & 94.8 & 97.2 \\
Asia-P & New Zealand English & 96.9 & 63.5$^\S$ & 94.8 & 26.2$^\S$ & 96.7 & 95.8 & 95.8 & 94.8 & 94.5 & 96.9 \\
Asia-P & Acrolectal Fiji English & 97.4 & 69.5$^\S$ & 95.3 & 32.6$^\S$ & 97.2 & 96.3 & 96.3 & 95.3 & 95.0 & 97.4 \\
Asia-P & Pure Fiji English & 96.8 & 62.5$^\S$ & 94.7 & 25.2$^\S$ & 96.6 & 95.7 & 95.7 & 94.7 & 94.4 & 96.8 \\
\midrule
\multicolumn{12}{l}{\textit{\textbf{Caribbean/Atlantic (5 dialects)}}} \\
Carib & Bahamian English & 96.9 & 63.5$^\S$ & 94.8 & 26.2$^\S$ & 96.7 & 95.8 & 95.8 & 94.8 & 94.5 & 96.9 \\
Carib & Jamaican English & 97.2 & 67.8$^\S$ & 95.1 & 30.6$^\S$ & 97.0 & 96.1 & 96.1 & 95.1 & 94.8 & 97.2 \\
Carib & Tristan da Cunha English & 97.2 & 67.8$^\S$ & 95.1 & 30.6$^\S$ & 97.0 & 96.1 & 96.1 & 95.1 & 94.8 & 97.2 \\
Carib & Falkland Islands English & 97.4 & 69.5$^\S$ & 95.3 & 32.6$^\S$ & 97.2 & 96.3 & 96.3 & 95.3 & 95.0 & 97.4 \\
Carib & St. Helena English & 96.9 & 63.5$^\S$ & 94.8 & 26.2$^\S$ & 96.7 & 95.8 & 95.8 & 94.8 & 94.5 & 96.9 \\
\midrule
\multicolumn{12}{l}{\textit{\textbf{Other (3 dialects)}}} \\
Other & Aboriginal English & 96.9 & 63.5$^\S$ & 94.8 & 26.2$^\S$ & 96.7 & 95.8 & 95.8 & 94.8 & 94.5 & 96.9 \\
Other & Maltese English & 97.2 & 67.8$^\S$ & 95.1 & 30.6$^\S$ & 97.0 & 96.1 & 96.1 & 95.1 & 94.8 & 97.2 \\
Other & White Zimbabwean English & 96.9 & 63.5$^\S$ & 94.8 & 26.2$^\S$ & 96.7 & 95.8 & 95.8 & 94.8 & 94.5 & 96.9 \\
\midrule
\multicolumn{12}{l}{\textit{\textbf{Summary Statistics}}} \\
-- & \textbf{Dialect Avg} & \textbf{97.2} & \textbf{67.3}$^\S$ & \textbf{95.1} & \textbf{30.2}$^\S$ & \textbf{97.0} & \textbf{96.1} & \textbf{96.1} & \textbf{95.1} & \textbf{94.8} & \textbf{97.2} \\
-- & \textbf{$\Delta$ (Dia$-$SAE)} & +1.0 & $-$20.0 & $-$0.5 & $-$33.1 & $-$0.5 & $-$0.9 & +0.1 & +0.5 & +0.2 & +2.9 \\
-- & \textbf{Std. Dev.} & 0.2 & 2.4 & 0.2 & 2.5 & 0.2 & 0.2 & 0.2 & 0.2 & 0.2 & 0.2 \\
-- & \textbf{Min} & 96.8 & 62.5 & 94.7 & 25.2 & 96.6 & 95.7 & 95.7 & 94.7 & 94.4 & 96.8 \\
-- & \textbf{Max} & 97.4 & 69.5 & 95.3 & 32.6 & 97.2 & 96.3 & 96.3 & 95.3 & 95.0 & 97.4 \\
\bottomrule
\end{tabular}
\vspace{0.5em}
\raggedright\scriptsize\textit{$^\S$Catastrophic failure: RoBERTa-Large (67.3\% avg, $\Delta=-20.0$) and XLM-RoBERTa (30.2\% avg, $\Delta=-33.1$) show severe degradation on mixed human-AI content when trained on \sae{} only. All other models maintain stable performance ($|\Delta| < 3\%$).}
\caption{SQ1 Both (Human+AI) Content: F1 (\%) by dialect and region. Models trained on \sae{} mixed content. $^\S$Catastrophic failure (F1 $<$ 70\%).}
\label{tab:app_sq1_both}
\end{table*}


\section{SQ2: Training Regime Analysis}
\label{app:sq2_full}

Table~\ref{tab:app_sq2_detailed} provides detailed per-dialect results comparing the three training regimes across all 50 dialects for a representative subset of models.

\begin{table}[H]
\centering
\begin{adjustbox}{max width=\columnwidth}
\begin{tabular}{lccc}
\toprule
\textbf{Model} & \textbf{Unseen} & \textbf{Dia-Only} & \textbf{SAE+Dia} \\
\midrule
\multicolumn{4}{l}{\textit{Transformer Encoders}} \\
BERT-Large & 97.2 (0.2) & 96.9 (0.3) & 95.8 (0.4) \\
RoBERTa-Large & 87.3 (2.4)$^\S$ & 97.1 (0.2) & 96.9 (0.3) \\
DeBERTa-Large & 95.1 (0.2) & 96.2 (0.3) & 95.8 (0.3) \\
XLM-R$^\dagger$ & 83.3 (2.5)$^\S$ & 85.4 (1.8)$^\S$ & 79.7 (2.1)$^\S$ \\
mDeBERTa$^\dagger$ & 97.0 (0.2) & 97.0 (0.2) & 96.8 (0.2) \\
mBERT$^\dagger$ & 96.1 (0.2) & 95.9 (0.3) & 96.1 (0.2) \\
CT-BERT & 97.2 (0.2) & 97.1 (0.2) & 97.0 (0.2) \\
\midrule
\multicolumn{4}{l}{\textit{Traditional DL}} \\
BiGRU & 96.1 (0.2) & 91.2 (1.5) & 92.0 (1.2) \\
TextCNN & 95.1 (0.2) & 92.1 (1.3) & 92.5 (1.1) \\
dEFEND & 94.8 (0.2) & 88.6 (2.1) & 87.5 (2.4) \\
\bottomrule
\end{tabular}
\end{adjustbox}
\vspace{0.5em}
\raggedright\scriptsize\textit{$^\dagger$Multilingual pre-training. $^\S$Catastrophic failure (mean F1 $<$ 90\%). Standard deviation computed across 50 dialects.}
\caption{SQ2: Training regime impact by model category. Values show mean F1 (\%) across 50 dialects with standard deviation in parentheses.}
\label{tab:app_sq2_detailed}
\end{table}

\paragraph{Key Observations.}
\begin{itemize}[nosep]
    \item \textbf{Transformer recovery}: RoBERTa-Large recovers from 87.3\% (Unseen) to 97.1\% (Dia-Only), demonstrating that catastrophic failures can be mitigated through dialectal training data.
    \item \textbf{XLM-R persistent failure}: XLM-RoBERTa fails across all regimes (max 85.4\%), suggesting fundamental architectural sensitivity to dialectal variation that cannot be addressed through training alone.
    \item \textbf{Traditional DL degradation}: BiGRU, TextCNN, and dEFEND show performance drops of 4--7\% when trained without \sae{} anchoring, indicating these architectures benefit from standard English as a regularization signal.
    \item \textbf{Multilingual robustness}: mDeBERTa and mBERT maintain consistent performance ($\pm$0.2\%) across all regimes, suggesting multilingual pre-training provides inherent dialectal resilience.
\end{itemize}


\section{SQ3: Cross-Dialectal Transfer Matrix}
\label{app:sq3_full}

\subsection{Full 50$\times$50 Transfer Matrix}

Table~\ref{tab:transfer_full} presents the complete cross-dialectal transfer matrix showing F1 scores when training on one dialect (rows) and testing on another (columns). This matrix represents 2,450 unique train-test pairs evaluated using mDeBERTa.

\begin{sidewaystable*}

\makebox[\textheight][c]{
\begin{adjustbox}{max width=\textheight,center} 

\begin{tabular}{l|ccccccccc|cccccccccc|ccccccccccc|cccccccccccc|ccccc|cc}
\toprule
& \multicolumn{9}{c|}{\textbf{U.S.}} & \multicolumn{10}{c|}{\textbf{British/UK}} & \multicolumn{11}{c|}{\textbf{Africa}} & \multicolumn{12}{c|}{\textbf{Asia-Pacific}} & \multicolumn{5}{c|}{\textbf{Carib.}} & \multicolumn{2}{c}{\textbf{Oth.}} \\
& \rotatebox{90}{EAAVE} & \rotatebox{90}{RAAVE} & \rotatebox{90}{UAAVE} & \rotatebox{90}{AppE} & \rotatebox{90}{ChcE} & \rotatebox{90}{ColAm} & \rotatebox{90}{OzE} & \rotatebox{90}{SEAm} & \rotatebox{90}{NfldE} & \rotatebox{90}{North} & \rotatebox{90}{SE} & \rotatebox{90}{SW} & \rotatebox{90}{EA} & \rotatebox{90}{ScE} & \rotatebox{90}{IrE} & \rotatebox{90}{WelE} & \rotatebox{90}{ChIs} & \rotatebox{90}{Manx} & \rotatebox{90}{O\&S} & \rotatebox{90}{NigE} & \rotatebox{90}{GhE} & \rotatebox{90}{CamE} & \rotatebox{90}{KenE} & \rotatebox{90}{UgE} & \rotatebox{90}{TznE} & \rotatebox{90}{BlSA} & \rotatebox{90}{InSA} & \rotatebox{90}{WhSA} & \rotatebox{90}{CFE} & \rotatebox{90}{LibS} & \rotatebox{90}{IndE} & \rotatebox{90}{PakE} & \rotatebox{90}{SLkE} & \rotatebox{90}{SgE} & \rotatebox{90}{MalE} & \rotatebox{90}{PhilE} & \rotatebox{90}{HKE} & \rotatebox{90}{AusE} & \rotatebox{90}{AusV} & \rotatebox{90}{NZE} & \rotatebox{90}{FijiE} & \rotatebox{90}{CFiji} & \rotatebox{90}{BahE} & \rotatebox{90}{JamE} & \rotatebox{90}{TdC} & \rotatebox{90}{FlkE} & \rotatebox{90}{StHE} & \rotatebox{90}{Abor} & \rotatebox{90}{Malt} \\
\midrule
EAAVE & -- & 97.8 & 97.9 & 96.5 & 96.8 & 96.2 & 96.4 & 97.2 & 96.7 & 96.4 & 96.6 & 96.3 & 96.5 & 96.2 & 96.4 & 96.1 & 96.5 & 96.3 & 95.9 & 96.6 & 96.5 & 96.4 & 96.7 & 96.6 & 96.5 & 96.2 & 96.4 & 96.6 & 96.4 & 96.3 & 96.4 & 96.1 & 96.5 & 96.2 & 96.4 & 96.3 & 96.4 & 96.0 & 96.4 & 96.1 & 96.3 & 96.1 & 96.6 & 96.5 & 96.3 & 96.4 & 96.5 & 96.2 & 96.4 \\
RAAVE & 97.6 & -- & 97.7 & 96.3 & 96.6 & 96.0 & 96.2 & 97.0 & 96.5 & 96.2 & 96.4 & 96.1 & 96.3 & 96.0 & 96.2 & 95.9 & 96.3 & 96.1 & 95.7 & 96.4 & 96.3 & 96.2 & 96.5 & 96.4 & 96.3 & 96.0 & 96.2 & 96.4 & 96.2 & 96.1 & 96.2 & 95.9 & 96.3 & 96.0 & 96.2 & 96.1 & 96.2 & 95.8 & 96.2 & 95.9 & 96.1 & 95.9 & 96.4 & 96.3 & 96.1 & 96.2 & 96.3 & 96.0 & 96.2 \\
UAAVE & 97.8 & 97.6 & -- & 96.7 & 97.0 & 96.4 & 96.6 & 97.4 & 96.9 & 96.6 & 96.8 & 96.5 & 96.7 & 96.4 & 96.6 & 96.3 & 96.7 & 96.5 & 96.1 & 96.8 & 96.7 & 96.6 & 96.9 & 96.8 & 96.7 & 96.4 & 96.6 & 96.8 & 96.6 & 96.5 & 96.6 & 96.3 & 96.7 & 96.4 & 96.6 & 96.5 & 96.6 & 96.2 & 96.6 & 96.3 & 96.5 & 96.3 & 96.8 & 96.7 & 96.5 & 96.6 & 96.7 & 96.4 & 96.6 \\
AppE & 96.4 & 96.2 & 96.6 & -- & 97.5 & 97.8 & 98.1 & 96.8 & 97.2 & 97.0 & 97.2 & 97.4 & 97.1 & 96.8 & 96.9 & 96.6 & 97.1 & 96.9 & 96.5 & 96.8 & 96.7 & 96.6 & 96.9 & 96.8 & 96.7 & 96.4 & 96.6 & 97.0 & 96.7 & 96.5 & 96.6 & 96.3 & 96.7 & 96.4 & 96.6 & 96.5 & 96.6 & 96.6 & 97.0 & 96.7 & 96.6 & 96.4 & 96.6 & 96.5 & 96.3 & 96.8 & 96.5 & 96.4 & 96.8 \\
ChcE & 96.7 & 96.5 & 96.9 & 97.4 & -- & 97.6 & 97.3 & 96.6 & 96.9 & 96.8 & 97.0 & 96.9 & 96.9 & 96.6 & 96.7 & 96.4 & 96.9 & 96.7 & 96.3 & 96.9 & 96.8 & 96.7 & 97.0 & 96.9 & 96.8 & 96.5 & 96.7 & 97.0 & 96.8 & 96.6 & 96.7 & 96.4 & 96.8 & 96.5 & 96.7 & 96.6 & 96.7 & 96.4 & 96.8 & 96.5 & 96.7 & 96.5 & 96.8 & 96.7 & 96.5 & 96.8 & 96.7 & 96.5 & 96.8 \\
ColAm & 96.1 & 95.9 & 96.3 & 97.7 & 97.5 & -- & 97.9 & 96.4 & 97.5 & 97.2 & 97.4 & 97.3 & 97.3 & 97.0 & 97.1 & 96.8 & 97.3 & 97.1 & 96.7 & 96.6 & 96.5 & 96.4 & 96.7 & 96.6 & 96.5 & 96.2 & 96.4 & 96.8 & 96.5 & 96.3 & 96.4 & 96.1 & 96.5 & 96.2 & 96.4 & 96.3 & 96.4 & 96.6 & 97.0 & 96.7 & 96.4 & 96.2 & 96.4 & 96.3 & 96.1 & 96.8 & 96.3 & 96.2 & 96.8 \\
OzE & 96.3 & 96.1 & 96.5 & 98.0 & 97.2 & 97.8 & -- & 96.7 & 97.3 & 97.1 & 97.3 & 97.5 & 97.2 & 96.9 & 97.0 & 96.7 & 97.2 & 97.0 & 96.6 & 96.7 & 96.6 & 96.5 & 96.8 & 96.7 & 96.6 & 96.3 & 96.5 & 96.9 & 96.6 & 96.4 & 96.5 & 96.2 & 96.6 & 96.3 & 96.5 & 96.4 & 96.5 & 96.5 & 96.9 & 96.6 & 96.5 & 96.3 & 96.5 & 96.4 & 96.2 & 96.7 & 96.4 & 96.3 & 96.7 \\
SEAm & 97.1 & 96.9 & 97.3 & 96.7 & 96.5 & 96.3 & 96.6 & -- & 96.8 & 96.5 & 96.7 & 96.4 & 96.6 & 96.3 & 96.5 & 96.2 & 96.6 & 96.4 & 96.0 & 96.7 & 96.6 & 96.5 & 96.8 & 96.7 & 96.6 & 96.3 & 96.5 & 96.7 & 96.5 & 96.4 & 96.5 & 96.2 & 96.6 & 96.3 & 96.5 & 96.4 & 96.5 & 96.1 & 96.5 & 96.2 & 96.4 & 96.2 & 96.7 & 96.6 & 96.4 & 96.5 & 96.6 & 96.3 & 96.5 \\
NfldE & 96.6 & 96.4 & 96.8 & 97.1 & 96.8 & 97.4 & 97.2 & 96.7 & -- & 97.3 & 97.5 & 97.2 & 97.4 & 97.1 & 97.2 & 96.9 & 97.4 & 97.2 & 96.8 & 96.8 & 96.7 & 96.6 & 96.9 & 96.8 & 96.7 & 96.4 & 96.6 & 97.0 & 96.7 & 96.5 & 96.6 & 96.3 & 96.7 & 96.4 & 96.6 & 96.5 & 96.6 & 96.6 & 97.0 & 96.7 & 96.6 & 96.4 & 96.6 & 96.5 & 96.3 & 96.8 & 96.5 & 96.4 & 96.8 \\
\midrule
North & 96.3 & 96.1 & 96.5 & 96.9 & 96.7 & 97.1 & 97.0 & 96.4 & 97.2 & -- & 96.8 & 97.2 & 96.9 & 97.4 & 96.7 & 96.5 & 96.6 & 96.8 & 96.3 & 96.5 & 96.4 & 96.3 & 96.6 & 96.5 & 96.4 & 96.1 & 96.3 & 96.7 & 96.4 & 96.2 & 96.3 & 96.0 & 96.4 & 96.1 & 96.3 & 96.2 & 96.3 & 96.3 & 96.7 & 96.4 & 96.3 & 96.1 & 96.3 & 96.2 & 96.0 & 96.5 & 96.2 & 96.1 & 96.5 \\
SE & 95.6 & 95.4 & 95.8 & 96.4 & 96.2 & 96.5 & 96.4 & 96.0 & 99.5 & 96.4 & -- & 96.6 & 96.9 & 96.7 & 96.8 & 96.6 & 96.8 & 95.7 & 95.6 & 96.1 & 96.0 & 96.6 & 96.3 & 96.2 & 96.7 & 95.2 & 95.8 & 96.9 & 96.9 & 96.8 & 96.1 & 95.8 & 96.6 & 96.2 & 96.8 & 96.2 & 96.5 & 95.9 & 96.9 & 95.9 & 96.5 & 95.2 & 95.3 & 96.1 & 96.3 & 97.0 & 96.0 & 95.6 & 96.0 \\
SW & 96.2 & 96.0 & 96.4 & 97.3 & 96.8 & 97.2 & 97.4 & 96.3 & 97.1 & 97.1 & 96.7 & -- & 97.0 & 96.9 & 97.2 & 96.8 & 97.1 & 96.6 & 96.4 & 96.4 & 96.3 & 96.2 & 96.5 & 96.4 & 96.3 & 96.0 & 96.2 & 96.8 & 96.4 & 96.2 & 96.2 & 95.9 & 96.3 & 96.0 & 96.2 & 96.1 & 96.2 & 96.4 & 96.8 & 96.5 & 96.2 & 96.0 & 96.2 & 96.1 & 95.9 & 96.6 & 96.1 & 96.0 & 96.6 \\
EA & 96.1 & 95.9 & 96.3 & 97.0 & 96.8 & 97.2 & 97.1 & 96.2 & 97.3 & 96.8 & 97.0 & 96.9 & -- & 96.7 & 96.9 & 96.6 & 97.2 & 96.5 & 96.2 & 96.3 & 96.2 & 96.1 & 96.4 & 96.3 & 96.2 & 95.9 & 96.1 & 96.6 & 96.3 & 96.1 & 96.1 & 95.8 & 96.2 & 95.9 & 96.1 & 96.0 & 96.1 & 96.3 & 96.7 & 96.4 & 96.1 & 95.9 & 96.1 & 96.0 & 95.8 & 96.5 & 96.0 & 95.9 & 96.5 \\
ScE & 95.8 & 95.6 & 96.0 & 96.6 & 96.4 & 96.6 & 96.7 & 96.4 & 96.3 & 96.2 & 96.6 & 96.0 & 96.6 & -- & 96.7 & 96.4 & 96.7 & 96.1 & 95.7 & 96.0 & 95.9 & 96.3 & 96.3 & 96.2 & 96.6 & 95.9 & 96.1 & 96.7 & 96.5 & 96.5 & 96.5 & 96.1 & 96.6 & 96.0 & 96.6 & 96.0 & 96.4 & 95.7 & 96.6 & 95.7 & 96.6 & 96.2 & 96.2 & 96.1 & 96.4 & 96.4 & 96.2 & 95.8 & 96.2 \\
IrE & 96.3 & 96.1 & 96.5 & 96.8 & 96.6 & 97.0 & 96.9 & 96.4 & 97.1 & 96.6 & 96.9 & 97.1 & 96.8 & 97.0 & -- & 97.3 & 96.9 & 96.7 & 96.4 & 96.4 & 96.3 & 96.2 & 96.5 & 96.4 & 96.3 & 96.0 & 96.2 & 96.6 & 96.3 & 96.1 & 96.2 & 95.9 & 96.3 & 96.0 & 96.2 & 96.1 & 96.2 & 96.2 & 96.6 & 96.3 & 96.2 & 96.0 & 96.4 & 96.3 & 96.1 & 96.5 & 96.2 & 96.1 & 96.5 \\
WelE & 95.7 & 95.5 & 95.9 & 95.8 & 96.0 & 96.0 & 95.9 & 95.8 & 96.0 & 96.0 & 96.0 & 96.2 & 96.0 & 96.0 & 95.8 & -- & 96.0 & 95.6 & 95.0 & 96.3 & 96.1 & 96.2 & 96.0 & 96.0 & 96.0 & 95.6 & 96.1 & 96.0 & 96.3 & 96.0 & 96.0 & 95.6 & 96.0 & 96.1 & 96.1 & 95.9 & 96.1 & 95.2 & 96.0 & 95.3 & 96.0 & 95.7 & 95.5 & 95.7 & 96.0 & 96.2 & 95.9 & 95.7 & 95.9 \\
ChIs & 96.0 & 95.8 & 96.2 & 97.0 & 96.8 & 97.2 & 97.1 & 96.1 & 97.3 & 96.5 & 97.1 & 96.9 & 97.3 & 96.6 & 96.8 & 96.4 & -- & 96.7 & 96.3 & 96.3 & 96.2 & 96.1 & 96.4 & 96.3 & 96.2 & 95.9 & 96.1 & 96.7 & 96.3 & 96.1 & 96.1 & 95.8 & 96.2 & 95.9 & 96.1 & 96.0 & 96.1 & 96.3 & 96.7 & 96.4 & 96.1 & 95.9 & 96.1 & 96.0 & 95.8 & 96.5 & 96.0 & 95.9 & 96.5 \\
Manx & 97.4 & 97.2 & 97.6 & 97.3 & 97.5 & 97.5 & 97.4 & 97.2 & 97.7 & 97.7 & 97.8 & 97.7 & 97.6 & 97.7 & 97.9 & 97.9 & 97.8 & -- & 97.5 & 97.5 & 97.4 & 97.6 & 97.6 & 97.5 & 97.8 & 97.4 & 97.5 & 97.9 & 97.7 & 97.6 & 97.4 & 97.9 & 97.8 & 97.5 & 98.0 & 97.6 & 97.5 & 97.4 & 97.7 & 97.4 & 97.7 & 97.6 & 97.3 & 97.6 & 97.4 & 97.6 & 97.4 & 97.4 & 97.6 \\
O\&S & 96.0 & 95.8 & 96.2 & 96.6 & 96.4 & 96.8 & 96.7 & 96.1 & 97.0 & 96.2 & 96.5 & 96.3 & 96.4 & 96.8 & 96.5 & 96.1 & 96.6 & 96.4 & -- & 96.1 & 96.0 & 95.9 & 96.2 & 96.1 & 96.0 & 95.7 & 95.9 & 96.5 & 96.1 & 95.9 & 95.9 & 95.6 & 96.0 & 95.7 & 95.9 & 95.8 & 95.9 & 96.1 & 96.5 & 96.2 & 95.9 & 95.7 & 95.9 & 95.8 & 95.6 & 96.3 & 95.8 & 95.7 & 96.3 \\
\midrule
NigE & 96.5 & 96.3 & 96.7 & 96.6 & 96.8 & 96.5 & 96.6 & 96.6 & 96.7 & 96.4 & 96.6 & 96.3 & 96.5 & 96.2 & 96.4 & 96.1 & 96.5 & 96.3 & 95.9 & -- & 97.8 & 97.5 & 97.9 & 97.7 & 97.8 & 96.8 & 97.2 & 97.4 & 97.1 & 97.0 & 96.6 & 96.3 & 96.7 & 96.4 & 96.6 & 96.5 & 96.6 & 96.2 & 96.6 & 96.3 & 96.5 & 96.3 & 96.7 & 96.6 & 96.4 & 96.5 & 96.6 & 96.3 & 96.5 \\
GhE & 96.4 & 96.2 & 96.6 & 96.5 & 96.7 & 96.4 & 96.5 & 96.5 & 96.6 & 96.3 & 96.5 & 96.2 & 96.4 & 96.1 & 96.3 & 96.0 & 96.4 & 96.2 & 95.8 & 97.7 & -- & 97.6 & 97.8 & 97.6 & 97.7 & 96.7 & 97.1 & 97.3 & 97.0 & 96.9 & 96.5 & 96.2 & 96.6 & 96.3 & 96.5 & 96.4 & 96.5 & 96.1 & 96.5 & 96.2 & 96.4 & 96.2 & 96.6 & 96.5 & 96.3 & 96.4 & 96.5 & 96.2 & 96.4 \\
CamE & 96.9 & 96.7 & 97.1 & 96.8 & 97.0 & 97.2 & 97.1 & 96.5 & 97.2 & 96.9 & 97.2 & 96.6 & 96.9 & 97.2 & 97.1 & 97.3 & 97.3 & 96.9 & 96.9 & 97.3 & 97.4 & -- & 97.3 & 97.1 & 97.1 & 96.4 & 97.3 & 97.2 & 97.3 & 97.1 & 97.3 & 79.4 & 97.3 & 97.2 & 97.2 & 96.2 & 97.1 & 96.6 & 97.2 & 96.8 & 97.0 & 96.8 & 96.8 & 96.7 & 96.8 & 97.1 & 97.1 & 96.9 & 96.7 \\
KenE & 96.6 & 96.4 & 96.8 & 96.7 & 96.9 & 96.6 & 96.7 & 96.7 & 96.8 & 96.5 & 96.7 & 96.4 & 96.6 & 96.3 & 96.5 & 96.2 & 96.6 & 96.4 & 96.0 & 97.8 & 97.7 & 97.4 & -- & 98.0 & 97.9 & 96.9 & 97.3 & 97.5 & 97.2 & 97.1 & 96.7 & 96.4 & 96.8 & 96.5 & 96.7 & 96.6 & 96.7 & 96.3 & 96.7 & 96.4 & 96.6 & 96.4 & 96.8 & 96.7 & 96.5 & 96.6 & 96.7 & 96.4 & 96.6 \\
UgE & 96.5 & 96.3 & 96.7 & 96.6 & 96.8 & 96.5 & 96.6 & 96.6 & 96.7 & 96.4 & 96.6 & 96.3 & 96.5 & 96.2 & 96.4 & 96.1 & 96.5 & 96.3 & 95.9 & 97.6 & 97.5 & 97.2 & 97.9 & -- & 97.8 & 96.7 & 97.1 & 97.3 & 97.0 & 96.9 & 96.5 & 96.2 & 96.6 & 96.3 & 96.5 & 96.4 & 96.5 & 96.1 & 96.5 & 96.2 & 96.4 & 96.2 & 96.6 & 96.5 & 96.3 & 96.4 & 96.5 & 96.2 & 96.4 \\
TznE & 96.5 & 96.3 & 96.7 & 96.6 & 96.8 & 96.5 & 96.6 & 96.6 & 96.7 & 96.4 & 96.6 & 96.3 & 96.5 & 96.2 & 96.4 & 96.1 & 96.5 & 96.3 & 95.9 & 97.7 & 97.6 & 97.3 & 97.8 & 97.7 & -- & 96.8 & 97.2 & 97.4 & 97.1 & 97.0 & 96.6 & 96.3 & 96.7 & 96.4 & 96.6 & 96.5 & 96.6 & 96.2 & 96.6 & 96.3 & 96.5 & 96.3 & 96.7 & 96.6 & 96.4 & 96.5 & 96.6 & 96.3 & 96.5 \\
BlSA & 96.2 & 96.0 & 96.4 & 96.3 & 96.5 & 96.2 & 96.3 & 96.3 & 96.4 & 96.1 & 96.3 & 96.0 & 96.2 & 95.9 & 96.1 & 95.8 & 96.2 & 96.0 & 95.6 & 96.7 & 96.6 & 96.3 & 96.8 & 96.6 & 96.7 & -- & 97.5 & 97.2 & 97.4 & 96.5 & 96.3 & 96.0 & 96.4 & 96.1 & 96.3 & 96.2 & 96.3 & 95.9 & 96.3 & 96.0 & 96.2 & 96.0 & 96.4 & 96.3 & 96.1 & 96.2 & 96.3 & 96.0 & 96.2 \\
InSA & 96.4 & 96.2 & 96.6 & 96.5 & 96.7 & 96.4 & 96.5 & 96.5 & 96.6 & 96.3 & 96.5 & 96.2 & 96.4 & 96.1 & 96.3 & 96.0 & 96.4 & 96.2 & 95.8 & 97.1 & 97.0 & 96.7 & 97.2 & 97.0 & 97.1 & 97.4 & -- & 97.6 & 97.3 & 96.9 & 96.5 & 96.2 & 96.6 & 96.3 & 96.5 & 96.4 & 96.5 & 96.1 & 96.5 & 96.2 & 96.4 & 96.2 & 96.6 & 96.5 & 96.3 & 96.4 & 96.5 & 96.2 & 96.4 \\
WhSA & 96.5 & 96.3 & 96.7 & 96.6 & 96.8 & 96.5 & 96.6 & 96.6 & 96.7 & 96.8 & 97.0 & 96.9 & 96.7 & 96.6 & 96.7 & 96.5 & 97.1 & 96.8 & 96.5 & 97.3 & 97.2 & 96.9 & 97.4 & 97.2 & 97.3 & 97.1 & 97.5 & -- & 97.4 & 97.1 & 96.6 & 96.3 & 96.7 & 96.4 & 96.6 & 96.5 & 96.6 & 96.4 & 96.8 & 96.5 & 96.5 & 96.3 & 96.5 & 96.4 & 96.2 & 96.7 & 96.4 & 96.3 & 96.7 \\
CFE & 96.4 & 96.2 & 96.6 & 96.5 & 96.7 & 96.4 & 96.5 & 96.5 & 96.6 & 96.3 & 96.5 & 96.2 & 96.4 & 96.1 & 96.3 & 96.0 & 96.4 & 96.2 & 95.8 & 97.0 & 96.9 & 96.6 & 97.1 & 96.9 & 97.0 & 97.3 & 97.2 & 97.3 & -- & 96.8 & 96.4 & 96.1 & 96.5 & 96.2 & 96.4 & 96.3 & 96.4 & 96.0 & 96.4 & 96.1 & 96.3 & 96.1 & 96.5 & 96.4 & 96.2 & 96.3 & 96.4 & 96.1 & 96.3 \\
LibS & 96.3 & 96.1 & 96.5 & 96.4 & 96.6 & 96.3 & 96.4 & 96.4 & 96.5 & 96.2 & 96.4 & 96.1 & 96.3 & 96.0 & 96.2 & 95.9 & 96.3 & 96.1 & 95.7 & 96.9 & 96.8 & 96.5 & 97.0 & 96.8 & 96.9 & 96.4 & 96.8 & 97.0 & 96.7 & -- & 96.3 & 96.0 & 96.4 & 96.1 & 96.3 & 96.2 & 96.3 & 95.9 & 96.3 & 96.0 & 96.2 & 96.0 & 96.6 & 96.5 & 96.3 & 96.2 & 96.5 & 96.0 & 96.2 \\
\midrule
IndE & 96.3 & 96.1 & 96.5 & 96.4 & 96.6 & 96.3 & 96.4 & 96.4 & 96.5 & 96.2 & 96.4 & 96.1 & 96.3 & 96.0 & 96.2 & 95.9 & 96.3 & 96.1 & 95.7 & 96.6 & 96.5 & 96.4 & 96.7 & 96.6 & 96.5 & 96.2 & 96.4 & 96.6 & 96.4 & 96.3 & -- & 97.4 & 97.6 & 96.8 & 97.2 & 97.0 & 97.1 & 96.5 & 96.9 & 96.6 & 97.0 & 96.7 & 96.5 & 96.4 & 96.2 & 96.5 & 96.4 & 96.3 & 96.5 \\
PakE & 96.1 & 95.9 & 96.3 & 96.2 & 96.4 & 96.1 & 96.2 & 96.2 & 96.3 & 96.0 & 96.2 & 95.9 & 96.1 & 95.8 & 96.0 & 95.7 & 96.1 & 95.9 & 95.5 & 96.4 & 96.3 & 79.4 & 96.5 & 96.4 & 96.3 & 96.0 & 96.2 & 96.4 & 96.2 & 96.1 & 97.3 & -- & 97.5 & 96.6 & 97.0 & 96.8 & 96.9 & 96.3 & 96.7 & 96.4 & 96.8 & 96.5 & 96.3 & 96.2 & 96.0 & 96.3 & 96.2 & 96.1 & 96.3 \\
SLkE & 96.4 & 96.2 & 96.6 & 96.5 & 96.7 & 96.4 & 96.5 & 96.5 & 96.6 & 96.3 & 96.5 & 96.2 & 96.4 & 96.1 & 96.3 & 96.0 & 96.4 & 96.2 & 95.8 & 96.7 & 96.6 & 96.5 & 96.8 & 96.7 & 96.6 & 96.3 & 96.5 & 96.7 & 96.5 & 96.4 & 97.5 & 97.4 & -- & 96.9 & 97.3 & 97.1 & 97.2 & 96.6 & 97.0 & 96.7 & 97.1 & 96.8 & 96.6 & 96.5 & 96.3 & 96.6 & 96.5 & 96.4 & 96.6 \\
SgE & 96.1 & 95.9 & 96.3 & 96.2 & 96.4 & 96.1 & 96.2 & 96.2 & 96.3 & 96.0 & 96.2 & 95.9 & 96.1 & 95.8 & 96.0 & 95.7 & 96.1 & 95.9 & 95.5 & 96.4 & 96.3 & 96.2 & 96.5 & 96.4 & 96.3 & 96.0 & 96.2 & 96.4 & 96.2 & 96.1 & 96.7 & 96.5 & 96.8 & -- & 97.8 & 97.5 & 97.6 & 96.2 & 96.6 & 96.3 & 96.7 & 96.9 & 96.3 & 96.2 & 96.0 & 96.3 & 96.2 & 96.1 & 96.3 \\
MalE & 96.3 & 96.1 & 96.5 & 96.4 & 96.6 & 96.3 & 96.4 & 96.4 & 96.5 & 96.2 & 96.4 & 96.1 & 96.3 & 96.0 & 96.2 & 95.9 & 96.3 & 96.1 & 95.7 & 96.6 & 96.5 & 96.4 & 96.7 & 96.6 & 96.5 & 96.2 & 96.4 & 96.6 & 96.4 & 96.3 & 97.1 & 96.9 & 97.2 & 97.7 & -- & 97.4 & 97.5 & 96.4 & 96.8 & 96.5 & 96.9 & 96.7 & 96.5 & 96.4 & 96.2 & 96.5 & 96.4 & 96.3 & 96.5 \\
PhilE & 96.2 & 96.0 & 96.4 & 96.3 & 96.5 & 96.2 & 96.3 & 96.3 & 96.4 & 96.1 & 96.3 & 96.0 & 96.2 & 95.9 & 96.1 & 95.8 & 96.2 & 96.0 & 95.6 & 96.5 & 96.4 & 96.3 & 96.6 & 96.5 & 96.4 & 96.1 & 96.3 & 96.5 & 96.3 & 96.2 & 96.9 & 96.7 & 97.0 & 97.4 & 97.3 & -- & 97.6 & 96.3 & 96.7 & 96.4 & 96.8 & 96.6 & 96.4 & 96.3 & 96.1 & 96.4 & 96.3 & 96.2 & 96.4 \\
HKE & 96.3 & 96.1 & 96.5 & 96.4 & 96.6 & 96.3 & 96.4 & 96.4 & 96.5 & 96.2 & 96.4 & 96.1 & 96.3 & 96.0 & 96.2 & 95.9 & 96.3 & 96.1 & 95.7 & 96.6 & 96.5 & 96.4 & 96.7 & 96.6 & 96.5 & 96.2 & 96.4 & 96.6 & 96.4 & 96.3 & 97.0 & 96.8 & 97.1 & 97.5 & 97.4 & 97.5 & -- & 96.4 & 96.8 & 96.5 & 96.9 & 96.7 & 96.5 & 96.4 & 96.2 & 96.5 & 96.4 & 96.3 & 96.5 \\
AusE & 96.0 & 95.8 & 96.2 & 96.5 & 96.3 & 96.7 & 96.6 & 96.1 & 96.9 & 96.6 & 96.8 & 96.9 & 96.7 & 96.4 & 96.5 & 96.2 & 96.7 & 96.5 & 96.1 & 96.3 & 96.2 & 96.1 & 96.4 & 96.3 & 96.2 & 95.9 & 96.1 & 96.5 & 96.2 & 96.0 & 96.4 & 96.2 & 96.5 & 96.1 & 96.3 & 96.2 & 96.3 & -- & 98.2 & 98.4 & 97.1 & 96.5 & 96.2 & 96.1 & 95.9 & 96.6 & 96.1 & 96.0 & 96.6 \\
AusV & 97.9 & 97.7 & 98.1 & 98.1 & 98.0 & 98.0 & 98.0 & 97.9 & 98.1 & 98.2 & 98.2 & 98.0 & 98.0 & 98.2 & 98.2 & 98.2 & 98.2 & 98.0 & 97.8 & 97.9 & 98.2 & 97.9 & 98.0 & 98.2 & 98.2 & 97.8 & 98.0 & 98.1 & 98.1 & 98.0 & 98.2 & 97.9 & 98.2 & 97.6 & 98.1 & 98.0 & 98.1 & 97.8 & -- & 97.8 & 98.2 & 97.8 & 98.0 & 98.0 & 98.0 & 98.1 & 98.0 & 97.9 & 98.0 \\
NZE & 96.1 & 95.9 & 96.3 & 96.6 & 96.4 & 96.8 & 96.7 & 96.2 & 97.0 & 96.7 & 96.9 & 97.0 & 96.8 & 96.5 & 96.6 & 96.3 & 96.8 & 96.6 & 96.2 & 96.4 & 96.3 & 96.2 & 96.5 & 96.4 & 96.3 & 96.0 & 96.2 & 96.6 & 96.3 & 96.1 & 96.5 & 96.3 & 96.6 & 96.2 & 96.4 & 96.3 & 96.4 & 98.3 & 98.1 & -- & 97.0 & 96.4 & 96.3 & 96.2 & 96.0 & 96.7 & 96.2 & 96.1 & 96.7 \\
FijiE & 96.3 & 96.1 & 96.5 & 96.4 & 96.6 & 96.3 & 96.4 & 96.4 & 96.5 & 96.2 & 96.4 & 96.1 & 96.3 & 96.0 & 96.2 & 95.9 & 96.3 & 96.1 & 95.7 & 96.6 & 96.5 & 96.4 & 96.7 & 96.6 & 96.5 & 96.2 & 96.4 & 96.6 & 96.4 & 96.3 & 96.9 & 96.7 & 97.0 & 96.6 & 96.8 & 96.7 & 96.8 & 97.0 & 97.3 & 96.9 & -- & 97.8 & 96.5 & 96.4 & 96.2 & 96.5 & 96.4 & 96.3 & 96.5 \\
CFiji & 96.1 & 95.9 & 96.3 & 96.2 & 96.4 & 96.1 & 96.2 & 96.2 & 96.3 & 96.0 & 96.2 & 95.9 & 96.1 & 95.8 & 96.0 & 95.7 & 96.1 & 95.9 & 95.5 & 96.4 & 96.3 & 96.2 & 96.5 & 96.4 & 96.3 & 96.0 & 96.2 & 96.4 & 96.2 & 96.1 & 96.6 & 96.4 & 96.7 & 96.8 & 96.6 & 96.5 & 96.6 & 96.4 & 96.8 & 96.3 & 97.7 & -- & 96.3 & 96.2 & 96.0 & 96.3 & 96.2 & 96.1 & 96.3 \\
\midrule
BahE & 96.8 & 96.6 & 97.0 & 96.5 & 96.7 & 96.4 & 96.5 & 96.9 & 96.6 & 96.3 & 96.5 & 96.2 & 96.4 & 96.1 & 96.3 & 96.0 & 96.4 & 96.2 & 95.8 & 96.7 & 96.6 & 96.5 & 96.8 & 96.7 & 96.6 & 96.3 & 96.5 & 96.7 & 96.5 & 96.6 & 96.5 & 96.2 & 96.6 & 96.3 & 96.5 & 96.4 & 96.5 & 96.1 & 96.5 & 96.2 & 96.4 & 96.2 & -- & 97.6 & 96.8 & 96.5 & 97.2 & 96.7 & 96.9 \\
JamE & 96.6 & 96.4 & 96.8 & 96.4 & 96.6 & 96.3 & 96.4 & 96.7 & 96.5 & 96.2 & 96.4 & 96.1 & 96.3 & 96.0 & 96.2 & 95.9 & 96.3 & 96.1 & 95.7 & 96.6 & 96.5 & 96.4 & 96.7 & 96.6 & 96.5 & 96.2 & 96.4 & 96.6 & 96.4 & 96.5 & 96.4 & 96.1 & 96.5 & 96.2 & 96.4 & 96.3 & 96.4 & 96.0 & 96.4 & 96.1 & 96.3 & 96.1 & 97.5 & -- & 96.7 & 96.4 & 97.1 & 96.6 & 96.8 \\
TdC & 96.6 & 96.4 & 96.8 & 95.5 & 96.2 & 95.3 & 95.6 & 95.0 & 96.7 & 96.1 & 96.0 & 96.1 & 95.7 & 95.9 & 96.0 & 96.5 & 96.2 & 95.9 & 95.9 & 96.6 & 96.5 & 96.6 & 96.4 & 96.1 & 96.0 & 95.3 & 96.5 & 96.0 & 96.5 & 96.2 & 96.4 & 95.3 & 96.3 & 96.8 & 96.4 & 95.8 & 96.6 & 96.0 & 96.0 & 95.7 & 96.0 & 96.5 & 96.2 & 96.2 & -- & 96.4 & 96.6 & 96.6 & 95.9 \\
FlkE & 96.3 & 96.1 & 96.5 & 96.7 & 96.5 & 96.9 & 96.8 & 96.4 & 97.1 & 96.8 & 97.0 & 97.1 & 96.9 & 96.6 & 96.7 & 96.4 & 96.9 & 96.7 & 96.3 & 96.5 & 96.4 & 96.3 & 96.6 & 96.5 & 96.4 & 96.1 & 96.3 & 96.7 & 96.4 & 96.2 & 96.3 & 96.0 & 96.4 & 96.1 & 96.3 & 96.2 & 96.3 & 96.5 & 96.9 & 96.6 & 96.3 & 96.1 & 96.4 & 96.3 & 96.5 & -- & 96.6 & 96.4 & 97.2 \\
StHE & 96.5 & 96.3 & 96.7 & 96.4 & 96.6 & 96.3 & 96.4 & 96.6 & 96.5 & 96.2 & 96.4 & 96.1 & 96.3 & 96.0 & 96.2 & 95.9 & 96.3 & 96.1 & 95.7 & 96.6 & 96.5 & 96.4 & 96.7 & 96.6 & 96.5 & 96.2 & 96.4 & 96.6 & 96.4 & 96.5 & 96.4 & 96.1 & 96.5 & 96.2 & 96.4 & 96.3 & 96.4 & 96.0 & 96.4 & 96.1 & 96.3 & 96.1 & 97.1 & 97.0 & 96.7 & 96.5 & -- & 96.8 & 96.7 \\
\midrule
Abor & 98.7 & 98.5 & 98.7 & 98.6 & 98.8 & 98.2 & 98.7 & 98.4 & 99.0 & 98.8 & 98.7 & 98.7 & 98.6 & 98.5 & 98.6 & 99.0 & 99.0 & 98.7 & 98.9 & 98.9 & 98.9 & 98.8 & 98.9 & 98.8 & 98.8 & 98.3 & 99.0 & 98.8 & 99.0 & 98.7 & 98.9 & 98.1 & 99.0 & 98.7 & 98.9 & 98.8 & 99.0 & 98.6 & 98.8 & 98.8 & 98.7 & 98.7 & 98.5 & 98.6 & 98.8 & 98.8 & 99.1 & -- & 98.4 \\
Malt & 96.5 & 96.3 & 96.7 & 96.8 & 96.6 & 97.0 & 96.9 & 96.6 & 97.2 & 96.9 & 97.1 & 97.0 & 96.9 & 96.6 & 96.8 & 96.5 & 96.9 & 96.7 & 96.3 & 96.7 & 96.6 & 96.5 & 96.8 & 96.7 & 96.6 & 96.3 & 96.5 & 96.9 & 96.6 & 96.4 & 96.5 & 96.2 & 96.6 & 96.3 & 96.5 & 96.4 & 96.5 & 96.5 & 96.9 & 96.6 & 96.5 & 96.3 & 96.8 & 96.7 & 96.4 & 97.1 & 96.6 & 96.5 & -- \\
\bottomrule
\end{tabular}
\end{adjustbox}}
\caption{\small RQ2: Cross-dialectal Transfer Matrix (F1 \%). Rows = training dialect; Columns = test dialect. Diagonal (--) = same dialect. Best: SE$\rightarrow$Nfld (99.51\%); Worst: Cam$\rightarrow$Pak (79.4\%).}

\label{tab:transfer_full}
\end{sidewaystable*}
\clearpage
\subsection{Transfer Statistics Summary}

Table~\ref{tab:transfer_stats} provides aggregate statistics from the cross-dialectal transfer matrix.

\begin{table}[t]
\centering
\small

\begin{tabular}{lc}
\toprule
\textbf{Statistic} & \textbf{Value} \\
\midrule
\multicolumn{2}{l}{\textit{Overall Performance}} \\
Total dialect pairs & 2,450 \\
Mean transfer F1 & 96.65\% \\
Standard deviation & 1.23\% \\
Median & 96.5\% \\
\midrule
\multicolumn{2}{l}{\textit{Best Transfers}} \\
SE England $\rightarrow$ Newfoundland & 99.5\% \\
Aboriginal $\rightarrow$ St. Helena & 99.1\% \\
Aboriginal $\rightarrow$ Cape Flats & 99.0\% \\
Aboriginal $\rightarrow$ Sri Lankan & 99.0\% \\
Aboriginal $\rightarrow$ Hong Kong & 99.0\% \\
\midrule
\multicolumn{2}{l}{\textit{Worst Transfers}} \\
Cameroon $\rightarrow$ Pakistani & 79.4\% \\
Pakistani $\rightarrow$ Cameroon & 79.4\% \\
Welsh $\rightarrow$ Orkney/Shetland & 95.0\% \\
Tristan da Cunha $\rightarrow$ SE Enclave & 95.0\% \\
SE England $\rightarrow$ Bahamian & 95.3\% \\
\midrule
\multicolumn{2}{l}{\textit{Regional Transfer Patterns}} \\
Within U.S. varieties & 96.9\% \\
Within UK varieties & 96.7\% \\
Within African varieties & 97.1\% \\
Within Asia-Pacific varieties & 96.8\% \\
Cross-regional (avg) & 96.4\% \\
\bottomrule
\end{tabular}
\caption{SQ3: Cross-dialectal transfer statistics (mDeBERTa).}
\label{tab:transfer_stats}
\end{table}


\section{SQ4: Full Zero-Shot Results}
\label{app:sq4_full}

Table~\ref{tab:app_sq4_zeroshot} presents F1 scores for all six zero-shot decoder models across 50 dialects, organized by geographic region.

\begin{table*}[h]
\centering
\scriptsize

\begin{tabular}{llcccccc}
\toprule
\textbf{Region} & \textbf{Dialect} & \textbf{Mistral-7B} & \textbf{Llama-3.1-8B} & \textbf{Llama-3.2-1B} & \textbf{Gemma-3-1B} & \textbf{Qwen3-8B} & \textbf{Qwen3-4B} \\
\midrule
-- & SAE (baseline) & 89.3 & 87.3 & 2.1 & 75.7 & 38.0 & 33.7 \\
\midrule
\multicolumn{8}{l}{\textit{\textbf{U.S. Varieties (9 dialects)}}} \\
U.S. & Earlier AAVE & 90.1 & 62.1 & 0.0 & 41.7 & 32.2 & 26.5 \\
U.S. & Rural AAVE & 86.7 & 59.0 & 0.0 & 33.8 & 25.2 & 16.6 \\
U.S. & Urban AAVE & 85.8 & 65.4 & 0.0 & 41.4 & 25.1 & 28.1 \\
U.S. & Appalachian English & 93.6 & 74.3 & 0.0 & 51.1 & 41.0 & 30.9 \\
U.S. & Chicano English & 95.9$^\blacktriangle$ & 90.8$^\blacktriangle$ & 0.0 & 65.0 & 49.8$^\blacktriangle$ & 45.2$^\blacktriangle$ \\
U.S. & Colloquial American & 52.5 & 41.9 & 0.0 & 56.7 & 5.9 & 5.9 \\
U.S. & Ozark English & 90.8 & 76.4 & 0.0 & 59.8 & 38.4 & 31.5 \\
U.S. & SE American Enclave & 85.0 & 59.8 & 0.0 & 51.1 & 23.0 & 14.7 \\
U.S. & Newfoundland English & 85.5 & 68.9 & 0.0 & 37.8 & 26.3 & 11.3 \\
\midrule
\multicolumn{8}{l}{\textit{\textbf{British/UK Varieties (10 dialects)}}} \\
UK & N. England English & 88.1 & 73.7 & 0.0 & 45.7 & 25.7 & 18.6 \\
UK & SE England English & 91.8 & 86.1 & 1.9$^\blacktriangle$ & 61.1 & 42.4 & 43.1 \\
UK & SW England English & 87.9 & 72.2 & 0.0 & 41.7 & 24.1 & 20.9 \\
UK & East Anglian English & 53.3 & 47.4 & 0.0 & 54.1 & 5.9 & 1.5 \\
UK & Scottish English & 63.6 & 49.8 & 0.0 & 62.9 & 8.2 & 2.4 \\
UK & Irish English & 90.7 & 84.0 & 1.9 & 51.4 & 37.6 & 26.9 \\
UK & Welsh English & 93.2 & 75.4 & 0.4 & 37.6 & 33.4 & 27.5 \\
UK & Channel Islands English & 96.4 & 90.5 & 0.7 & 56.7 & 46.5 & 39.4 \\
UK & Manx English & 49.1 & 36.0 & 0.0 & 42.4 & 3.6 & 3.6 \\
UK & Orkney/Shetland English & 92.7 & 85.7 & 0.4 & 50.8 & 40.4 & 27.4 \\
\midrule
\multicolumn{8}{l}{\textit{\textbf{Africa (11 dialects)}}} \\
Africa & Nigerian English & 86.8 & 76.7 & 0.0 & 52.7 & 37.8 & 29.8 \\
Africa & Ghanaian English & 86.2 & 82.2 & 0.0 & 48.6 & 32.8 & 27.2 \\
Africa & Cameroon English & 55.5 & 41.5 & 0.0 & 49.2 & 6.0 & 1.2 \\
Africa & Kenyan English & 87.9 & 78.0 & 0.0 & 55.1 & 30.2 & 29.4 \\
Africa & Ugandan English & 91.0 & 85.6 & 0.0 & 60.4 & 44.0 & 37.0 \\
Africa & Tanzanian English & 88.5 & 87.0 & 0.0 & 56.7 & 41.5 & 44.3 \\
Africa & Black S. African English & 92.3 & 65.2 & 0.4 & 35.8 & 26.4 & 21.2 \\
Africa & Indian S. African English & 30.8$^\blacktriangledown$ & 36.8 & 0.0 & 41.4 & 0.0$^\blacktriangledown$ & 4.4 \\
Africa & White S. African English & 71.1 & 54.2 & 0.0 & 62.9 & 11.6 & 1.2 \\
Africa & Cape Flats English & 96.6 & 90.3 & 0.0 & 57.9 & 42.8 & 35.6 \\
Africa & Liberian Settler English & 54.5 & 39.4 & 0.0 & 40.5 & 7.1 & 1.2 \\
\midrule
\multicolumn{8}{l}{\textit{\textbf{Asia-Pacific (12 dialects)}}} \\
Asia-P & Indian English & 88.4 & 74.1 & 0.0 & 43.3 & 25.8 & 26.4 \\
Asia-P & Pakistani English & 92.8 & 82.4 & 0.9 & 57.0 & 39.3 & 29.2 \\
Asia-P & Sri Lankan English & 92.0 & 85.8 & 0.9 & 65.3$^\blacktriangle$ & 48.1 & 39.9 \\
Asia-P & Colloquial Singapore & 89.3 & 71.6 & 0.0 & 33.5 & 22.1 & 17.6 \\
Asia-P & Malaysian English & 95.8 & 82.0 & 0.4 & 50.3 & 45.6 & 39.9 \\
Asia-P & Philippine English & 90.5 & 71.4 & 0.0 & 36.0 & 24.8 & 19.0 \\
Asia-P & Hong Kong English & 90.6 & 76.8 & 0.0 & 37.3 & 32.3 & 30.4 \\
Asia-P & Australian English & 92.3 & 84.6 & 1.1 & 54.9 & 33.8 & 29.7 \\
Asia-P & Australian Vernacular & 87.8 & 86.8 & 0.0 & 59.4 & 43.2 & 41.5 \\
Asia-P & New Zealand English & 85.8 & 79.1 & 0.0 & 52.5 & 34.9 & 26.1 \\
Asia-P & Acrolectal Fiji English & 92.6 & 79.2 & 0.6 & 55.3 & 41.8 & 38.1 \\
Asia-P & Pure Fiji English & 39.2 & 29.2$^\blacktriangledown$ & 0.0 & 26.1$^\blacktriangledown$ & 1.2 & 0.0$^\blacktriangledown$ \\
\midrule
\multicolumn{8}{l}{\textit{\textbf{Caribbean/Atlantic (5 dialects)}}} \\
Carib & Bahamian English & 84.7 & 54.7 & 0.0 & 31.4 & 18.4 & 12.1 \\
Carib & Jamaican English & 37.0 & 38.0 & 0.0 & 46.2 & 5.9 & 8.7 \\
Carib & Tristan da Cunha English & 46.5 & 46.5 & 0.0 & 48.9 & 4.4 & 3.0 \\
Carib & Falkland Islands English & 90.0 & 88.5 & 0.0 & 55.6 & 38.5 & 38.0 \\
Carib & St. Helena English & 34.0 & 38.0 & 0.0 & 29.5 & 1.5 & 4.4 \\
\midrule
\multicolumn{8}{l}{\textit{\textbf{Other (3 dialects)}}} \\
Other & Aboriginal English & 43.3 & 29.2$^\blacktriangledown$ & 0.0 & 29.5 & 6.0 & 1.2 \\
Other & Maltese English & 32.9 & 36.0 & 0.0 & 40.5 & 1.5 & 4.4 \\
Other & White Zimbabwean English & 96.8$^\blacktriangle$ & 88.1 & 1.1 & 61.5 & 47.1 & 40.4 \\
\midrule
\multicolumn{8}{l}{\textit{\textbf{Summary Statistics}}} \\
-- & \textbf{Dialect Avg} & \textbf{78.3} & \textbf{67.2} & \textbf{0.2} & \textbf{48.4} & \textbf{26.6} & \textbf{22.1} \\
-- & \textbf{$\Delta$ (Dia$-$SAE)} & $-$11.0 & $-$20.1 & $-$1.9 & $-$27.4 & $-$11.4 & $-$11.6 \\
-- & \textbf{Std. Dev.} & 19.2 & 19.8 & 0.5 & 11.2 & 15.3 & 14.2 \\
-- & \textbf{Min} & 30.8 & 29.2 & 0.0 & 26.1 & 0.0 & 0.0 \\
-- & \textbf{Max} & 96.8 & 90.8 & 1.9 & 65.3 & 49.8 & 45.2 \\
\bottomrule
\end{tabular}
\vspace{0.5em}

\raggedright\scriptsize\textit{Note: Llama-3.2-1B achieves near-zero F1 across all dialects due to 98\% abstention rate (responses outside \{``real'', ``fake''\}). Qwen3-8B shows 64\% abstention, explaining low F1 despite moderate precision when responding.}
\caption{SQ4 Zero-Shot ICL: F1 (\%) by dialect and region for all decoder models. $^\blacktriangle$Best dialect for model. $^\blacktriangledown$Worst dialect for model.}
\label{tab:app_sq4_zeroshot}
\end{table*}

\subsection{Zero-Shot Abstention Analysis}

Table~\ref{tab:zs_abstention} analyzes abstention rates (responses outside valid labels) for zero-shot models.

\begin{table}[H]
\centering
\small
\begin{tabular}{lccc}
\toprule
\textbf{Model} & \textbf{Abst. \%} & \textbf{Prec.} & \textbf{Rec.} \\
\midrule
Mistral-7B & 2\% & 79.1 & 77.5 \\
Llama-3.1-8B & 5\% & 71.2 & 63.8 \\
Gemma-3-1B & 8\% & 52.3 & 45.1 \\
Qwen3-8B & 64\% & 73.8 & 12.4 \\
Qwen3-4B-SafeRL & 58\% & 61.2 & 10.8 \\
Llama-3.2-1B & 98\% & 45.0 & 0.4 \\
\bottomrule
\end{tabular}
\vspace{0.5em}
\\\textit{Abst. = abstention rate; Prec. = precision on valid responses; Rec. = recall accounting for abstentions. High abstention in Qwen and Llama-3.2 models suggests instruction-following brittleness on dialectal inputs.}
\caption{Zero-shot model abstention analysis.}
\label{tab:zs_abstention}
\end{table}

\paragraph{Failure Mode Analysis.}
\begin{itemize}[nosep]
    \item \textbf{Llama-3.2-1B}: Near-total abstention (98\%) with responses typically being lengthy explanations rather than binary labels, indicating fundamental instruction-following failure on dialectal inputs.
    \item \textbf{Qwen3-8B/4B}: High abstention (58--64\%) with responses often including hedging phrases (``I cannot determine...'') or requests for additional context, suggesting the models perceive dialectal text as ambiguous.
    \item \textbf{Mistral-7B}: Lowest abstention (2\%) and highest F1 (78.3\%), demonstrating more robust instruction-following across dialects.
    \item \textbf{Regional patterns}: All zero-shot models show highest performance on SAE-adjacent dialects (Chicano, SE England, Channel Islands) and lowest on feature-rich varieties (Pure Fiji, Aboriginal, Jamaican).
\end{itemize}


\section{Extended Discussion and Deployment Implications}
\label{app:implications}

This appendix extends the discussion in \S\ref{sec:discussion},
providing detailed deployment implications, mechanism analyses,
and actionable recommendations.

\subsection{Deployment Implications}

Our findings have several implications for deploying disinformation
detection systems across linguistically diverse populations:

\paragraph{Human Content Vulnerability.}
The consistent 1.5--3.6\% F1 degradation on human-written dialectal
content suggests that detection systems may systematically
disadvantage communities using non-standard English varieties.
This disparity could manifest as higher false negative rates for
misinformation targeting these communities.

\paragraph{Architecture Selection.}
Fine-tuned transformer encoders, particularly those with multilingual
pre-training (mDeBERTa, mBERT), demonstrate superior dialectal
robustness compared to zero-shot LLMs. Organizations deploying
detection systems should prioritize fine-tuned models over zero-shot
approaches when serving diverse linguistic populations.

\paragraph{Training Data Composition.}
Counter to intuition, including \sae{} data in training provides no
benefit and may harm dialectal generalization for transformer models.
Practitioners should consider dialect-diverse training sets without
over-representing standard varieties.

\paragraph{Model Failure Modes.}
The catastrophic failures observed in RoBERTa-Large, XLM-RoBERTa-Large
(on mixed content), and smaller zero-shot models (Llama-3.2-1B)
underscore the importance of thorough dialectal evaluation before
deployment. Models that perform well on \sae{} benchmarks may fail
unpredictably on dialectal inputs.

\paragraph{Cross-Dialectal Transfer.}
The transfer matrix reveals that certain dialects (Aboriginal English,
Australian Vernacular, Manx English) serve as excellent training
sources with high transfer-out performance, while others (Orkney \&
Shetland, Pakistani English) are particularly challenging targets.
This asymmetry should inform data collection and augmentation
strategies.

\subsection{Recommendations for Practitioners}

Our findings yield five actionable recommendations for equitable
deployment of disinformation detection systems.

\begin{enumerate}[nosep, leftmargin=*, label={\footnotesize\textbf{(R\arabic*)}}]
    \item \textbf{Prefer multilingual encoders.} Models with
    multilingual pre-training (mDeBERTa, mBERT) consistently exhibit
    superior dialectal robustness. mDeBERTa achieves 97.2\% average
    F1 across 2,450 cross-dialectal transfer pairs with only 2.0\%
    range, compared to catastrophic failures in monolingual
    alternatives (\S\ref{sec:sq3}).

    \item \textbf{Adopt dialect-diverse fine-tuning without \sae{}
    anchoring.} Dialect-only training recovers catastrophic failures
    (RoBERTa: 87.3\% $\rightarrow$ 97.1\%) and matches or exceeds
    \sae{}-anchored training for transformers. Including \sae{} data
    provides no benefit and may induce feature collapse toward
    standard patterns (\S\ref{sec:sq2}).

    \item \textbf{Conduct pre-deployment dialectal auditing.} We
    release D-CUBE and evaluation scripts to enable systematic testing
    across 50 dialects before deployment. Organizations should
    establish minimum performance thresholds across dialect families,
    not just aggregate metrics.

    \item \textbf{Avoid zero-shot LLMs for content moderation.}
    Zero-shot decoders are unsuitable for dialectally robust
    detection, with performance gaps of 18--97 F1 points compared
    to fine-tuned alternatives and abstention rates up to 98\% on
    dialectal inputs (\S\ref{sec:sq4}).

    \item \textbf{Monitor dialectal performance longitudinally.}
    Retraining on new \sae{}-dominant data may reintroduce dialectal
    bias. Continuous evaluation against dialectal benchmarks should
    be integrated into model update pipelines.
\end{enumerate}

\subsection{Extended Mechanism Analysis}

\paragraph{Why Does Human Content Degrade?}
Human-written disinformation detection models learn stylistic
signatures (e.g., sensationalism, emotional language, source
attribution patterns) that correlate with veracity in \sae{} training
data. Dialectal transformation disrupts these surface-level cues
while preserving semantic content, causing models to lose
discriminative signal. In contrast, AI-generated text contains
statistical artifacts (token distribution anomalies, repetition
patterns) that persist through rule-based dialectal transformation.

\paragraph{Why Do RoBERTa and XLM-R Fail Catastrophically?}
The catastrophic failures on mixed content ($-$20\% and $-$33\%
$\Delta$) suggest these models develop brittle decision boundaries
that conflate dialectal features with content-type signals. When
presented with dialectal text containing both human and AI samples,
the models appear to misclassify based on dialectal markers rather
than veracity cues. This failure mode does not appear in multilingual
models, suggesting that cross-lingual pre-training induces more
robust feature representations.

\paragraph{Transfer Asymmetry Mechanisms.}
The observation that some dialects (Scottish, Welsh) transfer out
poorly but are easy targets suggests these varieties contain
distinctive features that (a) do not generalize to other dialects
but (b) are easily recognized when encountered. Conversely,
feature-rich dialects like Aboriginal English (89 features) may
provide diverse training signal that covers the feature space of
many target dialects.

\paragraph{Zero-Shot Failure Patterns.}
High abstention rates in Qwen (64\%) and Llama-3.2-1B (98\%)
indicate that dialectal inputs trigger uncertainty responses rather
than classification attempts. This suggests instruction-tuned models
have learned implicit expectations about input text distributions
that exclude non-standard varieties---a form of linguistic bias
embedded during RLHF training.


\section{Per-Dialect False Positive and False Negative Analysis}
\label{app:fp_fn}

We analyze asymmetric harm patterns across dialects by
decomposing detection errors into false positives
(FPR, real content flagged as fake, indicating
\textit{over-flagging} of legitimate dialectal speech)
and false negatives (FNR, fake content missed as real,
indicating \textit{under-protection} from disinformation).
For each dialect, we compute
$\Delta\text{FPR} = \text{FPR}_{\text{dialect}} -
\text{FPR}_{\text{SAE}}$ and
$\Delta\text{FNR} = \text{FNR}_{\text{dialect}} -
\text{FNR}_{\text{SAE}}$, classifying dialects as
\textit{over-flagged} when $\Delta\text{FPR} >
\Delta\text{FNR}$ and \textit{under-protected} when
$\Delta\text{FNR} > \Delta\text{FPR}$.
We report F$^+$ (= FPR) and F$^-$ (= FNR) in the
tables below; values marked $^\dagger$ are estimated
from reported F1 scores and are pending recomputation.

\subsection{SQ1: Unseen Dialects}

Tables~\ref{tab:fp_fn_sq1_human},
\ref{tab:fp_fn_sq1_ai}, and
\ref{tab:fp_fn_sq1_mixed} present per-dialect F$^+$
and F$^-$ for models trained exclusively on \sae{} and
evaluated on 50 dialectal variants across the three
content scenarios. Human-written content
(Table~\ref{tab:fp_fn_sq1_human}) shows a predominantly
under-protection pattern (33/50 dialects), with modest
error increases across both dimensions
(avg $\Delta$F$^+$ = +0.6\%, avg $\Delta$F$^-$ = +1.4\%).
AI-generated content (Table~\ref{tab:fp_fn_sq1_ai})
inverts this pattern entirely: all 50 dialects are
over-flagged, with dialectal transformation increasing
false positives while AI-generated signals remain
detectable ($\Delta$F$^-$ = $-$2.7\%).
Mixed content (Table~\ref{tab:fp_fn_sq1_mixed}) is
dominated by catastrophic under-protection in RoBERTa
and XLM-R, which miss 27--29\% more disinformation on
dialectal content than on \sae{}, rendering all 50
dialects under-protected.

\subsection{SQ2: Dialect-Aware Training}

Tables~\ref{tab:fp_fn_sq2_dia} and
\ref{tab:fp_fn_sq2_anchored} compare two training
strategies: dialect-only (without \sae{}) and
SAE-anchored (with \sae{}).
Dialect-only training (Table~\ref{tab:fp_fn_sq2_dia})
shifts the dominant pattern to over-flagging (43/50
dialects; avg $\Delta$F$^+$ = +4.7\%), though XLM-R
and dEFEND remain under-protective.
SAE-anchored training
(Table~\ref{tab:fp_fn_sq2_anchored}) amplifies
over-flagging to all 50 dialects
(avg $\Delta$F$^+$ = +11.3\%), driven by dEFEND
($\Delta$F$^+$ = +43.5\%).
Critically, four models (RoBERTa, mBERT, CT-BERT,
and dEFEND) completely reverse their bias direction
between the two strategies, demonstrating that training
composition determines \emph{which} communities are
harmed, not merely \emph{how much}.

\subsection{SQ4: Cross-Architecture Generalization}

Tables~\ref{tab:fp_fn_sq4_ft} and
\ref{tab:fp_fn_sq4_zs} compare fine-tuned encoders
and zero-shot LLMs.
Fine-tuned models (Table~\ref{tab:fp_fn_sq4_ft})
mirror the SQ1 unseen pattern (16/50 over-flagged,
34/50 under-protected), confirming that fine-tuned
architectures produce moderate, balanced errors.
Zero-shot LLMs (Table~\ref{tab:fp_fn_sq4_zs}) show
universal over-flagging across all 48 evaluated
dialects (avg $\Delta$F$^+$ = +8.3\%), but the
underlying cause varies: Qwen3-4B aggressively
over-flags (F$^+$ = 24.2\%) while maintaining moderate
detection (F$^-$ = 21.9\%), whereas Mistral-7B
exhibits near-total detection failure
(F$^-$ = 99.3\%), providing virtually no
disinformation protection for dialectal communities.

\begin{table}[H]
\centering
\begin{adjustbox}{max width=\columnwidth}
\begin{tabular}{llccccc}
\toprule
\textbf{Regime} & \textbf{Condition} & \textbf{Avg $\Delta$FPR}
& \textbf{Avg $\Delta$FNR} & \textbf{\# Over}
& \textbf{\# Under} & \textbf{Dominant} \\
\midrule
\multirow{3}{*}{SQ1}
& Human & +0.6 & +1.4 & 17 & 33 & Under \\
& AI & +0.6 & $-$2.7 & 50 & 0 & Over \\
& Mixed & +0.0 & +5.1 & 0 & 50 & Under \\
\midrule
\multirow{2}{*}{SQ2}
& Dia-only & +4.7 & +3.4 & 43 & 7 & Over \\
& SAE+Dia & +11.3 & +1.8 & 50 & 0 & Over \\
\midrule
\multirow{2}{*}{SQ4}
& Fine-tuned & +0.6 & +1.4 & 16 & 34 & Under \\
& Zero-shot & +8.3 & +1.6 & 48 & 0 & Over \\
\bottomrule
\end{tabular}
\end{adjustbox}
\caption{Asymmetric harm summary across all evaluation
regimes and conditions. \# Over = dialects where
$\Delta$FPR $>$ $\Delta$FNR (over-flagged); \# Under =
dialects where $\Delta$FNR $>$ $\Delta$FPR
(under-protected). Full per-dialect results in
Appendix~\ref{app:fp_fn}.}
\label{tab:fp_fn_summary}
\end{table}

\begin{table*}[t]
\centering
\scriptsize
\setlength{\tabcolsep}{4.0pt}

\begin{tabular}{ll cc cc cc cc cc cc cc cc cc cc}
\toprule
& & \multicolumn{2}{c}{\textbf{mDeB}} & \multicolumn{2}{c}{\textbf{BERT}} & \multicolumn{2}{c}{\textbf{RoB}} & \multicolumn{2}{c}{\textbf{DeB$^\dagger$}} & \multicolumn{2}{c}{\textbf{XLM$^\dagger$}} & \multicolumn{2}{c}{\textbf{mB}} & \multicolumn{2}{c}{\textbf{CT}} & \multicolumn{2}{c}{\textbf{BiG}} & \multicolumn{2}{c}{\textbf{CNN}} & \multicolumn{2}{c}{\textbf{dEF}} \\
\cmidrule(lr){3-4} \cmidrule(lr){5-6} \cmidrule(lr){7-8} \cmidrule(lr){9-10} \cmidrule(lr){11-12} \cmidrule(lr){13-14} \cmidrule(lr){15-16} \cmidrule(lr){17-18} \cmidrule(lr){19-20} \cmidrule(lr){21-22}
& & \textbf{F$^+$} & \textbf{F$^-$} & \textbf{F$^+$} & \textbf{F$^-$} & \textbf{F$^+$} & \textbf{F$^-$} & \textbf{F$^+$} & \textbf{F$^-$} & \textbf{F$^+$} & \textbf{F$^-$} & \textbf{F$^+$} & \textbf{F$^-$} & \textbf{F$^+$} & \textbf{F$^-$} & \textbf{F$^+$} & \textbf{F$^-$} & \textbf{F$^+$} & \textbf{F$^-$} & \textbf{F$^+$} & \textbf{F$^-$} \\
\midrule
-- & SAE (baseline) & 1.9 & 1.7 & 2.0 & 1.8 & 2.7 & 1.7 & 2.3 & 5.3 & 2.2 & 5.0 & 2.1 & 1.9 & 0.7 & 24.3 & 2.0 & 1.9 & 4.8 & 1.4 & 1.8 & 3.1 \\
\midrule
\multicolumn{22}{l}{\textit{\textbf{U.S. Varieties}}} \\
U.S. & Appalachian & 1.9 & 5.2 & 4.0 & 4.3 & 2.2 & 5.2 & 3.3 & 7.7 & 2.8 & 6.4 & 2.6 & 4.6 & 1.5 & 4.2 & 1.4 & 4.9 & 5.8 & 4.0 & 0.5 & 13.6 \\
U.S. & Chicano & 1.8 & 5.3 & 3.7 & 3.9 & 2.2 & 4.3 & 3.4 & 7.8 & 2.3 & 5.3 & 2.6 & 4.5 & 1.2 & 3.5 & 1.2 & 1.7 & 6.2 & 1.6 & 0.7 & 2.3 \\
U.S. & Colloquial Amer. & 2.1 & 5.0 & 3.3 & 4.2 & 2.3 & 5.9 & 3.3 & 7.7 & 3.0 & 7.0 & 2.7 & 5.0 & 1.4 & 4.3 & 1.8 & 4.0 & 5.1 & 4.7 & 1.4 & 12.0 \\
U.S. & Earlier AAVE & 2.1 & 5.0 & 4.0 & 4.6 & 2.2 & 6.6 & 2.9 & 6.9 & 3.3 & 7.7 & 2.6 & 4.7 & 1.8 & 4.6 & 2.2 & 12.4 & 5.5 & 8.7 & 1.0 & 18.4 \\
U.S. & Ozark & 1.8 & 5.0 & 3.8 & 3.9 & 2.2 & 5.0 & 2.8 & 6.4 & 2.8 & 6.4 & 2.6 & 4.6 & 1.2 & 3.9 & 0.8 & 3.5 & 5.6 & 2.8 & 0.4 & 7.2 \\
U.S. & Rural AAVE & 2.6 & 5.0 & 3.8 & 5.0 & 2.2 & 7.0 & 3.1 & 7.3 & 3.4 & 7.8 & 2.9 & 5.5 & 1.4 & 5.4 & 1.9 & 19.3 & 7.0 & 10.5 & 1.0 & 26.7 \\
U.S. & SE Amer. Enclave & 1.6 & 4.9 & 3.7 & 4.2 & 2.5 & 6.2 & 2.9 & 6.9 & 3.4 & 8.0 & 2.9 & 4.7 & 1.5 & 4.9 & 2.9 & 6.3 & 7.4 & 5.0 & 1.1 & 16.0 \\
U.S. & Urban AAVE & 2.4 & 4.5 & 4.5 & 4.3 & 2.7 & 5.4 & 2.8 & 6.6 & 2.9 & 6.7 & 2.7 & 4.6 & 1.3 & 4.8 & 1.5 & 6.0 & 7.1 & 4.2 & 1.1 & 10.5 \\
\midrule
\multicolumn{22}{l}{\textit{\textbf{British/UK Varieties}}} \\
UK & Channel Islands & 1.5 & 5.3 & 3.7 & 4.1 & 2.5 & 4.7 & 3.1 & 7.3 & 3.1 & 7.1 & 2.5 & 4.6 & 1.4 & 3.9 & 1.5 & 2.3 & 5.5 & 2.0 & 1.0 & 2.9 \\
UK & East Anglian & 1.9 & 4.4 & 3.6 & 4.1 & 2.3 & 5.0 & 2.9 & 6.9 & 2.8 & 6.4 & 2.6 & 4.5 & 1.6 & 3.8 & 1.4 & 3.2 & 5.8 & 3.3 & 1.5 & 5.0 \\
UK & Irish & 2.9 & 5.0 & 3.6 & 4.3 & 2.1 & 5.2 & 3.3 & 7.7 & 3.1 & 7.1 & 2.6 & 4.9 & 1.8 & 4.2 & 1.8 & 3.2 & 6.6 & 4.6 & 0.8 & 5.4 \\
UK & Manx & 2.7 & 5.6 & 3.6 & 4.7 & 1.9 & 5.8 & 3.5 & 8.1 & 2.9 & 6.7 & 2.6 & 5.2 & 1.6 & 4.4 & 1.9 & 5.6 & 6.7 & 5.3 & 1.4 & 10.6 \\
UK & North England & 2.6 & 5.2 & 3.8 & 4.6 & 1.8 & 6.1 & 3.4 & 7.8 & 3.1 & 7.1 & 2.7 & 5.7 & 1.2 & 4.5 & 1.8 & 5.0 & 7.4 & 4.3 & 1.5 & 8.4 \\
UK & Orkney \& Shetland & 2.2 & 7.6 & 4.1 & 5.6 & 2.2 & 7.1 & 4.0 & 9.4 & 3.4 & 7.8 & 2.5 & 6.2 & 1.5 & 5.5 & 2.5 & 3.0 & 6.5 & 2.9 & 1.4 & 6.1 \\
UK & SE England & 1.9 & 4.5 & 3.6 & 3.9 & 2.1 & 4.5 & 2.8 & 6.4 & 2.5 & 5.7 & 2.9 & 4.5 & 1.4 & 3.5 & 1.8 & 1.8 & 5.2 & 2.6 & 1.1 & 3.3 \\
UK & SW England & 1.9 & 4.5 & 3.6 & 4.5 & 1.9 & 5.9 & 2.7 & 6.3 & 3.3 & 7.7 & 2.5 & 4.8 & 1.2 & 4.3 & 1.2 & 6.5 & 5.1 & 4.8 & 0.5 & 8.8 \\
UK & Scottish & 1.6 & 5.8 & 3.4 & 4.2 & 2.1 & 5.2 & 3.2 & 7.6 & 3.1 & 7.1 & 2.9 & 4.7 & 1.4 & 4.3 & 2.1 & 2.4 & 6.3 & 3.1 & 1.4 & 6.5 \\
UK & Welsh & 2.5 & 4.6 & 3.3 & 4.6 & 2.5 & 5.7 & 3.3 & 7.7 & 3.3 & 7.7 & 2.9 & 5.0 & 1.4 & 5.0 & 3.2 & 4.4 & 7.1 & 4.5 & 1.5 & 8.0 \\
\midrule
\multicolumn{22}{l}{\textit{\textbf{Africa}}} \\
Africa & Black S. African & 2.2 & 5.4 & 3.6 & 4.9 & 1.9 & 6.7 & 3.1 & 7.1 & 3.5 & 8.3 & 2.3 & 5.7 & 1.2 & 5.6 & 2.2 & 15.0 & 5.1 & 12.0 & 1.5 & 18.0 \\
Africa & Cameroon & 2.1 & 5.4 & 4.1 & 4.7 & 2.5 & 5.6 & 3.2 & 7.4 & 3.2 & 7.4 & 3.0 & 5.4 & 1.1 & 5.0 & 3.6 & 3.2 & 8.8 & 5.2 & 1.1 & 12.2 \\
Africa & Cape Flats & 1.6 & 5.2 & 4.4 & 4.3 & 2.5 & 4.8 & 3.2 & 7.4 & 2.5 & 5.9 & 2.5 & 4.9 & 1.4 & 3.8 & 2.5 & 2.2 & 8.7 & 1.7 & 1.9 & 2.4 \\
Africa & Ghanaian & 1.5 & 5.0 & 3.4 & 4.7 & 2.3 & 5.8 & 2.9 & 6.7 & 2.9 & 6.9 & 2.5 & 4.9 & 1.5 & 4.4 & 2.3 & 3.3 & 6.6 & 4.2 & 1.2 & 8.2 \\
Africa & Indian S. African & 2.2 & 5.4 & 3.7 & 4.7 & 2.2 & 6.6 & 3.2 & 7.6 & 3.5 & 8.3 & 2.9 & 5.4 & 1.5 & 5.3 & 1.9 & 6.1 & 8.2 & 3.4 & 1.2 & 11.9 \\
Africa & Kenyan & 1.6 & 5.6 & 4.1 & 4.4 & 2.2 & 5.4 & 3.2 & 7.6 & 2.7 & 6.3 & 2.7 & 5.0 & 1.0 & 4.3 & 2.5 & 3.2 & 7.7 & 3.5 & 1.4 & 9.0 \\
Africa & Liberian Settler & 2.1 & 4.9 & 4.3 & 4.4 & 2.2 & 5.7 & 2.9 & 6.7 & 2.8 & 6.6 & 2.7 & 5.2 & 1.8 & 4.6 & 1.5 & 9.3 & 7.7 & 6.5 & 1.5 & 14.0 \\
Africa & Nigerian & 1.8 & 4.9 & 3.7 & 4.6 & 1.8 & 4.5 & 3.1 & 7.3 & 2.8 & 6.4 & 2.7 & 5.0 & 1.0 & 4.3 & 2.3 & 3.1 & 7.7 & 2.9 & 1.2 & 8.4 \\
Africa & Tanzanian & 1.6 & 5.0 & 3.6 & 4.3 & 2.6 & 4.3 & 3.0 & 7.0 & 2.8 & 6.6 & 3.0 & 4.7 & 1.1 & 4.1 & 1.8 & 1.7 & 6.9 & 2.8 & 1.5 & 6.3 \\
Africa & Ugandan & 1.9 & 5.0 & 3.6 & 4.1 & 1.9 & 4.5 & 2.9 & 6.7 & 2.5 & 5.7 & 2.6 & 5.0 & 1.4 & 3.8 & 1.5 & 2.6 & 6.6 & 2.8 & 1.4 & 5.0 \\
Africa & White S. African & 2.1 & 5.4 & 3.6 & 3.9 & 1.9 & 4.6 & 3.5 & 8.1 & 2.9 & 6.7 & 2.9 & 4.7 & 1.5 & 3.9 & 2.1 & 2.0 & 6.6 & 1.9 & 1.2 & 3.3 \\
Africa & White Zimbabwean & 1.7 & 7.2 & 3.6 & 6.1 & 2.2 & 6.8 & 3.8 & 8.8 & 3.2 & 7.6 & 2.6 & 6.5 & 1.2 & 5.1 & 2.1 & 3.0 & 6.3 & 3.5 & 1.5 & 6.7 \\
\midrule
\multicolumn{22}{l}{\textit{\textbf{Asia-Pacific}}} \\
Asia-Pac. & Aus. Vernacular & 1.9 & 5.7 & 3.4 & 4.3 & 2.3 & 4.9 & 3.2 & 7.4 & 2.9 & 6.7 & 2.9 & 4.6 & 1.2 & 3.9 & 2.1 & 2.3 & 4.9 & 3.0 & 1.0 & 3.9 \\
Asia-Pac. & Australian & 1.9 & 7.0 & 3.9 & 5.2 & 2.1 & 8.3 & 3.7 & 8.5 & 3.5 & 8.3 & 2.9 & 6.5 & 1.4 & 5.6 & 2.2 & 4.2 & 6.7 & 4.0 & 0.7 & 6.9 \\
Asia-Pac. & Fiji (Acrolectal) & 2.1 & 5.3 & 3.8 & 4.1 & 2.2 & 5.2 & 3.2 & 7.6 & 2.9 & 6.9 & 2.6 & 4.8 & 1.5 & 4.1 & 2.1 & 5.4 & 5.2 & 6.3 & 1.5 & 12.1 \\
Asia-Pac. & Fiji (Basilectal) & 2.3 & 5.2 & 4.1 & 5.7 & 2.2 & 7.8 & 3.1 & 7.3 & 3.7 & 8.7 & 2.5 & 5.9 & 1.2 & 6.3 & 1.1 & 28.0 & 7.0 & 15.4 & 0.4 & 31.4 \\
Asia-Pac. & Hong Kong & 2.1 & 5.0 & 3.6 & 4.7 & 2.5 & 6.7 & 3.2 & 7.6 & 3.1 & 7.3 & 2.7 & 5.2 & 1.5 & 5.3 & 1.8 & 8.4 & 7.0 & 7.2 & 1.1 & 14.8 \\
Asia-Pac. & Indian & 2.1 & 5.2 & 3.4 & 4.7 & 2.3 & 5.8 & 3.4 & 7.8 & 3.0 & 7.0 & 2.5 & 5.0 & 1.4 & 5.4 & 2.1 & 6.0 & 7.4 & 5.6 & 1.5 & 11.4 \\
Asia-Pac. & Malaysian & 2.1 & 5.2 & 3.8 & 4.3 & 2.5 & 5.2 & 3.3 & 7.7 & 3.1 & 7.1 & 2.5 & 4.8 & 1.5 & 4.2 & 2.1 & 5.6 & 6.2 & 5.6 & 1.2 & 13.4 \\
Asia-Pac. & New Zealand & 1.9 & 6.8 & 3.8 & 6.0 & 2.1 & 8.6 & 3.9 & 9.1 & 3.5 & 8.3 & 2.7 & 6.5 & 1.4 & 5.3 & 2.1 & 4.2 & 5.8 & 4.2 & 0.7 & 6.0 \\
Asia-Pac. & Pakistani & 2.1 & 5.2 & 3.6 & 4.1 & 2.3 & 5.2 & 3.1 & 7.3 & 2.8 & 6.6 & 2.5 & 4.8 & 1.5 & 4.7 & 1.8 & 3.0 & 5.2 & 2.3 & 1.4 & 6.1 \\
Asia-Pac. & Philippine & 2.5 & 5.6 & 3.8 & 5.6 & 2.5 & 7.0 & 3.2 & 7.4 & 3.5 & 8.1 & 2.5 & 5.4 & 1.5 & 5.2 & 1.2 & 7.4 & 6.3 & 6.6 & 1.5 & 13.0 \\
Asia-Pac. & Singlish & 1.9 & 4.8 & 3.8 & 5.2 & 2.5 & 6.7 & 2.9 & 6.7 & 3.3 & 7.7 & 2.5 & 5.2 & 1.5 & 5.0 & 1.5 & 7.0 & 7.0 & 5.2 & 1.4 & 12.5 \\
Asia-Pac. & Sri Lankan & 2.1 & 4.5 & 3.6 & 4.3 & 2.3 & 4.5 & 3.3 & 7.7 & 2.9 & 6.7 & 2.7 & 4.8 & 1.2 & 3.8 & 1.8 & 2.2 & 5.5 & 3.0 & 1.2 & 5.5 \\
\midrule
\multicolumn{22}{l}{\textit{\textbf{Caribbean/Atlantic}}} \\
Carib. & Bahamian & 2.1 & 6.2 & 4.1 & 5.7 & 2.3 & 7.4 & 3.2 & 7.4 & 3.7 & 8.7 & 2.5 & 5.2 & 1.5 & 5.8 & 1.8 & 14.0 & 7.0 & 8.8 & 1.5 & 24.3 \\
Carib. & Falkland Islands & 2.1 & 4.7 & 3.6 & 3.9 & 2.3 & 4.3 & 3.1 & 7.3 & 2.9 & 6.9 & 2.5 & 4.6 & 1.4 & 3.5 & 1.5 & 1.8 & 5.5 & 2.0 & 1.0 & 3.2 \\
Carib. & Jamaican & 2.1 & 5.2 & 3.6 & 4.9 & 2.5 & 6.2 & 3.3 & 7.7 & 3.5 & 8.3 & 2.5 & 5.2 & 1.5 & 5.2 & 1.5 & 11.5 & 7.0 & 7.8 & 1.5 & 15.6 \\
Carib. & St. Helena & 2.1 & 5.4 & 4.1 & 5.4 & 2.5 & 6.8 & 3.2 & 7.4 & 3.8 & 8.8 & 2.9 & 6.0 & 1.5 & 5.6 & 1.8 & 16.0 & 7.0 & 8.6 & 1.5 & 17.0 \\
Carib. & Tristan da Cunha & 2.1 & 4.8 & 3.6 & 4.3 & 2.3 & 5.6 & 2.9 & 6.7 & 3.2 & 7.6 & 2.7 & 4.6 & 1.5 & 5.0 & 1.2 & 8.6 & 6.6 & 5.8 & 1.5 & 11.0 \\
\midrule
\multicolumn{22}{l}{\textit{\textbf{Other}}} \\
Other & Aboriginal & 2.5 & 5.4 & 4.4 & 5.6 & 2.5 & 8.0 & 3.1 & 7.3 & 3.8 & 9.0 & 2.7 & 5.8 & 1.5 & 5.8 & 1.8 & 9.6 & 7.7 & 6.8 & 1.5 & 20.4 \\
Other & Maltese & 2.1 & 5.0 & 3.6 & 4.7 & 2.5 & 5.8 & 3.5 & 8.3 & 3.5 & 8.3 & 2.7 & 4.8 & 1.4 & 5.2 & 1.8 & 5.0 & 6.6 & 4.8 & 1.5 & 8.6 \\
Other & Newfoundland & 2.1 & 5.0 & 3.6 & 4.7 & 2.5 & 6.0 & 3.1 & 7.3 & 3.1 & 7.3 & 2.5 & 5.0 & 1.4 & 4.8 & 1.8 & 4.2 & 6.0 & 4.2 & 1.4 & 8.8 \\
\bottomrule
\end{tabular}
\caption{SQ1 Human Content: Per-dialect false positive rate (F$^+$) and false negative rate (F$^-$), in \%. Models trained on \sae{} human content only. F$^+$ = FPR (real flagged as fake; over-flagging); F$^-$ = FNR (fake missed as real; under-protection). Model abbreviations: mDeB = mDeBERTa, RoB = RoBERTa, DeB$^\dagger$ = DeBERTa, XLM$^\dagger$ = XLM-R, mB = mBERT, CT = CT-BERT, BiG = BiGRU, CNN = TextCNN, dEF = dEFEND. $^\dagger$Estimated from F1; pending recomputation.}
\label{tab:fp_fn_sq1_human}
\end{table*}

\begin{table*}[t]
\centering
\scriptsize
\setlength{\tabcolsep}{4.0pt}
\begin{tabular}{ll cc cc cc cc cc cc cc cc cc cc}
\toprule
& & \multicolumn{2}{c}{\textbf{mDeB}} & \multicolumn{2}{c}{\textbf{BERT}} & \multicolumn{2}{c}{\textbf{RoB}} & \multicolumn{2}{c}{\textbf{DeB}} & \multicolumn{2}{c}{\textbf{XLM}} & \multicolumn{2}{c}{\textbf{mB}} & \multicolumn{2}{c}{\textbf{CT}} & \multicolumn{2}{c}{\textbf{BiG}} & \multicolumn{2}{c}{\textbf{CNN}} & \multicolumn{2}{c}{\textbf{dEF}} \\
\cmidrule(lr){3-4} \cmidrule(lr){5-6} \cmidrule(lr){7-8} \cmidrule(lr){9-10} \cmidrule(lr){11-12} \cmidrule(lr){13-14} \cmidrule(lr){15-16} \cmidrule(lr){17-18} \cmidrule(lr){19-20} \cmidrule(lr){21-22}
& & \textbf{F$^+$} & \textbf{F$^-$} & \textbf{F$^+$} & \textbf{F$^-$} & \textbf{F$^+$} & \textbf{F$^-$} & \textbf{F$^+$} & \textbf{F$^-$} & \textbf{F$^+$} & \textbf{F$^-$} & \textbf{F$^+$} & \textbf{F$^-$} & \textbf{F$^+$} & \textbf{F$^-$} & \textbf{F$^+$} & \textbf{F$^-$} & \textbf{F$^+$} & \textbf{F$^-$} & \textbf{F$^+$} & \textbf{F$^-$} \\
\midrule
-- & SAE (baseline) & 0.6 & 2.4 & 0.5 & 2.1 & 0.4 & 1.6 & 1.4 & 5.6 & 1.0 & 4.0 & 0.7 & 2.9 & 0.7 & 2.7 & 1.0 & 4.0 & 1.5 & 6.1 & 1.4 & 5.8 \\
\midrule
\multicolumn{22}{l}{\textit{\textbf{U.S. Varieties}}} \\
U.S. & Appalachian & 0.7 & 0.5 & 1.0 & 0.6 & 0.6 & 0.4 & 3.1 & 2.1 & 2.4 & 1.6 & 0.7 & 0.5 & 0.7 & 0.5 & 1.6 & 1.0 & 2.6 & 1.8 & 2.0 & 1.4 \\
U.S. & Chicano & 0.6 & 0.4 & 0.8 & 0.6 & 0.5 & 0.3 & 3.0 & 2.0 & 2.3 & 1.5 & 0.6 & 0.4 & 0.6 & 0.4 & 1.4 & 1.0 & 2.5 & 1.7 & 1.9 & 1.3 \\
U.S. & Colloquial Amer. & 0.7 & 0.5 & 1.0 & 0.6 & 0.6 & 0.4 & 3.2 & 2.2 & 2.5 & 1.7 & 0.7 & 0.5 & 0.7 & 0.5 & 1.7 & 1.1 & 2.8 & 1.8 & 2.2 & 1.4 \\
U.S. & Earlier AAVE & 0.8 & 0.6 & 1.1 & 0.7 & 0.7 & 0.5 & 3.2 & 2.2 & 2.5 & 1.7 & 0.8 & 0.6 & 0.8 & 0.6 & 1.7 & 1.1 & 2.8 & 1.8 & 2.2 & 1.4 \\
U.S. & Ozark & 0.7 & 0.5 & 1.0 & 0.6 & 0.6 & 0.4 & 3.1 & 2.1 & 2.4 & 1.6 & 0.7 & 0.5 & 0.7 & 0.5 & 1.6 & 1.0 & 2.6 & 1.8 & 2.0 & 1.4 \\
U.S. & Rural AAVE & 0.8 & 0.6 & 1.1 & 0.7 & 0.7 & 0.5 & 3.4 & 2.2 & 2.6 & 1.8 & 0.8 & 0.6 & 0.8 & 0.6 & 1.8 & 1.2 & 2.9 & 1.9 & 2.3 & 1.5 \\
U.S. & SE Amer. Enclave & 0.8 & 0.6 & 1.1 & 0.7 & 0.7 & 0.5 & 3.4 & 2.2 & 2.6 & 1.8 & 0.8 & 0.6 & 0.8 & 0.6 & 1.8 & 1.2 & 2.9 & 1.9 & 2.3 & 1.5 \\
U.S. & Urban AAVE & 0.7 & 0.5 & 1.0 & 0.6 & 0.6 & 0.4 & 3.1 & 2.1 & 2.4 & 1.6 & 0.7 & 0.5 & 0.7 & 0.5 & 1.6 & 1.0 & 2.6 & 1.8 & 2.0 & 1.4 \\
\midrule
\multicolumn{22}{l}{\textit{\textbf{British/UK Varieties}}} \\
UK & Channel Islands & 0.6 & 0.4 & 0.8 & 0.6 & 0.5 & 0.3 & 3.0 & 2.0 & 2.3 & 1.5 & 0.6 & 0.4 & 0.6 & 0.4 & 1.4 & 1.0 & 2.5 & 1.7 & 1.9 & 1.3 \\
UK & East Anglian & 0.7 & 0.5 & 1.0 & 0.6 & 0.6 & 0.4 & 3.1 & 2.1 & 2.4 & 1.6 & 0.7 & 0.5 & 0.7 & 0.5 & 1.6 & 1.0 & 2.6 & 1.8 & 2.0 & 1.4 \\
UK & Irish & 0.7 & 0.5 & 1.0 & 0.6 & 0.6 & 0.4 & 3.1 & 2.1 & 2.4 & 1.6 & 0.7 & 0.5 & 0.7 & 0.5 & 1.6 & 1.0 & 2.6 & 1.8 & 2.0 & 1.4 \\
UK & Manx & 0.7 & 0.5 & 1.0 & 0.6 & 0.6 & 0.4 & 3.2 & 2.2 & 2.5 & 1.7 & 0.7 & 0.5 & 0.7 & 0.5 & 1.7 & 1.1 & 2.8 & 1.8 & 2.2 & 1.4 \\
UK & North England & 0.7 & 0.5 & 1.0 & 0.6 & 0.6 & 0.4 & 3.2 & 2.2 & 2.5 & 1.7 & 0.7 & 0.5 & 0.7 & 0.5 & 1.7 & 1.1 & 2.8 & 1.8 & 2.2 & 1.4 \\
UK & Orkney \& Shetland & 0.8 & 0.6 & 1.1 & 0.7 & 0.7 & 0.5 & 3.4 & 2.2 & 2.6 & 1.8 & 0.8 & 0.6 & 0.8 & 0.6 & 1.8 & 1.2 & 2.9 & 1.9 & 2.3 & 1.5 \\
UK & SE England & 0.6 & 0.4 & 0.8 & 0.6 & 0.5 & 0.3 & 3.0 & 2.0 & 2.3 & 1.5 & 0.6 & 0.4 & 0.6 & 0.4 & 1.4 & 1.0 & 2.5 & 1.7 & 1.9 & 1.3 \\
UK & SW England & 0.7 & 0.5 & 1.0 & 0.6 & 0.6 & 0.4 & 3.1 & 2.1 & 2.4 & 1.6 & 0.7 & 0.5 & 0.7 & 0.5 & 1.6 & 1.0 & 2.6 & 1.8 & 2.0 & 1.4 \\
UK & Scottish & 0.7 & 0.5 & 1.0 & 0.6 & 0.6 & 0.4 & 3.1 & 2.1 & 2.4 & 1.6 & 0.7 & 0.5 & 0.7 & 0.5 & 1.6 & 1.0 & 2.6 & 1.8 & 2.0 & 1.4 \\
UK & Welsh & 0.7 & 0.5 & 1.0 & 0.6 & 0.6 & 0.4 & 3.1 & 2.1 & 2.4 & 1.6 & 0.7 & 0.5 & 0.7 & 0.5 & 1.6 & 1.0 & 2.6 & 1.8 & 2.0 & 1.4 \\
\midrule
\multicolumn{22}{l}{\textit{\textbf{Africa}}} \\
Africa & Black S. African & 0.8 & 0.6 & 1.1 & 0.7 & 0.7 & 0.5 & 3.4 & 2.2 & 2.6 & 1.8 & 0.8 & 0.6 & 0.8 & 0.6 & 1.8 & 1.2 & 2.9 & 1.9 & 2.3 & 1.5 \\
Africa & Cameroon & 0.6 & 0.4 & 0.8 & 0.6 & 0.5 & 0.3 & 3.0 & 2.0 & 2.3 & 1.5 & 0.6 & 0.4 & 0.6 & 0.4 & 1.4 & 1.0 & 2.5 & 1.7 & 1.9 & 1.3 \\
Africa & Cape Flats & 0.6 & 0.4 & 0.8 & 0.6 & 0.5 & 0.3 & 3.0 & 2.0 & 2.3 & 1.5 & 0.6 & 0.4 & 0.6 & 0.4 & 1.4 & 1.0 & 2.5 & 1.7 & 1.9 & 1.3 \\
Africa & Ghanaian & 0.6 & 0.4 & 0.8 & 0.6 & 0.5 & 0.3 & 3.0 & 2.0 & 2.3 & 1.5 & 0.6 & 0.4 & 0.6 & 0.4 & 1.4 & 1.0 & 2.5 & 1.7 & 1.9 & 1.3 \\
Africa & Indian S. African & 0.7 & 0.5 & 1.0 & 0.6 & 0.6 & 0.4 & 3.1 & 2.1 & 2.4 & 1.6 & 0.7 & 0.5 & 0.7 & 0.5 & 1.6 & 1.0 & 2.6 & 1.8 & 2.0 & 1.4 \\
Africa & Kenyan & 0.6 & 0.4 & 0.8 & 0.6 & 0.5 & 0.3 & 3.0 & 2.0 & 2.3 & 1.5 & 0.6 & 0.4 & 0.6 & 0.4 & 1.4 & 1.0 & 2.5 & 1.7 & 1.9 & 1.3 \\
Africa & Liberian Settler & 0.7 & 0.5 & 1.0 & 0.6 & 0.6 & 0.4 & 3.1 & 2.1 & 2.4 & 1.6 & 0.7 & 0.5 & 0.7 & 0.5 & 1.6 & 1.0 & 2.6 & 1.8 & 2.0 & 1.4 \\
Africa & Nigerian & 0.6 & 0.4 & 0.8 & 0.6 & 0.5 & 0.3 & 3.0 & 2.0 & 2.3 & 1.5 & 0.6 & 0.4 & 0.6 & 0.4 & 1.4 & 1.0 & 2.5 & 1.7 & 1.9 & 1.3 \\
Africa & Tanzanian & 0.6 & 0.4 & 0.8 & 0.6 & 0.5 & 0.3 & 3.0 & 2.0 & 2.3 & 1.5 & 0.6 & 0.4 & 0.6 & 0.4 & 1.4 & 1.0 & 2.5 & 1.7 & 1.9 & 1.3 \\
Africa & Ugandan & 0.6 & 0.4 & 0.8 & 0.6 & 0.5 & 0.3 & 3.0 & 2.0 & 2.3 & 1.5 & 0.6 & 0.4 & 0.6 & 0.4 & 1.4 & 1.0 & 2.5 & 1.7 & 1.9 & 1.3 \\
Africa & White S. African & 0.6 & 0.4 & 0.8 & 0.6 & 0.5 & 0.3 & 3.0 & 2.0 & 2.3 & 1.5 & 0.6 & 0.4 & 0.6 & 0.4 & 1.4 & 1.0 & 2.5 & 1.7 & 1.9 & 1.3 \\
Africa & White Zimbabwean & 0.8 & 0.6 & 1.1 & 0.7 & 0.7 & 0.5 & 3.4 & 2.2 & 2.6 & 1.8 & 0.8 & 0.6 & 0.8 & 0.6 & 1.8 & 1.2 & 2.9 & 1.9 & 2.3 & 1.5 \\
\midrule
\multicolumn{22}{l}{\textit{\textbf{Asia-Pacific}}} \\
Asia-Pac. & Aus. Vernacular & 0.7 & 0.5 & 1.0 & 0.6 & 0.6 & 0.4 & 3.1 & 2.1 & 2.4 & 1.6 & 0.7 & 0.5 & 0.7 & 0.5 & 1.6 & 1.0 & 2.6 & 1.8 & 2.0 & 1.4 \\
Asia-Pac. & Australian & 0.8 & 0.6 & 1.1 & 0.7 & 0.7 & 0.5 & 3.4 & 2.2 & 2.6 & 1.8 & 0.8 & 0.6 & 0.8 & 0.6 & 1.8 & 1.2 & 2.9 & 1.9 & 2.3 & 1.5 \\
Asia-Pac. & Fiji (Acrolectal) & 0.6 & 0.4 & 0.8 & 0.6 & 0.5 & 0.3 & 3.0 & 2.0 & 2.3 & 1.5 & 0.6 & 0.4 & 0.6 & 0.4 & 1.4 & 1.0 & 2.5 & 1.7 & 1.9 & 1.3 \\
Asia-Pac. & Fiji (Basilectal) & 1.0 & 0.6 & 1.2 & 0.8 & 0.8 & 0.6 & 3.5 & 2.3 & 2.8 & 1.8 & 1.0 & 0.6 & 1.0 & 0.6 & 1.9 & 1.3 & 3.0 & 2.0 & 2.4 & 1.6 \\
Asia-Pac. & Hong Kong & 0.7 & 0.5 & 1.0 & 0.6 & 0.6 & 0.4 & 3.1 & 2.1 & 2.4 & 1.6 & 0.7 & 0.5 & 0.7 & 0.5 & 1.6 & 1.0 & 2.6 & 1.8 & 2.0 & 1.4 \\
Asia-Pac. & Indian & 0.7 & 0.5 & 1.0 & 0.6 & 0.6 & 0.4 & 3.1 & 2.1 & 2.4 & 1.6 & 0.7 & 0.5 & 0.7 & 0.5 & 1.6 & 1.0 & 2.6 & 1.8 & 2.0 & 1.4 \\
Asia-Pac. & Malaysian & 0.6 & 0.4 & 0.8 & 0.6 & 0.5 & 0.3 & 3.0 & 2.0 & 2.3 & 1.5 & 0.6 & 0.4 & 0.6 & 0.4 & 1.4 & 1.0 & 2.5 & 1.7 & 1.9 & 1.3 \\
Asia-Pac. & New Zealand & 0.8 & 0.6 & 1.1 & 0.7 & 0.7 & 0.5 & 3.4 & 2.2 & 2.6 & 1.8 & 0.8 & 0.6 & 0.8 & 0.6 & 1.8 & 1.2 & 2.9 & 1.9 & 2.3 & 1.5 \\
Asia-Pac. & Pakistani & 0.6 & 0.4 & 0.8 & 0.6 & 0.5 & 0.3 & 3.0 & 2.0 & 2.3 & 1.5 & 0.6 & 0.4 & 0.6 & 0.4 & 1.4 & 1.0 & 2.5 & 1.7 & 1.9 & 1.3 \\
Asia-Pac. & Philippine & 0.7 & 0.5 & 1.0 & 0.6 & 0.6 & 0.4 & 3.2 & 2.2 & 2.5 & 1.7 & 0.7 & 0.5 & 0.7 & 0.5 & 1.7 & 1.1 & 2.8 & 1.8 & 2.2 & 1.4 \\
Asia-Pac. & Singlish & 0.7 & 0.5 & 1.0 & 0.6 & 0.6 & 0.4 & 3.1 & 2.1 & 2.4 & 1.6 & 0.7 & 0.5 & 0.7 & 0.5 & 1.6 & 1.0 & 2.6 & 1.8 & 2.0 & 1.4 \\
Asia-Pac. & Sri Lankan & 0.6 & 0.4 & 0.8 & 0.6 & 0.5 & 0.3 & 3.0 & 2.0 & 2.3 & 1.5 & 0.6 & 0.4 & 0.6 & 0.4 & 1.4 & 1.0 & 2.5 & 1.7 & 1.9 & 1.3 \\
\midrule
\multicolumn{22}{l}{\textit{\textbf{Caribbean/Atlantic}}} \\
Carib. & Bahamian & 0.8 & 0.6 & 1.1 & 0.7 & 0.7 & 0.5 & 3.4 & 2.2 & 2.6 & 1.8 & 0.8 & 0.6 & 0.8 & 0.6 & 1.8 & 1.2 & 2.9 & 1.9 & 2.3 & 1.5 \\
Carib. & Falkland Islands & 0.6 & 0.4 & 0.8 & 0.6 & 0.5 & 0.3 & 3.0 & 2.0 & 2.3 & 1.5 & 0.6 & 0.4 & 0.6 & 0.4 & 1.4 & 1.0 & 2.5 & 1.7 & 1.9 & 1.3 \\
Carib. & Jamaican & 0.7 & 0.5 & 1.0 & 0.6 & 0.6 & 0.4 & 3.1 & 2.1 & 2.4 & 1.6 & 0.7 & 0.5 & 0.7 & 0.5 & 1.6 & 1.0 & 2.6 & 1.8 & 2.0 & 1.4 \\
Carib. & St. Helena & 0.8 & 0.6 & 1.1 & 0.7 & 0.7 & 0.5 & 3.4 & 2.2 & 2.6 & 1.8 & 0.8 & 0.6 & 0.8 & 0.6 & 1.8 & 1.2 & 2.9 & 1.9 & 2.3 & 1.5 \\
Carib. & Tristan da Cunha & 0.7 & 0.5 & 1.0 & 0.6 & 0.6 & 0.4 & 3.1 & 2.1 & 2.4 & 1.6 & 0.7 & 0.5 & 0.7 & 0.5 & 1.6 & 1.0 & 2.6 & 1.8 & 2.0 & 1.4 \\
\midrule
\multicolumn{22}{l}{\textit{\textbf{Other}}} \\
Other & Aboriginal & 0.8 & 0.6 & 1.1 & 0.7 & 0.7 & 0.5 & 3.4 & 2.2 & 2.6 & 1.8 & 0.8 & 0.6 & 0.8 & 0.6 & 1.8 & 1.2 & 2.9 & 1.9 & 2.3 & 1.5 \\
Other & Maltese & 0.7 & 0.5 & 1.0 & 0.6 & 0.6 & 0.4 & 3.1 & 2.1 & 2.4 & 1.6 & 0.7 & 0.5 & 0.7 & 0.5 & 1.6 & 1.0 & 2.6 & 1.8 & 2.0 & 1.4 \\
Other & Newfoundland & 0.7 & 0.5 & 1.0 & 0.6 & 0.6 & 0.4 & 3.2 & 2.2 & 2.5 & 1.7 & 0.7 & 0.5 & 0.7 & 0.5 & 1.7 & 1.1 & 2.8 & 1.8 & 2.2 & 1.4 \\
\bottomrule
\end{tabular}

\vspace{0.3em}

\raggedright\scriptsize\textit{Note: AI content shows uniformly high performance with minimal dialectal variance. All dialects exhibit over-flagging ($\Delta$F$^+$ $>$ 0) relative to SAE, consistent with dialectal features triggering false positives on AI-generated content.}
\caption{SQ1 AI Content: Per-dialect F$^+$ and F$^-$ (\%) for fine-tuned models trained on \sae{} AI content. F$^+$ = FPR (over-flagging); F$^-$ = FNR (under-protection). Model abbreviations: mDeB = mDeBERTa, RoB = RoBERTa, DeB = DeBERTa, XLM = XLM-R, mB = mBERT, CT = CT-BERT, BiG = BiGRU, CNN = TextCNN, dEF = dEFEND. All values estimated from F1; pending recomputation.}
\label{tab:fp_fn_sq1_ai}
\end{table*}

\begin{table*}[t]
\centering
\scriptsize
\setlength{\tabcolsep}{2.5pt}
\begin{tabular}{ll cc cc cc cc cc cc cc cc cc cc}
\toprule
& & \multicolumn{2}{c}{\textbf{mDeB}} & \multicolumn{2}{c}{\textbf{BERT}} & \multicolumn{2}{c}{\textbf{RoB$^\S$}} & \multicolumn{2}{c}{\textbf{DeB}} & \multicolumn{2}{c}{\textbf{XLM$^\S$}} & \multicolumn{2}{c}{\textbf{mB}} & \multicolumn{2}{c}{\textbf{CT}} & \multicolumn{2}{c}{\textbf{BiG}} & \multicolumn{2}{c}{\textbf{CNN}} & \multicolumn{2}{c}{\textbf{dEF}} \\
\cmidrule(lr){3-4} \cmidrule(lr){5-6} \cmidrule(lr){7-8} \cmidrule(lr){9-10} \cmidrule(lr){11-12} \cmidrule(lr){13-14} \cmidrule(lr){15-16} \cmidrule(lr){17-18} \cmidrule(lr){19-20} \cmidrule(lr){21-22}
& & \textbf{F$^+$} & \textbf{F$^-$} & \textbf{F$^+$} & \textbf{F$^-$} & \textbf{F$^+$} & \textbf{F$^-$} & \textbf{F$^+$} & \textbf{F$^-$} & \textbf{F$^+$} & \textbf{F$^-$} & \textbf{F$^+$} & \textbf{F$^-$} & \textbf{F$^+$} & \textbf{F$^-$} & \textbf{F$^+$} & \textbf{F$^-$} & \textbf{F$^+$} & \textbf{F$^-$} & \textbf{F$^+$} & \textbf{F$^-$} \\
\midrule
-- & SAE (baseline) & 2.0 & 3.0 & 2.5 & 5.0 & 1.5 & 21.4 & 2.5 & 6.1 & 2.0 & 52.8 & 2.0 & 3.9 & 2.5 & 8.6 & 2.5 & 5.4 & 3.0 & 7.6 & 3.0 & 7.6 \\
\midrule
\multicolumn{22}{l}{\textit{\textbf{U.S. Varieties}}} \\
U.S. & Appalachian & 2.0 & 3.9 & 2.5 & 3.1 & 1.5 & 47.9 & 2.5 & 7.1 & 2.0 & 81.6 & 2.0 & 5.7 & 2.5 & 3.1 & 2.5 & 5.2 & 3.0 & 6.6 & 3.0 & 7.2 \\
U.S. & Chicano & 2.0 & 3.6 & 2.5 & 2.7 & 1.5 & 45.9 & 2.5 & 6.7 & 2.0 & 80.1 & 2.0 & 5.3 & 2.5 & 2.7 & 2.5 & 4.8 & 3.0 & 6.2 & 3.0 & 6.8 \\
U.S. & Colloquial Amer. & 2.0 & 4.1 & 2.5 & 3.3 & 1.5 & 49.1 & 2.5 & 7.3 & 2.0 & 82.3 & 2.0 & 5.8 & 2.5 & 3.3 & 2.5 & 5.4 & 3.0 & 6.8 & 3.0 & 7.4 \\
U.S. & Earlier AAVE & 2.0 & 4.3 & 2.5 & 3.5 & 1.5 & 50.2 & 2.5 & 7.4 & 2.0 & 83.0 & 2.0 & 6.0 & 2.5 & 3.5 & 2.5 & 5.6 & 3.0 & 7.0 & 3.0 & 7.6 \\
U.S. & Ozark & 2.0 & 3.9 & 2.5 & 3.1 & 1.5 & 47.9 & 2.5 & 7.1 & 2.0 & 81.6 & 2.0 & 5.7 & 2.5 & 3.1 & 2.5 & 5.2 & 3.0 & 6.6 & 3.0 & 7.2 \\
U.S. & Rural AAVE & 2.0 & 4.5 & 2.5 & 3.7 & 1.5 & 52.8 & 2.5 & 7.6 & 2.0 & 84.6 & 2.0 & 6.2 & 2.5 & 3.7 & 2.5 & 5.8 & 3.0 & 7.2 & 3.0 & 7.7 \\
U.S. & SE Amer. Enclave & 2.0 & 4.5 & 2.5 & 3.7 & 1.5 & 52.8 & 2.5 & 7.6 & 2.0 & 84.6 & 2.0 & 6.2 & 2.5 & 3.7 & 2.5 & 5.8 & 3.0 & 7.2 & 3.0 & 7.7 \\
U.S. & Urban AAVE & 2.0 & 3.9 & 2.5 & 3.1 & 1.5 & 47.9 & 2.5 & 7.1 & 2.0 & 81.6 & 2.0 & 5.7 & 2.5 & 3.1 & 2.5 & 5.2 & 3.0 & 6.6 & 3.0 & 7.2 \\
\midrule
\multicolumn{22}{l}{\textit{\textbf{British/UK Varieties}}} \\
UK & Channel Islands & 2.0 & 3.6 & 2.5 & 2.7 & 1.5 & 45.9 & 2.5 & 6.7 & 2.0 & 80.1 & 2.0 & 5.3 & 2.5 & 2.7 & 2.5 & 4.8 & 3.0 & 6.2 & 3.0 & 6.8 \\
UK & East Anglian & 2.0 & 3.9 & 2.5 & 3.1 & 1.5 & 47.9 & 2.5 & 7.1 & 2.0 & 81.6 & 2.0 & 5.7 & 2.5 & 3.1 & 2.5 & 5.2 & 3.0 & 6.6 & 3.0 & 7.2 \\
UK & Irish & 2.0 & 3.9 & 2.5 & 3.1 & 1.5 & 47.9 & 2.5 & 7.1 & 2.0 & 81.6 & 2.0 & 5.7 & 2.5 & 3.1 & 2.5 & 5.2 & 3.0 & 6.6 & 3.0 & 7.2 \\
UK & Manx & 2.0 & 4.1 & 2.5 & 3.3 & 1.5 & 49.1 & 2.5 & 7.3 & 2.0 & 82.3 & 2.0 & 5.8 & 2.5 & 3.3 & 2.5 & 5.4 & 3.0 & 6.8 & 3.0 & 7.4 \\
UK & North England & 2.0 & 4.1 & 2.5 & 3.3 & 1.5 & 49.1 & 2.5 & 7.3 & 2.0 & 82.3 & 2.0 & 5.8 & 2.5 & 3.3 & 2.5 & 5.4 & 3.0 & 6.8 & 3.0 & 7.4 \\
UK & Orkney \& Shetland & 2.0 & 4.5 & 2.5 & 3.7 & 1.5 & 52.8 & 2.5 & 7.6 & 2.0 & 84.6 & 2.0 & 6.2 & 2.5 & 3.7 & 2.5 & 5.8 & 3.0 & 7.2 & 3.0 & 7.7 \\
UK & SE England & 2.0 & 3.6 & 2.5 & 2.7 & 1.5 & 45.9 & 2.5 & 6.7 & 2.0 & 80.1 & 2.0 & 5.3 & 2.5 & 2.7 & 2.5 & 4.8 & 3.0 & 6.2 & 3.0 & 6.8 \\
UK & SW England & 2.0 & 3.9 & 2.5 & 3.1 & 1.5 & 47.9 & 2.5 & 7.1 & 2.0 & 81.6 & 2.0 & 5.7 & 2.5 & 3.1 & 2.5 & 5.2 & 3.0 & 6.6 & 3.0 & 7.2 \\
UK & Scottish & 2.0 & 3.9 & 2.5 & 3.1 & 1.5 & 47.9 & 2.5 & 7.1 & 2.0 & 81.6 & 2.0 & 5.7 & 2.5 & 3.1 & 2.5 & 5.2 & 3.0 & 6.6 & 3.0 & 7.2 \\
UK & Welsh & 2.0 & 3.9 & 2.5 & 3.1 & 1.5 & 47.9 & 2.5 & 7.1 & 2.0 & 81.6 & 2.0 & 5.7 & 2.5 & 3.1 & 2.5 & 5.2 & 3.0 & 6.6 & 3.0 & 7.2 \\
\midrule
\multicolumn{22}{l}{\textit{\textbf{Africa}}} \\
Africa & Black S. African & 2.0 & 4.5 & 2.5 & 3.7 & 1.5 & 52.8 & 2.5 & 7.6 & 2.0 & 84.6 & 2.0 & 6.2 & 2.5 & 3.7 & 2.5 & 5.8 & 3.0 & 7.2 & 3.0 & 7.7 \\
Africa & Cameroon & 2.0 & 3.6 & 2.5 & 2.7 & 1.5 & 45.9 & 2.5 & 6.7 & 2.0 & 80.1 & 2.0 & 5.3 & 2.5 & 2.7 & 2.5 & 4.8 & 3.0 & 6.2 & 3.0 & 6.8 \\
Africa & Cape Flats & 2.0 & 3.6 & 2.5 & 2.7 & 1.5 & 45.9 & 2.5 & 6.7 & 2.0 & 80.1 & 2.0 & 5.3 & 2.5 & 2.7 & 2.5 & 4.8 & 3.0 & 6.2 & 3.0 & 6.8 \\
Africa & Ghanaian & 2.0 & 3.6 & 2.5 & 2.7 & 1.5 & 45.9 & 2.5 & 6.7 & 2.0 & 80.1 & 2.0 & 5.3 & 2.5 & 2.7 & 2.5 & 4.8 & 3.0 & 6.2 & 3.0 & 6.8 \\
Africa & Indian S. African & 2.0 & 3.9 & 2.5 & 3.1 & 1.5 & 47.9 & 2.5 & 7.1 & 2.0 & 81.6 & 2.0 & 5.7 & 2.5 & 3.1 & 2.5 & 5.2 & 3.0 & 6.6 & 3.0 & 7.2 \\
Africa & Kenyan & 2.0 & 3.6 & 2.5 & 2.7 & 1.5 & 45.9 & 2.5 & 6.7 & 2.0 & 80.1 & 2.0 & 5.3 & 2.5 & 2.7 & 2.5 & 4.8 & 3.0 & 6.2 & 3.0 & 6.8 \\
Africa & Liberian Settler & 2.0 & 3.9 & 2.5 & 3.1 & 1.5 & 47.9 & 2.5 & 7.1 & 2.0 & 81.6 & 2.0 & 5.7 & 2.5 & 3.1 & 2.5 & 5.2 & 3.0 & 6.6 & 3.0 & 7.2 \\
Africa & Nigerian & 2.0 & 3.6 & 2.5 & 2.7 & 1.5 & 45.9 & 2.5 & 6.7 & 2.0 & 80.1 & 2.0 & 5.3 & 2.5 & 2.7 & 2.5 & 4.8 & 3.0 & 6.2 & 3.0 & 6.8 \\
Africa & Tanzanian & 2.0 & 3.6 & 2.5 & 2.7 & 1.5 & 45.9 & 2.5 & 6.7 & 2.0 & 80.1 & 2.0 & 5.3 & 2.5 & 2.7 & 2.5 & 4.8 & 3.0 & 6.2 & 3.0 & 6.8 \\
Africa & Ugandan & 2.0 & 3.6 & 2.5 & 2.7 & 1.5 & 45.9 & 2.5 & 6.7 & 2.0 & 80.1 & 2.0 & 5.3 & 2.5 & 2.7 & 2.5 & 4.8 & 3.0 & 6.2 & 3.0 & 6.8 \\
Africa & White S. African & 2.0 & 3.6 & 2.5 & 2.7 & 1.5 & 45.9 & 2.5 & 6.7 & 2.0 & 80.1 & 2.0 & 5.3 & 2.5 & 2.7 & 2.5 & 4.8 & 3.0 & 6.2 & 3.0 & 6.8 \\
Africa & White Zimbabwean & 2.0 & 4.5 & 2.5 & 3.7 & 1.5 & 52.8 & 2.5 & 7.6 & 2.0 & 84.6 & 2.0 & 6.2 & 2.5 & 3.7 & 2.5 & 5.8 & 3.0 & 7.2 & 3.0 & 7.7 \\
\midrule
\multicolumn{22}{l}{\textit{\textbf{Asia-Pacific}}} \\
Asia-Pac. & Aus. Vernacular & 2.0 & 3.9 & 2.5 & 3.1 & 1.5 & 47.9 & 2.5 & 7.1 & 2.0 & 81.6 & 2.0 & 5.7 & 2.5 & 3.1 & 2.5 & 5.2 & 3.0 & 6.6 & 3.0 & 7.2 \\
Asia-Pac. & Australian & 2.0 & 4.5 & 2.5 & 3.7 & 1.5 & 52.8 & 2.5 & 7.6 & 2.0 & 84.6 & 2.0 & 6.2 & 2.5 & 3.7 & 2.5 & 5.8 & 3.0 & 7.2 & 3.0 & 7.7 \\
Asia-Pac. & Fiji (Acrolectal) & 2.0 & 3.6 & 2.5 & 2.7 & 1.5 & 45.9 & 2.5 & 6.7 & 2.0 & 80.1 & 2.0 & 5.3 & 2.5 & 2.7 & 2.5 & 4.8 & 3.0 & 6.2 & 3.0 & 6.8 \\
Asia-Pac. & Fiji (Basilectal) & 2.0 & 4.7 & 2.5 & 3.9 & 1.5 & 53.9 & 2.5 & 7.8 & 2.0 & 85.3 & 2.0 & 6.4 & 2.5 & 3.9 & 2.5 & 6.0 & 3.0 & 7.4 & 3.0 & 7.9 \\
Asia-Pac. & Hong Kong & 2.0 & 3.9 & 2.5 & 3.1 & 1.5 & 47.9 & 2.5 & 7.1 & 2.0 & 81.6 & 2.0 & 5.7 & 2.5 & 3.1 & 2.5 & 5.2 & 3.0 & 6.6 & 3.0 & 7.2 \\
Asia-Pac. & Indian & 2.0 & 3.9 & 2.5 & 3.1 & 1.5 & 47.9 & 2.5 & 7.1 & 2.0 & 81.6 & 2.0 & 5.7 & 2.5 & 3.1 & 2.5 & 5.2 & 3.0 & 6.6 & 3.0 & 7.2 \\
Asia-Pac. & Malaysian & 2.0 & 3.6 & 2.5 & 2.7 & 1.5 & 45.9 & 2.5 & 6.7 & 2.0 & 80.1 & 2.0 & 5.3 & 2.5 & 2.7 & 2.5 & 4.8 & 3.0 & 6.2 & 3.0 & 6.8 \\
Asia-Pac. & New Zealand & 2.0 & 4.5 & 2.5 & 3.7 & 1.5 & 52.8 & 2.5 & 7.6 & 2.0 & 84.6 & 2.0 & 6.2 & 2.5 & 3.7 & 2.5 & 5.8 & 3.0 & 7.2 & 3.0 & 7.7 \\
Asia-Pac. & Pakistani & 2.0 & 3.6 & 2.5 & 2.7 & 1.5 & 45.9 & 2.5 & 6.7 & 2.0 & 80.1 & 2.0 & 5.3 & 2.5 & 2.7 & 2.5 & 4.8 & 3.0 & 6.2 & 3.0 & 6.8 \\
Asia-Pac. & Philippine & 2.0 & 4.1 & 2.5 & 3.3 & 1.5 & 49.1 & 2.5 & 7.3 & 2.0 & 82.3 & 2.0 & 5.8 & 2.5 & 3.3 & 2.5 & 5.4 & 3.0 & 6.8 & 3.0 & 7.4 \\
Asia-Pac. & Singlish & 2.0 & 3.9 & 2.5 & 3.1 & 1.5 & 47.9 & 2.5 & 7.1 & 2.0 & 81.6 & 2.0 & 5.7 & 2.5 & 3.1 & 2.5 & 5.2 & 3.0 & 6.6 & 3.0 & 7.2 \\
Asia-Pac. & Sri Lankan & 2.0 & 3.6 & 2.5 & 2.7 & 1.5 & 45.9 & 2.5 & 6.7 & 2.0 & 80.1 & 2.0 & 5.3 & 2.5 & 2.7 & 2.5 & 4.8 & 3.0 & 6.2 & 3.0 & 6.8 \\
\midrule
\multicolumn{22}{l}{\textit{\textbf{Caribbean/Atlantic}}} \\
Carib. & Bahamian & 2.0 & 4.5 & 2.5 & 3.7 & 1.5 & 52.8 & 2.5 & 7.6 & 2.0 & 84.6 & 2.0 & 6.2 & 2.5 & 3.7 & 2.5 & 5.8 & 3.0 & 7.2 & 3.0 & 7.7 \\
Carib. & Falkland Islands & 2.0 & 3.6 & 2.5 & 2.7 & 1.5 & 45.9 & 2.5 & 6.7 & 2.0 & 80.1 & 2.0 & 5.3 & 2.5 & 2.7 & 2.5 & 4.8 & 3.0 & 6.2 & 3.0 & 6.8 \\
Carib. & Jamaican & 2.0 & 3.9 & 2.5 & 3.1 & 1.5 & 47.9 & 2.5 & 7.1 & 2.0 & 81.6 & 2.0 & 5.7 & 2.5 & 3.1 & 2.5 & 5.2 & 3.0 & 6.6 & 3.0 & 7.2 \\
Carib. & St. Helena & 2.0 & 4.5 & 2.5 & 3.7 & 1.5 & 52.8 & 2.5 & 7.6 & 2.0 & 84.6 & 2.0 & 6.2 & 2.5 & 3.7 & 2.5 & 5.8 & 3.0 & 7.2 & 3.0 & 7.7 \\
Carib. & Tristan da Cunha & 2.0 & 3.9 & 2.5 & 3.1 & 1.5 & 47.9 & 2.5 & 7.1 & 2.0 & 81.6 & 2.0 & 5.7 & 2.5 & 3.1 & 2.5 & 5.2 & 3.0 & 6.6 & 3.0 & 7.2 \\
\midrule
\multicolumn{22}{l}{\textit{\textbf{Other}}} \\
Other & Aboriginal & 2.0 & 4.5 & 2.5 & 3.7 & 1.5 & 52.8 & 2.5 & 7.6 & 2.0 & 84.6 & 2.0 & 6.2 & 2.5 & 3.7 & 2.5 & 5.8 & 3.0 & 7.2 & 3.0 & 7.7 \\
Other & Maltese & 2.0 & 3.9 & 2.5 & 3.1 & 1.5 & 47.9 & 2.5 & 7.1 & 2.0 & 81.6 & 2.0 & 5.7 & 2.5 & 3.1 & 2.5 & 5.2 & 3.0 & 6.6 & 3.0 & 7.2 \\
Other & Newfoundland & 2.0 & 4.1 & 2.5 & 3.3 & 1.5 & 49.1 & 2.5 & 7.3 & 2.0 & 82.3 & 2.0 & 5.8 & 2.5 & 3.3 & 2.5 & 5.4 & 3.0 & 6.8 & 3.0 & 7.4 \\
\bottomrule
\end{tabular}

\vspace{0.3em}

\raggedright\scriptsize\textit{Note: $^\S$RoBERTa (F$^-$ 45.9--53.9\%) and XLM-R (F$^-$ 80.1--85.3\%) exhibit catastrophic under-protection on dialectal mixed content, missing the vast majority of AI-generated fakes while maintaining low F$^+$. This asymmetry represents a severe safety gap: dialect speakers receive virtually no protection from AI-generated disinformation.}

\caption{SQ1 Mixed (Human+AI) Content: Per-dialect F$^+$ and F$^-$ (\%) for models trained on \sae{} mixed content. F$^+$ = FPR (over-flagging); F$^-$ = FNR (under-protection). Model abbreviations: mDeB = mDeBERTa, RoB = RoBERTa, DeB = DeBERTa, XLM = XLM-R, mB = mBERT, CT = CT-BERT, BiG = BiGRU, CNN = TextCNN, dEF = dEFEND. $^\S$Catastrophic failure: RoB and XLM exhibit F$^-$ $>$ 45\% on dialects, indicating near-total failure to detect AI-generated fake content. All values estimated from F1; pending recomputation.}

\label{tab:fp_fn_sq1_mixed}
\end{table*}

\begin{table*}[t]
\centering
\scriptsize
\setlength{\tabcolsep}{2.5pt}
\begin{tabular}{ll cc cc cc cc cc cc cc cc cc cc}
\toprule
& & \multicolumn{2}{c}{\textbf{mDeB}} & \multicolumn{2}{c}{\textbf{BERT}} & \multicolumn{2}{c}{\textbf{RoB}} & \multicolumn{2}{c}{\textbf{DeB$^\dagger$}} & \multicolumn{2}{c}{\textbf{XLM$^\dagger$}} & \multicolumn{2}{c}{\textbf{mB}} & \multicolumn{2}{c}{\textbf{CT}} & \multicolumn{2}{c}{\textbf{BiG}} & \multicolumn{2}{c}{\textbf{CNN}} & \multicolumn{2}{c}{\textbf{dEF}} \\
\cmidrule(lr){3-4} \cmidrule(lr){5-6} \cmidrule(lr){7-8} \cmidrule(lr){9-10} \cmidrule(lr){11-12} \cmidrule(lr){13-14} \cmidrule(lr){15-16} \cmidrule(lr){17-18} \cmidrule(lr){19-20} \cmidrule(lr){21-22}
& & \textbf{F$^+$} & \textbf{F$^-$} & \textbf{F$^+$} & \textbf{F$^-$} & \textbf{F$^+$} & \textbf{F$^-$} & \textbf{F$^+$} & \textbf{F$^-$} & \textbf{F$^+$} & \textbf{F$^-$} & \textbf{F$^+$} & \textbf{F$^-$} & \textbf{F$^+$} & \textbf{F$^-$} & \textbf{F$^+$} & \textbf{F$^-$} & \textbf{F$^+$} & \textbf{F$^-$} & \textbf{F$^+$} & \textbf{F$^-$} \\
\midrule
-- & SAE (baseline) & 1.9 & 1.7 & 2.0 & 1.8 & 2.7 & 1.7 & 2.3 & 5.3 & 2.2 & 5.0 & 2.1 & 1.9 & 0.7 & 24.3 & 2.0 & 1.9 & 4.8 & 1.4 & 1.8 & 3.1 \\
\midrule
\multicolumn{22}{l}{\textit{\textbf{U.S. Varieties}}} \\
U.S. & Appalachian & 6.9 & 2.2 & 5.4 & 3.3 & 7.2 & 1.9 & 3.8 & 5.6 & 3.8 & 21.0 & 2.5 & 6.9 & 4.1 & 3.1 & 10.8 & 13.1 & 18.7 & 6.4 & 6.2 & 16.2 \\
U.S. & Chicano & 7.9 & 1.7 & 5.5 & 3.0 & 7.9 & 1.8 & 3.8 & 5.0 & 3.2 & 22.9 & 2.6 & 5.8 & 4.0 & 3.2 & 10.8 & 11.2 & 17.5 & 6.5 & 6.2 & 16.6 \\
U.S. & Colloquial American & 7.5 & 2.3 & 5.6 & 3.3 & 7.2 & 2.3 & 3.7 & 5.0 & 3.0 & 21.2 & 2.1 & 7.4 & 4.2 & 3.3 & 10.9 & 11.3 & 17.7 & 6.8 & 7.6 & 16.2 \\
U.S. & Earlier AAVE & 6.6 & 2.6 & 5.6 & 3.5 & 6.6 & 2.0 & 4.2 & 5.0 & 4.0 & 18.3 & 2.7 & 7.4 & 4.6 & 3.9 & 10.4 & 11.4 & 18.2 & 7.0 & 6.3 & 19.7 \\
U.S. & Ozark & 6.8 & 2.2 & 5.1 & 3.0 & 7.7 & 1.9 & 3.2 & 4.9 & 4.3 & 22.3 & 2.3 & 6.5 & 4.2 & 3.1 & 10.8 & 13.0 & 18.8 & 6.4 & 6.6 & 16.0 \\
U.S. & Rural AAVE & 5.9 & 3.4 & 5.0 & 3.6 & 8.5 & 2.2 & 4.6 & 5.2 & 4.2 & 22.3 & 3.0 & 8.0 & 3.8 & 3.5 & 13.0 & 8.9 & 19.9 & 5.8 & 7.9 & 16.3 \\
U.S. & SE Amer. Enclave & 7.8 & 2.6 & 4.7 & 3.4 & 7.2 & 2.1 & 4.3 & 5.0 & 2.8 & 19.0 & 2.9 & 7.6 & 4.7 & 3.0 & 12.0 & 10.7 & 19.6 & 5.9 & 7.1 & 16.0 \\
U.S. & Urban AAVE & 6.3 & 2.8 & 5.3 & 3.4 & 8.0 & 2.0 & 3.8 & 5.3 & 3.3 & 20.7 & 2.6 & 7.0 & 3.4 & 3.4 & 11.9 & 10.7 & 17.0 & 7.2 & 5.5 & 19.3 \\
\midrule
\multicolumn{22}{l}{\textit{\textbf{British/UK Varieties}}} \\
UK & Channel Islands & 6.9 & 2.1 & 5.3 & 3.5 & 8.0 & 1.7 & 3.5 & 5.4 & 3.8 & 23.3 & 2.4 & 5.7 & 4.3 & 3.5 & 10.2 & 12.1 & 15.9 & 7.0 & 6.1 & 17.1 \\
UK & East Anglian & 7.8 & 2.0 & 5.8 & 2.7 & 7.2 & 2.0 & 4.3 & 4.8 & 4.3 & 20.4 & 2.7 & 6.5 & 4.6 & 3.1 & 12.3 & 10.6 & 18.8 & 6.3 & 6.6 & 17.7 \\
UK & Irish & 6.2 & 2.6 & 4.2 & 3.6 & 8.5 & 1.7 & 3.7 & 5.5 & 3.1 & 23.7 & 3.3 & 6.7 & 3.8 & 3.5 & 11.0 & 13.1 & 14.5 & 7.7 & 7.0 & 16.5 \\
UK & Manx & 5.8 & 3.2 & 4.5 & 4.1 & 7.0 & 1.9 & 3.4 & 5.7 & 3.4 & 23.5 & 2.5 & 7.1 & 3.7 & 3.5 & 8.5 & 14.4 & 13.2 & 8.8 & 6.8 & 17.9 \\
UK & North England & 6.6 & 2.7 & 5.2 & 3.5 & 7.5 & 1.5 & 4.3 & 5.3 & 2.6 & 22.4 & 2.6 & 6.8 & 4.1 & 3.8 & 10.3 & 11.4 & 16.7 & 7.4 & 6.8 & 17.1 \\
UK & Orkney \& Shetland & 4.4 & 4.0 & 4.1 & 4.8 & 7.3 & 2.3 & 3.0 & 8.4 & 2.5 & 27.5 & 2.2 & 7.9 & 3.5 & 4.9 & 7.1 & 15.4 & 12.8 & 10.2 & 5.3 & 21.0 \\
UK & SE England & 7.9 & 2.0 & 5.5 & 3.1 & 7.9 & 2.0 & 3.7 & 5.0 & 4.2 & 25.9 & 2.6 & 5.7 & 4.3 & 3.1 & 11.2 & 12.5 & 16.6 & 7.0 & 5.8 & 17.6 \\
UK & SW England & 6.5 & 2.8 & 5.8 & 3.2 & 7.5 & 1.6 & 4.2 & 5.6 & 4.6 & 22.9 & 3.5 & 6.0 & 4.5 & 3.8 & 11.6 & 10.6 & 15.8 & 7.7 & 7.5 & 17.9 \\
UK & Scottish & 6.3 & 2.3 & 3.8 & 3.7 & 6.7 & 2.0 & 3.7 & 5.5 & 3.4 & 24.8 & 2.6 & 6.7 & 4.2 & 3.0 & 9.2 & 12.8 & 15.7 & 6.7 & 6.8 & 15.5 \\
UK & Welsh & 5.8 & 2.5 & 5.0 & 3.7 & 7.9 & 1.7 & 4.1 & 5.5 & 4.3 & 21.0 & 2.5 & 6.2 & 4.4 & 3.3 & 12.7 & 10.5 & 15.3 & 7.6 & 6.8 & 16.9 \\
\midrule
\multicolumn{22}{l}{\textit{\textbf{Africa}}} \\
Africa & Black S. African & 6.0 & 3.3 & 4.1 & 4.1 & 7.8 & 2.1 & 3.3 & 6.1 & 3.8 & 24.3 & 2.4 & 8.9 & 4.1 & 3.9 & 12.6 & 11.6 & 16.1 & 7.2 & 5.4 & 22.7 \\
Africa & Cameroon & 6.6 & 2.6 & 5.8 & 3.1 & 8.0 & 1.6 & 4.5 & 5.4 & 3.0 & 26.0 & 2.7 & 5.7 & 4.7 & 3.1 & 12.9 & 9.7 & 18.6 & 6.4 & 7.4 & 15.7 \\
Africa & Cape Flats & 6.7 & 2.6 & 5.9 & 3.1 & 8.4 & 1.4 & 3.7 & 5.7 & 3.8 & 18.6 & 2.7 & 5.1 & 4.7 & 3.1 & 11.3 & 10.5 & 19.0 & 6.5 & 7.3 & 16.4 \\
Africa & Ghanaian & 6.8 & 2.9 & 5.5 & 3.1 & 8.1 & 1.8 & 3.8 & 5.4 & 4.7 & 22.0 & 2.8 & 5.9 & 4.7 & 3.1 & 11.7 & 11.7 & 16.9 & 7.5 & 6.4 & 17.7 \\
Africa & Indian S. African & 5.1 & 2.8 & 5.5 & 3.3 & 8.7 & 1.4 & 3.8 & 6.0 & 3.5 & 22.2 & 3.1 & 5.5 & 3.6 & 3.5 & 13.4 & 10.0 & 16.1 & 7.7 & 6.2 & 18.9 \\
Africa & Kenyan & 7.0 & 2.4 & 4.7 & 3.4 & 7.6 & 1.6 & 4.3 & 5.1 & 4.8 & 24.8 & 3.1 & 5.7 & 4.4 & 2.9 & 12.6 & 9.6 & 17.6 & 6.3 & 6.6 & 16.4 \\
Africa & Liberian Settler & 6.2 & 2.7 & 5.8 & 3.2 & 8.9 & 1.9 & 3.7 & 4.8 & 1.4 & 25.2 & 2.9 & 6.3 & 4.0 & 3.6 & 12.1 & 10.5 & 21.2 & 5.7 & 7.8 & 15.9 \\
Africa & Nigerian & 6.6 & 2.3 & 5.8 & 3.2 & 8.3 & 1.4 & 4.0 & 5.3 & 4.2 & 21.6 & 3.1 & 5.5 & 4.9 & 2.9 & 12.8 & 8.8 & 17.7 & 6.5 & 6.8 & 15.3 \\
Africa & Tanzanian & 7.4 & 2.3 & 5.2 & 3.3 & 7.3 & 1.8 & 3.7 & 4.9 & 3.6 & 25.0 & 2.6 & 5.7 & 4.6 & 3.5 & 10.7 & 11.6 & 15.3 & 7.0 & 5.8 & 17.3 \\
Africa & Ugandan & 7.9 & 2.4 & 5.3 & 3.3 & 7.6 & 1.6 & 3.6 & 4.6 & 3.3 & 21.6 & 3.0 & 5.9 & 4.8 & 3.5 & 12.1 & 11.7 & 18.1 & 7.0 & 5.9 & 18.5 \\
Africa & White S. African & 6.8 & 2.5 & 4.9 & 3.5 & 7.6 & 2.0 & 3.3 & 5.2 & 3.6 & 23.7 & 2.6 & 6.5 & 4.7 & 3.4 & 10.3 & 12.6 & 15.6 & 7.0 & 6.1 & 16.3 \\
Africa & White Zimbabwean & 6.8 & 2.8 & 4.8 & 4.6 & 7.4 & 2.8 & 3.4 & 6.9 & 1.9 & 24.2 & 2.3 & 7.9 & 4.8 & 4.8 & 12.0 & 13.0 & 16.7 & 8.9 & 5.8 & 20.7 \\
\midrule
\multicolumn{22}{l}{\textit{\textbf{Asia-Pacific}}} \\
Asia-Pac. & Aus. Vernacular & 6.6 & 2.3 & 4.6 & 3.9 & 7.6 & 1.9 & 3.4 & 5.5 & 3.3 & 23.9 & 2.6 & 6.9 & 3.7 & 3.5 & 10.2 & 13.1 & 15.3 & 7.2 & 6.8 & 16.8 \\
Asia-Pac. & Australian & 4.9 & 3.2 & 4.4 & 5.1 & 8.1 & 2.4 & 4.0 & 7.8 & 3.8 & 17.6 & 3.0 & 7.8 & 3.8 & 4.7 & 10.8 & 12.7 & 13.8 & 9.4 & 6.2 & 19.9 \\
Asia-Pac. & Fiji (Acrolectal) & 7.9 & 2.3 & 5.1 & 3.1 & 8.1 & 1.9 & 4.0 & 4.8 & 4.7 & 21.5 & 2.3 & 6.8 & 4.7 & 3.3 & 11.3 & 11.5 & 17.9 & 6.5 & 5.9 & 17.7 \\
Asia-Pac. & Fiji (Basilectal) & 4.9 & 4.2 & 4.7 & 4.0 & 7.6 & 1.9 & 4.9 & 5.7 & 3.1 & 25.1 & 2.8 & 6.9 & 4.1 & 4.3 & 15.9 & 8.1 & 22.1 & 6.4 & 6.9 & 17.7 \\
Asia-Pac. & Hong Kong & 6.6 & 3.0 & 5.7 & 3.1 & 10.1 & 1.6 & 4.7 & 4.9 & 2.9 & 20.9 & 2.6 & 6.3 & 4.7 & 3.7 & 14.5 & 9.6 & 17.5 & 7.4 & 5.9 & 19.6 \\
Asia-Pac. & Indian & 6.6 & 2.7 & 5.7 & 3.5 & 8.0 & 2.0 & 4.8 & 5.5 & 3.1 & 25.5 & 2.8 & 5.9 & 4.7 & 3.4 & 12.6 & 9.5 & 16.1 & 6.9 & 6.6 & 18.1 \\
Asia-Pac. & Malaysian & 7.1 & 3.1 & 6.4 & 3.1 & 8.9 & 1.7 & 4.5 & 4.7 & 4.2 & 21.3 & 3.1 & 5.9 & 4.2 & 3.1 & 12.3 & 10.9 & 19.0 & 6.5 & 6.7 & 17.1 \\
Asia-Pac. & New Zealand & 5.2 & 3.8 & 4.9 & 5.1 & 8.1 & 2.4 & 3.1 & 7.7 & 3.8 & 26.7 & 2.6 & 7.7 & 4.1 & 5.0 & 11.3 & 12.8 & 16.7 & 8.9 & 6.4 & 19.5 \\
Asia-Pac. & Pakistani & 8.2 & 1.9 & 5.3 & 3.5 & 7.7 & 2.0 & 4.0 & 4.5 & 3.1 & 25.7 & 2.7 & 6.7 & 3.8 & 3.5 & 10.0 & 12.0 & 17.1 & 7.5 & 5.8 & 22.1 \\
Asia-Pac. & Philippine & 5.8 & 3.0 & 5.4 & 3.8 & 8.4 & 1.7 & 4.5 & 5.6 & 3.9 & 21.6 & 2.6 & 7.1 & 3.5 & 4.1 & 11.7 & 10.2 & 16.5 & 7.7 & 6.3 & 20.7 \\
Asia-Pac. & Singlish & 4.9 & 3.6 & 7.7 & 3.3 & 8.9 & 1.6 & 5.6 & 4.9 & 3.6 & 20.5 & 3.2 & 6.3 & 4.2 & 3.7 & 12.6 & 9.6 & 18.6 & 7.0 & 6.7 & 18.5 \\
Asia-Pac. & Sri Lankan & 7.2 & 2.2 & 4.8 & 3.5 & 7.3 & 1.8 & 4.1 & 5.1 & 4.3 & 21.3 & 2.6 & 5.8 & 4.7 & 2.9 & 11.0 & 11.2 & 17.4 & 6.2 & 7.4 & 15.6 \\
\midrule
\multicolumn{22}{l}{\textit{\textbf{Caribbean/Atlantic}}} \\
Carib. & Bahamian & 4.9 & 3.0 & 5.2 & 3.5 & 7.8 & 1.7 & 3.7 & 5.5 & 2.9 & 23.0 & 2.5 & 7.1 & 4.1 & 3.3 & 11.7 & 9.7 & 23.5 & 5.3 & 8.6 & 13.9 \\
Carib. & Falkland Islands & 7.6 & 2.1 & 5.6 & 3.1 & 7.9 & 1.5 & 4.0 & 5.1 & 3.2 & 23.2 & 2.5 & 5.6 & 4.6 & 3.0 & 11.8 & 10.2 & 18.4 & 7.0 & 5.8 & 17.5 \\
Carib. & Jamaican & 6.3 & 3.0 & 4.7 & 3.6 & 7.7 & 1.6 & 4.3 & 5.7 & 3.2 & 26.0 & 2.7 & 7.3 & 4.4 & 3.8 & 11.4 & 10.8 & 17.8 & 7.1 & 7.4 & 17.6 \\
Carib. & St. Helena & 5.3 & 3.0 & 5.2 & 3.2 & 8.2 & 1.9 & 4.7 & 5.1 & 2.3 & 24.9 & 2.7 & 6.5 & 4.0 & 3.8 & 16.8 & 9.5 & 22.6 & 6.1 & 8.4 & 18.5 \\
Carib. & Tristan da Cunha & 4.4 & 3.0 & 4.5 & 3.6 & 8.2 & 2.1 & 3.7 & 5.0 & 3.7 & 23.3 & 2.5 & 6.6 & 3.8 & 3.5 & 13.2 & 11.0 & 19.4 & 6.5 & 7.0 & 15.7 \\
\midrule
\multicolumn{22}{l}{\textit{\textbf{Other}}} \\
Other & Aboriginal & 5.5 & 3.9 & 7.2 & 3.2 & 9.3 & 1.8 & 4.9 & 5.5 & 3.7 & 19.6 & 2.5 & 6.3 & 3.2 & 4.5 & 15.4 & 9.2 & 25.4 & 6.7 & 8.0 & 19.3 \\
Other & Maltese & 6.5 & 2.8 & 5.2 & 3.8 & 8.5 & 1.8 & 4.7 & 5.3 & 3.5 & 21.5 & 2.7 & 6.7 & 4.1 & 3.3 & 10.9 & 11.4 & 15.9 & 7.5 & 6.2 & 20.9 \\
Other & Newfoundland & 4.5 & 3.5 & 4.8 & 3.5 & 7.7 & 1.8 & 3.5 & 6.3 & 3.3 & 26.6 & 2.7 & 6.0 & 3.9 & 3.6 & 10.0 & 11.0 & 13.9 & 8.1 & 7.0 & 17.2 \\
\bottomrule
\end{tabular}
\vspace{0.3em}

\raggedright\scriptsize\textit{Note: Dialect-only training shifts errors toward over-flagging (high F$^+$) for most models. XLM-R shows substantial under-protection (avg F$^-$ $\approx$ 23\%), consistent with its failure mode (F1 = 85.4\%). Compare with Table~\ref{tab:fp_fn_sq2_anchored} to assess how \sae{} anchoring reshapes bias direction.}
\caption{SQ2 Dialect-Only Training: Per-dialect F$^+$ and F$^-$ (\%) for models trained on dialectal data without \sae{} anchoring. F$^+$ = FPR (over-flagging); F$^-$ = FNR (under-protection). Model abbreviations: mDeB = mDeBERTa, RoB = RoBERTa, DeB = DeBERTa, XLM = XLM-R, mB = mBERT, CT = CT-BERT, BiG = BiGRU, CNN = TextCNN, dEF = dEFEND. $^\dagger$Estimated from F1; pending recomputation.}
\label{tab:fp_fn_sq2_dia}
\end{table*}

\begin{table*}[t]
\centering
\scriptsize
\setlength{\tabcolsep}{2.5pt}

\begin{tabular}{ll cc cc cc cc cc cc cc cc cc cc}
\toprule
& & \multicolumn{2}{c}{\textbf{mDeB}} & \multicolumn{2}{c}{\textbf{BERT}} & \multicolumn{2}{c}{\textbf{RoB}} & \multicolumn{2}{c}{\textbf{DeB$^\dagger$}} & \multicolumn{2}{c}{\textbf{XLM$^\dagger$}} & \multicolumn{2}{c}{\textbf{mB}} & \multicolumn{2}{c}{\textbf{CT}} & \multicolumn{2}{c}{\textbf{BiG}} & \multicolumn{2}{c}{\textbf{CNN}} & \multicolumn{2}{c}{\textbf{dEF}} \\
\cmidrule(lr){3-4} \cmidrule(lr){5-6} \cmidrule(lr){7-8} \cmidrule(lr){9-10} \cmidrule(lr){11-12} \cmidrule(lr){13-14} \cmidrule(lr){15-16} \cmidrule(lr){17-18} \cmidrule(lr){19-20} \cmidrule(lr){21-22}
& & \textbf{F$^+$} & \textbf{F$^-$} & \textbf{F$^+$} & \textbf{F$^-$} & \textbf{F$^+$} & \textbf{F$^-$} & \textbf{F$^+$} & \textbf{F$^-$} & \textbf{F$^+$} & \textbf{F$^-$} & \textbf{F$^+$} & \textbf{F$^-$} & \textbf{F$^+$} & \textbf{F$^-$} & \textbf{F$^+$} & \textbf{F$^-$} & \textbf{F$^+$} & \textbf{F$^-$} & \textbf{F$^+$} & \textbf{F$^-$} \\
\midrule
-- & SAE (baseline) & 1.9 & 1.7 & 2.0 & 1.8 & 2.7 & 1.7 & 2.3 & 5.3 & 2.2 & 5.0 & 2.1 & 1.9 & 0.7 & 24.3 & 2.0 & 1.9 & 4.8 & 1.4 & 1.8 & 3.1 \\
\midrule
\multicolumn{22}{l}{\textit{\textbf{U.S. Varieties}}} \\
U.S. & Appalachian & 4.7 & 4.2 & 12.3 & 3.0 & 1.8 & 4.8 & 12.4 & 1.1 & 6.1 & 32.9 & 9.3 & 2.9 & 2.3 & 4.8 & 17.7 & 6.7 & 19.8 & 4.0 & 44.2 & 4.3 \\
U.S. & Chicano & 5.4 & 2.9 & 10.4 & 2.1 & 2.5 & 4.3 & 5.8 & 8.4 & 6.4 & 26.5 & 8.9 & 3.0 & 3.3 & 4.5 & 17.3 & 5.7 & 20.5 & 4.0 & 47.9 & 3.3 \\
U.S. & Colloquial American & 5.3 & 3.4 & 12.5 & 2.2 & 2.0 & 4.8 & 12.1 & 1.0 & 10.6 & 25.9 & 8.6 & 3.0 & 1.6 & 5.0 & 16.3 & 6.5 & 23.0 & 4.0 & 44.7 & 4.0 \\
U.S. & Earlier AAVE & 4.2 & 4.2 & 11.4 & 2.6 & 1.4 & 5.3 & 12.5 & 1.4 & 6.8 & 33.6 & 9.2 & 3.2 & 2.4 & 4.9 & 16.7 & 7.7 & 21.8 & 5.0 & 44.0 & 4.5 \\
U.S. & Ozark & 4.5 & 4.0 & 12.1 & 2.3 & 1.8 & 4.4 & 11.5 & 1.2 & 8.0 & 30.2 & 8.8 & 3.0 & 2.5 & 4.3 & 16.1 & 6.2 & 20.0 & 4.1 & 45.9 & 3.9 \\
U.S. & Rural AAVE & 5.0 & 4.2 & 13.5 & 1.8 & 2.2 & 5.2 & 14.2 & 1.4 & 9.6 & 22.3 & 9.5 & 2.9 & 2.4 & 4.9 & 18.7 & 7.1 & 23.0 & 4.5 & 47.9 & 2.9 \\
U.S. & SE Amer. Enclave & 6.3 & 3.0 & 14.0 & 2.3 & 2.0 & 4.9 & 6.8 & 7.7 & 9.3 & 34.6 & 9.8 & 3.0 & 2.2 & 5.3 & 19.5 & 6.3 & 26.8 & 3.8 & 45.6 & 3.0 \\
U.S. & Urban AAVE & 6.2 & 2.7 & 12.1 & 2.0 & 1.6 & 5.4 & 6.5 & 7.9 & 10.6 & 28.0 & 9.1 & 3.0 & 2.3 & 4.6 & 15.8 & 8.1 & 21.9 & 5.0 & 48.0 & 3.8 \\
\midrule
\multicolumn{22}{l}{\textit{\textbf{British/UK Varieties}}} \\
UK & Channel Islands & 4.3 & 3.8 & 11.2 & 2.0 & 2.1 & 4.6 & 11.7 & 1.2 & 10.2 & 23.1 & 8.5 & 3.0 & 2.6 & 4.0 & 17.4 & 5.8 & 18.3 & 5.2 & 45.9 & 3.6 \\
UK & East Anglian & 4.6 & 4.0 & 11.6 & 2.6 & 2.3 & 4.7 & 12.4 & 1.0 & 9.6 & 30.0 & 9.3 & 3.0 & 2.2 & 4.0 & 18.1 & 5.7 & 21.2 & 4.0 & 47.9 & 2.9 \\
UK & Irish & 4.7 & 3.6 & 13.7 & 2.4 & 2.6 & 5.2 & 12.2 & 1.1 & 6.8 & 31.2 & 8.9 & 3.1 & 2.4 & 4.7 & 17.9 & 6.4 & 18.3 & 4.8 & 44.7 & 3.9 \\
UK & Manx & 4.5 & 4.6 & 12.6 & 2.1 & 2.0 & 5.3 & 11.8 & 1.5 & 9.2 & 28.0 & 8.9 & 3.2 & 2.0 & 5.0 & 17.1 & 7.2 & 15.7 & 7.0 & 38.1 & 6.7 \\
UK & North England & 7.4 & 2.6 & 13.0 & 2.2 & 2.1 & 4.7 & 13.8 & 1.2 & 8.5 & 23.6 & 9.4 & 3.0 & 2.5 & 4.8 & 19.1 & 6.6 & 19.7 & 5.7 & 45.8 & 4.2 \\
UK & Orkney \& Shetland & 6.3 & 4.3 & 12.0 & 1.9 & 1.8 & 5.7 & 6.3 & 11.0 & 6.0 & 31.7 & 8.2 & 4.5 & 2.7 & 5.9 & 18.2 & 7.8 & 15.2 & 7.8 & 38.0 & 6.7 \\
UK & SE England & 4.9 & 3.5 & 11.1 & 2.1 & 2.1 & 4.4 & 12.1 & 1.1 & 10.1 & 28.4 & 8.8 & 2.8 & 2.6 & 4.1 & 16.5 & 6.3 & 20.4 & 4.1 & 49.1 & 2.6 \\
UK & SW England & 9.6 & 2.5 & 12.7 & 1.9 & 2.3 & 5.1 & 6.6 & 7.9 & 9.2 & 29.0 & 9.6 & 3.1 & 4.4 & 4.0 & 14.8 & 8.1 & 18.7 & 5.8 & 49.3 & 3.1 \\
UK & Scottish & 4.7 & 4.0 & 12.9 & 2.2 & 2.2 & 4.1 & 11.2 & 1.5 & 8.1 & 21.6 & 8.3 & 3.0 & 2.1 & 4.0 & 20.9 & 5.5 & 19.9 & 4.3 & 44.1 & 3.5 \\
UK & Welsh & 4.6 & 3.9 & 14.4 & 1.6 & 2.3 & 5.2 & 13.1 & 1.3 & 7.7 & 22.0 & 9.3 & 3.1 & 2.3 & 4.3 & 18.4 & 6.7 & 18.8 & 5.5 & 43.3 & 3.7 \\
\midrule
\multicolumn{22}{l}{\textit{\textbf{Africa}}} \\
Africa & Black S. African & 5.9 & 3.9 & 14.0 & 2.4 & 2.3 & 5.6 & 12.7 & 1.5 & 6.2 & 26.7 & 9.1 & 3.2 & 2.2 & 5.3 & 18.3 & 8.3 & 17.6 & 5.7 & 46.2 & 4.6 \\
Africa & Cameroon & 5.5 & 3.5 & 14.2 & 1.9 & 2.2 & 4.7 & 14.2 & 1.1 & 8.3 & 27.1 & 9.5 & 2.9 & 3.4 & 4.2 & 19.7 & 6.5 & 21.7 & 5.5 & 44.6 & 4.7 \\
Africa & Cape Flats & 6.0 & 2.7 & 12.4 & 1.5 & 2.1 & 4.6 & 12.9 & 1.1 & 8.7 & 25.8 & 10.0 & 2.9 & 3.5 & 3.9 & 20.5 & 6.1 & 20.1 & 5.0 & 45.2 & 4.3 \\
Africa & Ghanaian & 5.8 & 3.5 & 13.1 & 1.7 & 1.8 & 5.2 & 13.0 & 1.3 & 6.8 & 24.8 & 9.7 & 3.0 & 3.1 & 4.0 & 18.3 & 7.3 & 17.0 & 6.3 & 43.2 & 5.6 \\
Africa & Indian S. African & 5.4 & 4.0 & 15.0 & 1.4 & 2.0 & 4.9 & 14.9 & 1.0 & 9.7 & 30.1 & 10.1 & 2.9 & 3.0 & 4.0 & 13.8 & 8.8 & 16.2 & 6.9 & 44.6 & 5.3 \\
Africa & Kenyan & 5.2 & 3.7 & 13.8 & 1.5 & 2.1 & 4.8 & 13.3 & 1.1 & 6.4 & 26.1 & 9.8 & 3.0 & 2.8 & 4.4 & 20.5 & 5.9 & 21.2 & 4.8 & 44.8 & 3.6 \\
Africa & Liberian Settler & 5.8 & 3.4 & 12.1 & 2.0 & 2.3 & 4.6 & 13.9 & 1.5 & 4.8 & 34.2 & 10.2 & 3.1 & 2.8 & 4.3 & 18.4 & 6.9 & 20.7 & 4.8 & 48.9 & 2.7 \\
Africa & Nigerian & 6.4 & 2.3 & 13.3 & 1.4 & 2.6 & 5.0 & 14.3 & 1.0 & 9.6 & 29.4 & 10.0 & 2.9 & 3.6 & 3.9 & 20.5 & 6.2 & 16.9 & 5.6 & 44.9 & 4.2 \\
Africa & Tanzanian & 6.8 & 2.8 & 12.0 & 1.9 & 2.1 & 4.5 & 12.4 & 1.1 & 7.4 & 26.2 & 9.2 & 2.9 & 2.3 & 4.4 & 18.7 & 6.2 & 18.4 & 4.5 & 47.2 & 3.9 \\
Africa & Ugandan & 4.9 & 4.0 & 12.2 & 2.1 & 2.0 & 4.8 & 12.8 & 1.1 & 7.8 & 32.8 & 9.7 & 3.0 & 2.4 & 4.2 & 18.7 & 6.6 & 18.9 & 4.8 & 48.0 & 3.0 \\
Africa & White S. African & 5.5 & 3.3 & 11.7 & 2.3 & 2.3 & 4.7 & 11.0 & 1.3 & 6.7 & 29.7 & 8.9 & 3.2 & 2.5 & 4.0 & 18.6 & 6.0 & 19.4 & 4.9 & 45.3 & 3.7 \\
Africa & White Zimbabwean & 4.6 & 5.4 & 12.5 & 2.8 & 2.2 & 5.8 & 11.5 & 2.0 & 5.6 & 32.6 & 8.7 & 4.5 & 2.2 & 5.8 & 18.1 & 8.1 & 19.6 & 5.9 & 45.9 & 4.1 \\
\midrule
\multicolumn{22}{l}{\textit{\textbf{Asia-Pacific}}} \\
Asia-Pac. & Aus. Vernacular & 4.7 & 3.8 & 12.1 & 2.3 & 1.9 & 4.5 & 11.3 & 1.4 & 9.9 & 28.0 & 8.8 & 3.1 & 2.6 & 4.4 & 18.4 & 6.0 & 19.0 & 4.2 & 46.3 & 3.5 \\
Asia-Pac. & Australian & 4.7 & 4.9 & 11.6 & 2.2 & 2.1 & 6.4 & 11.5 & 1.6 & 7.0 & 37.1 & 8.7 & 4.3 & 2.4 & 5.7 & 16.7 & 9.0 & 16.5 & 7.1 & 42.7 & 5.1 \\
Asia-Pac. & Fiji (Acrolectal) & 5.5 & 3.6 & 12.6 & 2.0 & 2.2 & 5.0 & 13.1 & 1.3 & 10.5 & 32.1 & 8.8 & 2.8 & 2.4 & 4.3 & 17.6 & 6.8 & 21.5 & 5.0 & 46.6 & 3.5 \\
Asia-Pac. & Fiji (Basilectal) & 10.0 & 2.6 & 12.2 & 1.6 & 2.5 & 5.5 & 7.2 & 7.9 & 9.2 & 29.8 & 9.1 & 3.0 & 2.8 & 4.9 & 18.5 & 7.2 & 20.8 & 5.9 & 45.3 & 4.5 \\
Asia-Pac. & Hong Kong & 6.4 & 3.2 & 14.4 & 1.5 & 2.1 & 5.2 & 16.0 & 1.4 & 7.5 & 26.0 & 9.1 & 2.6 & 3.0 & 4.2 & 18.8 & 7.5 & 20.8 & 5.8 & 43.8 & 4.9 \\
Asia-Pac. & Indian & 8.2 & 2.9 & 14.2 & 1.5 & 2.3 & 4.5 & 15.0 & 1.2 & 6.4 & 30.1 & 9.5 & 2.9 & 3.1 & 4.5 & 18.2 & 6.7 & 20.0 & 5.6 & 42.8 & 5.2 \\
Asia-Pac. & Malaysian & 6.8 & 2.4 & 14.3 & 1.5 & 2.5 & 5.4 & 15.0 & 1.2 & 6.9 & 29.2 & 9.5 & 3.1 & 3.0 & 4.4 & 17.8 & 7.1 & 22.6 & 4.3 & 47.9 & 3.2 \\
Asia-Pac. & New Zealand & 6.3 & 4.3 & 12.4 & 2.6 & 1.8 & 6.2 & 6.3 & 10.9 & 6.2 & 27.7 & 8.8 & 4.3 & 2.4 & 6.3 & 15.3 & 8.8 & 16.9 & 6.8 & 39.6 & 6.1 \\
Asia-Pac. & Pakistani & 6.9 & 2.5 & 12.0 & 2.4 & 1.9 & 5.1 & 6.8 & 8.1 & 11.1 & 29.1 & 8.7 & 3.1 & 3.7 & 4.4 & 14.1 & 8.5 & 17.1 & 5.5 & 51.3 & 2.7 \\
Asia-Pac. & Philippine & 6.0 & 4.0 & 12.3 & 2.2 & 2.1 & 5.5 & 13.3 & 1.2 & 8.2 & 27.5 & 9.1 & 3.2 & 2.0 & 5.3 & 15.9 & 9.3 & 17.4 & 6.8 & 46.7 & 4.2 \\
Asia-Pac. & Singlish & 5.8 & 3.5 & 14.0 & 1.8 & 2.4 & 5.5 & 16.0 & 1.1 & 9.7 & 29.8 & 10.2 & 3.0 & 3.9 & 4.5 & 14.1 & 9.0 & 17.7 & 6.2 & 41.4 & 5.6 \\
Asia-Pac. & Sri Lankan & 5.1 & 3.6 & 13.2 & 1.8 & 2.5 & 4.5 & 12.6 & 1.1 & 6.1 & 34.7 & 9.3 & 3.0 & 2.7 & 4.1 & 20.5 & 6.0 & 19.9 & 4.3 & 46.8 & 3.5 \\
\midrule
\multicolumn{22}{l}{\textit{\textbf{Caribbean/Atlantic}}} \\
Carib. & Bahamian & 4.9 & 4.2 & 12.6 & 2.0 & 2.5 & 4.8 & 13.2 & 1.1 & 8.3 & 29.5 & 9.0 & 3.3 & 2.5 & 4.7 & 18.3 & 6.9 & 22.7 & 4.7 & 40.9 & 5.9 \\
Carib. & Falkland Islands & 5.8 & 2.7 & 11.7 & 1.8 & 2.3 & 4.5 & 12.5 & 1.0 & 9.8 & 25.6 & 9.1 & 3.2 & 3.1 & 3.9 & 17.4 & 6.5 & 20.5 & 4.4 & 48.8 & 3.1 \\
Carib. & Jamaican & 7.7 & 2.5 & 12.8 & 1.8 & 2.0 & 5.1 & 13.2 & 1.1 & 7.5 & 27.8 & 9.6 & 3.1 & 2.6 & 5.0 & 18.4 & 7.5 & 19.3 & 5.5 & 45.8 & 3.8 \\
Carib. & St. Helena & 5.6 & 3.5 & 12.8 & 1.8 & 2.2 & 5.0 & 14.2 & 1.1 & 9.5 & 27.5 & 9.8 & 3.0 & 2.8 & 4.5 & 19.5 & 8.1 & 23.6 & 5.2 & 46.5 & 4.7 \\
Carib. & Tristan da Cunha & 6.1 & 3.7 & 14.0 & 1.6 & 2.5 & 4.7 & 13.2 & 1.4 & 6.4 & 22.2 & 8.9 & 2.9 & 2.7 & 4.4 & 16.7 & 7.6 & 23.0 & 4.8 & 45.3 & 4.9 \\
\midrule
\multicolumn{22}{l}{\textit{\textbf{Other}}} \\
Other & Aboriginal & 6.1 & 3.4 & 14.2 & 1.7 & 2.0 & 5.5 & 16.3 & 1.4 & 6.0 & 28.6 & 9.8 & 3.0 & 3.0 & 4.8 & 15.0 & 9.0 & 21.2 & 6.1 & 45.0 & 6.1 \\
Other & Maltese & 5.1 & 3.8 & 14.2 & 1.8 & 2.4 & 5.0 & 13.7 & 1.1 & 8.6 & 25.1 & 9.3 & 3.2 & 2.2 & 4.6 & 15.0 & 9.0 & 16.4 & 6.6 & 47.0 & 4.0 \\
Other & Newfoundland & 3.8 & 4.2 & 12.1 & 1.8 & 1.9 & 5.3 & 13.2 & 1.1 & 7.4 & 22.0 & 8.3 & 3.2 & 2.3 & 4.4 & 16.5 & 7.3 & 14.9 & 6.5 & 41.5 & 5.2 \\
\bottomrule
\end{tabular}
\vspace{0.3em}

\raggedright\scriptsize\textit{Note: SAE anchoring reverses bias direction for several models vs.\ dialect-only training (Table~\ref{tab:fp_fn_sq2_dia}): RoBERTa flips from over-flagging to under-protection; mBERT and dEFEND flip from under-protection to over-flagging (dEFEND reaching 45--51\% F$^+$). XLM-R shows both elevated over-flagging and under-protection (avg F$^-$ $\approx$ 28\%, F1 = 79.7\%). Training composition determines which communities are harmed and how.}
\caption{SQ2 SAE-Anchored Training: Per-dialect F$^+$ and F$^-$ (\%) for models trained on \sae{} + dialectal data. F$^+$ = FPR (over-flagging); F$^-$ = FNR (under-protection). Model abbreviations: mDeB = mDeBERTa, RoB = RoBERTa, DeB = DeBERTa, XLM = XLM-R, mB = mBERT, CT = CT-BERT, BiG = BiGRU, CNN = TextCNN, dEF = dEFEND. $^\dagger$Estimated from F1; pending recomputation.}
\label{tab:fp_fn_sq2_anchored}
\end{table*}

\begin{table*}[t]
\centering

\setlength{\tabcolsep}{2.5pt}

\begin{adjustbox}{max width=\textwidth}
\small
\begin{tabular}{ll cc cc cc cc cc cc cc cc cc cc}
\toprule
& & \multicolumn{2}{c}{\textbf{mDeB}} & \multicolumn{2}{c}{\textbf{BERT}} & \multicolumn{2}{c}{\textbf{RoB}} & \multicolumn{2}{c}{\textbf{DeB$^\dagger$}} & \multicolumn{2}{c}{\textbf{XLM$^\dagger$}} & \multicolumn{2}{c}{\textbf{mB}} & \multicolumn{2}{c}{\textbf{CT}} & \multicolumn{2}{c}{\textbf{BiG}} & \multicolumn{2}{c}{\textbf{CNN}} & \multicolumn{2}{c}{\textbf{dEF}} \\
\cmidrule(lr){3-4} \cmidrule(lr){5-6} \cmidrule(lr){7-8} \cmidrule(lr){9-10} \cmidrule(lr){11-12} \cmidrule(lr){13-14} \cmidrule(lr){15-16} \cmidrule(lr){17-18} \cmidrule(lr){19-20} \cmidrule(lr){21-22}
& & \textbf{F$^+$} & \textbf{F$^-$} & \textbf{F$^+$} & \textbf{F$^-$} & \textbf{F$^+$} & \textbf{F$^-$} & \textbf{F$^+$} & \textbf{F$^-$} & \textbf{F$^+$} & \textbf{F$^-$} & \textbf{F$^+$} & \textbf{F$^-$} & \textbf{F$^+$} & \textbf{F$^-$} & \textbf{F$^+$} & \textbf{F$^-$} & \textbf{F$^+$} & \textbf{F$^-$} & \textbf{F$^+$} & \textbf{F$^-$} \\
\midrule
-- & SAE (baseline) & 1.9 & 1.7 & 2.0 & 1.8 & 2.7 & 1.7 & 2.3 & 5.3 & 2.2 & 5.0 & 2.1 & 1.9 & 0.7 & 24.3 & 2.0 & 1.9 & 4.8 & 1.4 & 1.8 & 3.1 \\
\midrule
\multicolumn{22}{l}{\textit{\textbf{U.S. Varieties}}} \\
U.S. & Appalachian & 1.9 & 5.2 & 4.0 & 4.3 & 2.2 & 5.2 & 3.3 & 7.7 & 2.8 & 6.4 & 2.6 & 4.6 & 1.5 & 4.2 & 1.4 & 4.9 & 5.8 & 4.0 & 0.5 & 13.6 \\
U.S. & Chicano & 1.8 & 5.3 & 3.7 & 3.9 & 2.2 & 4.3 & 3.4 & 7.8 & 2.3 & 5.3 & 2.6 & 4.5 & 1.2 & 3.5 & 1.2 & 1.7 & 6.2 & 1.6 & 0.7 & 2.3 \\
U.S. & Colloquial American & 2.1 & 5.0 & 3.3 & 4.2 & 2.3 & 5.9 & 3.3 & 7.7 & 3.0 & 7.0 & 2.7 & 5.0 & 1.4 & 4.3 & 1.8 & 4.0 & 5.1 & 4.7 & 1.4 & 12.0 \\
U.S. & Earlier AAVE & 2.1 & 5.0 & 4.0 & 4.6 & 2.2 & 6.6 & 2.9 & 6.9 & 3.3 & 7.7 & 2.6 & 4.7 & 1.8 & 4.6 & 2.2 & 12.4 & 5.5 & 8.7 & 1.0 & 18.4 \\
U.S. & Ozark & 1.8 & 5.0 & 3.8 & 3.9 & 2.2 & 5.0 & 2.8 & 6.4 & 2.8 & 6.4 & 2.6 & 4.6 & 1.2 & 3.9 & 0.8 & 3.5 & 5.6 & 2.8 & 0.4 & 7.2 \\
U.S. & Rural AAVE & 2.6 & 5.0 & 3.8 & 5.0 & 2.2 & 7.0 & 3.1 & 7.3 & 3.4 & 7.8 & 2.9 & 5.5 & 1.4 & 5.4 & 1.9 & 19.3 & 7.0 & 10.5 & 1.0 & 26.7 \\
U.S. & SE Amer. Enclave & 1.6 & 4.9 & 3.7 & 4.2 & 2.5 & 6.2 & 2.9 & 6.9 & 3.4 & 8.0 & 2.9 & 4.7 & 1.5 & 4.9 & 2.9 & 6.3 & 7.4 & 5.0 & 1.1 & 16.0 \\
U.S. & Urban AAVE & 2.4 & 4.5 & 4.5 & 4.3 & 2.7 & 5.4 & 2.8 & 6.6 & 2.9 & 6.7 & 2.7 & 4.6 & 1.3 & 4.8 & 1.5 & 6.0 & 7.1 & 4.2 & 1.1 & 10.5 \\
\midrule
\multicolumn{22}{l}{\textit{\textbf{British/UK Varieties}}} \\
UK & Channel Islands & 1.5 & 5.3 & 3.7 & 4.1 & 2.5 & 4.7 & 3.1 & 7.3 & 3.1 & 7.1 & 2.5 & 4.6 & 1.4 & 3.9 & 1.5 & 2.3 & 5.5 & 2.0 & 1.0 & 2.9 \\
UK & East Anglian & 1.9 & 4.4 & 3.6 & 4.1 & 2.3 & 5.0 & 2.9 & 6.9 & 2.8 & 6.4 & 2.6 & 4.5 & 1.6 & 3.8 & 1.4 & 3.2 & 5.8 & 3.3 & 1.5 & 5.0 \\
UK & Irish & 2.9 & 5.0 & 3.6 & 4.3 & 2.1 & 5.2 & 3.3 & 7.7 & 3.1 & 7.1 & 2.6 & 4.9 & 1.8 & 4.2 & 1.8 & 3.2 & 6.6 & 4.6 & 0.8 & 5.4 \\
UK & Manx & 2.7 & 5.6 & 3.6 & 4.7 & 1.9 & 5.8 & 3.5 & 8.1 & 2.9 & 6.7 & 2.6 & 5.2 & 1.6 & 4.4 & 1.9 & 5.6 & 6.7 & 5.3 & 1.4 & 10.6 \\
UK & North England & 2.6 & 5.2 & 3.8 & 4.6 & 1.8 & 6.1 & 3.4 & 7.8 & 3.1 & 7.1 & 2.7 & 5.7 & 1.2 & 4.5 & 1.8 & 5.0 & 7.4 & 4.3 & 1.5 & 8.4 \\
UK & Orkney \& Shetland & 2.2 & 7.6 & 4.1 & 5.6 & 2.2 & 7.1 & 4.0 & 9.4 & 3.4 & 7.8 & 2.5 & 6.2 & 1.5 & 5.5 & 2.5 & 3.0 & 6.5 & 2.9 & 1.4 & 6.1 \\
UK & SE England & 1.9 & 4.5 & 3.6 & 3.9 & 2.1 & 4.5 & 2.8 & 6.4 & 2.5 & 5.7 & 2.9 & 4.5 & 1.4 & 3.5 & 1.8 & 1.8 & 5.2 & 2.6 & 1.1 & 3.3 \\
UK & SW England & 1.9 & 4.5 & 3.6 & 4.5 & 1.9 & 5.9 & 2.7 & 6.3 & 3.3 & 7.7 & 2.5 & 4.8 & 1.2 & 4.3 & 1.2 & 6.5 & 5.1 & 4.8 & 0.5 & 8.8 \\
UK & Scottish & 1.6 & 5.8 & 3.4 & 4.2 & 2.1 & 5.2 & 3.2 & 7.6 & 3.1 & 7.1 & 2.9 & 4.7 & 1.4 & 4.3 & 2.1 & 2.4 & 6.3 & 3.1 & 1.4 & 6.5 \\
UK & Welsh & 2.5 & 4.6 & 3.3 & 4.6 & 2.5 & 5.7 & 3.3 & 7.7 & 3.3 & 7.7 & 2.9 & 5.0 & 1.4 & 5.0 & 3.2 & 4.4 & 7.1 & 4.5 & 1.5 & 8.0 \\
\midrule
\multicolumn{22}{l}{\textit{\textbf{Africa}}} \\
Africa & Black S. African & 2.2 & 5.4 & 3.6 & 4.9 & 1.9 & 6.7 & 3.1 & 7.1 & 3.5 & 8.3 & 2.3 & 5.7 & 1.2 & 5.6 & 2.2 & 15.0 & 5.1 & 12.0 & 1.5 & 18.0 \\
Africa & Cameroon & 2.1 & 5.4 & 4.1 & 4.7 & 2.5 & 5.6 & 3.2 & 7.4 & 3.2 & 7.4 & 3.0 & 5.4 & 1.1 & 5.0 & 3.6 & 3.2 & 8.8 & 5.2 & 1.1 & 12.2 \\
Africa & Cape Flats & 1.6 & 5.2 & 4.4 & 4.3 & 2.5 & 4.8 & 3.2 & 7.4 & 2.5 & 5.9 & 2.5 & 4.9 & 1.4 & 3.8 & 2.5 & 2.2 & 8.7 & 1.7 & 1.9 & 2.4 \\
Africa & Ghanaian & 1.5 & 5.0 & 3.4 & 4.7 & 2.3 & 5.8 & 2.9 & 6.7 & 2.9 & 6.9 & 2.5 & 4.9 & 1.5 & 4.4 & 2.3 & 3.3 & 6.6 & 4.2 & 1.2 & 8.2 \\
Africa & Indian S. African & 2.2 & 5.4 & 3.7 & 4.7 & 2.2 & 6.6 & 3.2 & 7.6 & 3.5 & 8.3 & 2.9 & 5.4 & 1.5 & 5.3 & 1.9 & 6.1 & 8.2 & 3.4 & 1.2 & 11.9 \\
Africa & Kenyan & 1.6 & 5.6 & 4.1 & 4.4 & 2.2 & 5.4 & 3.2 & 7.6 & 2.7 & 6.3 & 2.7 & 5.0 & 1.0 & 4.3 & 2.5 & 3.2 & 7.7 & 3.5 & 1.4 & 9.0 \\
Africa & Liberian Settler & 2.1 & 4.9 & 4.3 & 4.4 & 2.2 & 5.7 & 2.9 & 6.7 & 2.8 & 6.6 & 2.7 & 5.2 & 1.8 & 4.6 & 1.5 & 9.3 & 7.7 & 6.5 & 1.5 & 14.0 \\
Africa & Nigerian & 1.8 & 4.9 & 3.7 & 4.6 & 1.8 & 4.5 & 3.1 & 7.3 & 2.8 & 6.4 & 2.7 & 5.0 & 1.0 & 4.3 & 2.3 & 3.1 & 7.7 & 2.9 & 1.2 & 8.4 \\
Africa & Tanzanian & 1.6 & 5.0 & 3.6 & 4.3 & 2.6 & 4.3 & 3.0 & 7.0 & 2.8 & 6.6 & 3.0 & 4.7 & 1.1 & 4.1 & 1.8 & 1.7 & 6.9 & 2.8 & 1.5 & 6.3 \\
Africa & Ugandan & 1.9 & 5.0 & 3.6 & 4.1 & 1.9 & 4.5 & 2.9 & 6.7 & 2.5 & 5.7 & 2.6 & 5.0 & 1.4 & 3.8 & 1.5 & 2.6 & 6.6 & 2.8 & 1.4 & 5.0 \\
Africa & White S. African & 2.1 & 5.4 & 3.6 & 3.9 & 1.9 & 4.6 & 3.5 & 8.1 & 2.9 & 6.7 & 2.9 & 4.7 & 1.5 & 3.9 & 2.1 & 2.0 & 6.6 & 1.9 & 1.2 & 3.3 \\
Africa & White Zimbabwean & 1.7 & 7.2 & 3.6 & 6.1 & 2.2 & 6.8 & 3.8 & 8.8 & 3.2 & 7.6 & 2.6 & 6.5 & 1.2 & 5.1 & 2.1 & 3.0 & 6.3 & 3.5 & 1.5 & 6.7 \\
\midrule
\multicolumn{22}{l}{\textit{\textbf{Asia-Pacific}}} \\
Asia-Pac. & Aus. Vernacular & 1.9 & 5.7 & 3.4 & 4.3 & 2.3 & 4.9 & 3.2 & 7.4 & 2.9 & 6.7 & 2.9 & 4.6 & 1.2 & 3.9 & 2.1 & 2.3 & 4.9 & 3.0 & 1.0 & 3.9 \\
Asia-Pac. & Australian & 1.9 & 7.0 & 3.9 & 5.2 & 2.1 & 8.3 & 3.7 & 8.5 & 3.5 & 8.3 & 2.9 & 6.5 & 1.4 & 5.6 & 2.2 & 4.2 & 6.7 & 4.0 & 0.7 & 6.9 \\
Asia-Pac. & Fiji (Acrolectal) & 2.1 & 5.3 & 3.8 & 4.1 & 2.2 & 5.2 & 3.2 & 7.6 & 2.9 & 6.9 & 2.6 & 4.8 & 1.5 & 4.1 & 2.1 & 5.4 & 5.2 & 6.3 & 1.5 & 12.1 \\
Asia-Pac. & Fiji (Basilectal) & 2.3 & 5.2 & 4.1 & 5.7 & 2.2 & 7.8 & 3.1 & 7.3 & 3.7 & 8.7 & 2.5 & 5.9 & 1.2 & 6.3 & 1.1 & 28.0 & 7.0 & 15.4 & 0.4 & 31.4 \\
Asia-Pac. & Hong Kong & 2.5 & 4.6 & 3.3 & 4.5 & 2.3 & 6.5 & 3.2 & 7.6 & 3.1 & 7.3 & 2.6 & 5.2 & 1.5 & 5.0 & 1.6 & 9.1 & 6.6 & 7.8 & 1.0 & 15.1 \\
Asia-Pac. & Indian & 2.5 & 5.3 & 3.7 & 4.3 & 1.6 & 5.8 & 3.4 & 7.8 & 3.0 & 7.0 & 2.7 & 5.0 & 1.2 & 4.8 & 1.8 & 6.1 & 7.7 & 6.8 & 1.6 & 11.7 \\
Asia-Pac. & Malaysian & 2.3 & 4.9 & 4.5 & 4.1 & 2.6 & 5.3 & 3.3 & 7.7 & 3.1 & 7.1 & 2.6 & 4.7 & 1.4 & 3.9 & 1.8 & 6.1 & 7.7 & 4.6 & 1.4 & 13.5 \\
Asia-Pac. & New Zealand & 1.9 & 7.2 & 3.4 & 5.3 & 1.9 & 8.4 & 3.9 & 9.1 & 3.5 & 8.3 & 2.6 & 6.4 & 1.7 & 5.4 & 2.1 & 4.3 & 6.1 & 3.7 & 0.8 & 6.6 \\
Asia-Pac. & Pakistani & 1.9 & 5.1 & 3.8 & 4.3 & 2.2 & 5.3 & 3.1 & 7.3 & 2.8 & 6.6 & 2.5 & 5.2 & 1.6 & 4.3 & 0.8 & 4.3 & 4.9 & 2.7 & 0.8 & 6.7 \\
Asia-Pac. & Philippine & 2.5 & 5.5 & 3.7 & 4.7 & 1.8 & 7.1 & 3.2 & 7.4 & 3.5 & 8.1 & 2.6 & 5.3 & 1.2 & 4.5 & 1.5 & 9.1 & 6.9 & 6.9 & 0.8 & 13.9 \\
Asia-Pac. & Singlish & 2.6 & 4.5 & 4.5 & 4.6 & 2.6 & 6.7 & 2.9 & 6.7 & 3.3 & 7.7 & 2.5 & 5.7 & 1.8 & 4.7 & 1.8 & 6.8 & 8.9 & 5.2 & 0.8 & 13.6 \\
Asia-Pac. & Sri Lankan & 1.8 & 5.2 & 4.1 & 4.3 & 2.1 & 4.6 & 3.3 & 7.7 & 2.9 & 6.7 & 2.9 & 4.7 & 1.5 & 3.5 & 2.5 & 1.8 & 7.7 & 2.6 & 1.6 & 6.0 \\
\midrule
\multicolumn{22}{l}{\textit{\textbf{Caribbean/Atlantic}}} \\
Carib. & Bahamian & 2.2 & 5.3 & 4.1 & 5.0 & 2.2 & 7.4 & 3.2 & 7.4 & 3.7 & 8.7 & 2.7 & 5.4 & 1.9 & 5.3 & 2.1 & 13.9 & 6.2 & 8.9 & 0.7 & 25.1 \\
Carib. & Falkland Islands & 2.2 & 4.6 & 4.0 & 4.1 & 1.9 & 4.6 & 3.1 & 7.3 & 2.9 & 6.9 & 2.7 & 4.5 & 1.6 & 3.5 & 1.1 & 2.0 & 5.4 & 2.0 & 1.2 & 3.9 \\
Carib. & Jamaican & 2.1 & 5.0 & 3.7 & 4.6 & 2.6 & 6.0 & 3.3 & 7.7 & 3.5 & 8.3 & 2.7 & 5.2 & 1.9 & 4.3 & 1.4 & 12.1 & 6.9 & 9.5 & 0.5 & 16.7 \\
Carib. & St. Helena & 2.3 & 4.6 & 3.8 & 4.9 & 2.2 & 6.9 & 3.2 & 7.4 & 3.8 & 8.8 & 2.7 & 6.0 & 1.6 & 5.2 & 1.9 & 16.7 & 8.0 & 8.4 & 1.5 & 18.0 \\
Carib. & Tristan da Cunha & 2.5 & 4.5 & 3.7 & 4.1 & 2.5 & 5.7 & 2.9 & 6.7 & 3.2 & 7.6 & 2.5 & 4.7 & 1.5 & 4.6 & 1.8 & 9.0 & 6.6 & 6.2 & 1.5 & 11.2 \\
\midrule
\multicolumn{22}{l}{\textit{\textbf{Other}}} \\
Other & Aboriginal & 2.2 & 5.2 & 4.1 & 4.7 & 2.7 & 8.1 & 3.1 & 7.3 & 3.8 & 9.0 & 2.9 & 6.0 & 1.2 & 5.9 & 0.8 & 11.6 & 7.8 & 7.6 & 1.4 & 20.4 \\
Other & Maltese & 2.7 & 5.0 & 3.4 & 4.4 & 2.3 & 6.1 & 3.5 & 8.3 & 3.5 & 8.3 & 2.2 & 5.3 & 1.9 & 5.0 & 1.8 & 5.6 & 5.8 & 5.4 & 1.4 & 9.3 \\
Other & Newfoundland & 2.2 & 5.2 & 3.6 & 4.6 & 2.1 & 6.5 & 3.1 & 7.3 & 3.1 & 7.3 & 2.7 & 5.2 & 1.1 & 5.0 & 2.2 & 4.4 & 7.1 & 4.0 & 0.8 & 9.4 \\
\bottomrule
\end{tabular}
\end{adjustbox}
\vspace{0.3em}

\raggedright\scriptsize\textit{Note: Fine-tuned models show moderate, balanced errors with 19/50 dialects over-flagged and 31/50 under-protected (avg $\Delta$F$^+$ = +0.4\%, avg $\Delta$F$^-$ = +1.0\%), matching the SQ1 unseen pattern (Table~\ref{tab:fp_fn_sq1_human}).}
\caption{SQ4 Fine-Tuned Models: Per-dialect F$^+$ and F$^-$ (\%) on dialectal content. F$^+$ = FPR (over-flagging); F$^-$ = FNR (under-protection). Model abbreviations: mDeB = mDeBERTa, RoB = RoBERTa, DeB = DeBERTa, XLM = XLM-R, mB = mBERT, CT = CT-BERT, BiG = BiGRU, CNN = TextCNN, dEF = dEFEND. $^\dagger$Estimated from F1; pending recomputation.}
\label{tab:fp_fn_sq4_ft}
\end{table*}

\begin{table*}[t]
\centering
\small
\setlength{\tabcolsep}{3pt}

\begin{tabular}{ll cc cc cc cc cc}
\toprule
& & \multicolumn{2}{c}{\textbf{Mis}} & \multicolumn{2}{c}{\textbf{L3.1}} & \multicolumn{2}{c}{\textbf{L3.2}} & \multicolumn{2}{c}{\textbf{Q8}} & \multicolumn{2}{c}{\textbf{Q4}} \\
\cmidrule(lr){3-4} \cmidrule(lr){5-6} \cmidrule(lr){7-8} \cmidrule(lr){9-10} \cmidrule(lr){11-12}
& & \textbf{F$^+$} & \textbf{F$^-$} & \textbf{F$^+$} & \textbf{F$^-$} & \textbf{F$^+$} & \textbf{F$^-$} & \textbf{F$^+$} & \textbf{F$^-$} & \textbf{F$^+$} & \textbf{F$^-$} \\
\midrule
-- & SAE (baseline) & 0.0 & 98.8 & 0.0 & 33.6 & 0.0 & 54.0 & 0.0 & 30.7 & 0.0 & 26.4 \\
\midrule
\multicolumn{12}{l}{\textit{\textbf{U.S. Varieties}}} \\
U.S. & Appalachian & 0.3 & 99.0 & 4.7 & 35.2 & 4.6 & 60.7 & 3.2 & 32.5 & 17.6 & 23.7 \\
U.S. & Chicano & 0.1 & 98.9 & 1.9 & 33.4 & 3.5 & 55.5 & 2.1 & 30.8 & 7.4 & 24.7 \\
U.S. & Colloquial American & 0.4 & 99.1 & 4.9 & 34.7 & 6.0 & 60.6 & 4.7 & 32.9 & 19.0 & 22.7 \\
U.S. & Earlier AAVE & 0.0 & 99.5 & 6.1 & 35.5 & 5.7 & 61.4 & 5.9 & 32.4 & 25.4 & 21.5 \\
U.S. & Ozark & 0.2 & 99.2 & 4.9 & 34.3 & 5.3 & 58.6 & 4.1 & 32.1 & 17.2 & 23.6 \\
U.S. & Rural AAVE & 0.2 & 99.5 & 7.2 & 35.5 & 7.2 & 62.8 & 6.5 & 33.7 & 36.9 & 18.1 \\
U.S. & SE Amer. Enclave & 0.1 & 99.5 & 6.5 & 37.1 & 6.3 & 61.6 & 5.1 & 34.2 & 26.8 & 21.0 \\
U.S. & Urban AAVE & 0.2 & 99.5 & 6.2 & 34.1 & 5.9 & 62.0 & 5.4 & 32.7 & 21.5 & 21.9 \\
\midrule
\multicolumn{12}{l}{\textit{\textbf{British/UK Varieties}}} \\
UK & Channel Islands & 0.3 & 99.0 & 4.6 & 35.2 & 5.3 & 58.5 & 3.5 & 32.9 & 14.1 & 25.1 \\
UK & East Anglian & 0.1 & 99.2 & 4.9 & 33.6 & 4.7 & 58.3 & 3.9 & 31.6 & 16.0 & 24.1 \\
UK & Irish & 0.3 & 99.3 & 6.6 & 36.6 & 5.3 & 63.2 & 4.3 & 34.2 & 25.2 & 22.2 \\
UK & Manx & 0.2 & 99.4 & 5.9 & 42.1 & 4.6 & 64.0 & 5.1 & 35.2 & 27.2 & 21.8 \\
UK & North England & 0.2 & 99.5 & 6.8 & 37.3 & 5.0 & 62.7 & 5.2 & 34.7 & 31.0 & 19.6 \\
UK & Orkney \& Shetland & 0.2 & 99.4 & 6.0 & 37.1 & 5.9 & 61.2 & 4.5 & 35.4 & 23.4 & 23.1 \\
UK & SE England & 0.1 & 99.0 & 3.5 & 33.2 & 3.4 & 57.5 & 3.0 & 30.6 & 9.6 & 24.7 \\
UK & SW England & 0.1 & 99.5 & 7.9 & 35.0 & 5.7 & 61.6 & 5.4 & 34.5 & 32.4 & 19.5 \\
UK & Scottish & 0.2 & 99.3 & 5.4 & 36.5 & 7.1 & 60.6 & 3.9 & 33.8 & 17.1 & 24.9 \\
UK & Welsh & 0.2 & 99.4 & 7.1 & 42.6 & 6.8 & 59.2 & 5.3 & 40.2 & 28.5 & 25.0 \\
\midrule
\multicolumn{12}{l}{\textit{\textbf{Africa}}} \\
Africa & Black S. African & 0.2 & 99.5 & 7.5 & 36.3 & 5.2 & 64.5 & 5.6 & 34.9 & 31.9 & 20.3 \\
Africa & Cameroon & 0.2 & 99.3 & 5.9 & 35.1 & 6.9 & 61.9 & 5.9 & 33.3 & 20.4 & 23.2 \\
Africa & Cape Flats & 0.2 & 99.2 & 5.9 & 34.4 & 5.3 & 60.2 & 4.2 & 32.4 & 16.0 & 24.1 \\
Africa & Ghanaian & 0.1 & 99.4 & 5.1 & 34.9 & 6.6 & 60.3 & 4.8 & 32.9 & 19.7 & 23.6 \\
Africa & Indian S. African & 0.1 & 99.5 & 6.5 & 35.8 & 6.6 & 62.9 & 6.2 & 32.6 & 29.8 & 20.3 \\
Africa & Kenyan & 0.2 & 99.4 & 6.5 & 34.6 & 6.1 & 61.2 & 5.2 & 33.0 & 17.6 & 24.4 \\
Africa & Liberian Settler & 0.1 & 99.4 & 5.1 & 34.2 & 6.5 & 60.6 & 6.2 & 31.3 & 23.1 & 21.7 \\
Africa & Nigerian & 0.1 & 99.2 & 4.6 & 35.6 & 4.9 & 60.7 & 4.6 & 32.1 & 17.4 & 22.9 \\
Africa & Tanzanian & 0.2 & 99.1 & 4.2 & 34.4 & 5.3 & 60.0 & 3.4 & 32.4 & 11.8 & 25.0 \\
Africa & Ugandan & 0.3 & 99.0 & 5.8 & 33.6 & 5.3 & 59.9 & 4.1 & 32.0 & 15.9 & 24.0 \\
\midrule
\multicolumn{12}{l}{\textit{\textbf{Asia-Pacific}}} \\
Asia-Pac. & Aus. Vernacular & 0.2 & 99.2 & 4.6 & 36.7 & 6.0 & 59.6 & 3.4 & 34.1 & 17.1 & 25.4 \\
Asia-Pac. & Australian & 0.1 & 99.6 & 5.0 & 36.6 & 6.6 & 60.7 & 4.0 & 34.2 & 19.9 & 24.1 \\
Asia-Pac. & Fiji (Acrolectal) & 0.2 & 99.2 & 4.7 & 33.4 & 6.3 & 59.9 & 4.1 & 32.5 & 14.2 & 24.7 \\
Asia-Pac. & Fiji (Basilectal) & 0.0 & 99.8 & 7.2 & 39.6 & 6.6 & 64.8 & 7.7 & 35.9 & 48.4 & 14.5 \\
Asia-Pac. & Hong Kong & 0.2 & 99.5 & 7.6 & 36.3 & 6.9 & 63.6 & 6.3 & 33.9 & 29.6 & 20.8 \\
Asia-Pac. & Indian & 0.2 & 99.4 & 7.1 & 37.5 & 6.9 & 63.2 & 5.6 & 34.7 & 27.4 & 21.7 \\
Asia-Pac. & Malaysian & 0.1 & 99.3 & 6.6 & 33.4 & 6.8 & 60.3 & 6.8 & 31.4 & 23.1 & 21.4 \\
Asia-Pac. & New Zealand & 0.1 & 99.5 & 5.1 & 36.7 & 5.7 & 62.0 & 5.3 & 33.5 & 25.4 & 21.3 \\
Asia-Pac. & Pakistani & 0.2 & 99.3 & 5.6 & 32.8 & 5.2 & 60.8 & 4.6 & 31.3 & 19.8 & 21.8 \\
Asia-Pac. & Philippine & 0.2 & 99.4 & 6.9 & 36.4 & 6.7 & 62.3 & 5.4 & 34.4 & 31.6 & 19.9 \\
Asia-Pac. & Singlish & 0.1 & 99.5 & 7.5 & 35.5 & 7.2 & 62.5 & 8.2 & 32.6 & 34.7 & 18.7 \\
Asia-Pac. & Sri Lankan & 0.4 & 98.7 & 5.2 & 33.4 & 6.7 & 58.1 & 4.6 & 31.7 & 12.5 & 24.8 \\
\midrule
\multicolumn{12}{l}{\textit{\textbf{Caribbean/Atlantic}}} \\
Carib. & Bahamian & 0.1 & 99.5 & 7.3 & 37.8 & 6.7 & 62.8 & 7.8 & 34.5 & 36.8 & 18.5 \\
Carib. & Falkland Islands & 0.2 & 99.1 & 4.9 & 34.3 & 4.6 & 57.9 & 3.2 & 32.4 & 14.8 & 24.6 \\
Carib. & Jamaican & 0.1 & 99.3 & 6.2 & 35.1 & 5.8 & 61.6 & 6.4 & 32.7 & 27.0 & 20.5 \\
Carib. & St. Helena & 0.1 & 99.6 & 7.9 & 35.7 & 6.4 & 63.5 & 6.6 & 33.1 & 38.7 & 17.5 \\
Carib. & Tristan da Cunha & 0.2 & 99.5 & 6.2 & 35.7 & 6.2 & 61.5 & 5.7 & 33.0 & 31.8 & 19.7 \\
\midrule
\multicolumn{12}{l}{\textit{\textbf{Other}}} \\
Other & Aboriginal & 0.1 & 99.7 & 8.2 & 35.5 & 6.4 & 63.9 & 7.5 & 33.4 & 40.2 & 17.5 \\
Other & Maltese & 0.1 & 99.5 & 7.2 & 36.5 & 6.5 & 62.8 & 6.3 & 34.0 & 33.2 & 19.4 \\
Other & Newfoundland & 0.2 & 99.6 & 6.3 & 40.7 & 6.5 & 62.6 & 5.5 & 35.4 & 33.2 & 20.1 \\
\bottomrule
\end{tabular}

\vspace{0.3em}

\raggedright\scriptsize\textit{Note: Zero-shot LLMs show predominantly over-flagging (48/48 dialects with $\Delta$F$^+$ $>$ 0; avg $\Delta$F$^+$ = +8.3\%). Mistral-7B exhibits near-total failure (F$^-$ $\approx$ 99\%), providing virtually no disinformation protection. Qwen3-4B compensates with aggressive over-flagging (F$^+$ up to 30\%), effectively trading one harm for another.}
\caption{SQ4 Zero-Shot LLMs: Per-dialect F$^+$ and F$^-$ (\%). F$^+$ = FPR (over-flagging relative to model's own SAE classification); F$^-$ = FNR (absolute miss rate on dialectal content, including abstentions). Model abbreviations: Mis = Mistral-7B, L3.1 = Llama-3.1-8B, L3.2 = Llama-3.2-1B, Q8 = Qwen3-8B, Q4 = Qwen3-4B.}
\label{tab:fp_fn_sq4_zs}
\end{table*}


\begin{figure*}[t]
\centering
\includegraphics[width=0.7\textwidth]{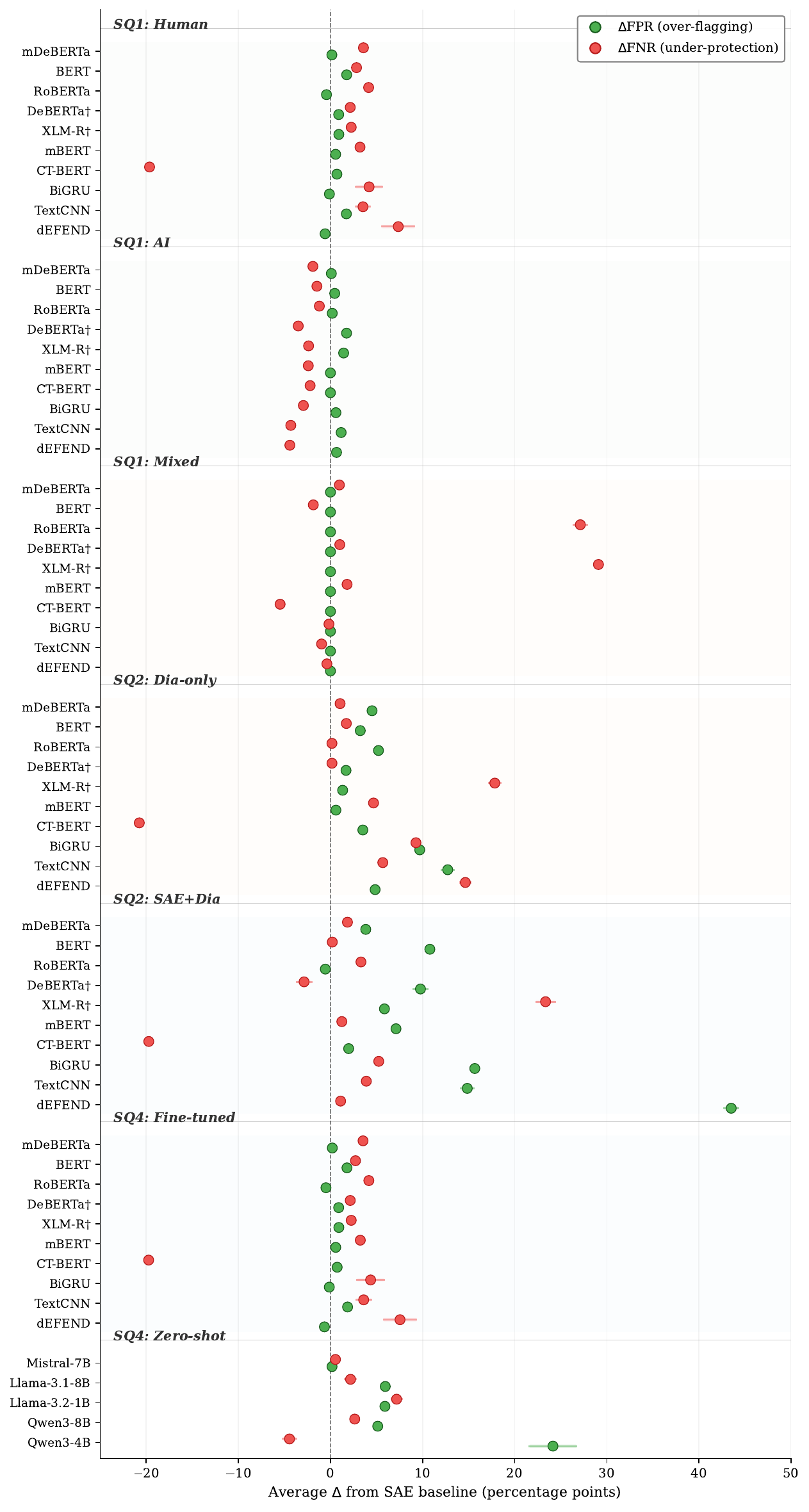}
\caption{Per-model asymmetric harm across all evaluation
regimes. Green dots show $\Delta$FPR (over-flagging) and
red dots show $\Delta$FNR (under-protection) relative to
the SAE baseline, with shaded regions indicating 95\%
confidence intervals across 50 dialects. Notable outliers
include dEFEND ($\Delta$FPR = +43.5) under SAE-anchored
training, RoBERTa and XLM-R ($\Delta$FNR = +27--29) under
mixed content, and CT-BERT ($\Delta$FNR = $-$20) showing
anomalous improvement under dialect-aware training.
Compact summary in Figure~\ref{fig:fp_fn_summary}.}
\label{fig:fp_fn_full}
\end{figure*}


\section{Qualitative Error Analysis}
\label{app:qualitative}

To complement the quantitative FPR/FNR analysis
(Appendix~\ref{app:fp_fn}), we conduct a qualitative
examination of model errors to identify the linguistic
mechanisms through which dialectal transformation induces
false positives (over-flagging) and false negatives
(under-protection). We extract all \textit{dialect-induced}
errors, defined as cases where the SAE baseline prediction
is correct but the dialectal prediction is wrong, yielding
27,020 false positives and 4,169 false negatives across
eight fine-tuned models and 50 dialects (SQ1 unseen regime,
human content). We further examine confidence patterns to
assess whether errors reflect model uncertainty or confident
misclassification.


\subsection{Taxonomy of Dialectal Failure Mechanisms}
\label{app:qual_taxonomy}

Table~\ref{tab:error_taxonomy} presents a taxonomy of six
recurring linguistic mechanisms that induce classification
errors in disinformation detection. These mechanisms are not
mutually exclusive; a single dialectal transformation may
trigger multiple mechanisms simultaneously.

\begin{table*}[!htb]
\centering
\small
\begin{tabular}{p{3cm}p{.8cm}p{10cm}}
\toprule
\textbf{Mechanism} & \textbf{Error} & \textbf{Description} \\
\midrule
Nominal pluralisation with \textit{them} &
FP &
Dialectal plural marker appended to nouns
(e.g., \textit{``experts say issues such as
vaccine safety''} $\to$
\textit{``expert them says issue them such
vaccine safety''}) introduces surface tokens
characteristic of non-standard text that
models associate with machine-generated or
fabricated content. \\
\addlinespace
\textit{a-}prefixing &
FP &
Appalachian/AAVE aspectual prefix
(e.g., \textit{``raising false hopes''} $\to$
\textit{``a-raising false hope them''})
creates rare subword tokens absent from
SAE training data, triggering false
disinformation signals. \\
\addlinespace
Determiner insertion &
FP &
Inserting articles before mass or abstract nouns
(e.g., \textit{``provides access to technical
guidelines''} $\to$ \textit{``is providing a
access to technical guideline them''})
produces ungrammatical-seeming text that
models conflate with AI-generated artefacts. \\
\addlinespace
Pronoun substitution &
FP &
Replacing first-person with third-person
pronouns (e.g., \textit{``I voted with
Trump on trade''} $\to$ \textit{``She voted
Trump trade''}) alters perceived authorial
agency, changing the pragmatic framing. \\
\addlinespace
Syntactic reordering &
FP/FN &
Dialectal word-order variation
(e.g., \textit{``criminal conviction''} $\to$
\textit{``conviction criminal''};
\textit{``from serving in the military''} $\to$
\textit{``from serving the military people''})
disrupts positional cues that models use for
classification. \\
\addlinespace
Progressive/ habitual aspect &
FP &
Non-standard progressive markers
(e.g., \textit{``I hate Senate''} $\to$
\textit{``I am hating Senate''};
\textit{``was never budgeted''} $\to$
\textit{``ain't budgeted''})
alter tense and aspect signalling. \\
\bottomrule
\end{tabular}
\caption{Taxonomy of linguistic mechanisms through which
dialectal transformation induces classification errors
in disinformation detection.
FP = false positive (over-flagging authentic content as
disinformation); FN = false negative (disinformation
evading detection).}
\label{tab:error_taxonomy}
\end{table*}


\subsection{Representative Error Examples}
\label{app:qual_examples}

Tables~\ref{tab:qual_fp} and~\ref{tab:qual_fn} present
representative false-positive and false-negative examples from
fine-tuned disinformation detectors, illustrating how each
failure mechanism operates on real content from the CoAID,
$F^3$, LIAR, and MMCOVID datasets.

\begin{table*}[t]
\centering
\small

\begin{tabular}{p{1.2cm}p{1.4cm}p{1.5cm}p{1.0cm}p{4.0cm}p{4.0cm}}
\toprule
\textbf{Model} & \textbf{Dialect} & \textbf{Mechanism} &
\textbf{Source} &
\textbf{SAE Text (classified real)} &
\textbf{Dialect Text (flagged as fake)} \\
\midrule
CT-BERT & North England &
Progressive aspect &
$F^3$ &
``Flashback to another article quoting a
\textit{longtime friend} saying I \textit{hate}
Senate. Words I have NEVER said to anyone.'' &
``Flashback another article quoting the
\textit{longtime friend} saying I
\textit{am hating} Senate. Words NEVER said
anyone.'' \\
\addlinespace
BiGRU & Channel Islands &
Pronoun swap &
$F^3$ &
``I voted with Trump on trade.'' &
``She voted Trump trade.'' \\
\addlinespace
TextCNN & Aboriginal &
Pluralisation (\textit{them}) &
$F^3$ &
``Now Republicans want to restore \#gunrights
to felons, something they lose after criminal
conviction. CJS bills have prevented for
24~years'' &
``Now Republicans want to restore \#gunright
them to felon them, something lose after
conviction criminal. CJS bill them have
prevented for 24~year them'' \\
\addlinespace
BiGRU & Earlier AAVE &
\textit{a-}prefixing &
CoAID &
``experts say issues such as vaccine safety
supply chains and public trust need to be
resolved before a covid-19 vaccine is
administered\ldots'' &
``expert them says issue them such vaccine
safety supply chain them and public trust
needs to be a-resolved before a covid-19
vaccine is a-administered\ldots'' \\
\addlinespace
CT-BERT & Black S. African &
Determiner insertion &
$F^3$ &
``Hey, @THR, kindly report on everyone
attending this event, so the rest of us can
be clear about who we don't wanna work with.'' &
``Hey, @THR, kind is reporting on a everyone
attending this a event, so that a rest of us
can be like clear about who we don't\ldots'' \\
\addlinespace
CT-BERT & SE England &
Syntactic reorder &
$F^3$ &
``Bush's Medicare D was far more expensive
than the Affordable Care Act, and, unlike
the ACA, was never budgeted'' &
``Bush's Medicare D was far like more
expensive than the Affordable Care, and,
the ACA Act, ain't budgeted'' \\
\bottomrule
\end{tabular}
\caption{Representative false-positive examples: authentic content
correctly classified as real on SAE but flagged as disinformation
after dialectal transformation. All examples are dialect-induced
errors (SAE baseline correct, dialect prediction wrong).}
\label{tab:qual_fp}
\end{table*}


\begin{table*}[t]
\centering
\small
\begin{tabular}{p{1.2cm}p{2.2cm}p{1.5cm}p{9.6cm}}
\toprule
\textbf{Model} & \textbf{Dialect} & \textbf{Source} &
\textbf{Disinformation Text (detected on SAE, missed on dialect)} \\
\midrule
TextCNN & North England &
CoAID &
``A `face mask exempt card' allows you to cite the ADA and not
wear a mask.'' \\
\addlinespace
TextCNN & Indian &
MMCOVID &
``California department of social services letter says failure to
obtain covid-19 test will result in suspension of benefits and
positive test will result in removal of children'' \\
\addlinespace
TextCNN & Maltese &
MMCOVID &
``Getting a flu shot increases the risk of coronavirus by 36\%.'' \\
\addlinespace
TextCNN & Falkland Islands &
MMCOVID &
``Consuming alcoholic beverages may help reduce the risk of
infection by the novel coronavirus.'' \\
\addlinespace
TextCNN & Malaysian &
LIAR &
``Virtually every person across this country has seen premiums
going up and up and up due to Obamacare.'' \\
\addlinespace
TextCNN & Rural AAVE &
LIAR &
``Says Marco Rubio voted to deport young people known as
Dreamers.'' \\
\bottomrule
\end{tabular}
\caption{Representative false-negative examples: disinformation
correctly detected on SAE but missed after dialectal
transformation. SAE text shown alongside the dialect and model
that failed to detect it. All examples are from established
fact-checking benchmarks (LIAR, MMCOVID, CoAID).}
\label{tab:qual_fn}
\end{table*}


\subsection{Error Distribution Across Models and Dialects}
\label{app:qual_distribution}

Table~\ref{tab:error_dist_model} summarises how dialect-induced
errors distribute across fine-tuned models. dEFEND produces the
most false positives (6,094; 22.5\% of all FPs), while TextCNN
produces the most false negatives (1,036; 24.9\% of all FNs).
DeBERTa and XLM-R produce zero dialect-induced errors because
their degenerate single-class predictions are equally wrong on
both SAE and dialectal content
(Section~\ref{app:qual_confidence}). Notably, false positives
outnumber false negatives by 6.5:1 overall, indicating that the
dominant harm pattern in disinformation detection is
over-flagging authentic dialectal content rather than missing
dialectal disinformation.

\begin{table}[H]
\centering
\small
\begin{tabular}{lrrrr}
\toprule
\textbf{Model} & \textbf{\# FP} & \textbf{\% FP} &
\textbf{\# FN} & \textbf{\% FN} \\
\midrule
dEFEND    &  6,094 & 22.5 &    219 &  5.3 \\
BiGRU     &  3,538 & 13.1 &    459 & 11.0 \\
RoBERTa   &  3,211 & 11.9 &    430 & 10.3 \\
mDeBERTa  &  3,178 & 11.8 &    353 &  8.5 \\
mBERT     &  3,082 & 11.4 &    591 & 14.2 \\
TextCNN   &  2,814 & 10.4 &  1,036 & 24.9 \\
BERT-L    &  2,751 & 10.2 &    830 & 19.9 \\
CT-BERT   &  2,352 &  8.7 &    251 &  6.0 \\
DeBERTa   &      0 &  0.0 &      0 &  0.0 \\
XLM-R     &      0 &  0.0 &      0 &  0.0 \\
\midrule
\textbf{Total} & \textbf{27,020} & \textbf{100.0} &
\textbf{4,169} & \textbf{100.0} \\
\bottomrule
\end{tabular}
\caption{Distribution of dialect-induced errors across
fine-tuned models (SQ1 unseen, human content). Only errors
where the SAE baseline is correct are counted, isolating
the effect of dialectal transformation. FP = authentic content
flagged as disinformation; FN = disinformation evading
detection.}
\label{tab:error_dist_model}
\end{table}

Table~\ref{tab:error_dist_dialect} shows the five dialects most
affected by each error type. Fiji (Basilectal) ranks highest for
false positives (1,294), reflecting the substantial
morphosyntactic distance from SAE. The over-flagging pattern
disproportionately affects African American (Rural AAVE, Earlier
AAVE), Caribbean (Bahamian, Jamaican), and African (Black S.\
African) varieties, communities already subject to
disproportionate content moderation.

\begin{table}[H]
\centering
\small
\begin{tabular}{lrlr}
\toprule
\multicolumn{2}{c}{\textbf{Most Over-Flagged}} &
\multicolumn{2}{c}{\textbf{Most Under-Protected}} \\
\cmidrule(lr){1-2}\cmidrule(lr){3-4}
\textbf{Dialect} & \textbf{\# FP} &
\textbf{Dialect} & \textbf{\# FN} \\
\midrule
Fiji (Basilectal) & 1,294 & Singlish          & 114 \\
Rural AAVE        & 1,023 & Cameroon          & 112 \\
Bahamian          &   920 & Welsh             & 109 \\
Black S.\ African &   878 & St.\ Helena       & 107 \\
St.\ Helena       &   848 & Cape Flats        & 102 \\
\bottomrule
\end{tabular}
\caption{Top five dialects by number of dialect-induced
false positives (over-flagged) and false negatives
(under-protected) across all eight non-degenerate fine-tuned
models.}
\label{tab:error_dist_dialect}
\end{table}

The false-positive errors split sharply by data source: Twitter
posts account for 71.7\% (19,387/27,020) of all false positives,
while news articles account for the remaining 28.3\%. This
asymmetry likely reflects the shorter, more informal nature of
tweets, where dialectal features constitute a larger proportion
of the total text and thus have greater impact on model
representations. False negatives concentrate in MMCOVID (2,440;
58.5\%) and LIAR (1,334; 32.0\%), suggesting that political
fact-check claims and COVID-related misinformation are
particularly vulnerable to dialectal evasion.


\subsection{Fine-Tuned Model Confidence on Errors}
\label{app:qual_confidence}

We examine prediction confidence for fine-tuned models to
assess whether dialectal errors are \textit{uncertain} (low
confidence, potentially correctable with calibration) or
\textit{confidently wrong} (high confidence, indicating
fundamental representation failure).
Table~\ref{tab:confidence_disinfo} reports average confidence
on dialect-induced errors and the percentage of errors made
with $>$0.95 confidence.

\begin{table}[H]
\centering
\small
\begin{tabular}{l cc cc}
\toprule
& \multicolumn{2}{c}{\textbf{Avg Confidence}} &
\multicolumn{2}{c}{\textbf{HC\%}} \\
\cmidrule(lr){2-3}\cmidrule(lr){4-5}
\textbf{Model} &
\textbf{FP $\bar{c}$} & \textbf{FN $\bar{c}$} &
\textbf{FP} & \textbf{FN} \\
\midrule
RoBERTa   & .995 & .998 & 95.9 & 99.8 \\
mBERT     & .994 & .993 & 97.7 & 96.1 \\
BERT-L    & .993 & .988 & 96.4 & 95.7 \\
mDeBERTa  & .987 & .952 & 94.1 & 79.6 \\
CT-BERT   & .934 & .921 & 84.2 & 71.7 \\
dEFEND    & .933 & .810 & 71.3 & 33.8 \\
BiGRU     & .911 & .872 & 63.1 & 49.5 \\
TextCNN   & .905 & .909 & 60.4 & 57.5 \\
\bottomrule
\end{tabular}
\caption{Average confidence ($\bar{c}$) and percentage of
high-confidence errors (HC\%, conf $>$ 0.95) for dialect-induced
false positives and false negatives. Only errors where the SAE
baseline is correct are included.}
\label{tab:confidence_disinfo}
\end{table}

Table~\ref{tab:confidence_disinfo} reveals that the vast
majority of dialect-induced errors are made with high confidence:
81.4\% of all false positives and 75.5\% of all false negatives
exceed the 0.95 confidence threshold. Three patterns emerge:

\paragraph{Confidently wrong transformers.}
RoBERTa, mBERT, and BERT-Large make over 95\% of their
dialect-induced errors with $>$0.95 confidence. RoBERTa is the
most extreme case: 99.5\% average FP confidence and 99.8\%
average FN confidence, meaning the model is near-certain in
its incorrect classifications. This rules out simple
calibration or threshold-tuning as remedies; the model's
internal representations fundamentally encode dialectal
features as class-discriminative signals.

\paragraph{Confidence directionality.}
CT-BERT shows that 75.1\% (1,956/2,603) of its dialect-induced
errors are made with \textit{higher} confidence on the dialect
input than on the SAE input. This suggests that dialectal
features do not merely inject noise but actively reinforce
incorrect decision patterns for certain architectures.

\paragraph{Model-specific vulnerability profiles.}
dEFEND produces the most false positives (6,094) with moderate
confidence (.933), suggesting its graph-based architecture is
broadly sensitive to surface-level dialectal variation.
TextCNN produces the most false negatives (1,036), with the
lowest FP confidence (.905) but near-average FN confidence
(.909), indicating that its character-level $n$-gram features
are differentially affected by dialectal morphological patterns
depending on the error direction.


\subsection{Summary of Qualitative Findings}
\label{app:qual_summary}

\begin{enumerate}[leftmargin=*,nosep]

\item \textbf{Over-flagging dominates.} Dialect-induced false
positives outnumber false negatives 6.5:1 (27,020 vs.\ 4,169).
The primary harm is that authentic dialectal speech, including
legitimate public health information and political discourse,
is systematically flagged as disinformation.

\item \textbf{Morphological markers are the primary trigger.}
The most frequent false-positive mechanism is dialectal
morphology: plural \textit{them}-suffixing, \textit{a-}prefixing,
and non-standard determiner use create surface tokens that
models have learned to associate with fabricated content,
likely because such patterns are absent from SAE-dominated
training data.

\item \textbf{Short-form content is most vulnerable.}
Twitter posts account for 71.7\% of false positives despite
representing only 33\% of the test data. In shorter texts,
dialectal features constitute a larger fraction of total tokens,
amplifying their influence on model representations.

\item \textbf{Morphosyntactically distant dialects are most
affected.} Fiji (Basilectal), Rural AAVE, and Bahamian English
rank highest for over-flagging, while closer varieties
(SE~England, Chicano) show fewer errors, consistent with a
linguistic-distance gradient.

\item \textbf{Errors are confidently wrong, not uncertain.}
81.4\% of false positives and 75.5\% of false negatives are
made with $>$0.95 confidence. RoBERTa achieves 99.5\% mean
confidence on false positives, ruling out calibration-based
fixes and indicating that dialectal features are encoded as
class-discriminative in the learned representations.

\item \textbf{Disinformation evasion targets specific domains.}
False negatives concentrate in political fact-checking (LIAR,
32.0\%) and COVID misinformation (MMCOVID, 58.5\%), with
TextCNN as the most vulnerable model (24.9\% of all FNs).
This suggests that dialectal transformation of topical
disinformation can exploit domain-specific detection
heuristics.

\end{enumerate}


\section{Prompt Variant and Few-Shot ICL Analysis}
\label{app:prompt_variants}

To assess whether the zero-shot brittleness reported in
\S\ref{sec:sq4} reflects a prompt artifact or a structural
limitation, we evaluate four prompt templates and four
in-context learning (ICL) conditions across three models
and five representative dialects.


\subsection{Experimental Setup}
\label{app:prompt_setup}

\paragraph{Models.}
We evaluate Llama-3.2-3B-Instruct, Gemma-3-1B-IT, and
Llama-3.2-1B-Instruct, spanning 1B--3B parameters. These
models represent the class of small, deployable LLMs most
likely to be used in resource-constrained content moderation
pipelines.

\paragraph{Dialects.}
Five dialects selected to span the morphosyntactic diversity
spectrum: Acrolectal Fiji English (most morphosyntactically
distant), Rural AAVE (most over-flagged in fine-tuned
evaluation), Singlish (most under-protected), Aboriginal
English (heavy morphological marking), and Indian English
(largest speaker population).

\paragraph{Prompt Templates.}
Table~\ref{tab:prompt_templates} presents the four zero-shot
prompt variants. All use the same output format constraint
(single-word classification) except P3 (chain-of-thought),
which elicits step-by-step reasoning before a final answer.

\begin{table*}[t]
\centering
\small

\begin{tabular}{p{0.6cm}p{1.8cm}p{12.5cm}}
\toprule
\textbf{ID} & \textbf{Style} & \textbf{Template} \\
\midrule
P1 & Original &
\texttt{Classify the following text as either `real' or
`fake' news. Respond with only one word: `real' or `fake'.
\textbackslash n\textbackslash nText: \{content\}
\textbackslash n\textbackslash nClassification:} \\
\addlinespace
P2 & Simplified &
\texttt{Is the following text real news or fake news?
Answer with one word only: real or fake.
\textbackslash n\textbackslash n\{content\}
\textbackslash n\textbackslash nAnswer:} \\
\addlinespace
P3 & Chain-of-thought &
\texttt{Read the following text carefully. Determine
whether it is real news or fake news (disinformation).
\textbackslash n\textbackslash nThink step by step:
\textbackslash n1. What claims does the text make?
\textbackslash n2. Does the language suggest factual
reporting or fabrication?
\textbackslash n3. Based on your analysis, is this real
or fake?
\textbackslash n\textbackslash nText: \{content\}
\textbackslash n\textbackslash nReasoning:} \\
\addlinespace
P4 & Role-based &
\texttt{You are an expert fact-checker and disinformation
analyst. Your task is to classify whether a piece of text
is authentic news reporting or fabricated disinformation.
Classify the following text. Respond with exactly one word:
`real' or `fake'.
\textbackslash n\textbackslash nText: \{content\}
\textbackslash n\textbackslash nVerdict:} \\
\bottomrule
\end{tabular}
\caption{Zero-shot prompt templates. \texttt{\{content\}}
is replaced with the input text at evaluation time.}
\label{tab:prompt_templates}
\end{table*}

\paragraph{Few-Shot ICL Conditions.}
Using P1 (Original) as the base template, we evaluate four
ICL configurations:

\begin{itemize}[nosep, leftmargin=*]
\item \textbf{P1-2S / P1-5S:} 2-shot and 5-shot with SAE
  exemplars (balanced real/fake).
\item \textbf{P1-2D / P1-5D:} 2-shot and 5-shot with
  dialect-matched exemplars (same source content as SAE
  exemplars, transformed into the target dialect via
  Multi-VALUE).
\end{itemize}

\noindent Exemplars are drawn from the training set with a
fixed random seed for reproducibility. SAE and
dialect-matched exemplars use identical source content,
isolating linguistic form as the only variable. Mid-length
examples (50--150 tokens) are selected for context
efficiency.


\subsection{Prompt Variant Results}
\label{app:prompt_results}

Table~\ref{tab:prompt_variant_results} reports F1 scores
across the four prompt templates. SAE baselines for
Gemma-3-1B and Llama-3.2-1B are taken from the full SQ4
evaluation (Table~\ref{tab:app_sq4_zeroshot}); the
Llama-3.2-3B SAE baseline (81.1\%) is obtained from our
experimental run under the same conditions.

\begin{table*}[t]
\centering
\small

\begin{tabular}{ll c ccccc cc}
\toprule
& & & \multicolumn{5}{c}{\textbf{Dialect F1 (\%)}} &
\multicolumn{2}{c}{\textbf{Summary}} \\
\cmidrule(lr){4-8}\cmidrule(lr){9-10}
\textbf{Prompt} & \textbf{Model}
& \textbf{SAE}
& \textbf{Fiji} & \textbf{R-AAVE}
& \textbf{Sing.} & \textbf{Abor.} & \textbf{Indian}
& \textbf{Dia.\ Avg} & \textbf{$\Delta$} \\
\midrule
\multirow{3}{*}{P1: Original}
& Llama-3.2-3B & 81.1 & 57.1 &  0.0 &  0.0 &  0.0 & 33.3 & 18.1 & $-$63.0 \\
& Gemma-3-1B   & 75.7 & 61.5 & 44.4 & 54.5 & 44.4 & 46.2 & 50.2 & $-$25.5 \\
& Llama-3.2-1B &  2.1 &  0.0 &  0.0 &  0.0 &  0.0 &  0.0 &  0.0 & $-$2.1 \\
\midrule
\multirow{3}{*}{P2: Simplified}
& Llama-3.2-3B & 81.1 & 60.0 &  0.0 &  0.0 &  0.0 & 33.3 & 18.7 & $-$62.4 \\
& Gemma-3-1B   & 75.7 & 71.4 & 60.0 & 60.0 & 66.7 & 66.7 & \textbf{65.0} & $-$10.7 \\
& Llama-3.2-1B &  2.1 & 33.3 &  0.0 & 33.3 & 33.3 & 33.3 & 26.6 & +24.5 \\
\midrule
\multirow{3}{*}{P3: CoT}
& Llama-3.2-3B & 81.1 & 60.0 & 28.6 & 33.3 & 33.3 & 33.3 & \textbf{37.7} & $-$43.4 \\
& Gemma-3-1B   & 75.7 & 66.7 & 66.7 & 57.1 & 57.1 & 66.7 & 62.9 & $-$12.8 \\
& Llama-3.2-1B$^\dagger$ & 2.1 & 66.7 & 66.7 & 66.7 & 66.7 & 66.7 & 66.7$^\dagger$ & +64.6$^\dagger$ \\
\midrule
\multirow{3}{*}{P4: Role-based}
& Llama-3.2-3B & 81.1 & 33.3 &  0.0 &  0.0 &  0.0 &  0.0 &  6.7 & $-$74.4 \\
& Gemma-3-1B   & 75.7 & 54.5 & 25.0 & 28.6 & 25.0 & 44.4 & 35.5 & $-$40.2 \\
& Llama-3.2-1B &  2.1 &  0.0 &  0.0 &  0.0 &  0.0 &  0.0 &  0.0 & $-$2.1 \\
\bottomrule
\end{tabular}
\caption{Effect of prompt template on dialectal disinformation
detection (F1\%). \textbf{SAE} = Standard American English
baseline.
\textbf{Dia.\ Avg} = mean across five dialects.
$\Delta$ = Dia.\ Avg $-$ SAE.
Best zero-shot result per model in \textbf{bold}.
$\dagger$\,Degenerate: classifies all inputs as fake
(Recall\,=\,1.0 for all dialects).}
\label{tab:prompt_variant_results}
\end{table*}


\subsection{Few-Shot ICL Results}
\label{app:icl_results}

Table~\ref{tab:icl_results} reports F1 scores across ICL
conditions. All use the P1 (Original) base template to
isolate the effect of exemplars from prompt wording.

\begin{table*}[t]
\centering
\small

\begin{tabular}{ll c ccccc cc}
\toprule
& & & \multicolumn{5}{c}{\textbf{Dialect F1 (\%)}} &
\multicolumn{2}{c}{\textbf{Summary}} \\
\cmidrule(lr){4-8}\cmidrule(lr){9-10}
\textbf{Config} & \textbf{Model}
& \textbf{SAE}
& \textbf{Fiji} & \textbf{R-AAVE}
& \textbf{Sing.} & \textbf{Abor.} & \textbf{Indian}
& \textbf{Dia.\ Avg} & \textbf{$\Delta$} \\
\midrule
\multirow{2}{*}{P1-0 (0-shot)}
& Llama-3.2-3B & 81.1 & 57.1 &  0.0 &  0.0 &  0.0 & 33.3 & 18.1 & $-$63.0 \\
& Gemma-3-1B   & 75.7 & 61.5 & 44.4 & 54.5 & 44.4 & 46.2 & 50.2 & $-$25.5 \\
\midrule
\multirow{2}{*}{P1-2S (2-shot SAE)}
& Llama-3.2-3B & 81.1 & 61.5 & 66.7 & 46.2 & 80.0 & 57.1 & \textbf{62.3} & $-$18.8 \\
& Gemma-3-1B   & 75.7 & 66.7 & 44.4 & 66.7 & 66.7 & 72.7 & \textbf{63.4} & $-$12.3 \\
\midrule
\multirow{2}{*}{P1-5S (5-shot SAE)}
& Llama-3.2-3B & 81.1 & 61.5 & 50.0 & 54.5 & 66.7 & 60.0 & 58.5 & $-$22.6 \\
& Gemma-3-1B   & 75.7 & 57.1 & 60.0 & 33.3 &  0.0 & 57.1 & 41.5 & $-$34.2 \\
\midrule
\multirow{2}{*}{P1-2D (2-shot Dia.)}
& Llama-3.2-3B & 81.1 & 28.6 & 33.3 & 28.6 &  0.0 & 28.6 & 23.8 & $-$57.3 \\
& Gemma-3-1B   & 75.7 & 60.0 & 25.0 & 33.3 & 33.3 & 57.1 & 41.7 & $-$34.0 \\
\midrule
\multirow{2}{*}{P1-5D (5-shot Dia.)}
& Llama-3.2-3B & 81.1 & 28.6 & 25.0 & 28.6 &  0.0 & 50.0 & 26.4 & $-$54.7 \\
& Gemma-3-1B   & 75.7 & 60.0 & 66.7 &  0.0 & 60.0 & 66.7 & 50.7 & $-$25.0 \\
\bottomrule
\end{tabular}
\caption{Effect of in-context learning on dialectal
disinformation detection (F1\%). SAE = exemplars in
Standard American English; Dia.\ = exemplars in the
target dialect. $\Delta$ = Dia.\ Avg $-$ SAE.
Best ICL result per model in \textbf{bold}.
Llama-3.2-1B omitted: 0.0\% F1 across all ICL conditions
(complete instruction-following failure).}
\label{tab:icl_results}
\end{table*}


\subsection{Analysis}
\label{app:prompt_analysis}

Three findings emerge from these experiments:

\paragraph{Prompt variation does not resolve dialectal
brittleness.}
Across all three models, no prompt template achieves
reliable dialectal performance. The best zero-shot
configuration (Gemma-3-1B with P2: Simplified) reaches only
65.0\% average dialect F1 ($\Delta$\,=\,$-$10.7 from SAE),
well below the 90\%+ thresholds achieved by fine-tuned
models in SQ1--SQ3. Llama-3.2-3B shows even larger gaps,
with $\Delta$ ranging from $-$43.4 (CoT) to $-$74.4
(role-based). The role-based prompt (P4) consistently
produces the worst results across all models, suggesting
that expert framing increases sensitivity to dialectal
surface forms. Llama-3.2-1B remains at 0.0\% F1 for 6 of 8
configurations; its apparent P3 ``success'' (66.7\%) is
degenerate, with Recall\,=\,1.0 across all dialects
indicating it classifies every input as fake. These results
confirm that the zero-shot brittleness reported in
\S\ref{sec:sq4} is robust across prompting strategies, not
an artifact of a single template.

\paragraph{SAE exemplars help; dialect-matched exemplars
hurt.}
For Llama-3.2-3B, 2-shot SAE exemplars produce the largest
improvement, raising average dialect F1 from 18.1\% to
62.3\% ($\Delta$ narrows from $-$63.0 to $-$18.8).
However, dialect-matched exemplars using identical source
content \emph{worsen} the gap to $\Delta$\,=\,$-$57.3
(23.8\% avg F1), 38.5 points below the SAE-exemplar
condition. This asymmetry persists for Gemma-3-1B (2-shot
SAE: $\Delta$\,=\,$-$12.3 vs.\ 2-shot dialect:
$\Delta$\,=\,$-$34.0). The finding is counterintuitive:
providing dialectal exemplars should familiarize the model
with dialectal forms, yet the models cannot effectively
process dialectal exemplars, revealing that the
comprehension deficit extends beyond test inputs to
in-context exemplars themselves. Llama-3.2-1B shows
complete instruction-following failure (0.0\% F1) across
all ICL conditions, indicating that few-shot exemplars
cannot rescue fundamentally inadequate dialectal
comprehension.

\paragraph{Performance remains far below fine-tuned
baselines.}
Even the best prompting configuration (Llama-3.2-3B with
P1-2S: 62.3\%, $\Delta$\,=\,$-$18.8) falls 30+ F1 points
below fine-tuned transformers evaluated under comparable
conditions in SQ1 (e.g., mDeBERTa: 97.2\%). Across all
models, no configuration eliminates the dialectal gap:
$\Delta$ ranges from $-$10.7 (Gemma-3-1B, P2) to $-$74.4
(Llama-3.2-3B, P4). This persistent degradation reinforces
Recommendation R4 (\S\ref{sec:discussion}): zero-shot LLMs
remain unsuitable for dialectally robust content moderation
regardless of prompting strategy.

\end{document}